\documentclass[10pt,journal,cspaper,compsoc]{IEEEtran}
 \usepackage[pdftex]{graphicx}
 \usepackage[cmex10]{amsmath}
 \usepackage{array}

 \usepackage{epsfig}
 \usepackage{amssymb}
 \usepackage{multirow}
 \usepackage{rotating}
 \usepackage{subfigure}
 \usepackage{url}
 \usepackage{color}
 \usepackage{multicol}

\ifCLASSOPTIONcompsoc
  \usepackage[nocompress]{cite}
\else
  \usepackage{cite}
\fi

\ifCLASSINFOpdf
\else
\fi

\begin{document}

\title{Structured Depth Prediction in Challenging Monocular Video Sequences}

\author{Miaomiao~Liu,~\IEEEmembership{Member,~IEEE,}
        Mathieu~Salzmann,~\IEEEmembership{Member,~IEEE,}
        and~Xuming~He,~\IEEEmembership{Member,~IEEE}
\IEEEcompsocitemizethanks{\IEEEcompsocthanksitem M. Liu, M. Salzmann, and X. He are with NICTA, Canberra, Australia.\protect\\
E-mail: \{miaomiao.liu, mathieu.salzmann, xuming.he\}@nicta.com.au
}
\thanks{Manuscript received April 19, 2005; revised September 17, 2014.}}

\markboth{Journal of \LaTeX\ Class Files,~Vol.~13, No.~9, September~2014}%
{Shell \MakeLowercase{\textit{et al.}}: Bare Advanced Demo of IEEEtran.cls for Journals}
\IEEEcompsoctitleabstractindextext{%
	\begin{abstract}
In this paper, we tackle the problem of estimating the depth of a scene from a monocular video sequence. In particular, we handle challenging scenarios, such as non-translational camera motion and dynamic scenes, where traditional structure from motion and motion stereo methods do not apply. To this end, we first study the problem of depth estimation from a single image. In this context, we exploit the availability of a pool of images for which the depth is known, and formulate monocular depth estimation as a discrete-continuous optimization problem, where the continuous variables encode the depth of the superpixels in the input image, and the discrete ones represent relationships between neighboring superpixels. The solution to this discrete-continuous optimization problem is obtained by performing inference in a graphical model using particle belief propagation. To handle video sequences, we then extend our single image model to a two-frame one that naturally encodes short-range temporal consistency and inherently handles dynamic objects. Based on the prediction of this model, we then introduce a fully-connected pairwise CRF that accounts for longer range spatio-temporal interactions throughout a video. We demonstrate the effectiveness of our model in both the indoor and outdoor scenarios.
\end{abstract}

}

\maketitle

\IEEEdisplaynontitleabstractindextext

%
\IEEEpeerreviewmaketitle

\newcommand{\Perp}{\perp\!\!\! \perp}

\newcommand{\comment}[1]{}

\newcommand{\bK}{\mathbf{K}}
\newcommand{\bX}{\mathbf{X}}
\newcommand{\bY}{\mathbf{Y}}
\newcommand{\bk}{\mathbf{k}}
\newcommand{\bx}{\mathbf{x}}
\newcommand{\bbx}{\bar{\mathbf{x}}}
\newcommand{\by}{\mathbf{y}}
\newcommand{\bhy}{\hat{\mathbf{y}}}
\newcommand{\bty}{\tilde{\mathbf{y}}}
\newcommand{\bG}{\mathbf{G}}
\newcommand{\bI}{\mathbf{I}}
\newcommand{\bg}{\mathbf{g}}
\newcommand{\bS}{\mathbf{S}}
\newcommand{\bs}{\mathbf{s}}
\newcommand{\bM}{\mathbf{M}}
\newcommand{\bw}{\mathbf{w}}
\newcommand{\eye}{\mathbf{I}}
\newcommand{\bU}{\mathbf{U}}
\newcommand{\bV}{\mathbf{V}}
\newcommand{\bW}{\mathbf{W}}
\newcommand{\bn}{\mathbf{n}}
\newcommand{\bv}{\mathbf{v}}
\newcommand{\bq}{\mathbf{q}}
\newcommand{\bR}{\mathbf{R}}
\newcommand{\bi}{\mathbf{i}}
\newcommand{\bj}{\mathbf{j}}
\newcommand{\bp}{\mathbf{p}}
\newcommand{\bt}{\mathbf{t}}
\newcommand{\bJ}{\mathbf{J}}
\newcommand{\bu}{\mathbf{u}}
\newcommand{\bB}{\mathbf{B}}
\newcommand{\bD}{\mathbf{D}}
\newcommand{\bz}{\mathbf{z}}
\newcommand{\bP}{\mathbf{P}}
\newcommand{\bC}{\mathbf{C}}
\newcommand{\bA}{\mathbf{A}}
\newcommand{\bZ}{\mathbf{Z}}
\newcommand{\bff}{\mathbf{f}}
\newcommand{\bF}{\mathbf{F}}
\newcommand{\bo}{\mathbf{o}}
\newcommand{\bc}{\mathbf{c}}
\newcommand{\bm}{\mathbf{m}}
\newcommand{\bT}{\mathbf{T}}
\newcommand{\bQ}{\mathbf{Q}}
\newcommand{\bL}{\mathbf{L}}
\newcommand{\bl}{\mathbf{l}}
\newcommand{\ba}{\mathbf{a}}
\newcommand{\bE}{\mathbf{E}}
\newcommand{\bH}{\mathbf{H}}
\newcommand{\bd}{\mathbf{d}}
\newcommand{\br}{\mathbf{r}}
\newcommand{\be}{\mathbf{e}}
\newcommand{\bhe}{\hat{\mathbf{e}}}
\newcommand{\bb}{\mathbf{b}}
\newcommand{\bh}{\mathbf{h}}
\newcommand{\bhh}{\hat{\mathbf{h}}}
\newcommand{\beps}{\boldsymbol{\epsilon}}

\newcommand{\lambdahat}{\hat{\lambda}}
\newcommand{\lambdae}{{\lambda}^E}
\newcommand{\lambdaev}{{\boldsymbol{\lambda}}^E}
\newcommand{\lambdai}{{\lambda}^I}
\newcommand{\lambdaiv}{{\boldsymbol{\lambda}}^I}

\newcommand{\btheta}{\boldsymbol{\theta}}
\newcommand{\bpi}{\boldsymbol{\pi}}
\newcommand{\bphi}{\boldsymbol{\phi}}
\newcommand{\bPhi}{\boldsymbol{\Phi}}
\newcommand{\bmu}{\boldsymbol{\mu}}
\newcommand{\bSigma}{\boldsymbol{\Sigma}}
\newcommand{\bGamma}{\boldsymbol{\Gamma}}
\newcommand{\bbeta}{\boldsymbol{\beta}}
\newcommand{\bomega}{\boldsymbol{\omega}}
\newcommand{\blambda}{\boldsymbol{\lambda}}
\newcommand{\bkappa}{\boldsymbol{\kappa}}
\newcommand{\btau}{\boldsymbol{\tau}}
\newcommand{\balpha}{\boldsymbol{\alpha}}
\def\bgamma{\boldsymbol\gamma}

\newcommand{\argmin}{\operatornamewithlimits{argmin}}
\newcommand{\minimize}{\operatornamewithlimits{min}}
\newcommand{\minimizem}{\operatornamewithlimits{minimize}}
\newcommand{\maximize}{\operatornamewithlimits{max}}

 \newcommand{\ikron}[1] {\bI\otimes #1}
  \newcommand{\val}{\bar{\bx}}
    \newcommand{\train}[1]{{\phi(\bx_{#1})}}
    \newcommand{\ikronval}[1]{(\ikron{\phi(\val_{#1}))}}
\newcommand{\ikronvalT}[1]{(\ikron{\phi(\val_{#1})^T)}}
\newcommand{\ikrontrainT}{(\ikron{\train{i}^T)}}
\newcommand{\ikrontrain}[1]{(\ikron{\train{#1})}}
\newcommand{\ikrontrainAT}{(\ikron{\phi(\bx)^T)}}
\newcommand{\ikrontrainA}{(\ikron{\phi(\bx))}}
  \newcommand{\half}{\frac{1}{2}}
  \newcommand{\con}{C^{(c,u)}}
  \newcommand{\ineqcon}{D^{(d,v)}}
\newcommand{\conv}{C^{(u)}}
  \newcommand{\ineqconv}{D^{(v)}}
    \newcommand{\ig}{\frac{1}{\gamma}}
      \newcommand{\Bi}{\bB^{-1}}
 \newcommand{\kernel}{\bK}    
 \newcommand{\ikrontestT}{(\ikron{\test^T)}}
   \newcommand{\test}{\phi(\bx_*)}

 \newcommand{\pardev}[2]{\frac{\partial #1}{\partial #2}}
  \newcommand{\dw}{\delta\bw}
  \newcommand{\dW}{\delta\bW}
    \newcommand{\deps}{\delta \beps}
  
    \newcommand{\lab}{\mathcal{L}}
 	     \newcommand{\unlab}{\mathcal{U}}

\newcommand{\argmax}[1]{\underset{#1}{\mathrm{argmax}} \:}
\def\eop {{\noindent\framebox[0.5em]{\rule[0.25ex]{0em}{0.75ex}}}}

\ifCLASSOPTIONcompsoc
\IEEEraisesectionheading{\section{Introduction}\label{sec:introduction}}
\else
\section{Introduction}
\label{sec:introduction}
\fi

\IEEEPARstart{D}{epth} estimation from a monocular video sequence is a fundamental problem in computer vision, with potential applications in many domains such as video editing, automatic 3D movie generation, and image-based modeling. Most existing methods, such as structure from motion (SFM)~\cite{Snavely06} and motion stereo~\cite{ZhangConsis2009}, are only applicable to static scenes and require the camera motion to introduce parallax. In contrast, in this paper, we tackle the challenging problem of depth estimation from videos acquired by non-translational cameras and depicting dynamic scenes, where SFM and motion stereo do not apply. 

While there has been relatively little work on video depth estimation in these challenging conditions~\cite{Karsch12,Karsch14}, much progress has been recently made on the related task of single image depth estimation.
In particular, simple geometric assumptions (i.e., box models) have proven effective to estimate the layout of a room~\cite{Hedau10,Lee10,Schwing12}. Similarly, for outdoor scenes, the Manhattan, or blocks world, assumption has been utilized to perform 3D scene layout estimation~\cite{Gupta10}. These models, however, are restricted to represent simple structures, and are therefore ill-suited to perform detailed 3D reconstruction.
By contrast, several methods have been proposed to directly estimate the depth of image (super)pixels~\cite{Saxena07,Saxena09}. In this context, it was shown that exploiting additional sources of information, such as user annotations~\cite{Russell09}, semantic labels~\cite{Liu10}, or the presence of repetitive structures~\cite{Wu11}, could help improving reconstruction accuracy. To tackle the cases where such additional information is not available, several nonparametric approaches were recently introduced~\cite{Karsch12,KonradSPIE, KonradDCINE}. Given an input image, these approaches proceed by retrieving similar images in a pool of images for which the depth is known. The depth maps of the retrieved candidates are then employed in conjunction with smoothness constraints to estimate a depth map. While this has achieved some success, as suggested in~\cite{Yamaguchi12} in the context of stereo, the gradient-aware smoothing strategy often poorly reflects the real 3D scene observed in the image. 

In this paper, inspired by this recent progress, we tackle the video depth estimation problem by first introducing an approach to single image depth estimation. In particular, we propose to model depth estimation as a discrete-continuous optimization problem, which, in addition to the standard continuous variables that encode the depth of the superpixels in the input image, makes use of discrete variables that allow us to model complex relationships between neighboring superpixels. This lets us express depth estimation as inference in a discrete-continuous graphical model, which can be performed using particle belief propagation.


We then extend our approach to handle monocular video-based dynamic scene depth estimation. To this end, we follow a similar structured prediction strategy as for the single image case, but further reason about frame-to-frame motion and model the static and dynamic structure of the scene within a single image and across the frames of a video. The resulting framework allows us to inherently {\bf (i)} account for moving objects; and {\bf(ii)} model complex interactions between different regions in the scene, both spatially and temporally.

More specifically, we extend our single image model to a two-frame conditional random field (CRF) to jointly estimate depth and motion in two consecutive frames of a sequence. This model lets us encode spatial coherence via geometric relationships between neighboring image superpixels, such as occlusion and co-planarity, as well as impose short-term temporal consistency in depth estimation. Furthermore, it explicitly accounts for moving objects by exploiting a motion-aware energy that enforces depth coherence between these objects and their surrounding environment.

To obtain a coherent depth sequence throughout a video, we introduce a second CRF that encodes long-range spatio-temporal consistency of the pixel depths within and across frames. To this end, we make use of our two-frame CRF depth estimates and construct a fully connected pairwise graph over all the pixels in a sequence. With Gaussian edge potentials, inference in such a CRF can be performed efficiently~\cite{Krah11}. Our fully-connected model inherently represents the spatio-temporal scene structure over the pixels of a video. As a consequence, not only does it yield temporal smoothness, but it also helps improving the accuracy of depth boundary estimation.

We demonstrate the effectiveness of our single image and video-based depth estimation methods on NYUv2, Make3D, and MSR-V3D. Our experiments evidence the benefits of our discrete-continuous model for single image depth estimation, as well as the benefits of our video-based approach over single frame depth estimation methods and over the state-of-the-art video-based depth estimation technique of~\cite{Karsch12,Karsch14}.

\section{Related Work}
\label{sec:related}

Depth estimation of video sequences has attracted a lot of attention in the literature. For instance, structure-from-motion (SFM) was introduced to obtain the camera parameters and sparse 3D points from a monocular video sequence. Given the estimated camera parameters and sparse 3D structure, dense depth maps can be computed either by using traditional multi-view stereo techniques~\cite{ZhangHand2011,ZhangConsis2009}, or by propagating the sparse depths to each pixel in the video sequence according to user-defined segments~\cite{WardKB2011}. Most existing methods assume that the video sequence is acquired by a camera whose motion introduces sufficient parallax for the depth to be reliably estimated. As a consequence, SFM will typically fail for non-translational cameras.

To address this limitation, it was shown that depth estimation of single static images can have a beneficial impact on the depth estimation of non-translational video sequences and of dynamic videos~\cite{Karsch12,Karsch14}.  In recent years, single image depth estimation has attracted increasing attention in the community. For indoor scenes, effective techniques have been proposed to estimate the layout of rooms. These methods typically rely on box-shaped models, and try to fit the box edges to those observed in the image~\cite{Hedau10,Lee10,Schwing12}. The same simple geometric prior, blocks world, was exploited in outdoor scenes~\cite{Gupta10}. In~\cite{Hoiem05}, a more accurate geometric model was employed, but the results remain only a rough estimate of surface normals.

The simple geometric models described above do not allow us to obtain a detailed 3D description of the scene. In contrast, several methods have proposed to directly estimate the depth of image (super)pixels. Since a single image does typically not provide enough information to estimate depth, other sources of information have been exploited. In particular, in~\cite{Russell09}, depth was predicted from user annotations. In~\cite{Liu10}, this was achieved by making use of semantic class labels. Alternatively, the presence of repetitive structures in the scene was also employed for 3D reconstruction~\cite{Wu11}. With the recent popularity of depth sensors, sparse depths have also been used to estimate denser depth maps~\cite{Aodha12}. 

To tackle less constrained scenario of single image depth estimation in the absence of additional information, the common approach consists of exploiting training data. In this context, several methods that learn local depth predictors have been proposed. In particular, in~\cite{Ladicky14}, specific classifiers were trained for different semantic classes at canonical depths, thus allowing to predict depth at pixel level. In~\cite{Eigen14,Facvpr15}, deep networks were trained from a large collection of image-depth pairs to estimate the depth of the pixels in an input image. While these method have proven effective, they fail to explicitly model the relationships of neighboring regions in the input image.

By contrast, other techniques have proposed to explicitly model these relationships. In particular, the pioneering work of~\cite{Saxena07,Saxena09} modeled depth estimation with a Markov random field whose edges encoded a simple smoothness term between image superpixels. In~\cite{Liu10}, a similar CRF was employed, but relying on better geometric priors and semantic class labels. Besides graphical models, other methods have been proposed to infer depth from a single image in a structured manner. In particular, several non-parametric techniques that transfer depth from training image-depth pairs were introduced recently~\cite{Karsch12,KonradSPIE,KonradDCINE}. More specifically, depth in the input image is estimated by first retrieving similar images in the training set, and combining the depth maps of the retrieved images via a nonlinear, continuous optimization problem that encourages the resulting depth map to be smooth. 

Our approach is close in spirit to that of~\cite{Karsch12,KonradSPIE,KonradDCINE} in the sense that we also make use of a nonparametric method to retrieve candidate depth maps. However, we avoid the warping process of~\cite{Karsch12,KonradSPIE}, which is computationally expensive and does not necessarily improve the quality of the candidates. Furthermore, as~\cite{Saxena07,Saxena09,Liu10}, we formulate depth estimation as inference in a graphical model. By contrast, however, we treat depth as continuous variables and introduce additional discrete ones that allow us to model more complex relationships between neighboring superpixels. As a result, beyond explicitly modeling the dependencies between neighboring superpixels, our discrete-continuous CRF lets us encode more complex relationships, thus better respecting depth discontinuities than the above-mentioned two lines of research.

Our goal here is to go beyond single image depth estimation and handle monocular videos of dynamic scenes acquired by non-translational cameras. To the best of our knowledge, only~\cite{Karsch12,Karsch14} have tackled this challenging scenario. Unfortunately,~\cite{Karsch12,Karsch14} formulate depth estimation as a continuous optimization problem, which makes it hard to inherently account for moving objects, as well as to encode complex priors modeling spatio-temporal discontinuities.

Here, by contrast, we follow a structured prediction approach that lets us reason about geometry and motion jointly, and thus model complex spatio-temporal relationships in a unified framework. In particular, our two-frame CRF allows us to inherently account for moving objects and their interactions with their surroundings, and our fully-connected model encodes long-range spatio-temporal connections throughout a video.

\section{Depth Estimation from a Single Image}
\label{sec:method}
We now describe our approach to depth estimation from a single image. To this end, we first derive the Conditional Random Field (CRF) that defines our problem, and discuss the inference method that we use. We then define the different potentials utilized in our model.

\subsection{A Discrete-Continuous CRF}

Our goal is to estimate the depth of the pixels observed in a single image depicting a general scene. We formulate this problem in terms of superpixels, making the common assumption that each superpixel is planar. The pose of a superpixel is then expressed in terms of the depth of its centroid and its plane normal. Furthermore, we make use of additional discrete variables that encode the relationship of two neighboring superpixels. In particular, here, we consider 4 types of relationships encoding the fact that the two superpixels {\bf (i)} belong to the same object; {\bf (ii)} belong to two different but connected objects; {\bf (iii)} belong to two objects that form a left occlusion; and {\bf (iv)} belong to two objects that form a right occlusion. Here, the notion of left and right occlusions follows the formalism of~\cite{Hoiem07} based on edge directions. Given these variables, we express depth estimation as an inference problem in a discrete-continuous CRF.

More specifically, let $\bY = \{\by_1,\dots,\by_S\}$ be the set of continuous variables, where each $\by_i\in \mathbb{R}^4$ concatenates the centroid depth and plane normal of superpixel $i$, and where $S$ is the total number of superpixels in the input image. Furthermore, let $\bE = \{\be_{p}\}_{p\in\mathcal{E}}$ be the set of discrete variables, where each $\be_{p}\in \{so,co,lo,ro\}$, which indicates same object ($so$), connected but different objects ($co$), left occlusion ($lo$) and right occlusion ($ro$), respectively. $\mathcal{E}$ is the set of pairs of superpixels that share a common boundary.

Given these variables, we then form a CRF, where the joint distribution over the random variables factorizes into a product of non-negative potentials. This joint distribution can be written as
\begin{equation*}
	p(\bY,\bE) = \frac{1}{Z} \prod_i\Psi_i(\by_i) \prod_{\alpha} \Psi_{\alpha}(\by_{\alpha},\be_{\alpha}) \prod_{\beta}\Psi_{\beta}(\be_{\beta}) \;,
\end{equation*}
where $Z$ is a normalization constant, i.e., the partition function, $\Psi_i$ is a unary potential function over the continuous variables that defines the data term for depth, and $\Psi_{\alpha}$ and $\Psi_{\beta}$ are potentials over mixed variables and discrete variables, respectively, which encode the smoothness and consistency between depth and edge types.

Inference in the graphical model is then performed by computing a MAP estimate. By working with negative log potential functions, e.g., $\phi_i(\by_i) = -\ln\left(\Psi_i(\by_i)\right)$, this can be expressed as the optimization problem
\begin{align}\label{eq:map_si}
	&\left(\bY^*,\bE^*\right) \\&= \argmin_{\bY,\bE} \sum_i \phi_i(\by_i) + \sum_{\alpha}\phi_{\alpha}(\by_{\alpha},\be_{\alpha}) + \sum_{\beta}\phi_{\beta}(\be_{\beta}) \;\nonumber.
\end{align}
The potentials that we use here are discussed in Section~\ref{sec:potentials_si}.

To handle mixed discrete and continuous variables, we make use of particle (convex) belief propagation (PCBP)~\cite{Peng11}, which lets us obtain an approximate solution to the optimization problem~\eqref{eq:map_si}. More specifically, PCBP proceeds by iteratively solving the following steps:
\begin{enumerate}
	\item Draw $N_s$ random samples $\by_i^j\;,\;1\leq j \leq N_s$ around the previous MAP solution for each variable $\by_i$.
	\item Compute the (approximate) MAP solution of the discrete CRF formed by the discrete variables $\{\be_p\}$ and by utilizing the random samples $\{\by_i^j\}$ as discrete states for the variables $\{\by_i\}$.
\end{enumerate}
In practice, we draw samples for the plane normal of the superpixels according to a Fisher-Bingham distribution, which forces them to have unit norm. Samples for the depth of the centroid of each superpixel are drawn according to a Gaussian distribution. At each iteration, we tighten the sampling around the previous MAP solution. The approximate MAP of the discrete CRF is obtained by distributed convex belief propagation~\cite{Schwing11}.

In this work, we make use of a nonparametric approach to obtain a reasonable initialization for the algorithm. In particular, we retrieve the $K$ images most similar to the input image from a set of images for which the depth is known. To this end, we perform a nearest-neighbor search based on concatenated GIST~\cite{Olivagist2001}, PHOG~\cite{annaphog2007} and Object Bank~\cite{lijia2010} features and directly make use the depth of the retrieved images, i.e., in contrast to~\cite{Karsch12,KonradSPIE}, we do not warp the depth of the retrieved images. The retrieved $K$ depth maps then directly act as states in the first round of PCBP, i.e., no random samples are used in this round.

In the next section, we describe the specific potentials used in the optimization problem~\eqref{eq:map_si}.

\subsection{Depth and Occlusion Potentials}
\label{sec:potentials_si}

The objective function in~\eqref{eq:map_si} contains three different types of potentials involving, respectively, continuous variables only, discrete and continuous variables, and discrete variables only. Below, we discuss the functions used in these three different types of potentials.

\vspace{0.2cm}
\noindent {\bf Potentials for continuous variables:}\mbox{}\\
The potentials involving purely continuous variables are unary potentials, and are of two different kinds. For the first one, we exploit the $K$ candidates retrieved by the image-based nearest-neighbor strategy mentioned in the previous section. The first potential encodes the fact that the final depth should remain close to at least one candidate. To this end, we make use of the squared depth difference. More specifically, assuming a calibrated camera (i.e., known camera intrinsics), the depth $d_i^{\bu_j}$ of pixel $\bu_j = (u_j,v_j)$ in superpixel $i$ can be obtained by intersecting the visual ray passing through $\bu_j$ with the plane defined by $\by_i$. This lets us write the potential
\begin{equation}
\phi^c_i(\by_i) = \min_{k=1}^K \frac{1}{N^p_i} \sum_{j=1}^{N^p_i} (d_i^{\bu_j}(\by_i) - d_{k,i}^{\bu_j})^2\;,
\label{eq:pot_depth_knn}
\end{equation}
where $N^p_i$ is the number of pixels in superpixel $i$, and $d_{k,i}^{\bu_j}$ denotes the depth of the $k^{th}$ candidate for superpixel $i$ at pixel $\bu_j$. In practice, instead of directly using the candidate depth, we fit a plane to the candidate superpixels and use the intersection of this plane with the visual rays. This provides some robustness to noise in the candidates.

As a second unary potential for the continuous variables, we also make use of the candidate depths, but in a less direct manner. More specifically, we train 4 different Gaussian Process (GP) regressors, each corresponding to one dimension of the variable $\by_i$. The input to each regressor is composed of the corresponding measurement of the candidates for superpixel $i$. We found these inputs to be more reliable than image features. For each GP, we used an RBF kernel with width set to the median squared distance computed over all the training samples. For more details on GP regression, we refer the reader to~\cite{Rasmussen06}. Given the regressed value $\by^r_i$ for superpixel $i$, we compute the depth $d_{r,i}^{\bu_j}$ at each pixel $\bu_j$ in the same manner as before, and write the potential
\begin{equation}
\phi^p_i(\by_i) = \frac{w_p}{N^p_i} \sum_{j=1}^{N^p_i} (d_i^{\bu_j}(\by_i) - d_{r,i}^{\bu_j})^2\;,
\label{eq:pot_depth_gp}
\end{equation}
where $w_p$ is the weight of this potential relative to $\phi^c_i(\by_i)$. In practice, we also use the regressed value $\by^r_i$ as a state for superpixel $i$ in the first round of PCBP where no sampling is performed.

\vspace{0.2cm}
\noindent {\bf Potential for mixed variables:}\mbox{}\\
Our model also exploits a potential that involves both continuous and discrete variables. In particular, we define a potential that encodes the compatibility of two superpixels that share a common boundary and the corresponding discrete variable. This potential can be expressed as
\begin{align}
	&\phi^m_{i,j}(\by_i,\by_j,\be_{i,j}) = w_m \times \label{eq:pot_comp}\\ &\left\{\begin{array}{lc} 
		g_{i,j}\|\bn_i - \bn_j\|^2 & \\ + \frac{1}{N^b_{i,j}}\sum\limits_{m=1}^{N^b_{i,j}} (d_i^{\bu_m}(\by_i)- d_j^{\bu_m}(\by_j))^2   &{\rm if}\; \be_{i,j} = so \\
		\frac{1}{N^b_{i,j}} \sum\nolimits_{m=1}^{N^b_{i,j}} (d_i^{\bu_m}(\by_i)- d_j^{\bu_m}(\by_j))^2 & {\rm if}\; \be_{i,j} = co \\ \phi^o_{i,j}(\by_i,\by_j,\be_{i,j}) & {\rm otherwise,} \end{array}\right. \nonumber
\end{align}
where $w_m$ is the weight of this potential, $\bn_i$ is the plane normal of superpixel $i$, i.e., 3 components of $\by_i$, $N^b_{i,j}$ is the number of pixels shared along the boundary between superpixel $i$ and superpixel $j$, and $g_{i,j}$ is a weight based on the image gradient at the boundary between superpixel $i$ and $j$, i.e., $g_{i,j} = \exp(-\mu_{i,j}/\sigma)$, with $\mu_{i,j}$ the mean gradient along the boundary between the two superpixels. To handle the occlusion cases, the function $\phi^o_{i,j}(\by_i,\by_j,\be_{i,j})$ assigns a cost 0 if the two superpixels are in a configuration that agrees with the state of $\be_{i,j}$, i.e., left occlusion or right occlusion, and a cost $\theta_{max}$ otherwise. While this potential depends on three variables, it remains fast to compute, since $\be_{i,j}$ can only take four states.

\vspace{0.2cm}
\noindent{\bf Potentials for discrete variables:}\mbox{}\\
Finally, we use two different potentials that only involve discrete variables. The first one is a unary potential that makes use of a classifier trained to discriminate between occlusion (i.e., $lo\cup ro$) and non-occlusion (i.e., $so\cup co$) cases. To this end, we utilize the image-based occlusion cues introduced in~\cite{Hoiem07} and employ a binary boosted decision tree classifier. Given the prediction of the classifier $\bhe_p$, our potential function takes the form
\begin{equation}
\phi^u_{p}(\be_p) = \left\{ \begin{array}{c} -\theta_e \;\; {\rm if}\; \be_p \; {\rm agrees\; with}\; \bhe_p \\ \hspace{-1cm}\theta_e \;\; {\rm otherwise}\;, \end{array}\right.
\label{eq:pot_edge_class}
\end{equation}
where $\theta_e$ is a parameter of our model.
Note that distinguishing between all four types of edge variables proved too unreliable, which motivated our decision to only consider occlusion vs. non-occlusion.

The second purely discrete potential is similar to the junction feasibility potential used in~\cite{Yamaguchi12} for stereo. More specifically, it encodes information about whether the junction between three edge variables is physically possible, or not. Therefore, this potential takes the form
\begin{equation}
\phi^h_{p,q,r}(\be_p,\be_q,\be_r) = \left\{ \begin{array}{c} \theta_{max} \;\; {\rm if\; impossible\; case} \\ \hspace{-1.5cm} 0 \;\; {\rm otherwise}\;. \end{array}\right.
\label{eq:pot_edge_valid}
\end{equation}
Here, we employed the same impossible cases as in~\cite{Yamaguchi12} for our 4 states, assuming that superpixels from connected but different objects typically form a hinge, while superpixels from the same object are mostly coplanar. Note that, here, we only considered junctions of three superpixels, since junctions of four occur very rarely. However, 4-junctions could easily be introduced in our framework.

As shown by our experiments, our discrete-continuous approach to single image depth estimation typically helps better modeling the relationships between neighboring superpixels, thus improving depth maps accuracy. However, in the presence of videos, it does not enforce any temporal coherence, and thus may yield unnatural and jerky motion. In the following two sections, we therefore introduce an approach to encoding short-term and long-term temporal coherence in our framework.

\section{A Two-Frame CRF for Depth Estimation}

Ultimately, goal is to estimate the depth map of a generic video sequence including, e.g., dynamic objects and non-translational camera motion. As a first step towards this goal, we extend our single-image framework to modeling short-range motion coherence. In particular, we consider the case of two consecutive frames, $I_0$ and $I_1$, with a calibrated camera. In this setting, we perform depth estimation by modeling the depth of superpixels in the first image $I_0$, the geometric relationships of such superpixels, and their motion from $I_0$ to $I_1$.

More specifically, given an over-segmentation of $I_0$ into superpixels, each superpixel is modeled as a plane in 3D. As before, we make use of one plane variable ${\bf y}$ to encode the depth of a superpixel in $I_0$, and one edge type variable {\bf e} to encode the semantic and geometric relationship between two neighboring superpixels. Here, however, to better account for dynamic scenes, we consider five types of relationships encoding the fact that two neighboring superpixels {\bf (i)} belong to the same foreground object (${\bf e} = sfo$); {\bf (ii)} belong to the same background object (${\bf e} = sbo$); {\bf (iii)} belong to two connected but different objects (${\bf e} = co$); {\bf (iv)-(v)} belong to two objects that form a left/right occlusion (${\bf e} = \{lo,ro\}$). To model depth in the second frame $I_1$, we assume that each plane undergoes rigid motion, parametrized by a rotation matrix ${\bf r}$ and a translation vector ${\bf t}$. 

Let $(\bY, \bR, \bT) = \{(\by_i,\br_i,\bt_i)|i\in \{1,\cdots,S\}\}$ be the set of variables encoding the depth and motion information of the $S$ superpixels in $I_0$, and let ${\bf E} = \{{\bf e}_{p}\}_{p\in\mathcal{E}}$ be the set of edge type variables for these superpixels, where $\mathcal{E}$ contains all pairs of superpixels that share a common boundary. We formulate depth estimation as inference in a CRF built over $\bY$, $\bR$, $\bT$, and ${\bf E}$. In particular, we express the joint distribution of these variables in our two-frame CRF as
\begin{align}
&p(\bY,\bR, \bT, \bE)
=\frac{1}{Z}\prod_i\Psi^y_i(\by_i)\Psi^r_i(\br_i)\Psi^t_i(\bt_i)\Psi^{yrt}_i(\by_i,\br_i,\bt_i)\nonumber\\
&\cdot\prod_{\alpha}\Psi^{ye}_{\alpha}(\by_{\alpha},\be_{\alpha})\Psi^{re}_{\alpha}(\br_{\alpha},\be_{\alpha})\Psi^{te}_{\alpha}(\bt_{\alpha},\be_{\alpha})\cdot\prod_{\beta}\Psi^e_{\beta}(\be_{\beta}),\nonumber
\end{align}
where $Z$ is a normalization constant, i.e., the partition function, $\{\Psi^{\{y,r,t,yrt\}}_i\}$ are potential functions that define the data term for the depth and motion variables, $\{\Psi^{\{y,r,t\}e}_{\alpha}\}$ are potential functions that encode smoothness and consistency of depth and motion for neighboring superpixels, and $\{\Psi^{e}_{\beta}\}$ are potential functions that define the data term for the edge type variables and also enforce the edge types to take valid configurations. The specific form of our potential functions will be discussed in details in Section~\ref{sec:potentials}.

Depth is then reconstructed by computing a MAP estimate of this joint distribution. By working with negative log potential functions, e.g., $\phi^y_i(\by_i) = -\ln\left(\Psi^y_i(\by_i)\right)$, this can be expressed as the optimization problem
\begin{align}\label{eq:map}
\min_{\bY,\bR,\bT,\bE} &\sum_i \phi^y_i(\by_i) +\phi^r_i(\br_i) + \phi^t_i(\bt_i) + \phi^{yrt}_i(\by_i,\br_i,\bt_i)\nonumber\\
&+ \sum_{\alpha}\phi^{ye}_{\alpha}(\by_{\alpha},\be_{\alpha})+\phi^{re}_{\alpha}(\br_{\alpha},\be_{\alpha})+\phi^{te}_{\alpha}(\bt_{\alpha},\be_{\alpha})\nonumber\\
&+\sum_{\beta}\phi^{e}_{\beta}(\be_{\beta}).
\end{align}

Note that, while $\bE$ are discrete variables, $(\bY, \bR,\bT)$ are continuous ones. Since we now have more continuous variables than in our single image model discussed in Section~\ref{sec:method}, we propose to avoid the relatively expensive process of iteratively sampling particles for each variable by making use of a non-parametric method to discretize these continuous variables. Details are provided in below. In practice, we make use of the distributed convex belief propagation (DCBP) algorithm of~\cite{Schwing11} to perform inference in our CRF and thus minimize this energy.

  More specifically for the discretization of the continuous variables, we retrieve the $K$ image pairs (consecutive frames) most similar to the two input frames from a set of video sequences for which the depths are known. To this end, we perform a nearest-neighbor search based on concatenated GIST, PHOG, Object Bank and optical flow features. We then directly use the depth of the retrieved images (i.e., estimated plane normal and centroid depth for each superpixel) as states for $\bY$. To obtain states for $\bR$ and $\bT$, we compute the relative rotation matrix and translation vector between each superpixel in the first frame of the retrieved pairs and its corresponding superpixel in the second retrieved frame. The correspondence between two superpixels is determined by optical flow. Furthermore, to be able to model the static portions of the scene, we add a~\emph{no-motion} state, i.e., ${\bf r}_i = \bI$ and ${\bf t}_i = {\bf 0}$, for the motion variables, where $\bI$ denotes the identity matrix. To have a broad variety of candidates, we only retrieve at most one pair of frames from each video sequence in the database.

While this discretization procedure gives good states for the scene background, it may not be precise enough to model the depth of moving foreground objects. Therefore, we make use of the following motion-aware discretization procedure: We train a motion classifier to distinguish between superpixels belonging to moving foreground and to background (i.e., moving and static background). To this end, we utilize image-based features consisting of local superpixel features and a global image feature. The local features include RGB color mean and standard deviation, superpixel centroid location and optical flow mean magnitude. The global feature is taken as the histogram of optical flow magnitude computed over the image. We then employ a binary boosted decision tree classifier. Given the classification result, we assign each superpixel a binary variable $m_i$ to denote its motion status. The states of the variables for the superpixels assigned to the background class (i.e., $m_i = 0$) are then set according to the procedure described above, and those for the moving foreground superpixels (i.e., $m_i = 1$) are sampled uniformly within the range of possible depth values.


\subsection{Potentials}\label{sec:potentials}
The objective function in~\eqref{eq:map} involves data terms for depth, motion and  edge type variables and regularization terms for mixed variables. Below, we discuss the functions used in those different types of potentials.

\vspace{0.2cm}
\subsubsection{ Data Terms for Depth and Motion Variables}
\label{subsec:dataterm}
The potentials involving the depth and motion variables can be grouped into the energy
\begin{align}
E_D = \sum_i \phi^y_i(\by_i)+\phi^r_i(\br_i) +\phi^t_i(\bt_i)+\phi^{yrt}_i(\by_i,\br_i,\bt_i), \nonumber
\end{align}
which consists of three different types of potentials: a unary potential for the depth variables, unary potentials for the motion variables and a high-order potential for mixed depth and motion variables. 

\vspace{0.2cm}
\noindent{\bf Unary potentials for the depth variables:}\mbox{}\\
Similarly to the single image case, we define the unary potentials for the depth variables, $\phi^y_i(\by_i)$, based on the retrieved $K$ depth candidates described above. In particular, we express this potential as
\begin{equation}
\phi^y_i(\by_i) = (\phi^c_i(\by_i) + \phi^p_i(\by_i))\cdot(1-m_i), 
\end{equation}
where the two terms $\phi^c_i(\by_i)$ and $\phi^p_i(\by_i)$ are defined in the same manner as the candidate-based and regression-based potentials of Eqs.~\ref{eq:pot_depth_knn} and~\ref{eq:pot_depth_gp}, but where the use of the binary variable $m_i$ indicates that this potential will not apply to the superpixels classified as moving foreground. In practice, we also use the regressed value $\by^r_i$ (see Section~\ref{sec:potentials_si}, above Eq.~\ref{eq:pot_depth_gp}) as a state for superpixel $i$ if it belongs to the background area (i.e., if $m_i = 0$).

\vspace{0.2cm}
\noindent{\bf Unary potentials for the motion variables:}\mbox{}\\
We define the unary potentials for the motion variables, $\phi^r_i({\bf r}_i)$ and $\phi^t_i({\bf t}_i)$, using our motion prior $m_i$. These potentials encode the fact that the foreground is moving and the background remains static. In particular, we write them as
\begin{eqnarray}
\phi^r_{i}({\bf r}_i) &=& \left\{ \begin{array}{c} 0 \;\; \hspace{0.1cm} {\rm if}\; m_i = 0, \; \|   {\bf r}_i - {\bf I}\|^2 \leq \epsilon_r\\
\theta_r \;\; {\rm if}\; m_i = 0,\; \|{\bf r}_i - {\bf I}\|^2 >\epsilon_r \\
0 \;\;\hspace{0.1cm} {\rm if}\; m_i = 1, \; \|   {\bf r}_i - {\bf I}\|^2 \geq \epsilon_r\\
\theta_r \;\; {\rm if}\; m_i = 1, \; \|{\bf r}_i - {\bf I}\|^2 <\epsilon_r \end{array}\right.\\
\phi^t_{i}({\bf t}_i) &=&\left\{ \begin{array}{c} 0 \;\;\hspace{0.1cm} {\rm if}\; m_i = 0, \; \|   {\bf t}_i\|^2 \leq \epsilon_t\\
\theta_t \;\;{\rm if}\; m_i = 0, \; \|{\bf t}_i\|^2 >\epsilon_t \\
0 \;\; \hspace{0.1cm} {\rm if}\; m_i = 1, \; \|{\bf t}_i\|^2 \geq \epsilon_t\\
\theta_t \;\;{\rm if}\; m_i = 1,\; \|{\bf t}_i\|^2 <\epsilon_t \end{array}\right.
\end{eqnarray}
where $\theta_r$, $\theta_t$, $\epsilon_r$, and $\epsilon_t$ are parameters of our model.

\vspace{0.2cm}
\noindent {\bf High-order potentials for the depth and motion variables:}\\
The high-order potential $\phi^{yrt}_i$ involves the depth variable ${\bf y}_i$ and the motion variables ${\bf r}_i$ and ${\bf t}_i$. It consists of two kinds of potentials,
\begin{align}
\phi^{yrt}_i({\bf y}_i,{\bf r}_i,{\bf t}_i) = \phi_i^d({\bf y}_i,{\bf r}_i,{\bf t}_i)\cdot(1- m_i) + \phi_i^a({\bf y}_i,{\bf r}_i,{\bf t}_i), \nonumber
\end{align}
the first of which is only effective when superpixel $i$ belongs to the background.

The first potential, $\phi_i^d$, models the assumption that the depth of the superpixels of $I_0$ after a rigid transformation should be close to at least one candidate of its corresponding patch in $I_1$. More specifically, given the motion parameters $\br_i$ and $\bt_i$ we can compute a homography $H(\by_i,\br_i,\bt_i)$ that maps the pixels $\{\bu_j^0\}$ in superpixel $i$ to a set of pixels $\{\bu_j^1\}$ in $I_1$. The depth $d^{\bu_j^1}(\by_i,\br_i,\bt_i)$ is then obtained by intersecting the visual ray passing through $\bu_j^1$ with the plane $\by_i$ after applying the rigid transformation $\br_i$ and $\bt_i$. The potential can then be written as
\begin{equation}
\phi_i^d(\by_i,\br_i,\bt_i) =w_d\cdot \min_{k=1}^K \frac{1}{N^p_i} \sum_{j=1}^{N^p_i} (d^{\bu_j^1}(\by_i,\br_i,\bt_i) - d_{k}^{\bu_j^1})^2\;,
\end{equation}
where $w_d$ is a model parameter, $N^p_i$ is the number of pixels in superpixel $i$, and $d_{k}^{\bu_j^1}$ denotes the $k^{th}$ candidate for pixel $\bu_j^1$ in frame $I_1$. 

The second potential, $\phi_i^a$, models the assumption that the corresponding points across frames have similar appearance. We can thus define the optical flow-induced appearance constraints across frames as
\begin{equation}
\phi_i^a(\by_i,\br_i,\bt_i) = \frac{w_a}{N^p_i} \sum_{j=1}^{N^p_i} \left(I_0^{\bu_j^0}-I_1^{\bu_j^1}\right)^2\;,
\end{equation}
where $w_a$ is a model parameter.

\vspace{0.2cm}
\subsubsection{ Regularization Terms for Mixed Variables}
The potentials involving the mixed variables are regularization terms, and can be grouped into the energy
\begin{equation}
E_{\alpha}= \phi^{ye}_{\alpha}(\by_{\alpha},\be_{\alpha})+\phi^{re}_{\alpha}(\br_{\alpha},\be_{\alpha})+\phi^{te}_{\alpha}(\bt_{\alpha},\be_{\alpha}). \nonumber
\end{equation}
These potentials encourage depth to be smooth while respecting discontinuities, as well as the 3D motion field defined by ${\bf r}$ and ${\bf t}$ to be smooth. 

The first potential, $\phi^{ye}_{\alpha}(\by_{\alpha},\be_{\alpha})$, encodes the compatibility of two neighboring superpixels with the corresponding discrete variable. In other words, this potential is similar to the one given in Eq.~\ref{eq:pot_comp} for our single image model, but adapted to the slightly different states for the edge-type variables in our two-frame CRF. It is therefore expressed as
\begin{align*}
&\phi^{ye}_{i,j}(\by_i,\by_j,\be_{i,j}) = w_{ye} \times \\ &\left\{\begin{array}{lc} 
\|\bn_i - \bn_j\|^2 & \\ + \frac{1}{N^b_{i,j}}\sum\limits_{m=1}^{N^b_{i,j}} (d_i^{\bu_m}(\by_i)- d_j^{\bu_m}(\by_j))^2   &{\rm if}\; \be_{i,j} = sof\\
g_{i,j}\|\bn_i - \bn_j\|^2 & \\ + \frac{1}{N^b_{i,j}}\sum\limits_{m=1}^{N^b_{i,j}} (d_i^{\bu_m}(\by_i)- d_j^{\bu_m}(\by_j))^2   &{\rm if}\; \be_{i,j} = sob\\
\frac{1}{N^b_{i,j}} \sum\nolimits_{m=1}^{N^b_{i,j}} (d_i^{\bu_m}(\by_i)- d_j^{\bu_m}(\by_j))^2 & {\rm if}\; \be_{i,j} = co \\ \phi^o_{i,j}(\by_i,\by_j,\be_{i,j}) & {\rm otherwise,} \end{array}\right. \nonumber
\end{align*}
where $w_{ye}$ is the weight of this potential, $\bn_i$ is the plane normal of superpixel $i$, i.e., 3 components of $\by_i$, $N^b_{i,j}$ is the number of pixels shared along the boundary between superpixel $i$ and superpixel $j$, and $g_{i,j}$ is a weight based on the image gradient at the boundary between superpixel $i$ and $j$, i.e., $g_{i,j} = \exp(-\mu_{i,j}/\sigma)$, with $\mu_{i,j}$ the mean gradient along the boundary between the two superpixels. To handle the occlusion cases, the function $\phi^o_{i,j}(\by_i,\by_j,\be_{i,j})$ assigns a cost 0 if the two superpixels are in a configuration that agrees with the state of $\be_{i,j}$, i.e., left occlusion or right occlusion, and a cost $\theta_{max}$ otherwise. As in the single image case, while this potential depends on three variables, it remains fast to compute, since $\be_{i,j}$ can only take five states.

The potentials $\phi^{re}_{\alpha}(\br_{\alpha},\be_{\alpha})$ and $\phi^{te}_{\alpha}(\bt_{\alpha},\be_{\alpha})$ encode the fact that the motion of neighboring superpixels should agree with the motion defined by the discrete variable $\be_{i,j}$. They specifically encourage two neighboring superpixels belonging to the same background object to undergo the same motion, i.e., same rotation and translation. We thus define these potentials as
\begin{align*}
\phi^{re}_{i,j}(\br_i,\br_j,\be_{i,j})&=&w_{re} \times \left\{\begin{array}{lc} 
\|\br_i-\br_j\|^2&\hspace{-0.1cm} {\rm if}\; \be_{i,j} = sob\\ \theta_{m}  &\hspace{-0.1cm}{\rm otherwise,} \end{array}\right. \nonumber\\
\phi^{te}_{i,j}(\bt_i,\bt_j,\be_{i,j})&=&w_{te} \times \left\{\begin{array}{lc} 
\|\bt_i-\bt_j\|^2&\hspace{-0.1cm}{\rm if}\; \be_{i,j} = sob\\ \theta_{m}  & \hspace{-0.1cm}{\rm otherwise,} \end{array}\right. \nonumber
\end{align*}
where $w_{re}$, $w_{te}$, and $\theta_m$ denote model parameters.\\

\subsubsection{Potentials for Edge Type Variables}

\label{subsec:disc_dataterm}
We finally define two potentials for the edge type variables, which lets us write 
\begin{equation}
\phi^e_{\beta} = \phi^u_p(\be_p)+\phi^h_{p,q,l}(\be_p,\be_q,\be_l)\;.
\end{equation} 

The first potential, $\phi^u_p(\be_p)$, is a unary potential that has the same form as the one given in Eq.~\ref{eq:pot_edge_class} for the single image case and makes use of a similar edge-type classifier. Here, however, to account for the different states of our edge-type variables, we consider a three-class classifier, where the classes represent foreground object ($sof$), background non-occlusion ($sob\cup co$) and background occlusion ($lo\cup ro$). Furthermore, we also make use of our motion classifier in the following manner: Given the estimated foreground area, we detect its bottom boundary $\be_{mb}$, i.e., where the foreground region connects to the background. We then enforce $\be_{mb} = co$.  Jointly, our three-class classifier and our foreground-background constraints let us define an estimate $\bhe_p$ for the edge type variable $\be_p$, which we employ in the same manner as in Eq.~\ref{eq:pot_edge_class}.


The second potential is a junction feasibility potentials and has the same form as the one in Eq.~\ref{eq:pot_edge_valid}. Here, to accommodate our slightly different states for the edge-type variables, we defined the impossible cases by assuming that superpixels belonging to the same objects are mostly coplanar, independently of whether they are part of the foreground or of the background, and, as before, that superpixels belonging to different connected objects typically form a hinge. In practice, as for single image depth estimation, we only considered junctions of three superpixels, which are the most common ones in real scenarios.


\section{Depth Estimation for Videos}
\label{sec:full}
The method described in the previous section lets us estimate a depth map in each frame of a video sequence, while explicitly handling moving objects and accounting for short-range temporal smoothness. To achieve long-range smoothness across the video sequence, we propose to make use of a fully-connected CRF with Gaussian edge potentials. To this end, we construct a CRF over variables encoding the depth of each individual pixel in the video sequence. We then define the Gibbs energy of the CRF as
\begin{equation}
E(\bx) = \sum_{i}\phi_u(x_i) + \sum_{i < j}\phi_p(x_i,x_j),
\end{equation}
where $\phi_u$ defines the unary term and $\phi_p$ defines the pairwise regularization term. The variable $x_i$ denotes the depth of pixel $i$, where, for ease of notation, we ignore the dependence on the frame index. 

The unary term $\phi_u(x_i)$ encodes the fact that the final depth of pixel $x_i$ should be close to the depth estimated by our two-frame CRF according to the squared difference. The pairwise term encourages pixels similar in appearance and location to have similar depth. It is defined as
\begin{equation}
\phi_p(x_i,x_j) = \mu(x_i,x_j)\underbrace{\sum_{m=1}^{K}\omega^{(m)}k^{(m)}({\bf f}_i,{\bf f}_j)}_{k({\bf f}_i,{\bf f}_j)},
\end{equation} 
where $k^{(m)}({\bf f}_i,{\bf f}_j)$ is a Gaussian kernel, ${\bf f}_i$ and ${\bf f}_j$ are feature vectors, and $\mu$ is a label compatibility function (in practice, we used a Potts model, i.e., $\mu(x_i,x_j)=[x_i \neq x_j]$). Here we make use of the following two kernels, defined in terms of the color vectors $\bc_i$ and $\bc_j$, the positions $\bp_i$ and $\bp_j$, and the frame indices $q_i$ and $q_j$:
\begin{align}
\hspace{-0.3cm}k({\bf f}_i, {\bf f}_j) &= \omega^{(1)}{\rm{exp}}(-\frac{\|\bp_i-\bp_j\|^2}{2\theta_{\alpha}^2}-\frac{(q_i-q_j)^2}{2\theta_q^2}-\frac{\|\bc_i-\bc_j\|^2}{2\theta_{\beta}^2})\nonumber\\
&+\omega^{(2)}{\rm{exp}}(-\frac{\|\bp_i-\bp_j\|^2}{2\theta_{\gamma}^2} - \frac{(q_i-q_j)^2}{2\theta_{\gamma}^2})\;,
\end{align}
where the first kernel encodes a notion of appearance similarity of two pixels, and the second one purely encourages smoothness. Note that our use of the frame index $q_i$ as an additional feature for pixel $i$ lets us encode temporal smoothness. The state space of $x_i$ is formed by uniformly sampling depth between $0.5$m and $10$m at an interval of $0.05$m. Following~\cite{Krah11}, inference in such a fully-connected CRF can be performed efficiently using a mean-field method with an efficient Gaussian filtering step. More details on the inference procedure are provided in~\cite{Krah11}.

By working at pixel-level and accounting for long-range connections, this fully-connected model provides temporal smoothness to the results of our two-frame CRF, and can additionally reduce the artifacts introduced by our previous superpixel representation. Nonetheless, it benefits from the motion-awareness of our two-frame model, and thus still respects the discontinuities of the depth map.

\section{Experiments}
We now present our experimental results on depth estimation in outdoor and indoor scenes. In particular, we first evaluate our single image depth estimation method on two publicly available datasets: the Make3D range image dataset~\cite{Saxena09}, and the NYUv2 Kinect dataset~\cite{Silberman12}. We then evaluate our method for depth estimation from challenging monocular video sequence method on MSR-V3D~\cite{Karsch12}, which, to the best of our knowledge is the only available dataset depicting the kind of scenarios that we are interested in.

For our quantitative evaluation, we report the following three commonly-used error metrics:
\begin{itemize}
	\item average relative error ({\bf rel}): $\frac{1}{N}\sum_{\bu}\frac{|g_{\bu}-d_{\bu}|}{g_{\bu}}$,
	\item average ${\bf log}_{10}$ error: $\frac{1}{N}\sum_{\bu}|log_{10} g_{\bu} - log_{10} d_{\bu}|$,
	\item root mean squared error ({\bf rms}): $\sqrt[]{\frac{1}{N}\sum_{\bu}(g_{\bu}-d_{\bu})^2}$, 
\end{itemize}
where $g_{\bu}$ is the ground-truth depth at pixel $\bu$, $d_{\bu}$ is the corresponding estimated depth, and $N$ denotes the total number of pixels in all the images.
In the following, we make use of the DepthTransfer method of~\cite{Karsch12,Karsch14} as our main baseline, since it also applies to both the single image and the video scenarios, and is representative of the state-of-the-art on the datasets we employ.

\subsection{Evaluation of Single Image Depth Estimation}

We first evaluate our single image method introduced in Section~\ref{sec:method}. In the following, we refer to this method as Ours-1F, to indicate that it works using one single frame. In addition to the DepthTransfer baseline~\cite{Karsch12}, we report the results of {\bf (i)} our unary terms only;  {\bf (ii)} our GP depth regressors;  {\bf (iii)} our model without discrete variables and with the same pairwise term as the $e_{i,j}=so$ case; and  {\bf (iv)} the approximate MAP obtained by our model before sampling particles using PCBP.

\begin{figure*}[t!] 
	\centering
	\begin{small}
		\begin{tabular}{cccccc}
			\hspace{-0.0cm}\begin{sideways}\hspace{1.2cm}{\bf Image}\end{sideways} &
			\hspace{-0.0cm}\includegraphics[width=0.15\linewidth]{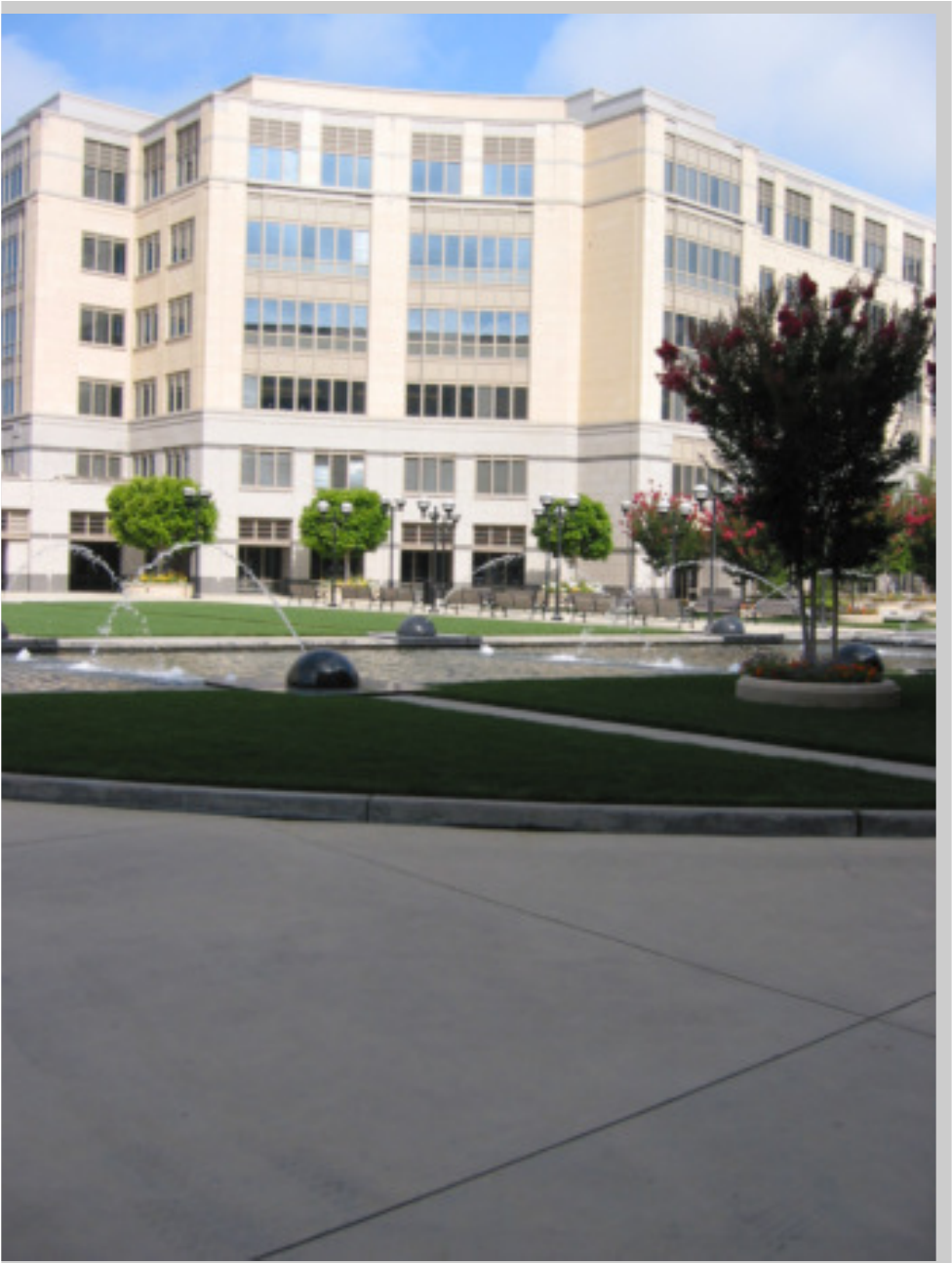} &
			\hspace{-0.0cm}\includegraphics[width=0.15\linewidth]{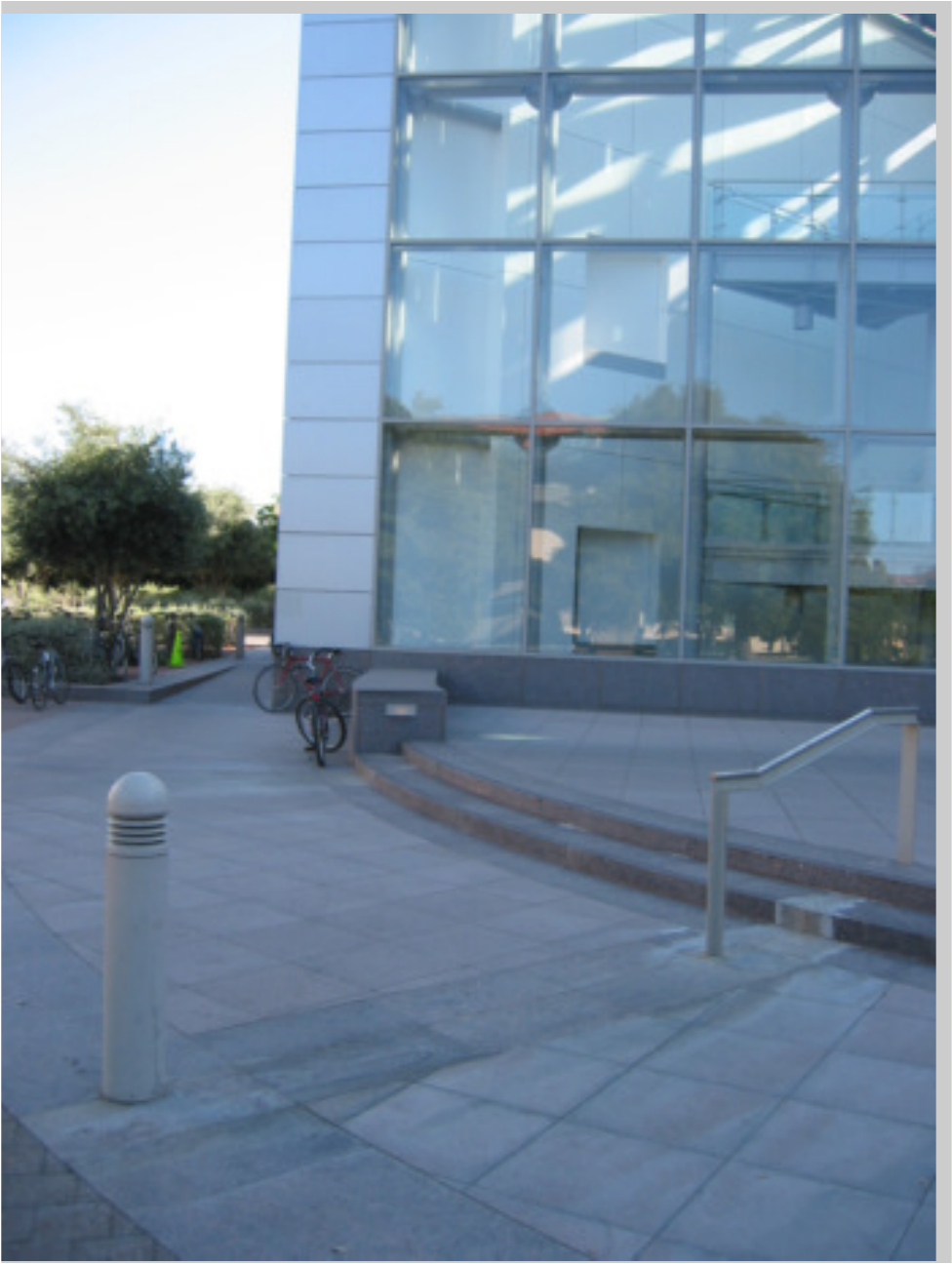} &
			\hspace{-0.0cm}\includegraphics[width=0.15\linewidth]{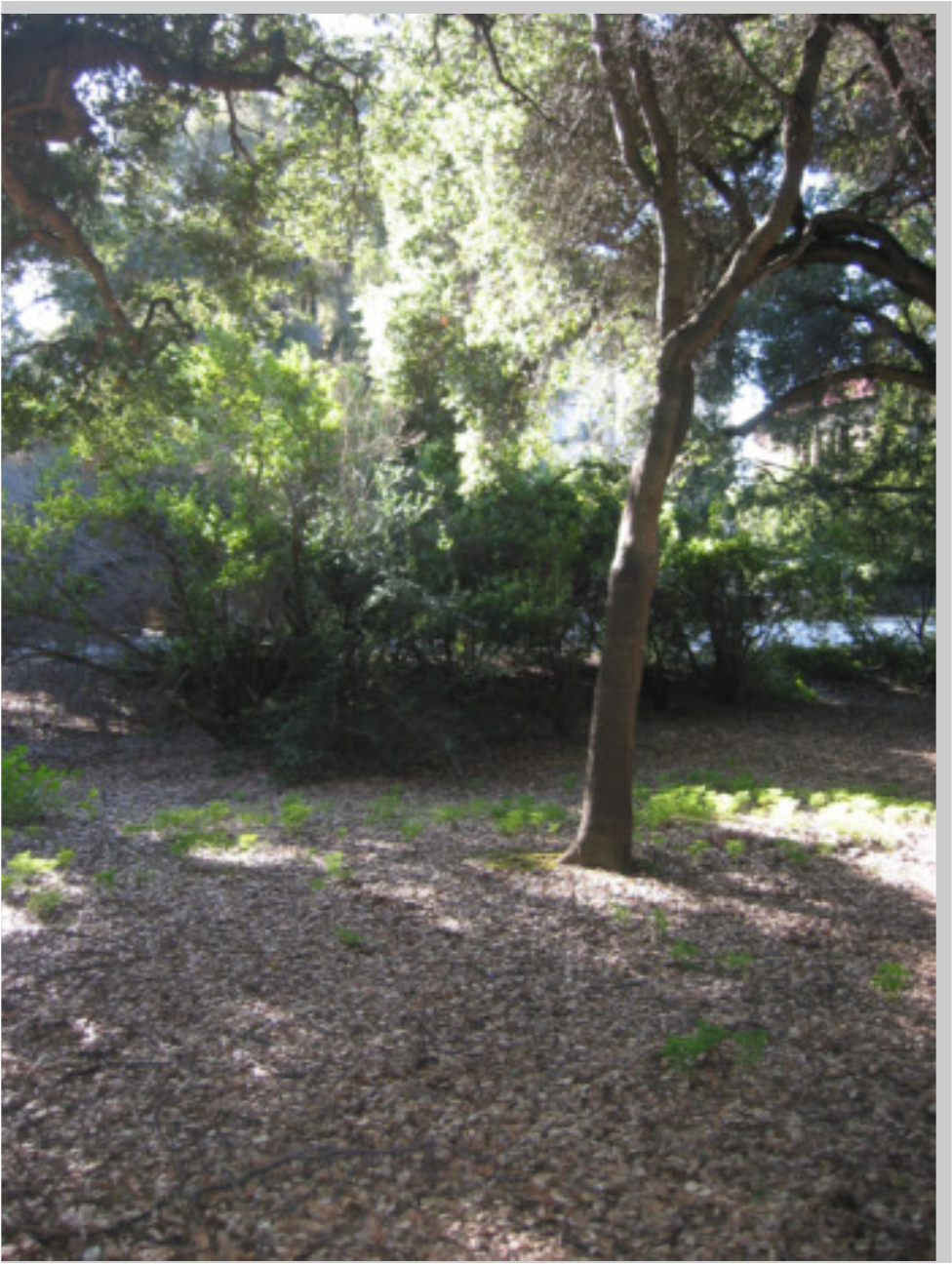} &
			\hspace{-0.0cm}\includegraphics[width=0.15\linewidth]{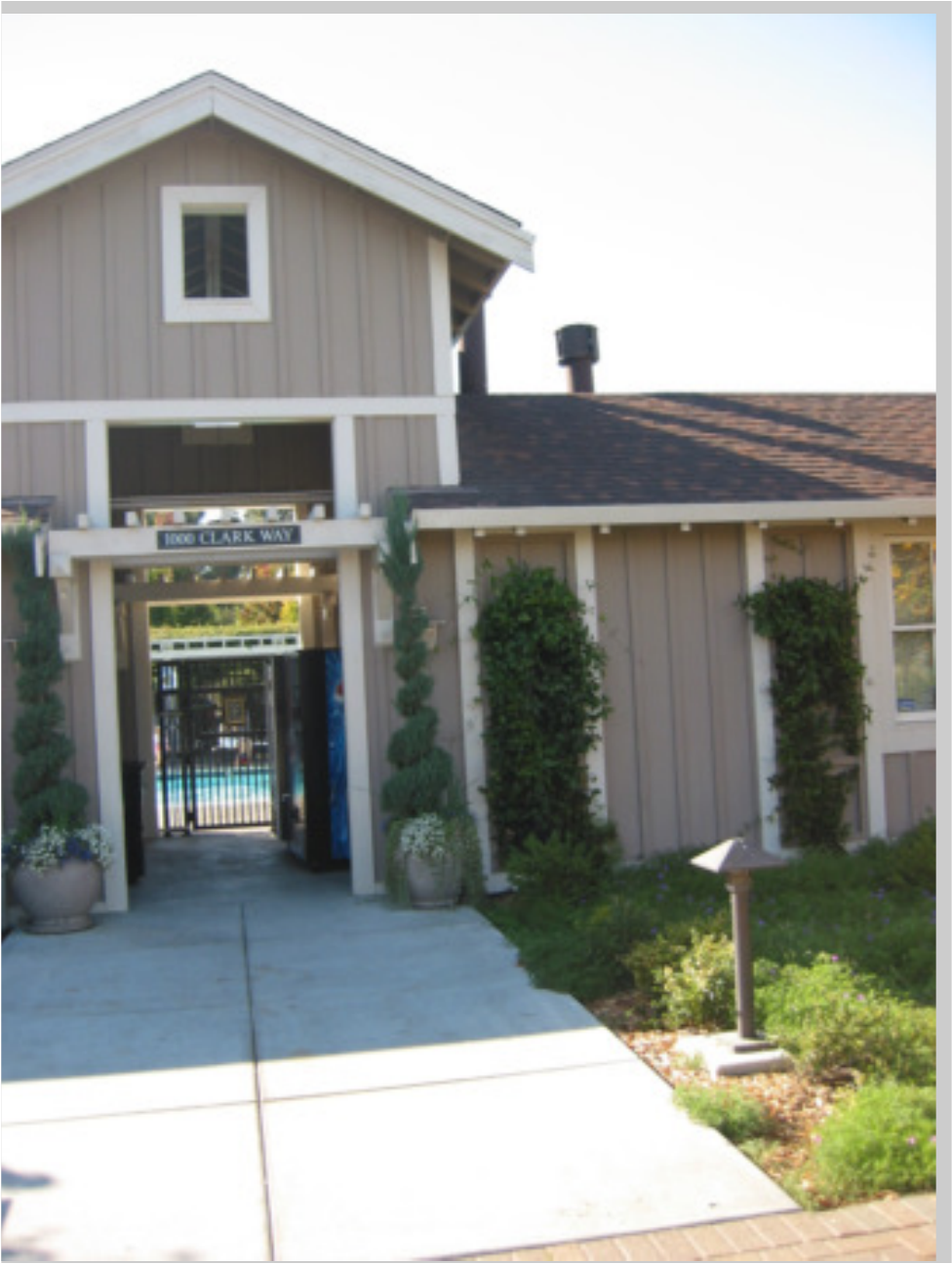} &
			\hspace{-0.0cm}\includegraphics[width=0.15\linewidth]{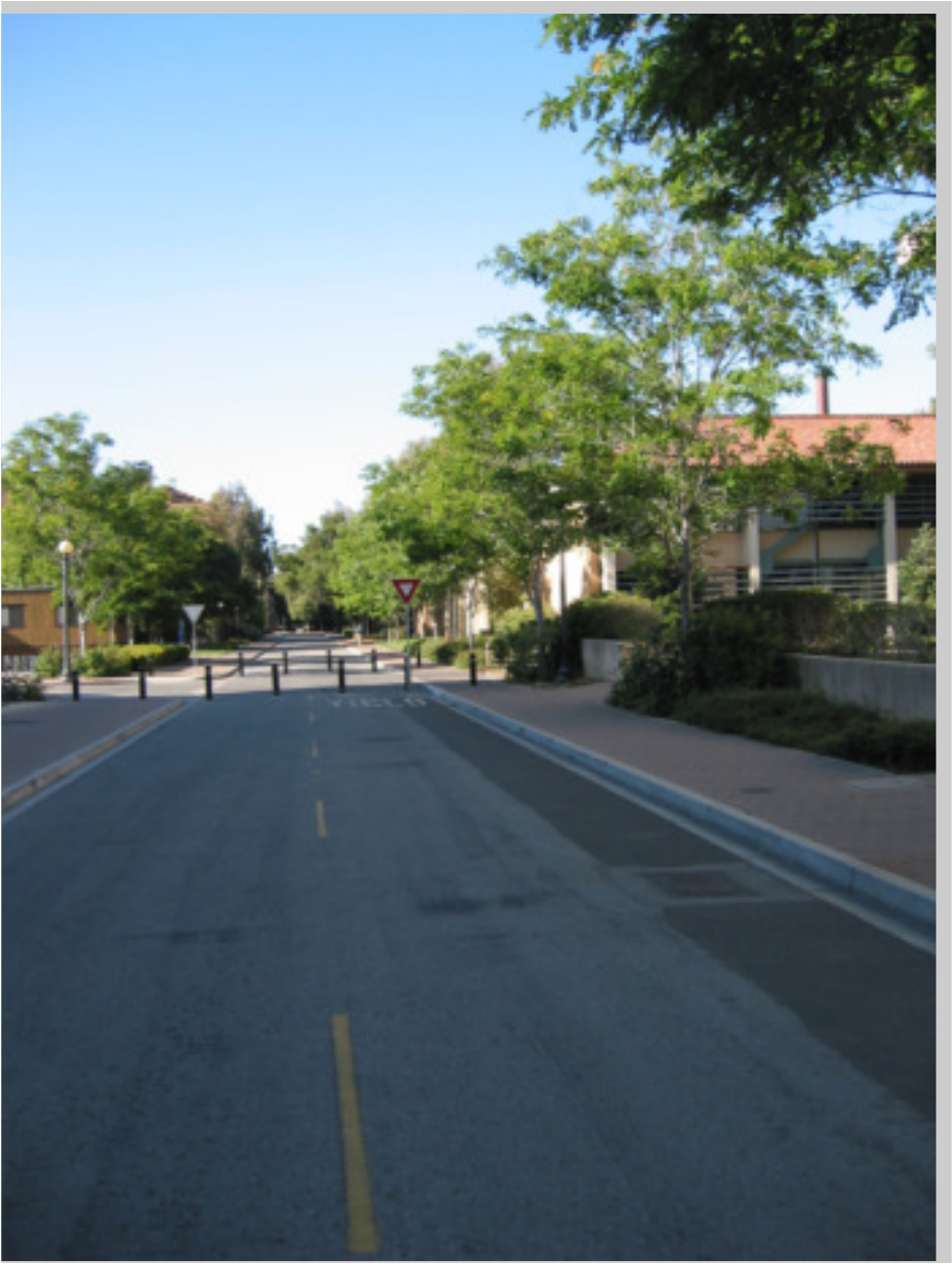} \\
			\hspace{-0.0cm}\begin{sideways}\hspace{0.8cm}{\bf Ground-truth}\end{sideways} &
			\hspace{-0.0cm}\includegraphics[width=0.15\linewidth]{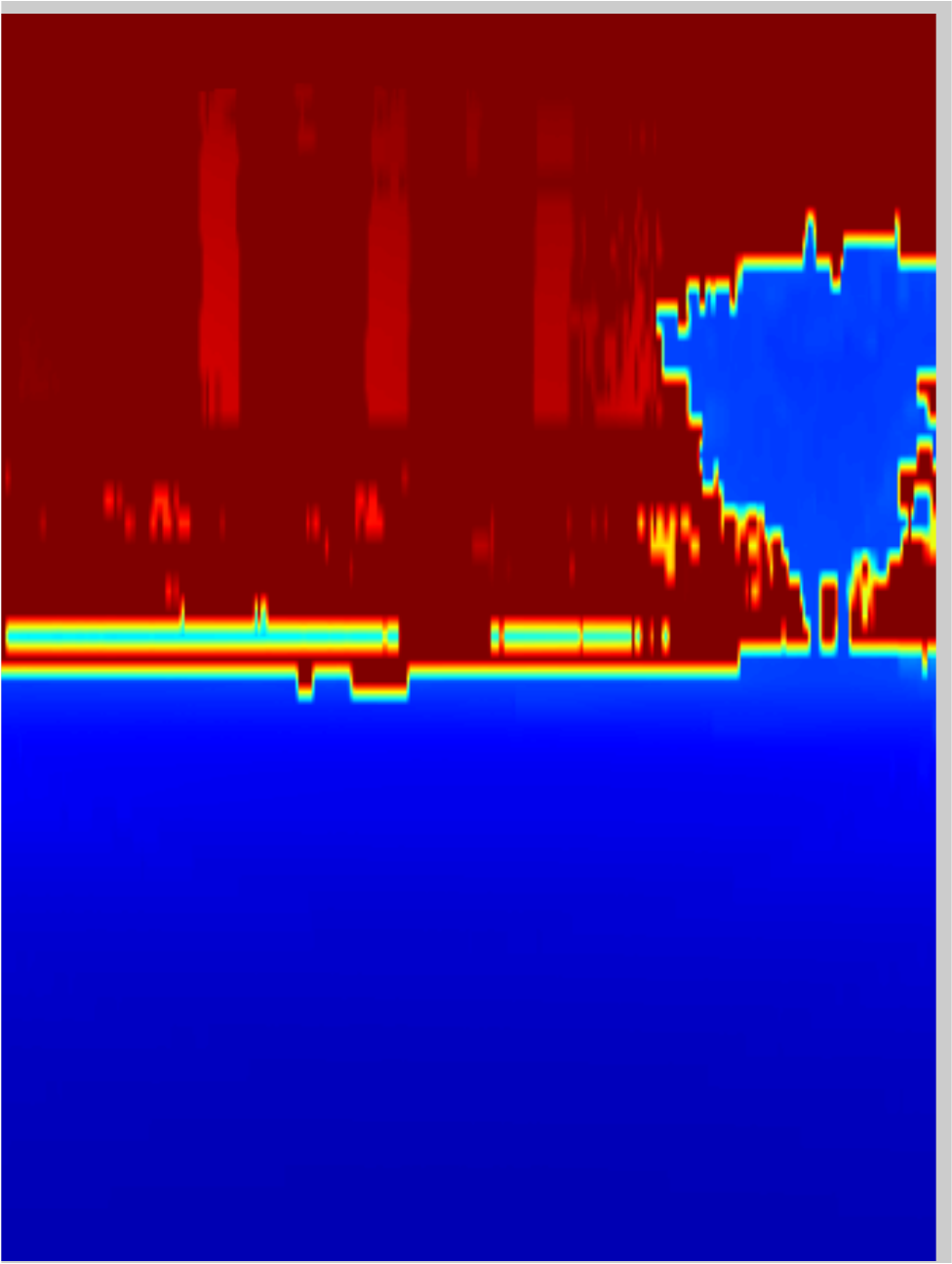} &
			\hspace{-0.0cm}\includegraphics[width=0.15\linewidth]{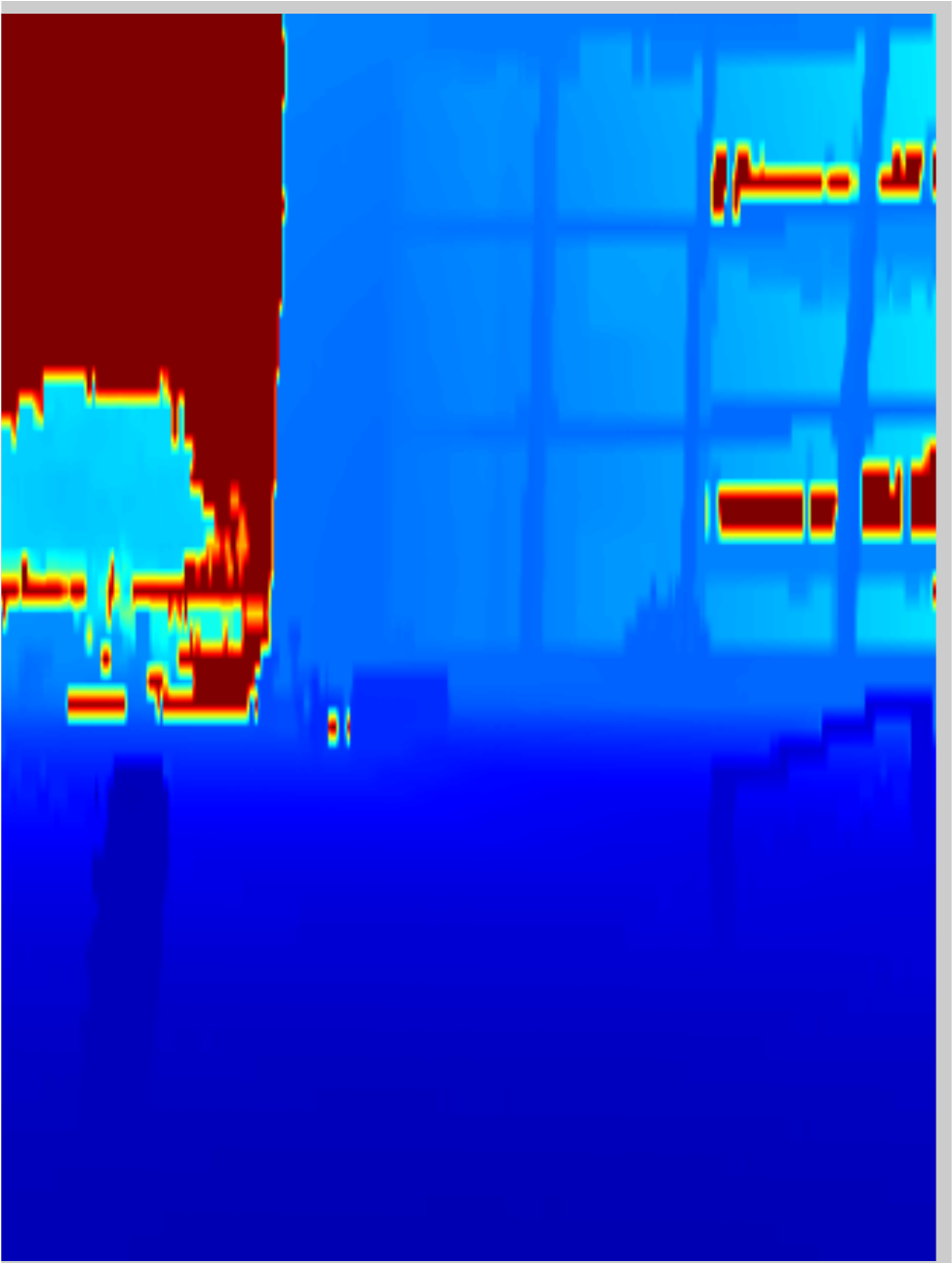} &
			\hspace{-0.0cm}\includegraphics[width=0.15\linewidth]{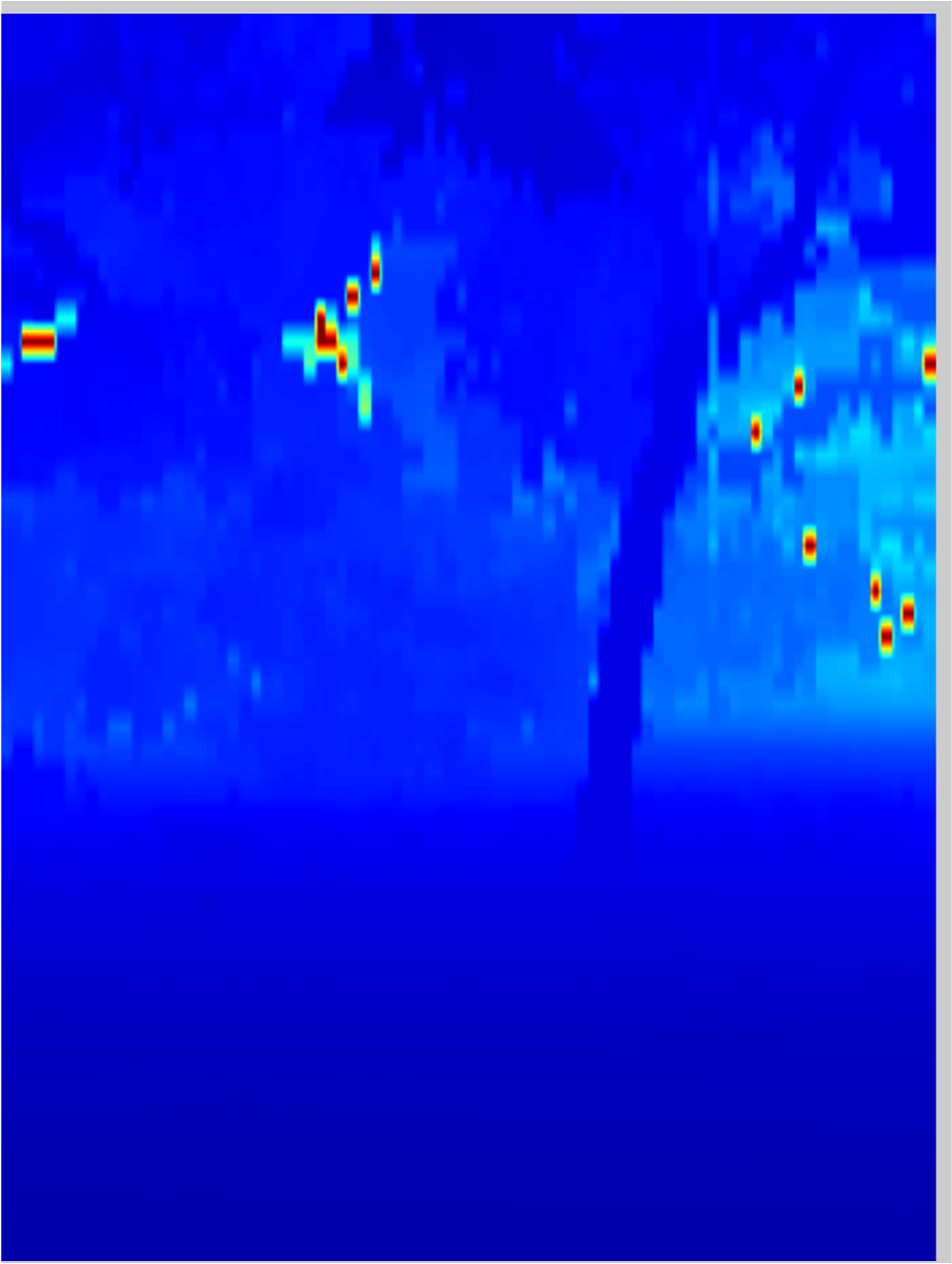} &
			\hspace{-0.0cm}\includegraphics[width=0.15\linewidth]{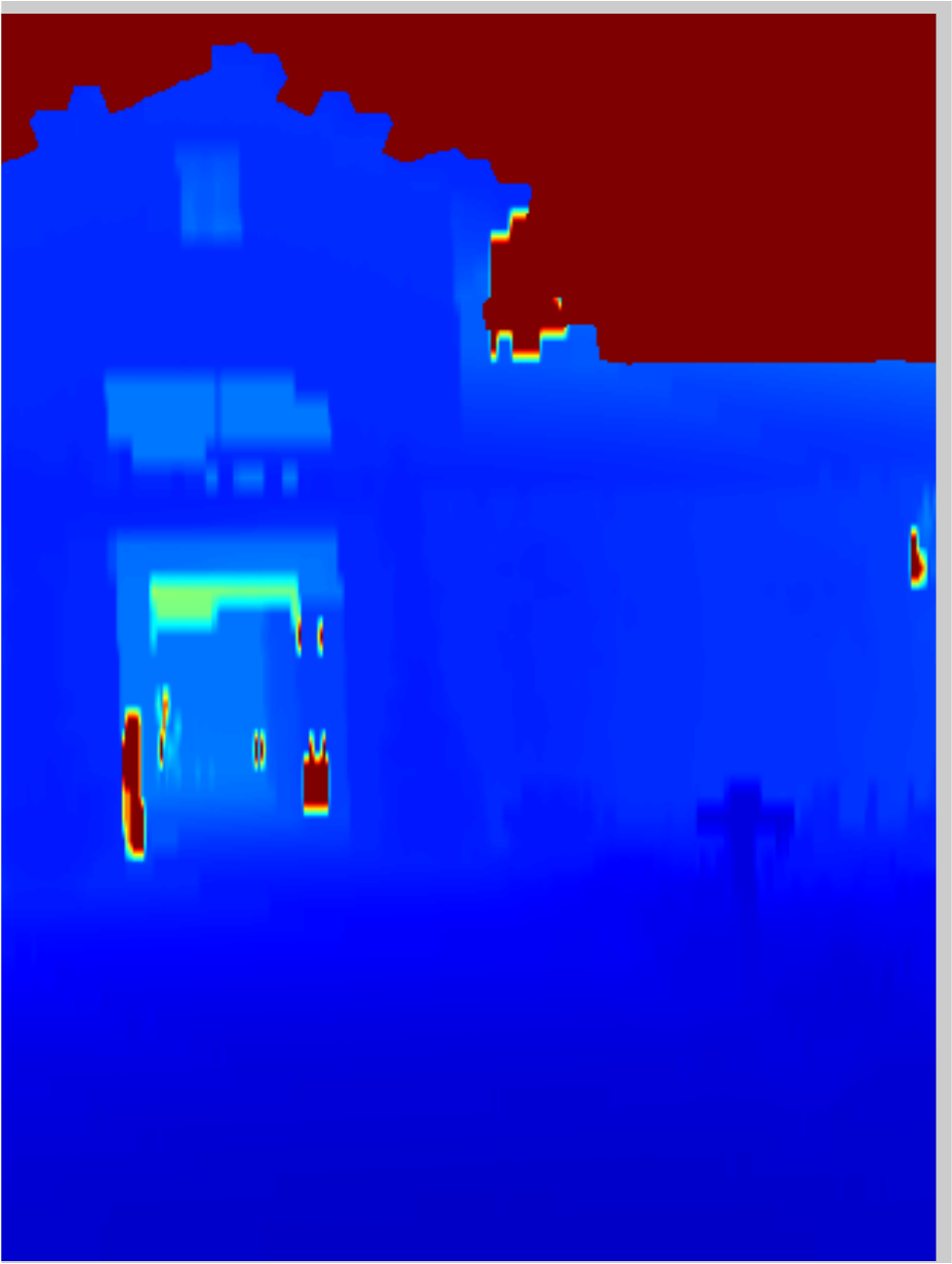} &
			\hspace{-0.0cm}\includegraphics[width=0.15\linewidth]{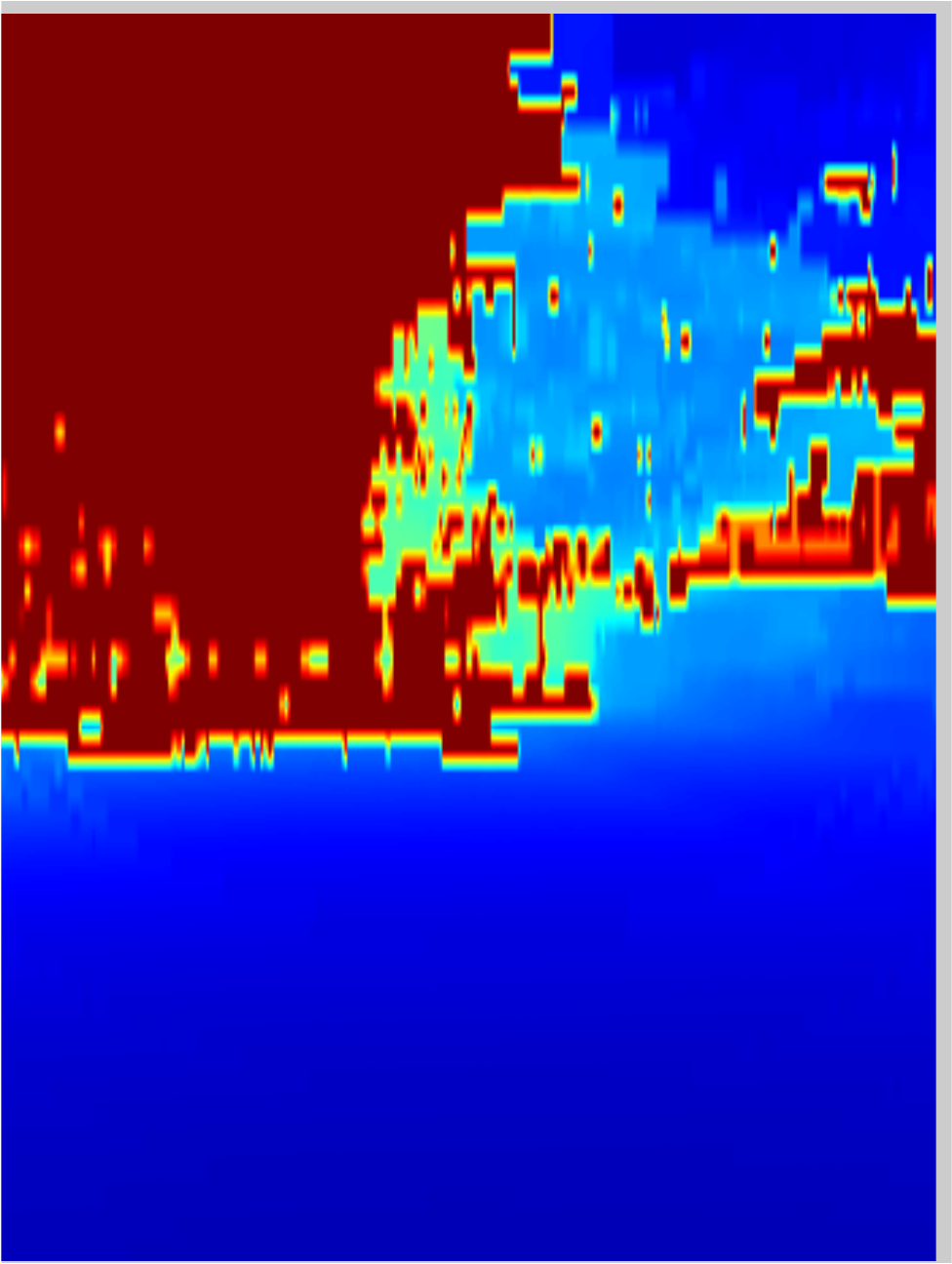} \\
			\hspace{-0.0cm}\begin{sideways}\hspace{0.4cm}{\bf DepthTransfer~\cite{Karsch12}}\end{sideways} &
			\hspace{-0.0cm}\includegraphics[width=0.15\linewidth]{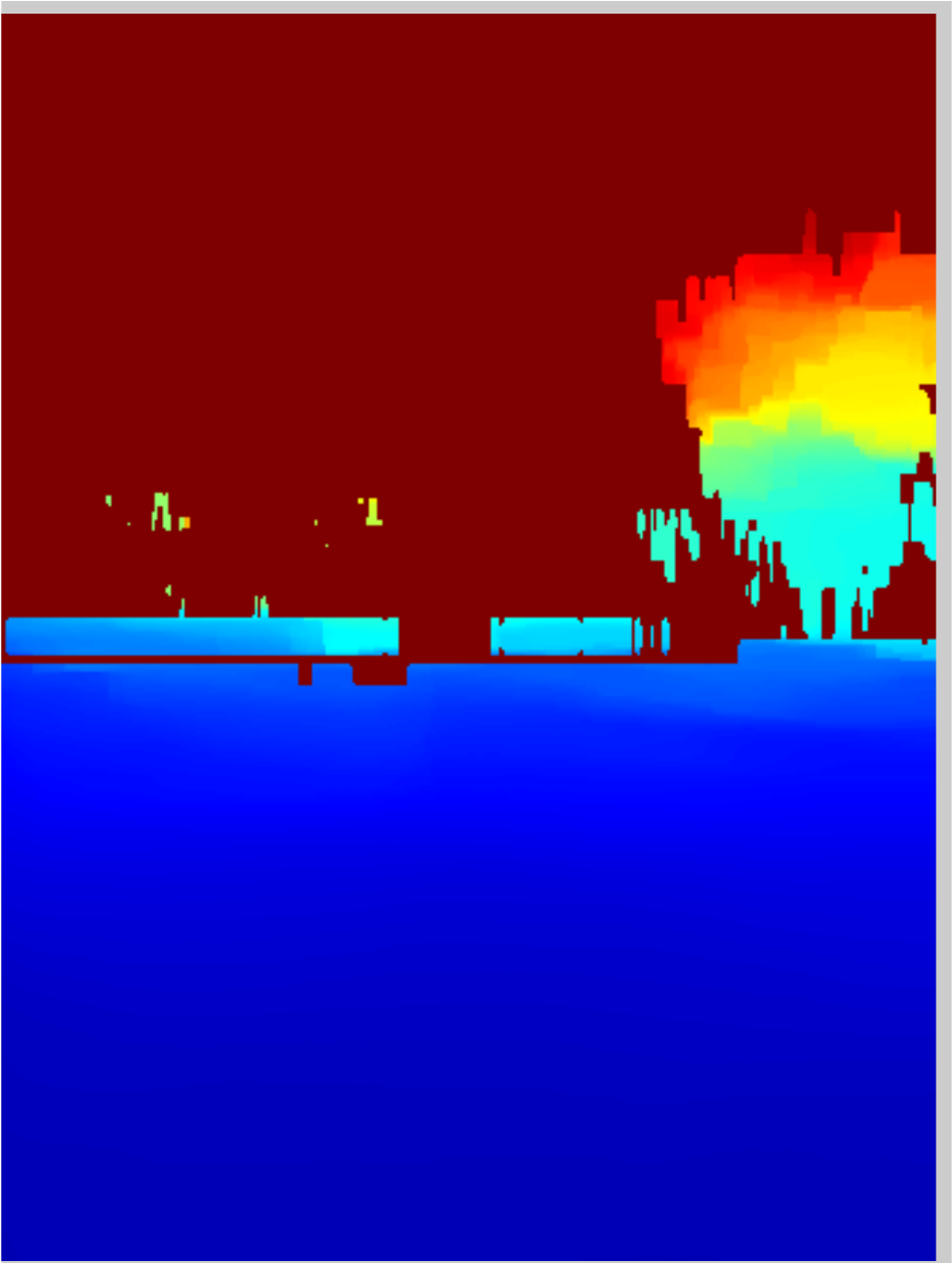} &
			\hspace{-0.0cm}\includegraphics[width=0.15\linewidth]{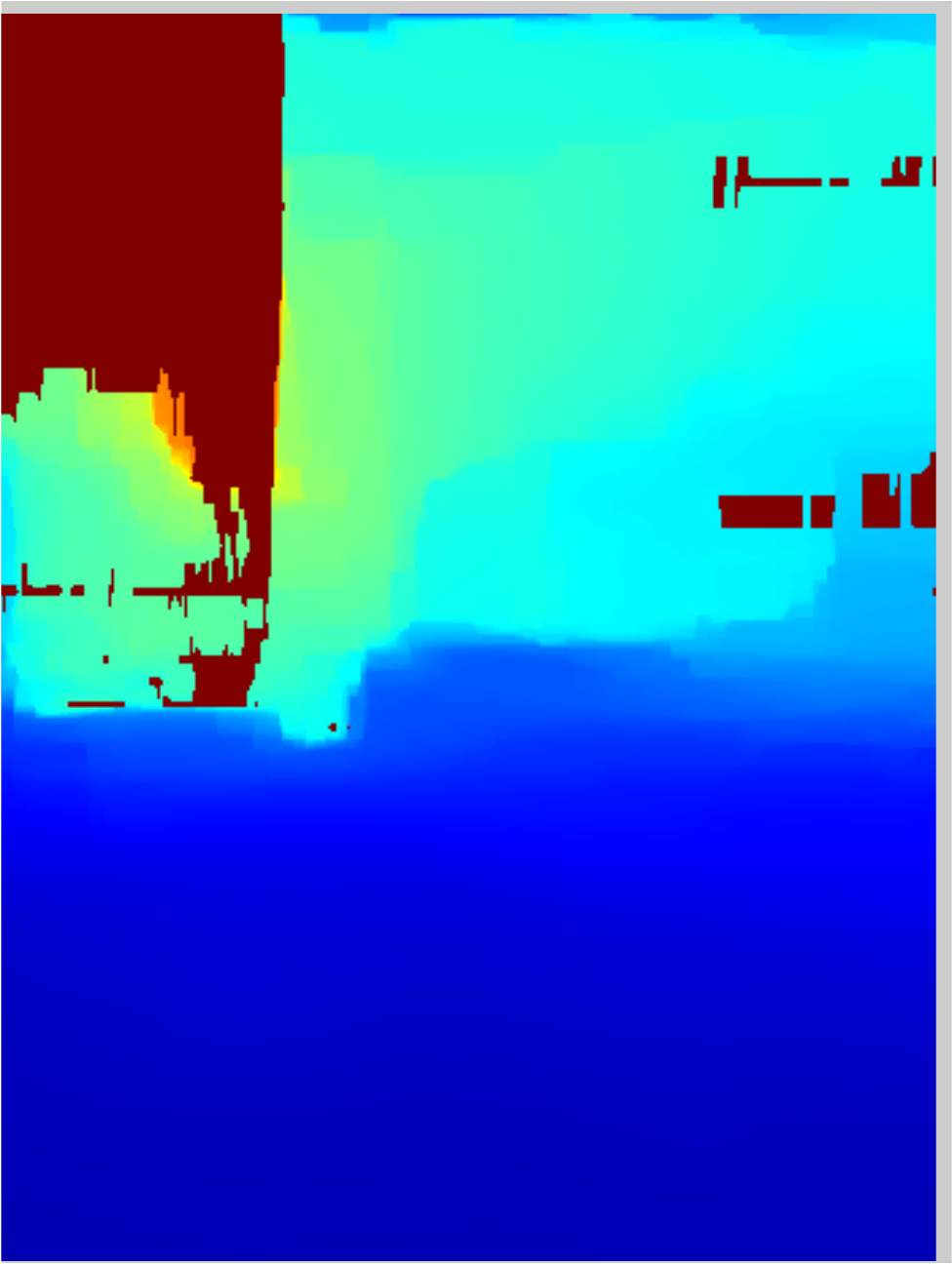} &
			\hspace{-0.0cm}\includegraphics[width=0.15\linewidth]{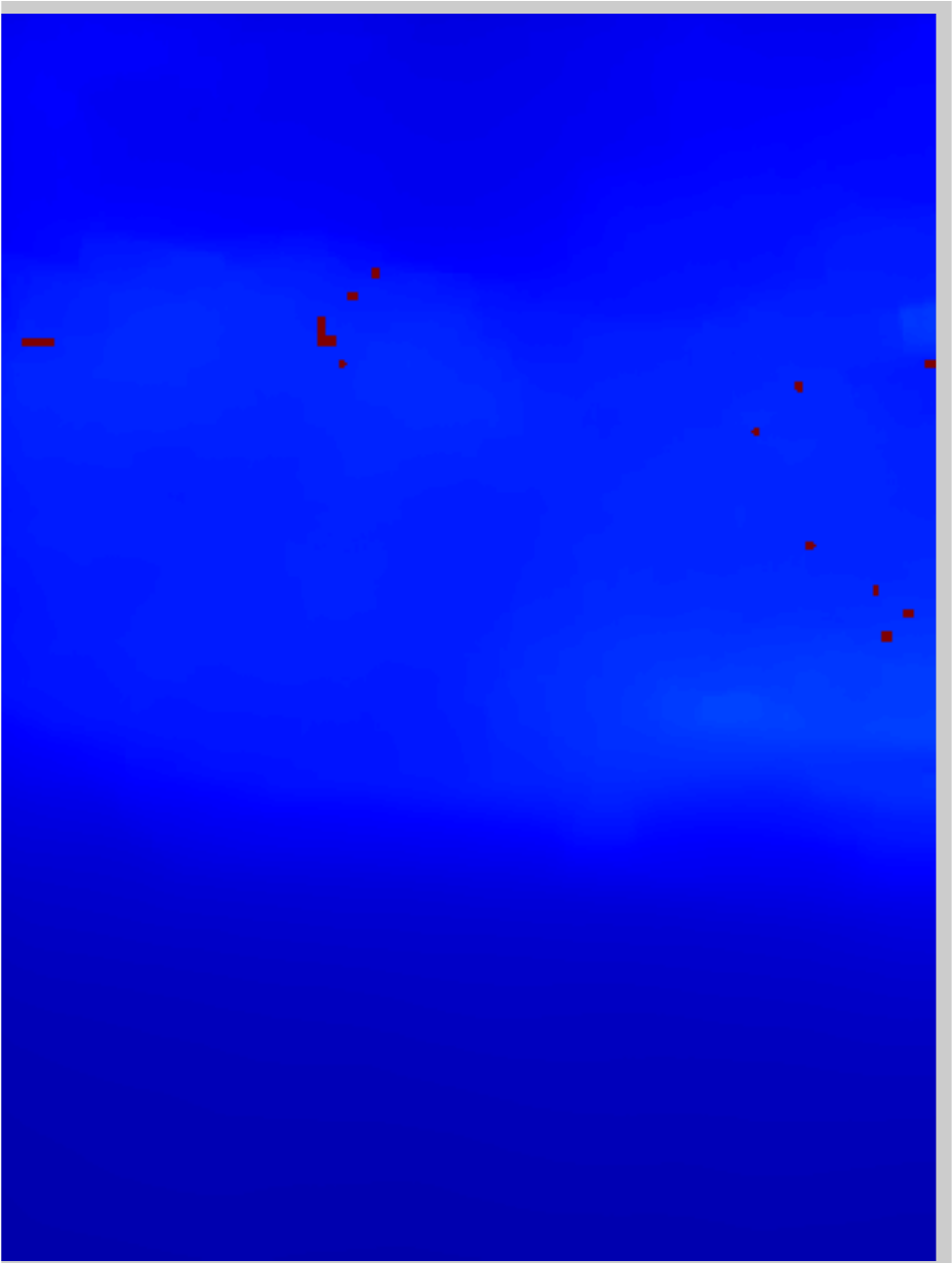} &
			\hspace{-0.0cm}\includegraphics[width=0.15\linewidth]{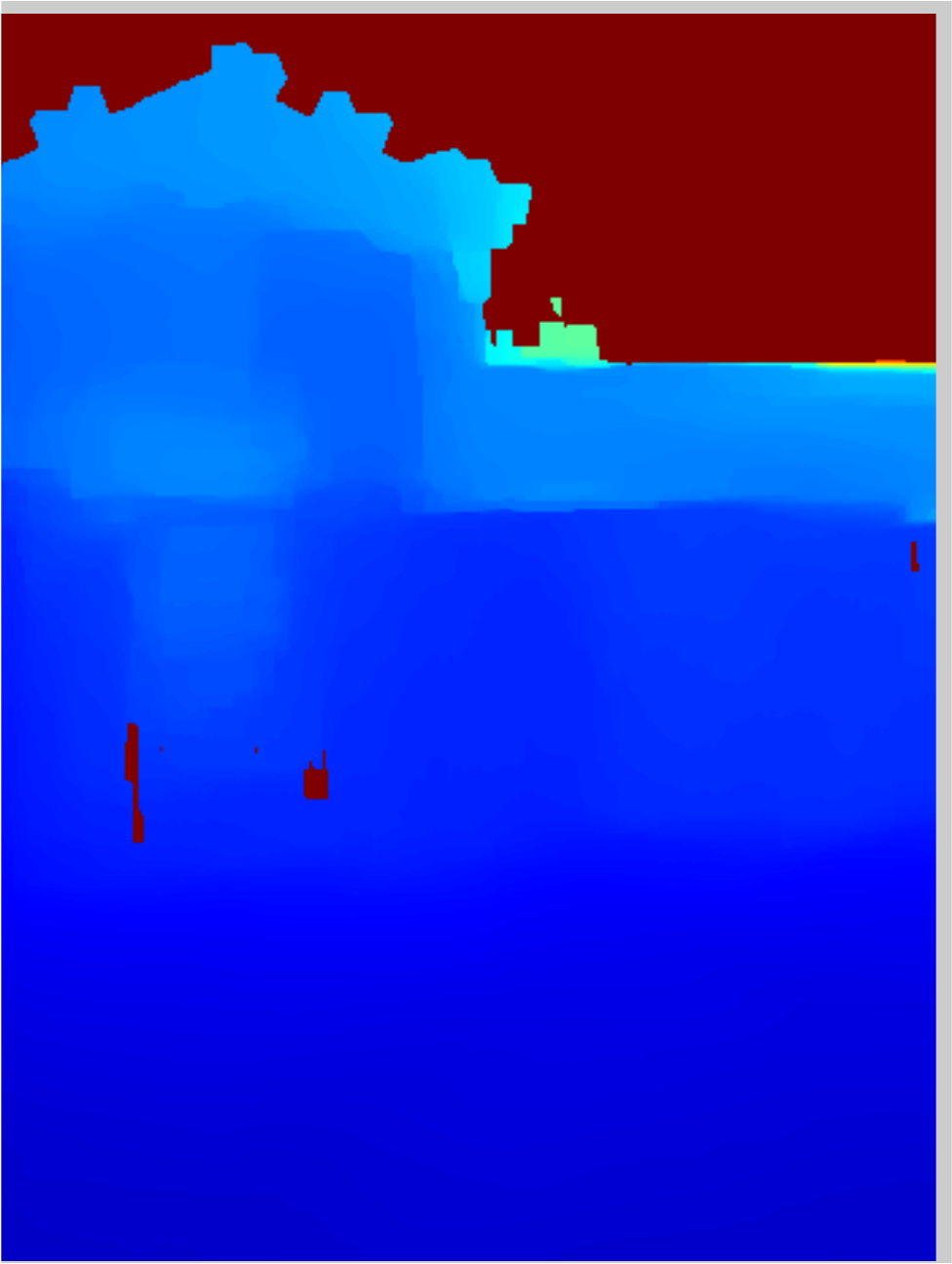} &
			\hspace{-0.0cm}\includegraphics[width=0.15\linewidth]{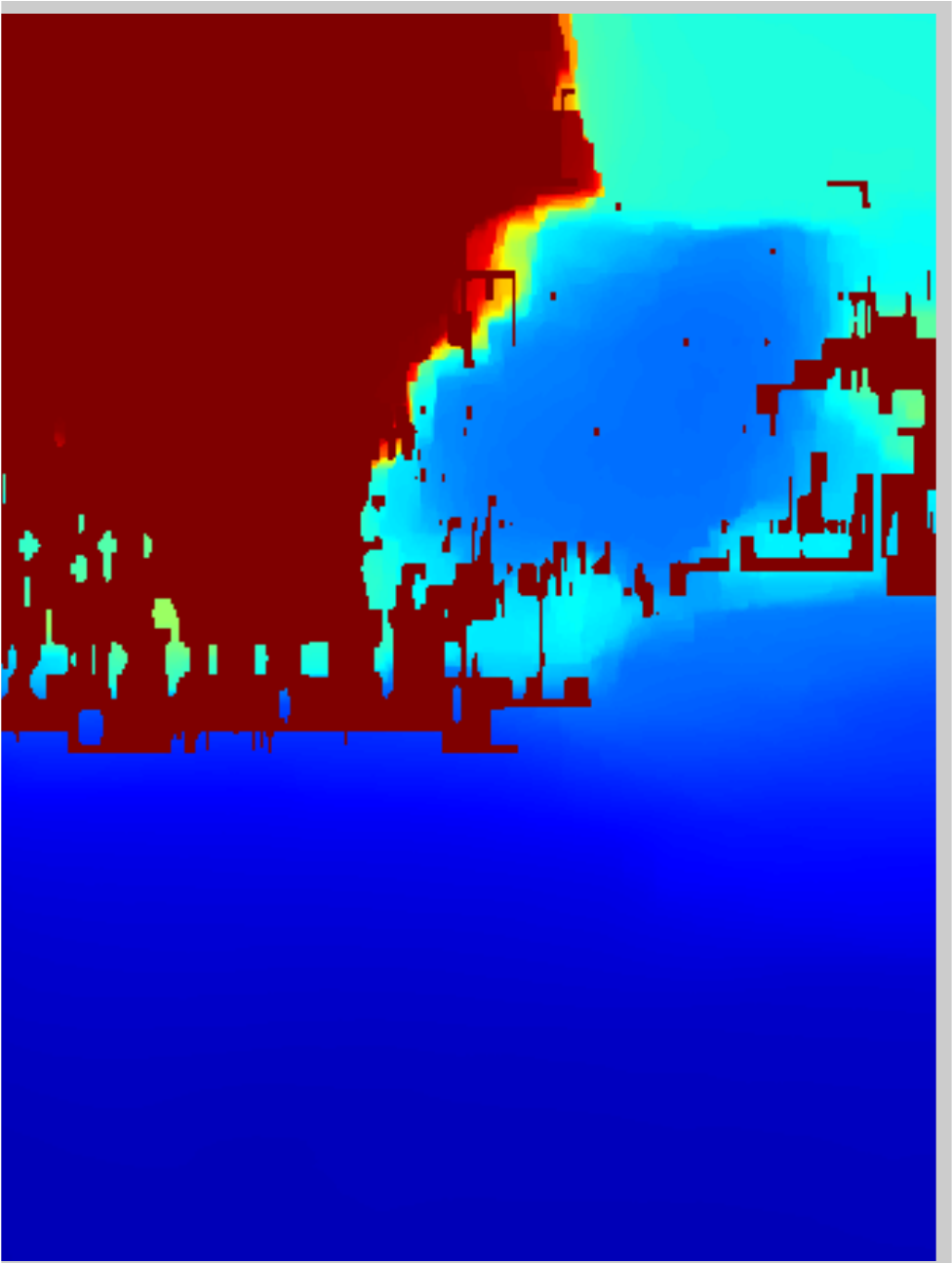} \\
			\hspace{-0.0cm}\begin{sideways}\hspace{1.1cm}{\bf Ours-1F}\end{sideways} &
			\hspace{-0.0cm}\includegraphics[width=0.15\linewidth]{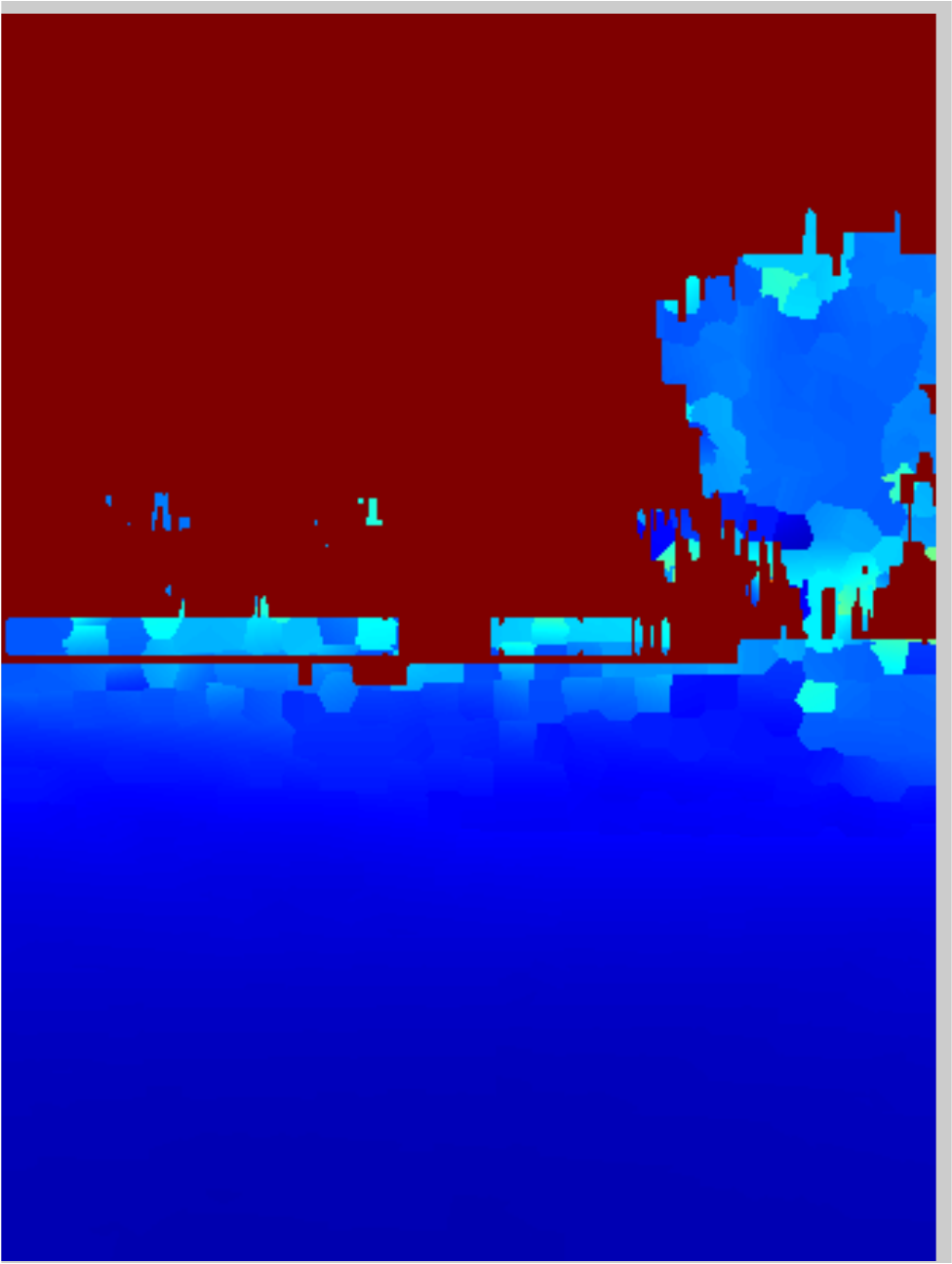} &
			\hspace{-0.0cm}\includegraphics[width=0.15\linewidth]{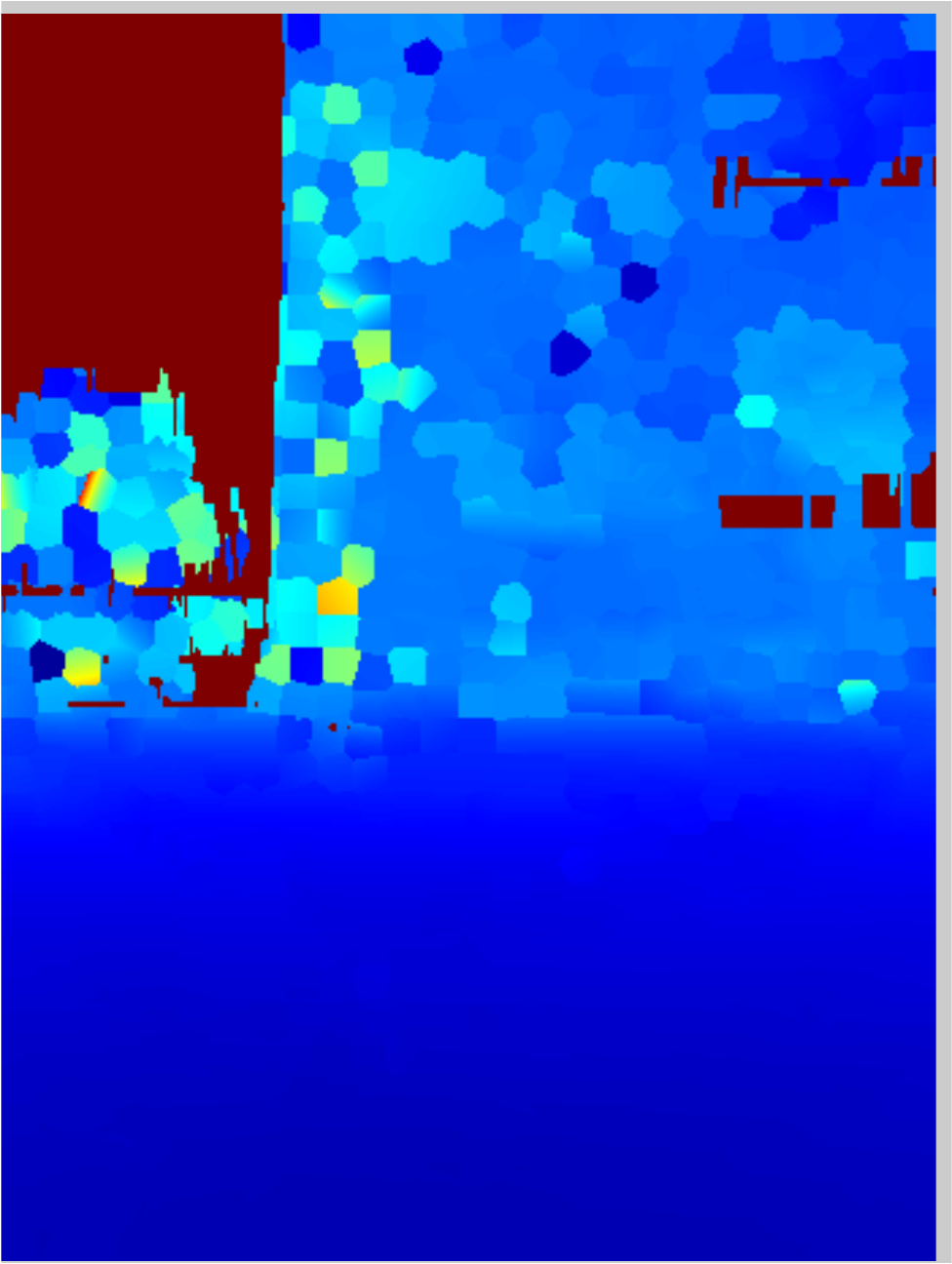} &
			\hspace{-0.0cm}\includegraphics[width=0.15\linewidth]{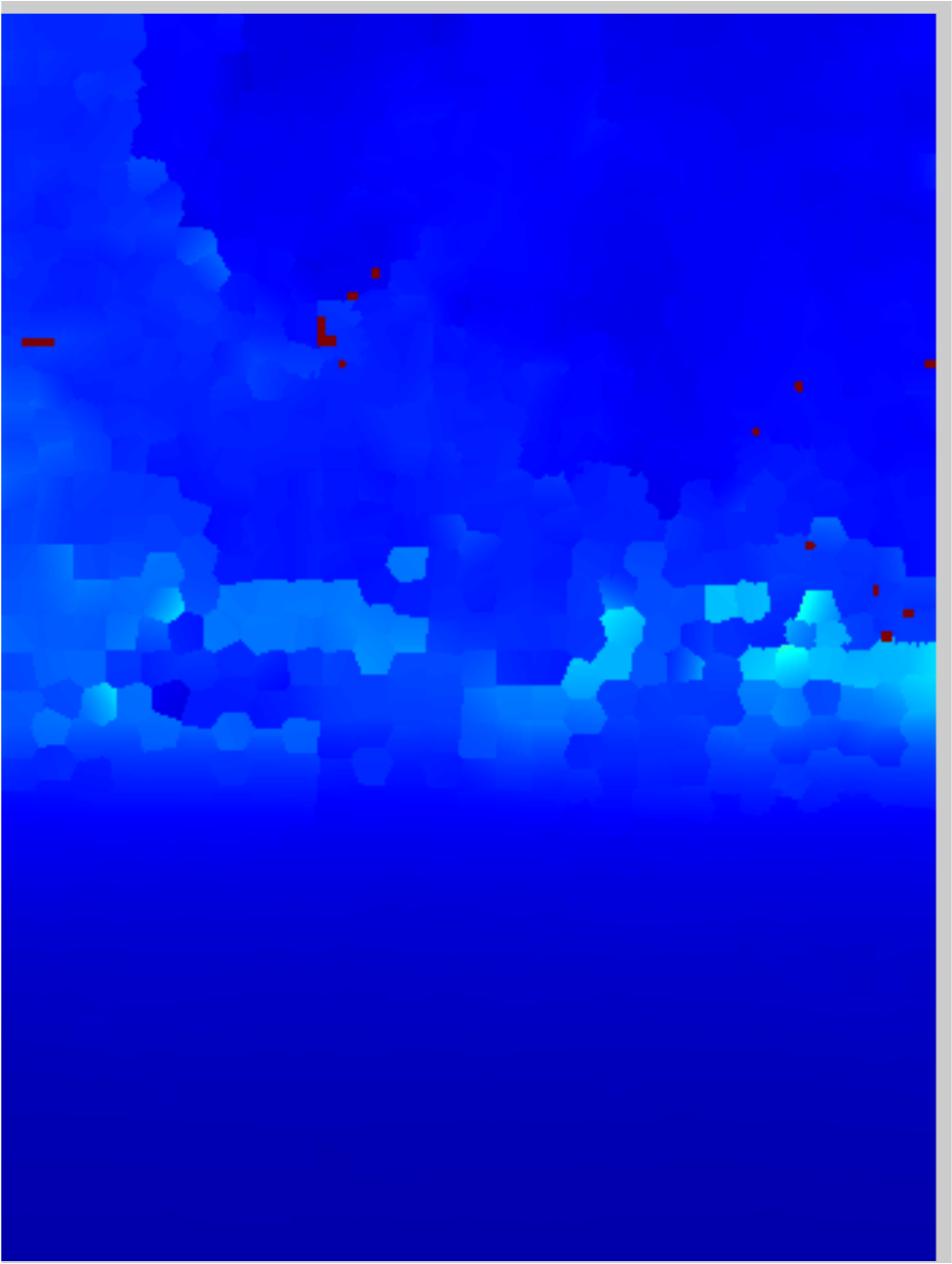} &
			\hspace{-0.0cm}\includegraphics[width=0.15\linewidth]{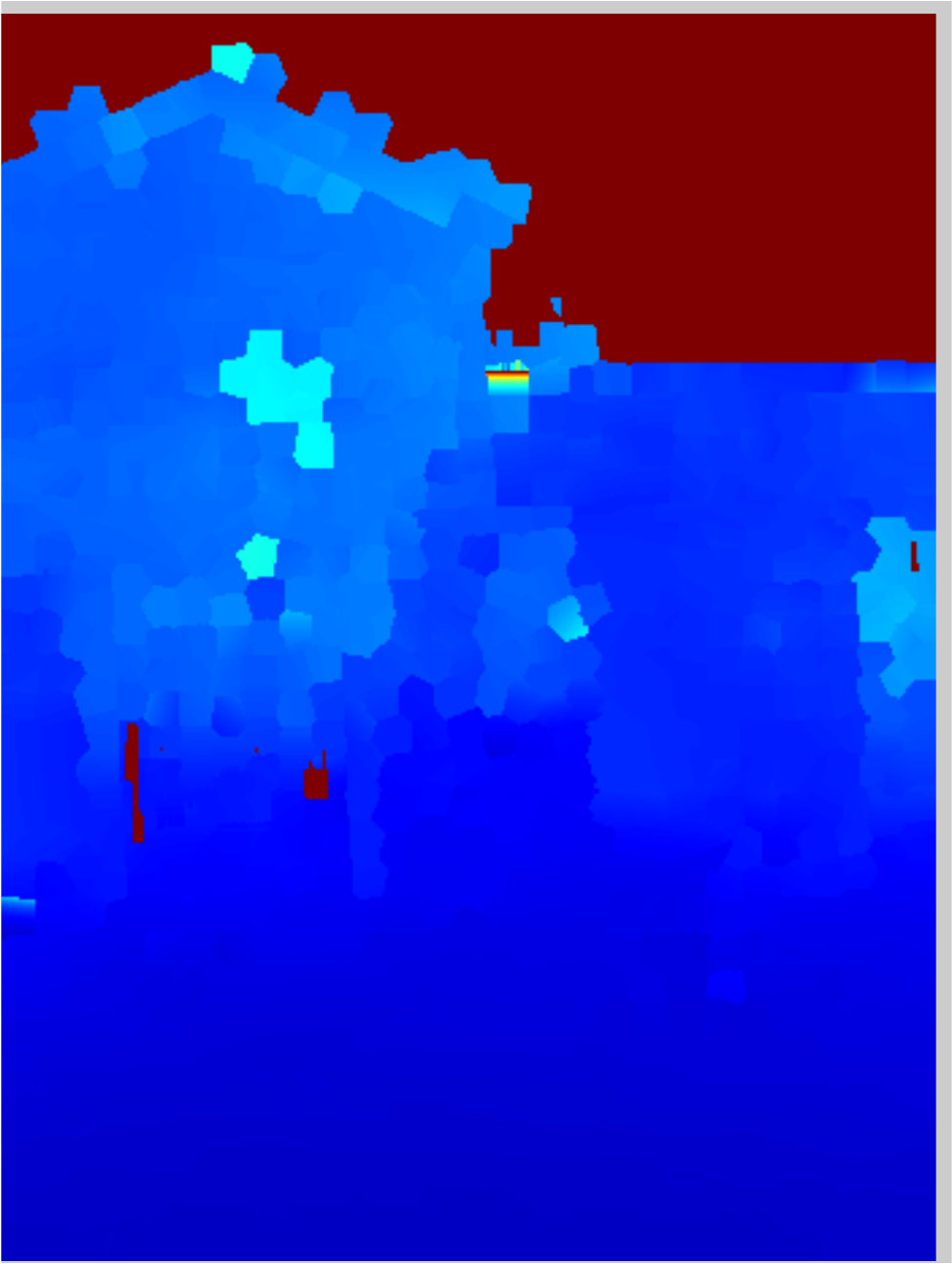} &
			\hspace{-0.0cm}\includegraphics[width=0.15\linewidth]{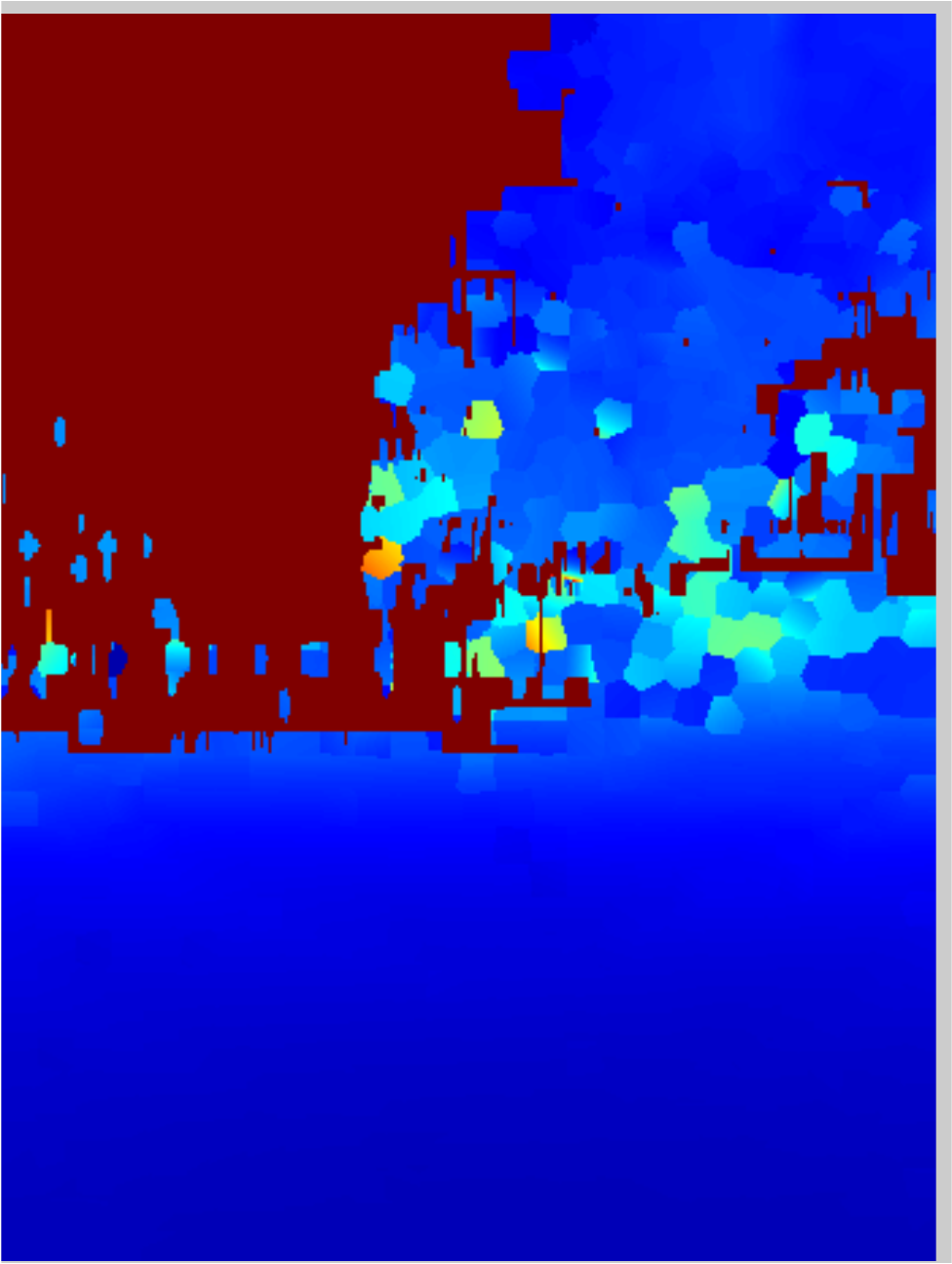} \\
		\end{tabular}
	\end{small}
	\vspace{-0.1cm}
	\caption{Make3D: Qualitative comparison of the depths estimated with DepthTransfer~\cite{Karsch12} and with our single-image method. Color indicates depth (red is far, blue is close).}
	\label{fig:make3dcomp}
\end{figure*}


For both datasets, i.e., Make3D and NYUv2, we used SLIC~\cite{Achanta12} to compute the superpixels. For each test image, we retrieved $K=7$ candidates from the training images. The parameters of our model were set using a small validation set of 10 images from the NYUv2 training set and kept the same in both experiments. The specific values of these parameters were $w_p = 1$, $w_m = 10$, $\theta_e = 10$, and $\theta_{max} = 20$. Note that these parameters could, in principle be learned. However, our approach proved robust enough for us to just have to search for their correct order of magnitude. In the following experiments, we performed two iterations of PCBP with $N_s = 20$ particles at each iteration.

\begin{table}[t!]
	\centering
	\begin{tabular}{|c |c c| c| c|}
		\hline
		Method & & ${\bf rel}$ & ${\bf log}_{10}$ & ${\bf rms}$ \\ 
		\hline
		\multirow{2}{*}{DepthTransfer~\cite{Karsch12}} &${\bf C_1}$&0.355 & {\bf 0.127} &{\bf  9.2}\\
		& ${\bf C_2}$&0.361&0.148&15.1 \\ 
		\hline
		\multirow{2}{*}{Ours-1F}&${\bf C_1}$& {\bf 0.335} & 0.137 & 9.49\\
		&${\bf C_2}$& {\bf 0.338}&{\bf 0.134} &{\bf 12.6} \\
		\hline
	\end{tabular}
	\vspace{0.2cm}
	\caption{Make3D: Depth reconstruction errors for DepthTransfer~\cite{Karsch12} and for our method evaluated according to two criteria (${\bf C_1}$ and ${\bf C_2}$, see text for details.)}
	\label{table:make3dcomp}
\end{table}

\begin{table}[t!]
	\centering
	\begin{tabular}{|c |c| c| c|}
		\hline
		Method & ${\bf rel}$ & ${\bf log}_{10}$ & ${\bf rms}$ \\ 
		\hline
		Unary & 0.352 & 0.142 & 9.61\\
		GP regression & 0.547 & 0.175 & 10.5 \\ 
		No discrete variables & 0.326 & 0.147 & 9.932\\
		No sampling & 0.337 &  0.139 & 9.54 \\
		Ours-1F  & 0.335 &  0.137 & 9.49\\
		\hline
	\end{tabular}
	\vspace{0.2cm}
	\caption{Make3D: Comparison of our final results with those obtained with unary terms only, with our GP depth regressors only, using a model without discrete edge type variables, and after the first round of PCBP where no sampling is involved.}
	\label{table:make3d_terms}
\end{table}

\subsubsection{Outdoor Scene Reconstruction: Make3D}

The Make3D dataset contains 534 images with corresponding depth maps, partitioned into 400 training images and 134 test images. All the images were resized to 460$\times$345 pixels in order to preserve the aspect ratio of the original images. Since the true focal length of the camera is unknown, we assume a reasonable value of 500 for the resized images. Due to the limited range and resolution of the sensor used to collect the ground-truth, far away pixels were arbitrarily set to depth 80 in the original dataset. To take this, as well as the effect of interpolation when resizing the images, into account in our evaluation, we report errors based on two different criteria: (${\bf C_1}$) Errors are computed in the regions with ground-truth depth less than 70; (${\bf C_2}$) Errors are computed in the entire image. In this second scenario, to reduce the effect of meaningless candidates in sky regions, we used a classifier to label sky pixels and forced the depth of the corresponding superpixels to take the value $(0,0,1,80)$. Note that the same two criteria (${\bf C_1}$ and ${\bf C_2}$) were used to evaluate the baseline. 

\begin{figure*}[t!] 
	\centering
	\begin{tabular}{ccccc}
		\hspace{0.7cm}\includegraphics[width=0.15\linewidth]{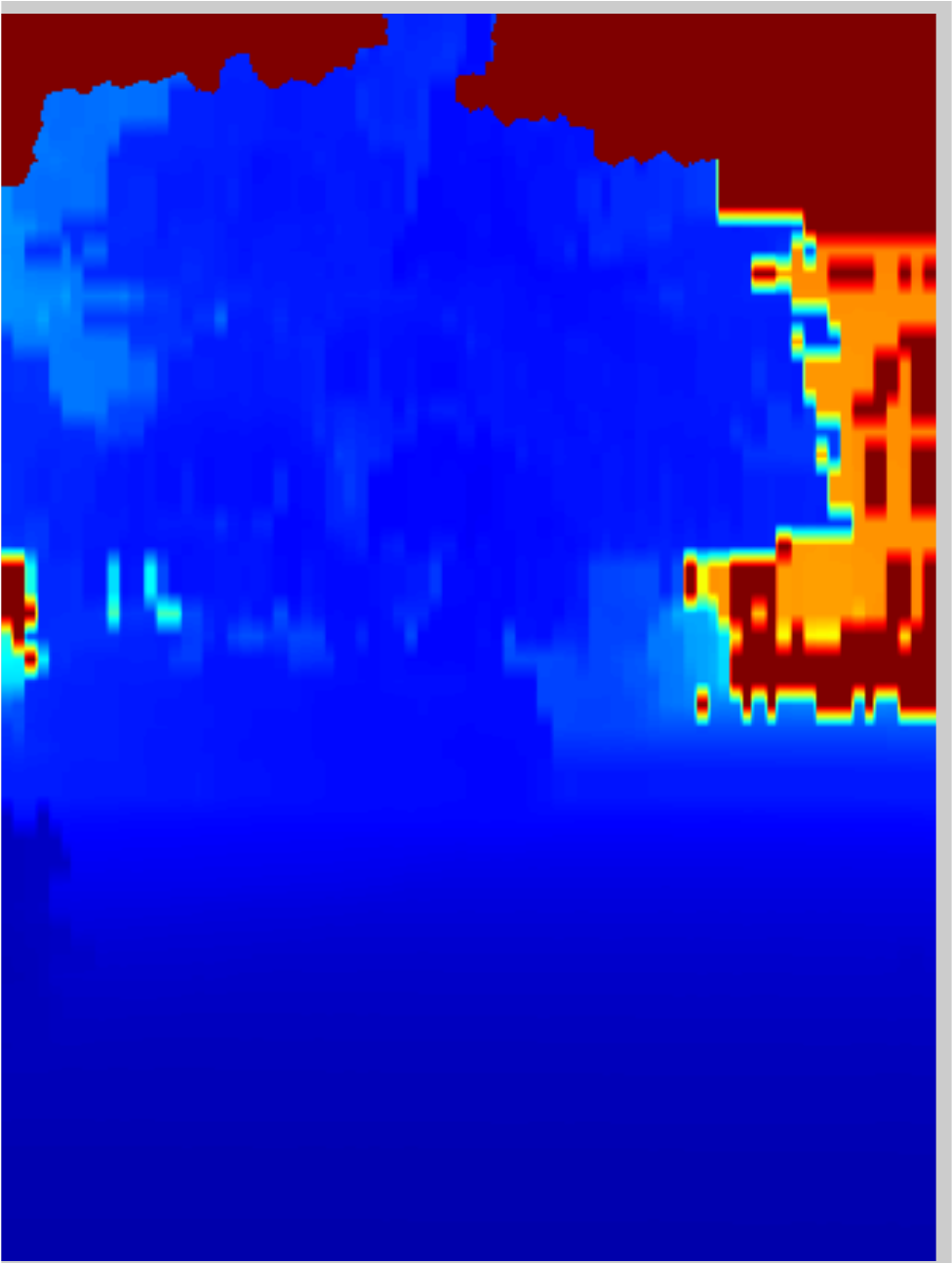} &
		\hspace{-0.0cm}\includegraphics[width=0.15\linewidth]{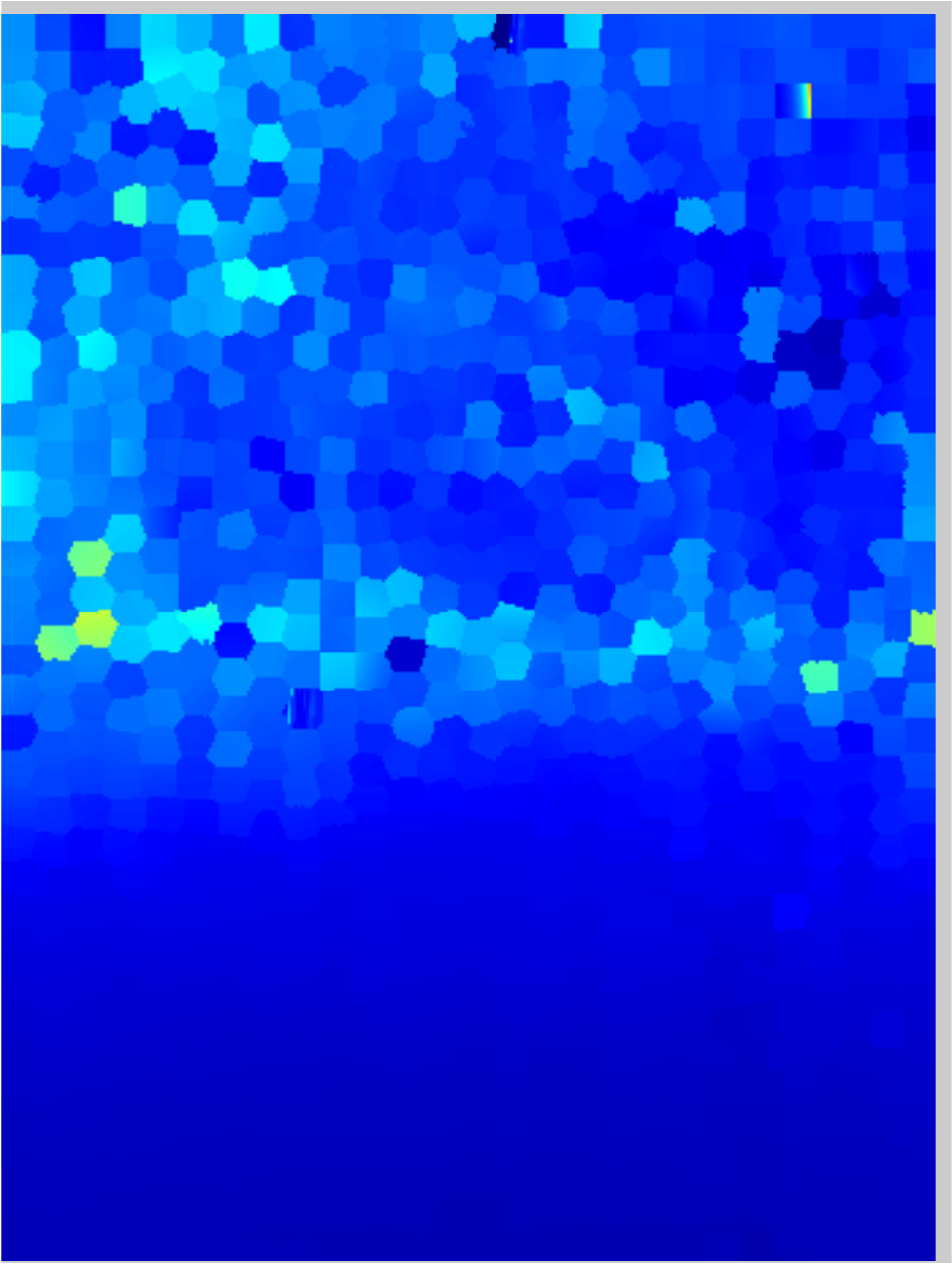} &
		\hspace{-0.0cm}\includegraphics[width=0.15\linewidth]{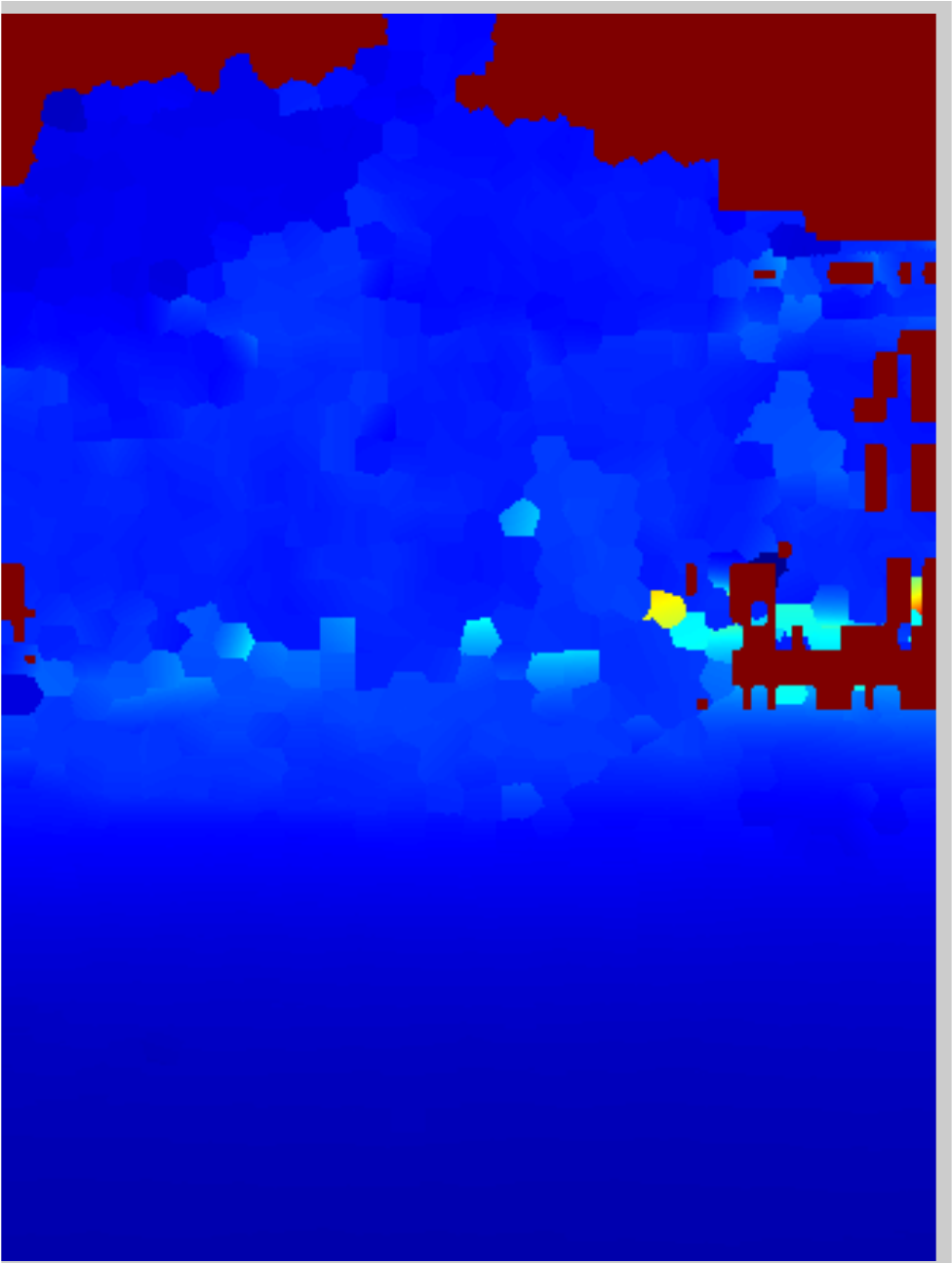} &
		\hspace{-0.0cm}\includegraphics[width=0.15\linewidth]{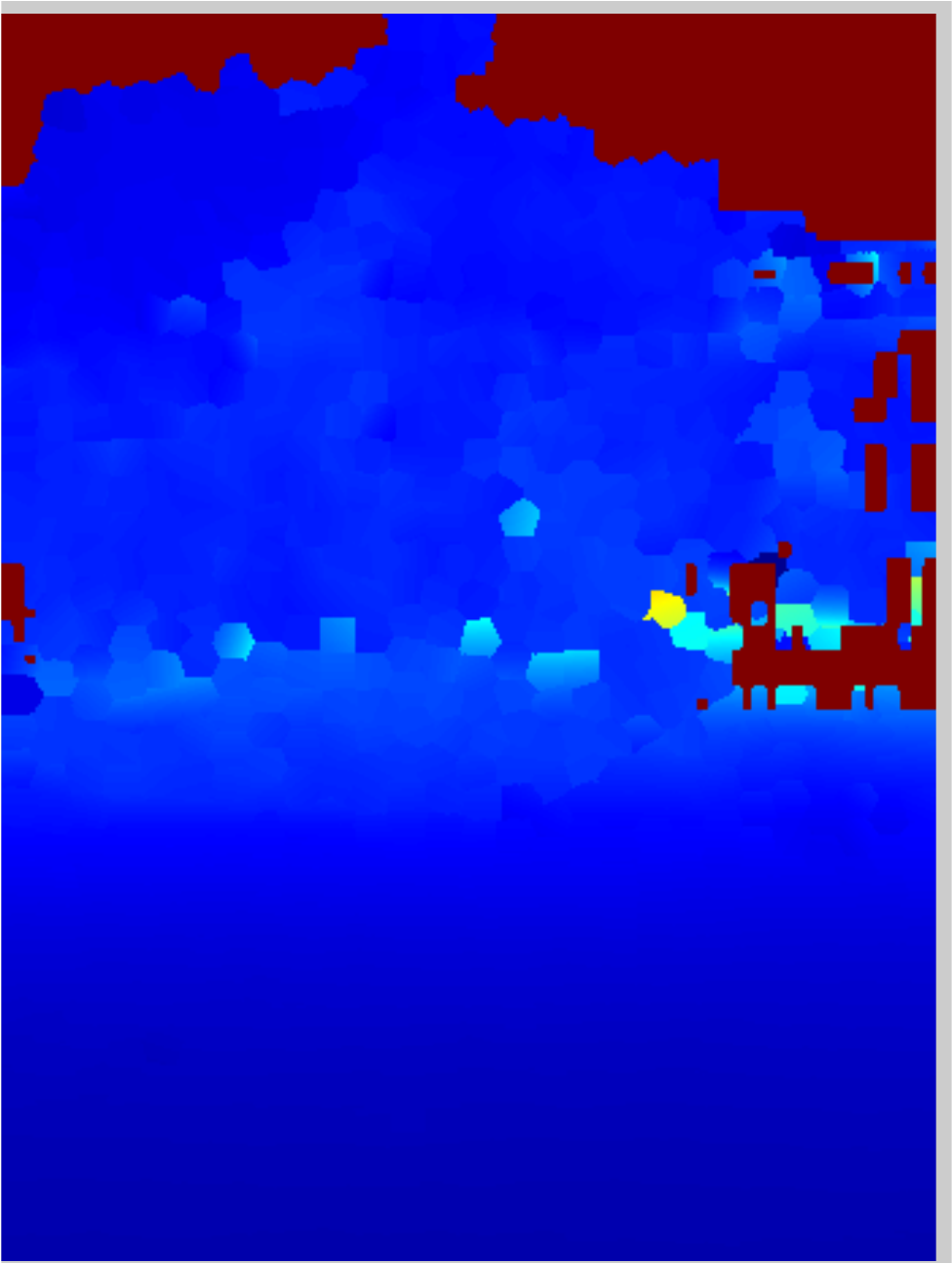} &
		\hspace{-0.0cm}\includegraphics[width=0.15\linewidth]{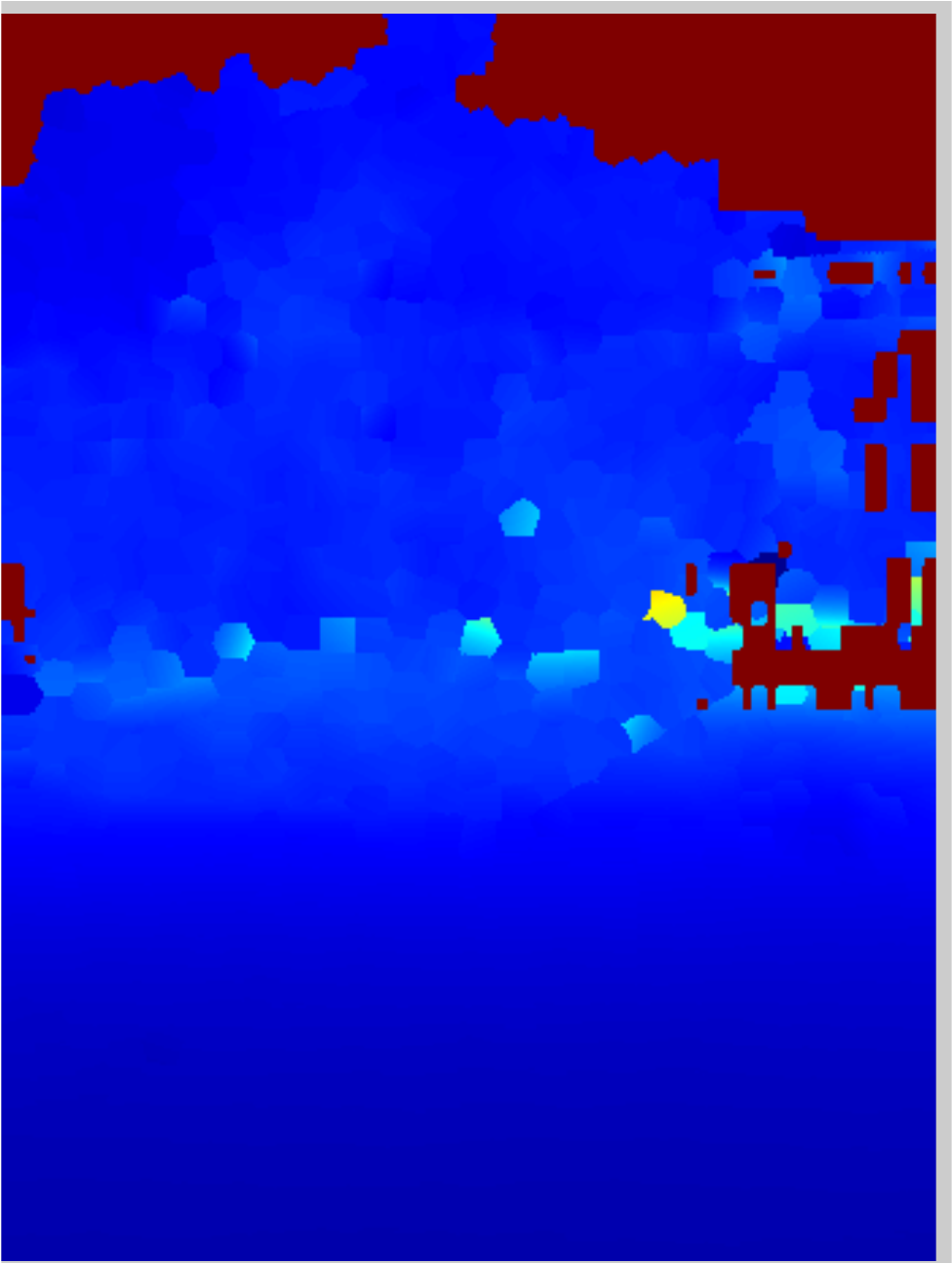} \\
		\hspace{0.7cm} Ground-truth & \hspace{-0.0cm} Regression & \hspace{-0.0cm} No sampling & \hspace{-0.0cm} Sampling round 1 & \hspace{-0.0cm}Final depth
	\end{tabular}
	\vspace{-0.2cm}
	\caption{Make3D: Depth maps at different stages of our approach.}
	\label{fig:m3d_evolution}
\end{figure*}

In Table~\ref{table:make3dcomp}, we compare the results of our approach with those obtained by DepthTransfer~\cite{Karsch12}. Note that, using criteria ${\bf C_1}$, we outperform the baseline in terms of relative error and perform slightly worse for the other metrics. Using criteria ${\bf C_2}$, we outperform the baseline for all metrics. Fig.~\ref{fig:make3dcomp} provides a qualitative comparison of our depth maps with those estimated by DepthTransfer~\cite{Karsch12} for some images of the dataset. Note that depth transfer tends to over-smooth the depth maps and, e.g., merge foreground objects with the background. Thanks to our discrete variables, our approach better respects the discontinuities in the scene. 

In Table~\ref{table:make3d_terms}, we show the results obtained with some of the parts of our model. Note that, even though the sampling in PCBP does not seem to have a great impact on the errors, it helps smoothing the depth maps and thus makes them look more realistic. This is evidenced by Fig.~\ref{fig:m3d_evolution}, where we show the depth maps at different stages of our approach. Note that the influence of each stage is more easily seen on the NYUv2 dataset (see Fig.~\ref{fig:nyu_evolution}) for which the overall depth range is smaller. 

\begin{figure*}[ht]
	\begin{small}
		\begin{tabular}{cccccc}
			\hspace{-0.5cm}\begin{sideways}\hspace{0.8cm}{\bf Image}\end{sideways} &
			\hspace{-0.37cm}\includegraphics[width=0.2\linewidth]{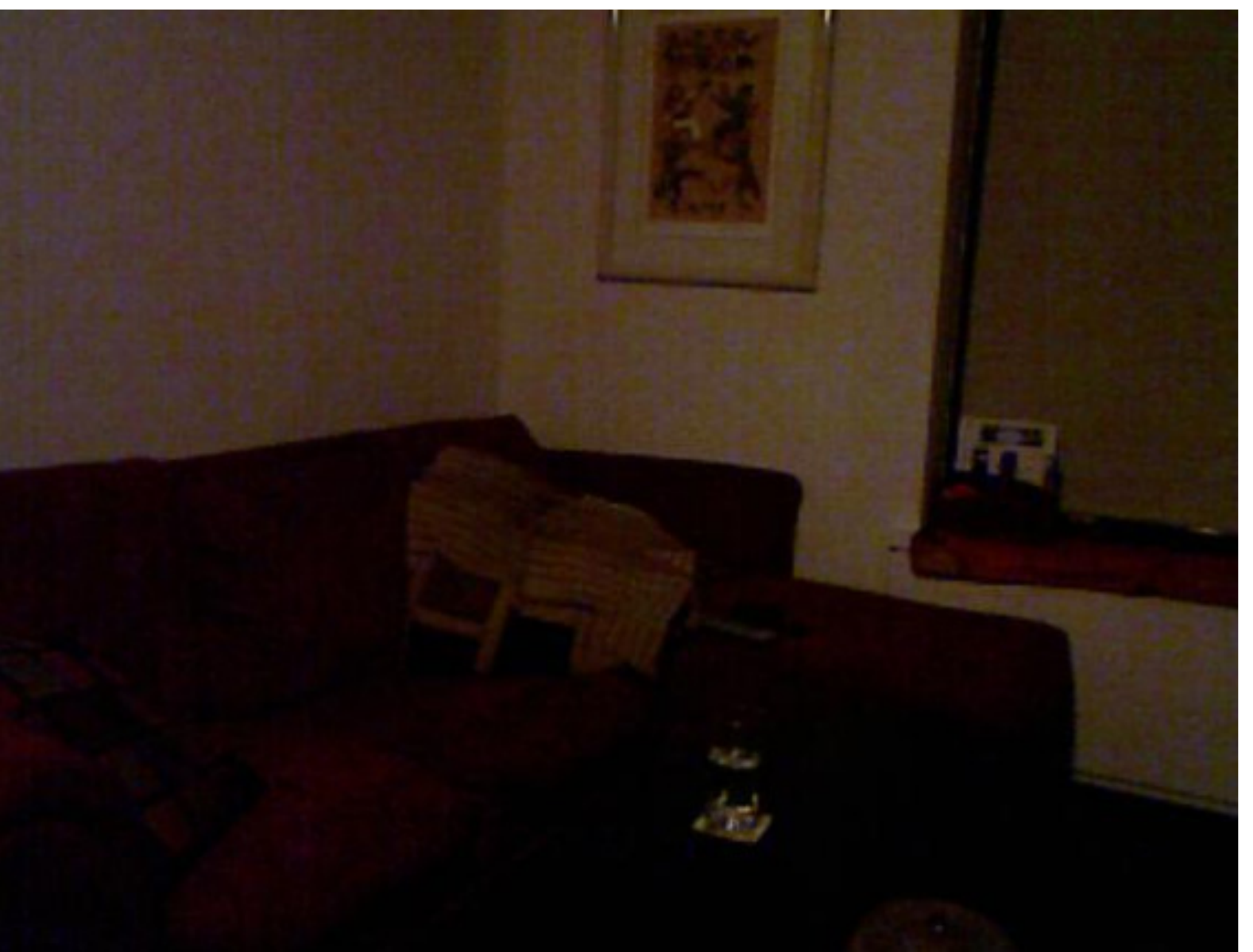} &
			\hspace{-0.37cm}\includegraphics[width=0.2\linewidth]{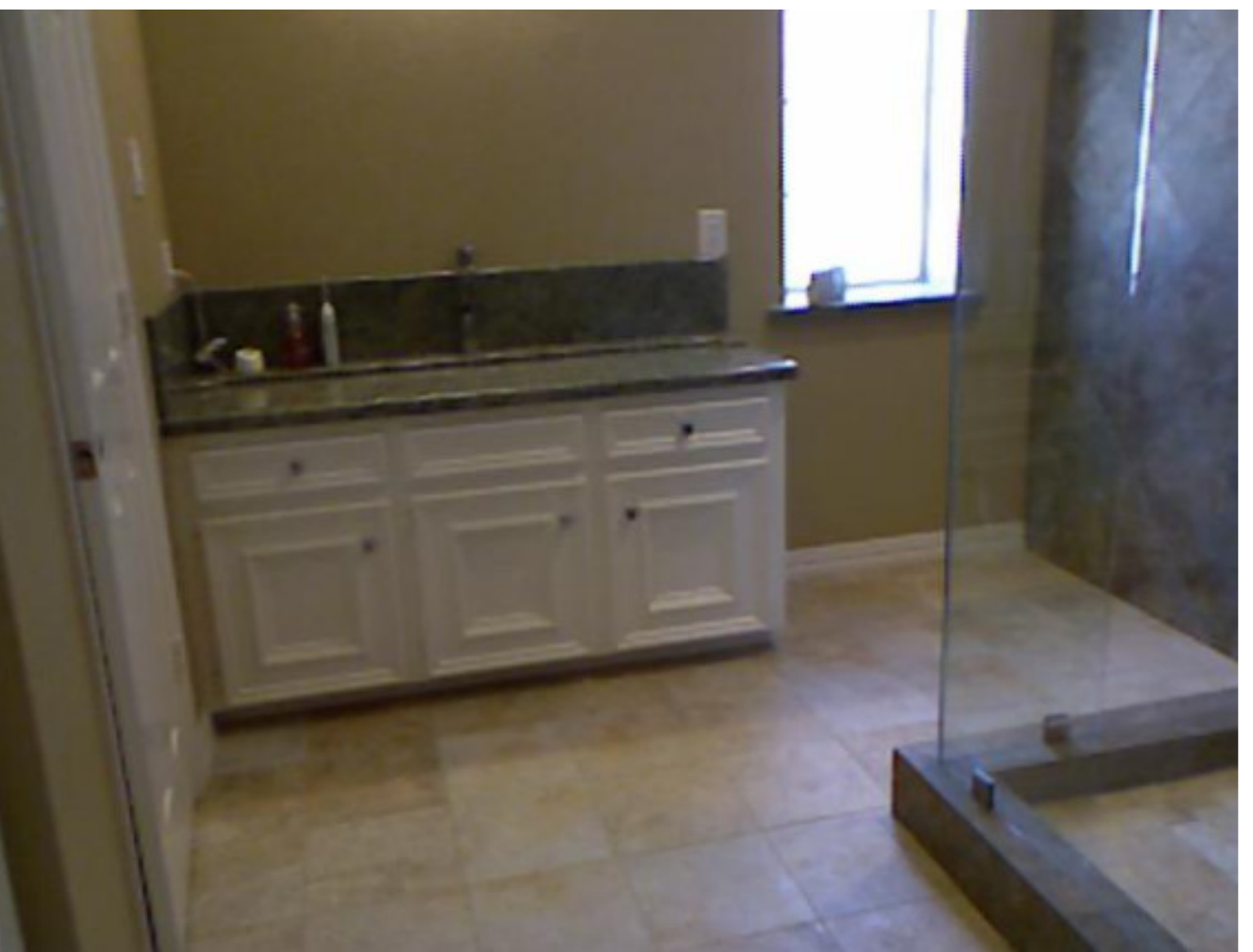} &
			\hspace{-0.37cm}\includegraphics[width=0.2\linewidth]{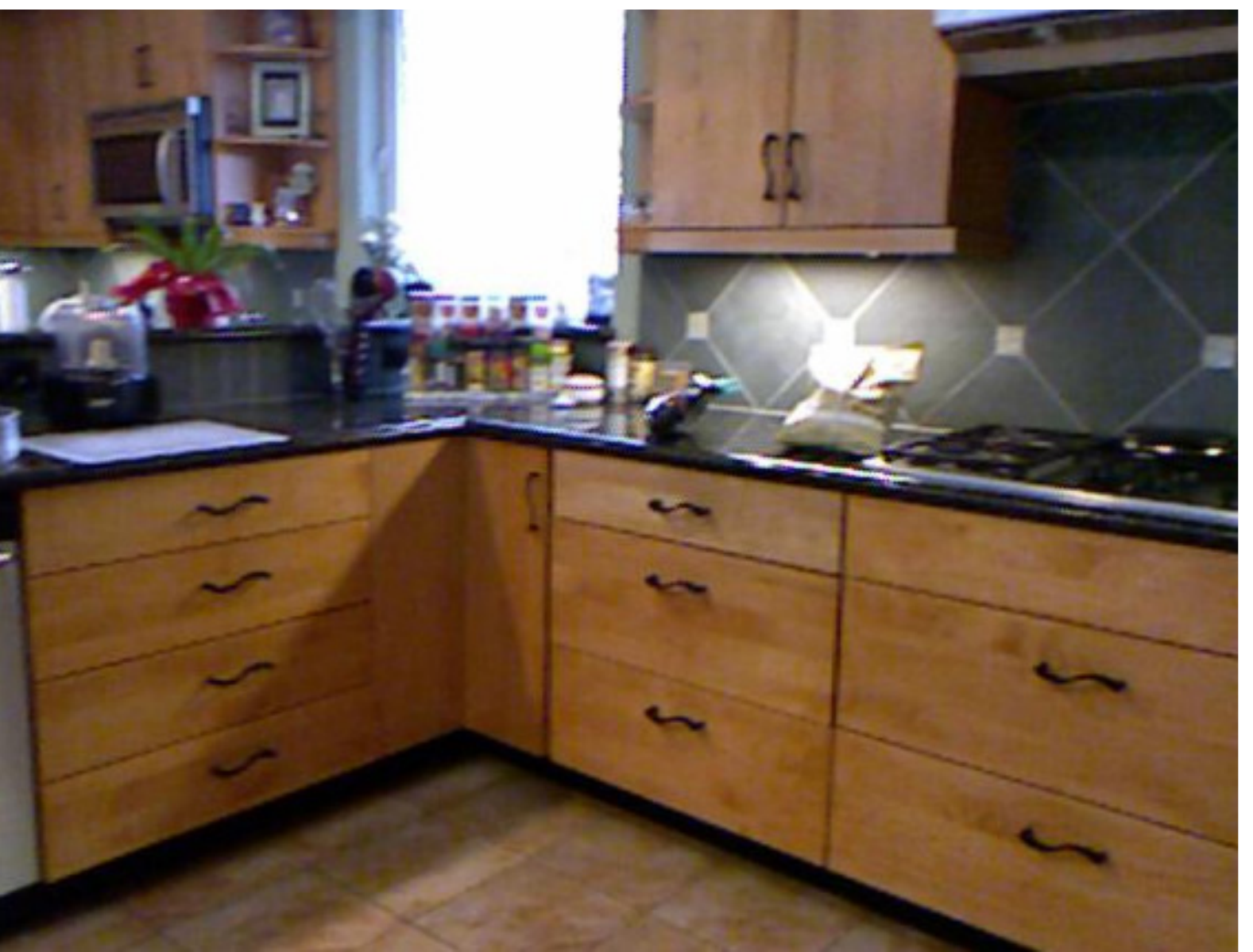} &
			\hspace{-0.37cm}\includegraphics[width=0.2\linewidth]{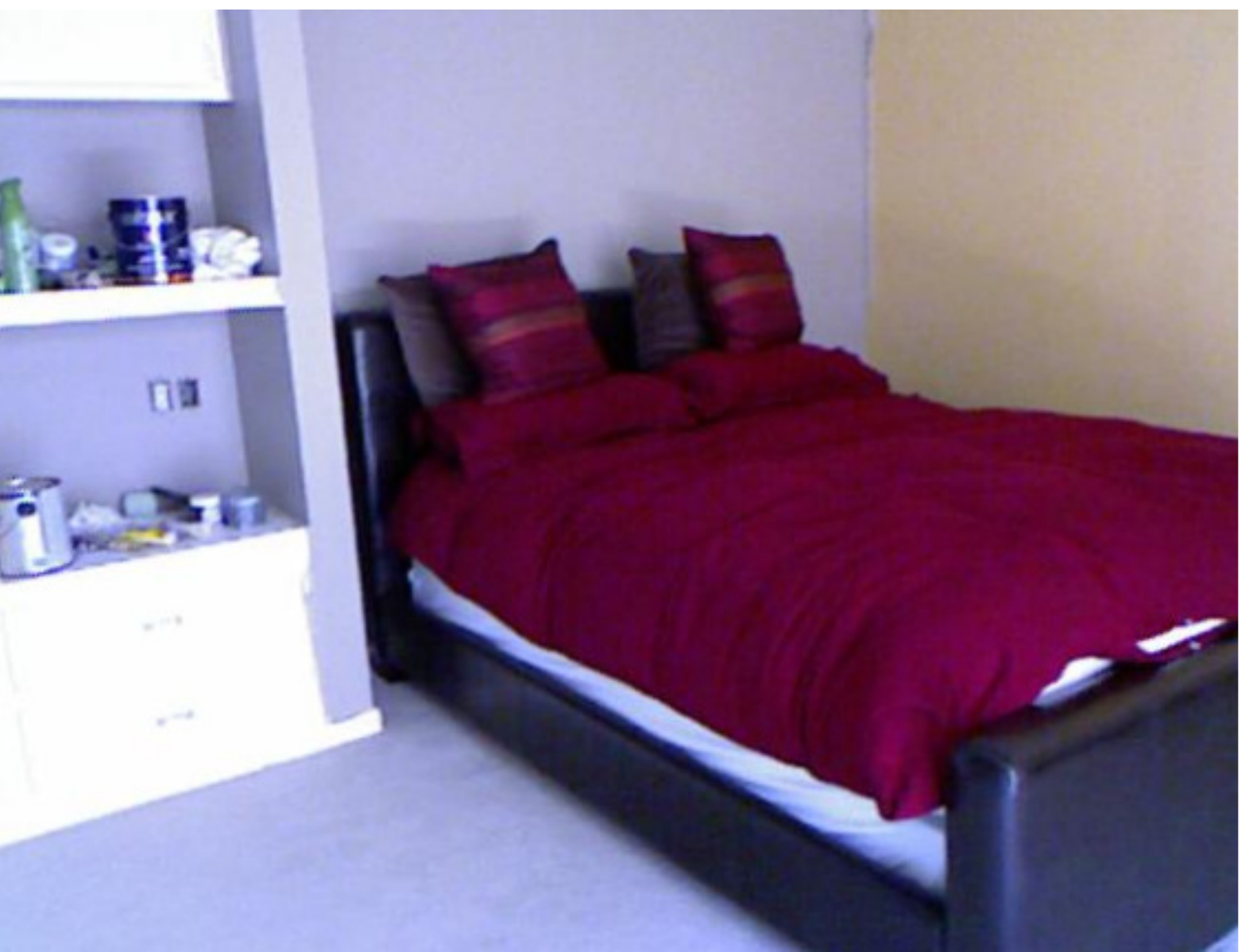} &
			\hspace{-0.37cm}\includegraphics[width=0.2\linewidth]{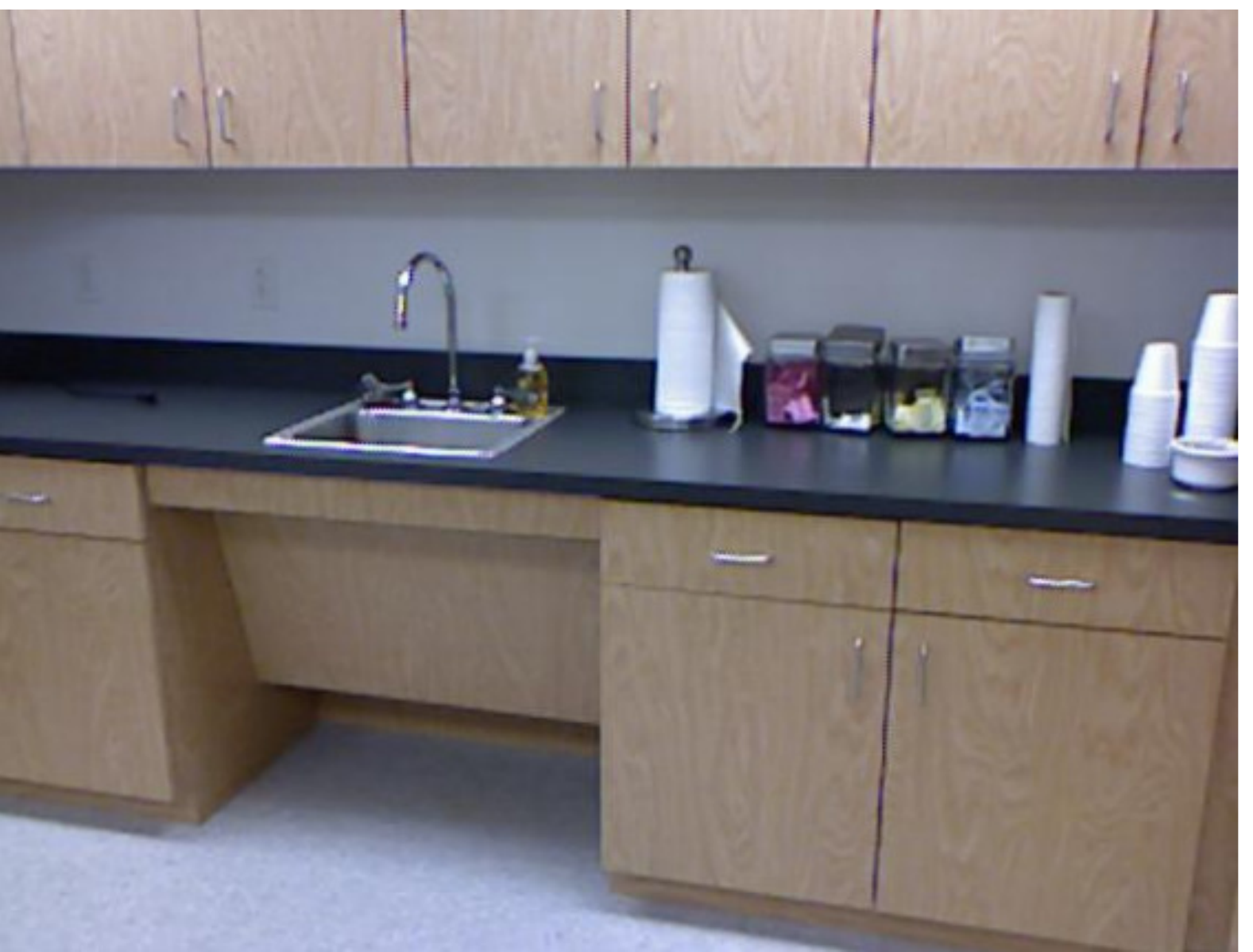} \\
			\hspace{-0.5cm}\begin{sideways}\hspace{0.4cm}{\bf Ground-truth}\end{sideways} &
			\hspace{-0.37cm}\includegraphics[width=0.2\linewidth]{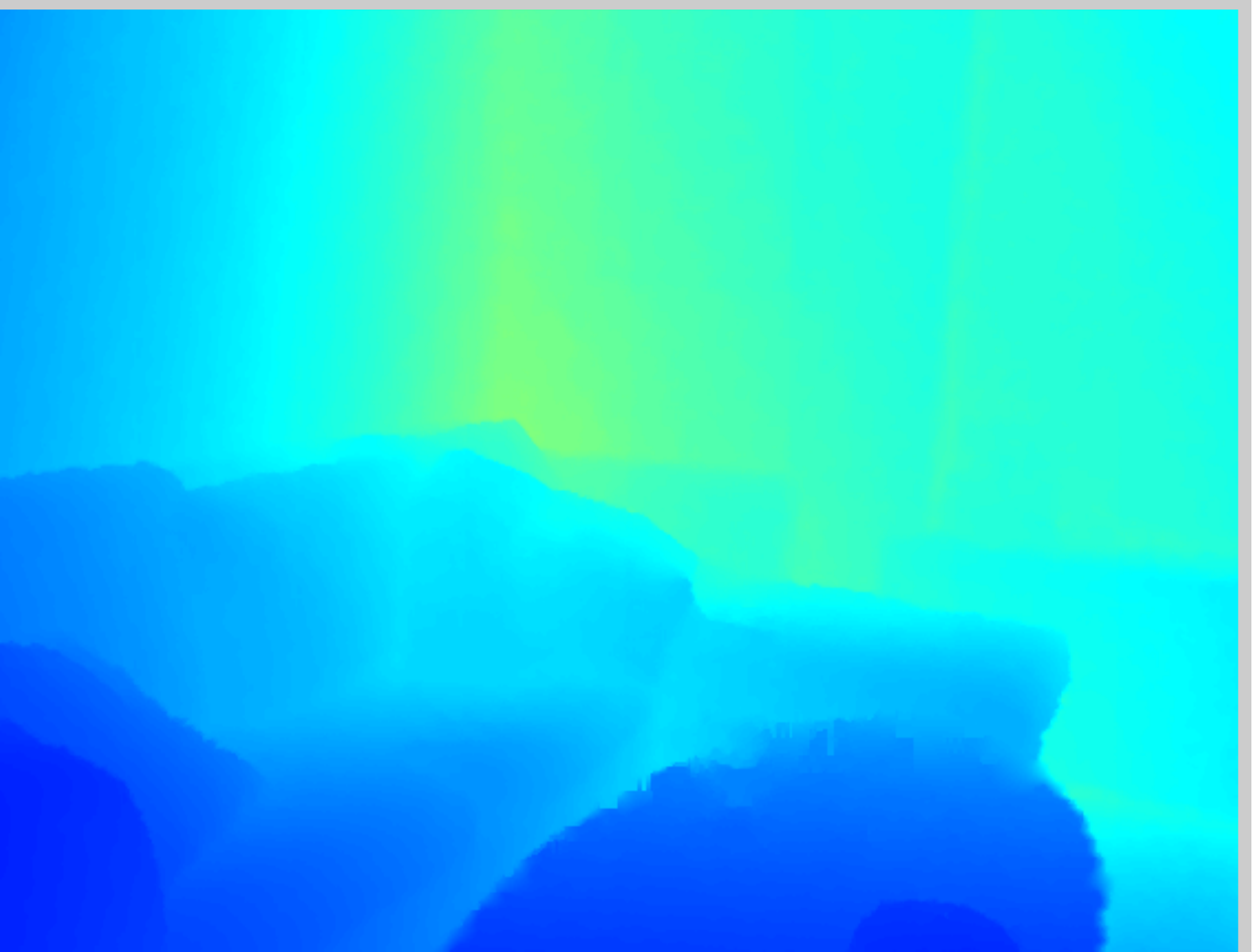} &
			\hspace{-0.37cm}\includegraphics[width=0.2\linewidth]{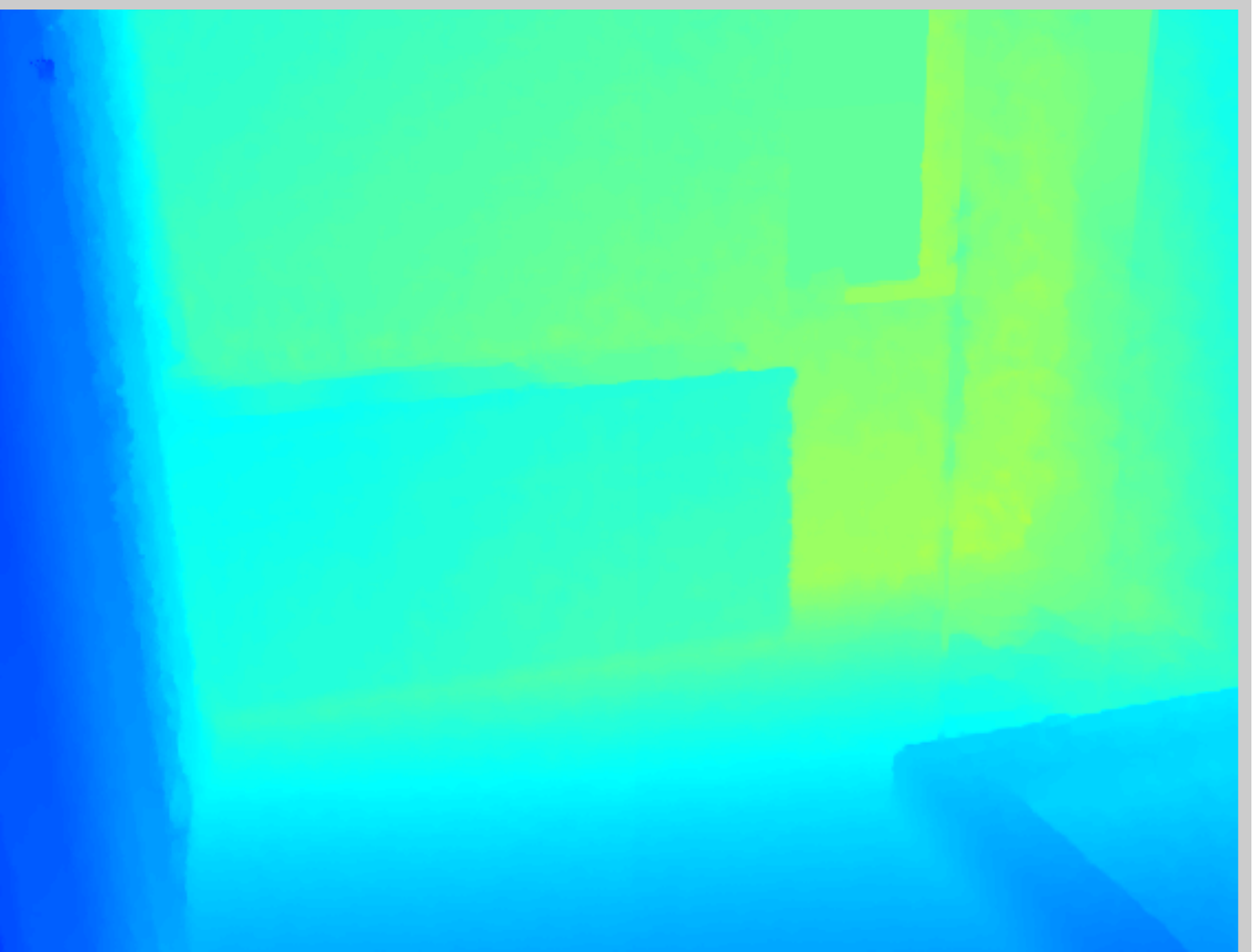} &
			\hspace{-0.37cm}\includegraphics[width=0.2\linewidth]{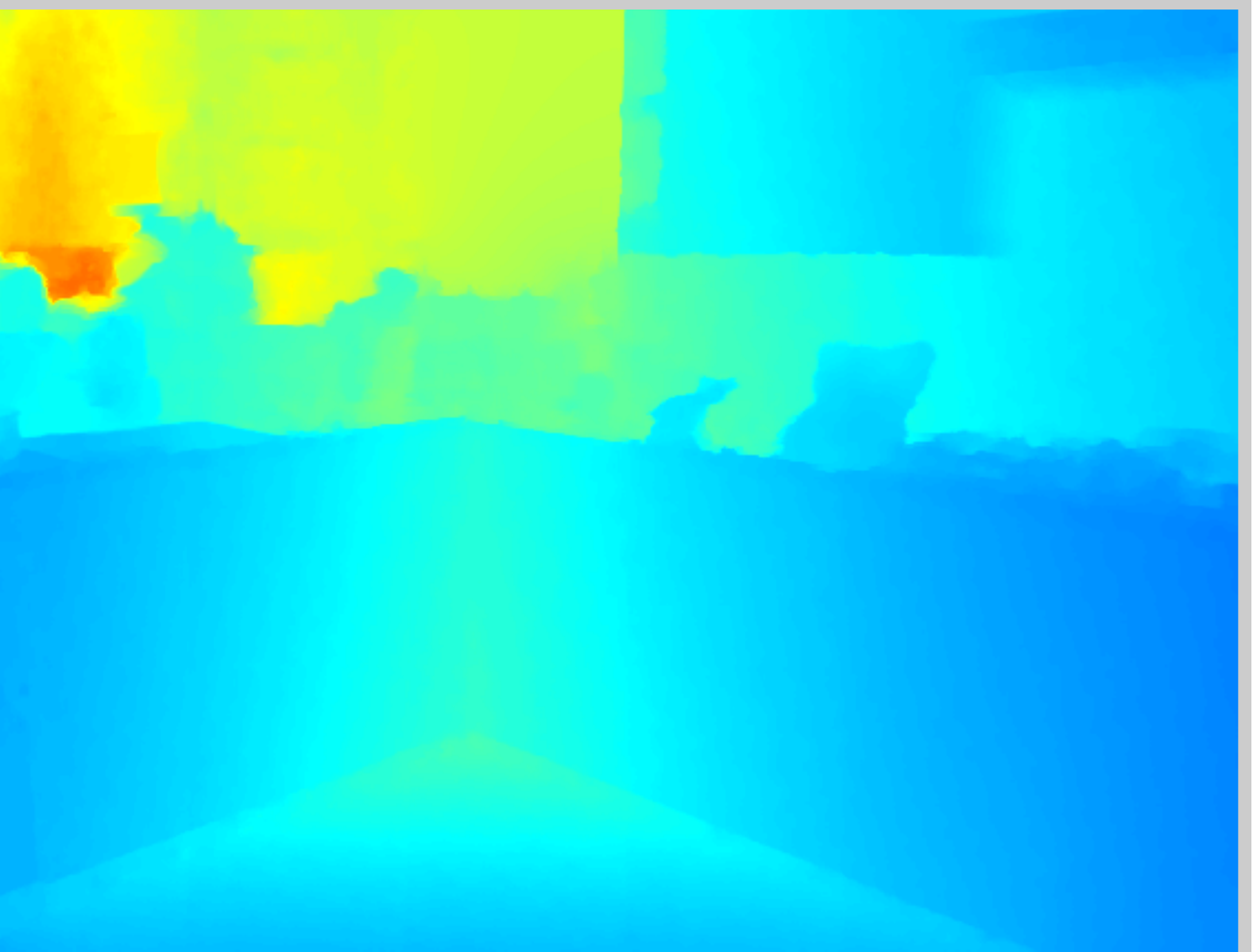} &
			\hspace{-0.37cm}\includegraphics[width=0.2\linewidth]{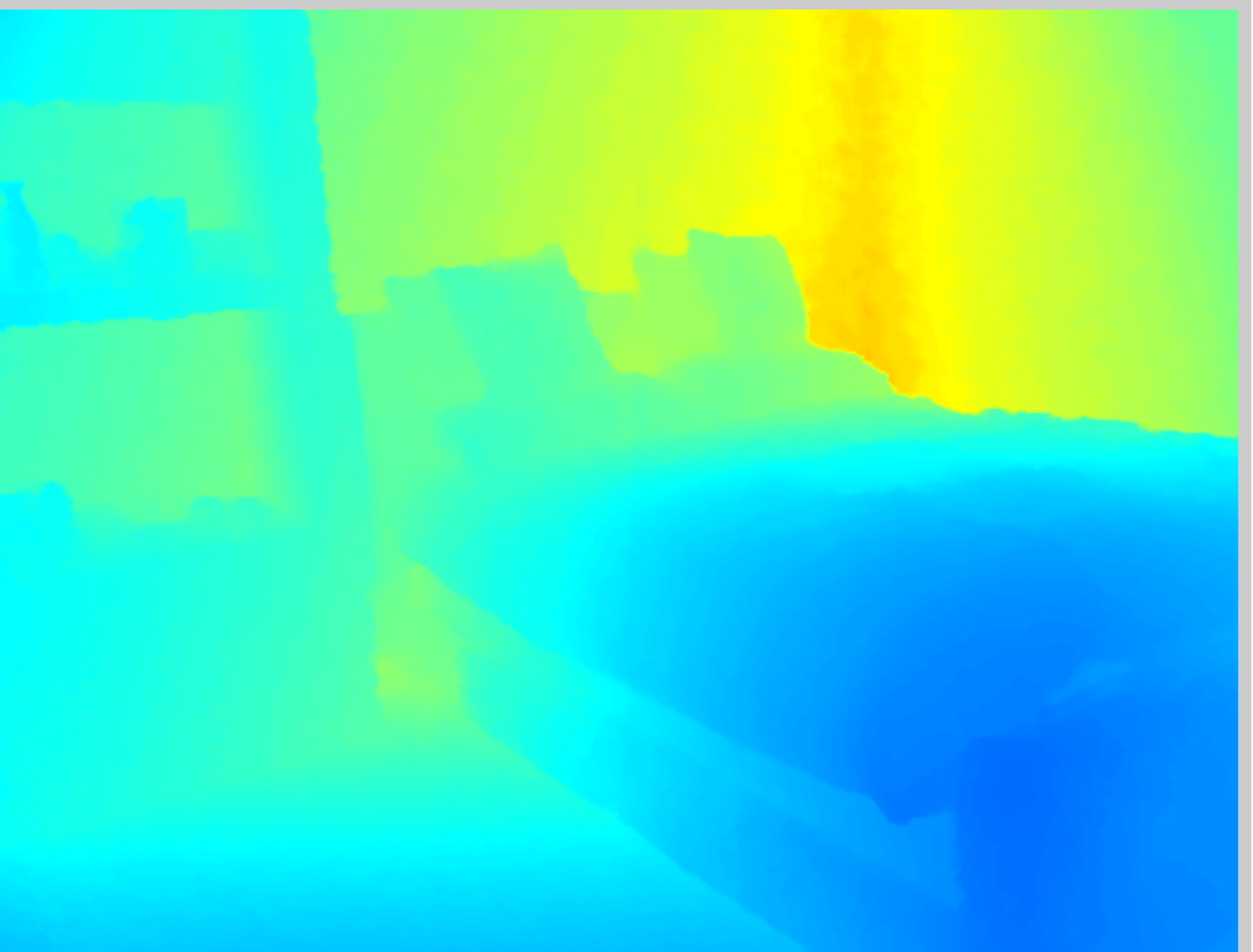} &
			\hspace{-0.37cm}\includegraphics[width=0.2\linewidth]{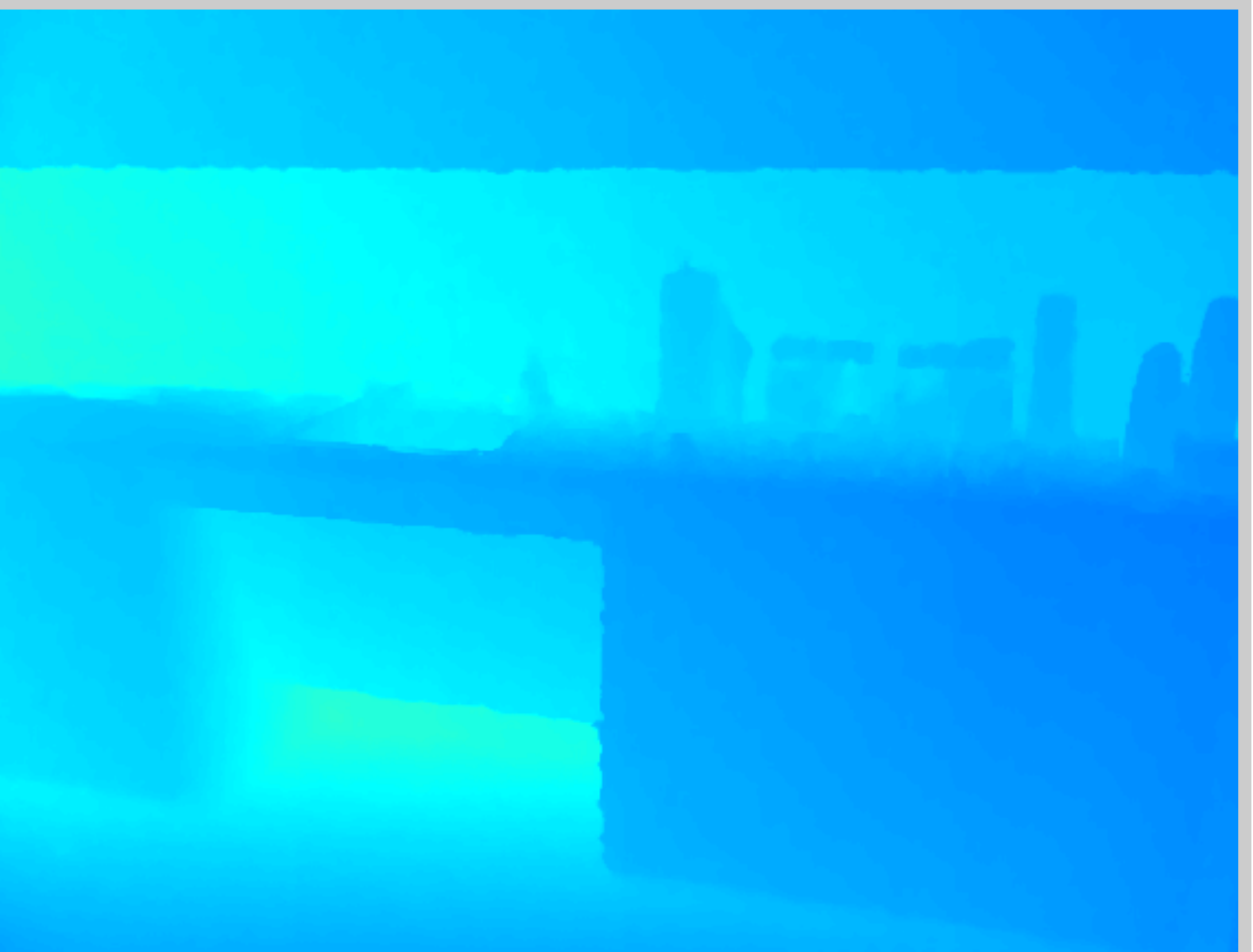} \\
			\hspace{-0.5cm}\begin{sideways}\hspace{0.0cm}{\bf DepthTransfer~\cite{Karsch12}}\end{sideways} &
			\hspace{-0.37cm}\includegraphics[width=0.2\linewidth]{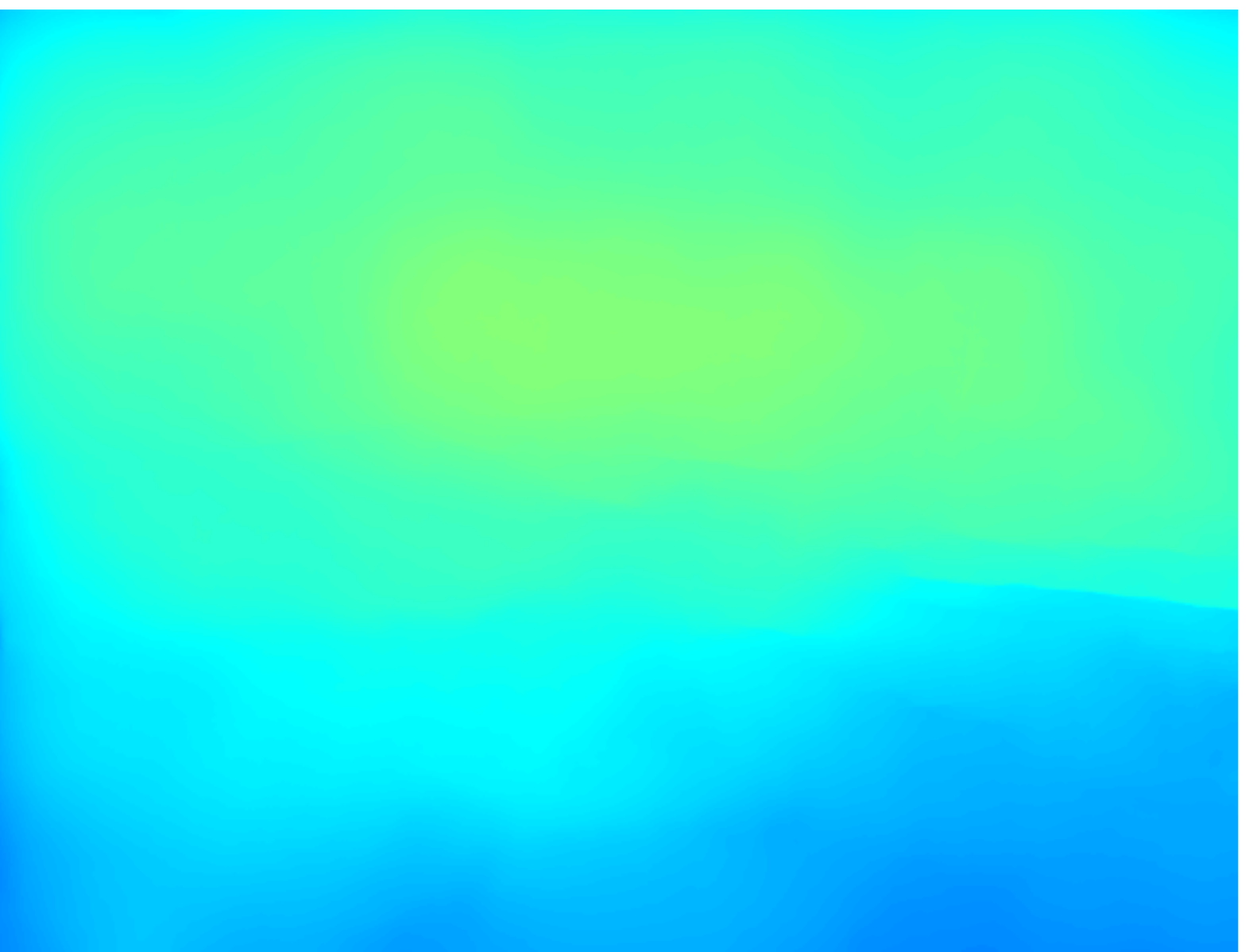} &
			\hspace{-0.37cm}\includegraphics[width=0.2\linewidth]{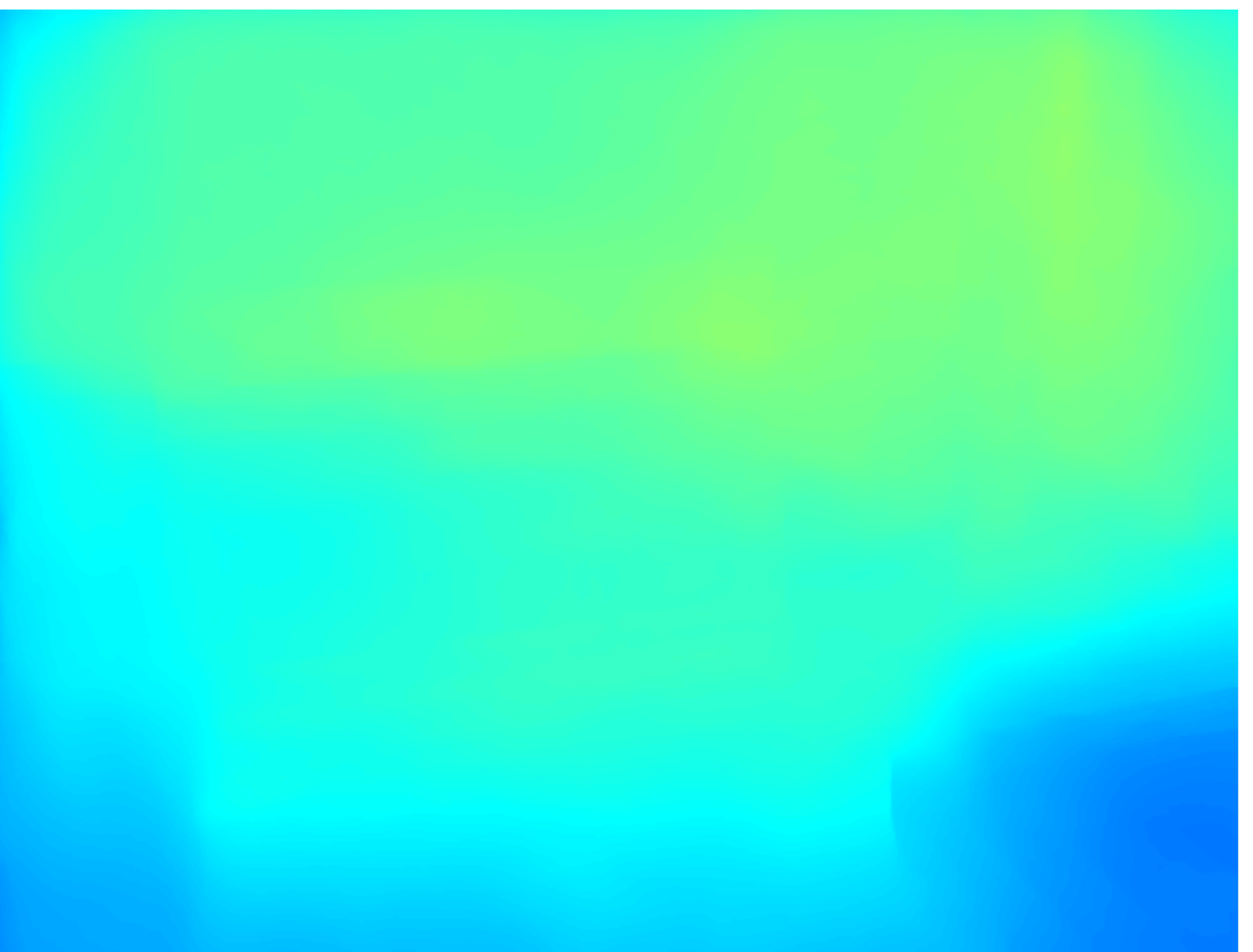} &
			\hspace{-0.37cm}\includegraphics[width=0.2\linewidth]{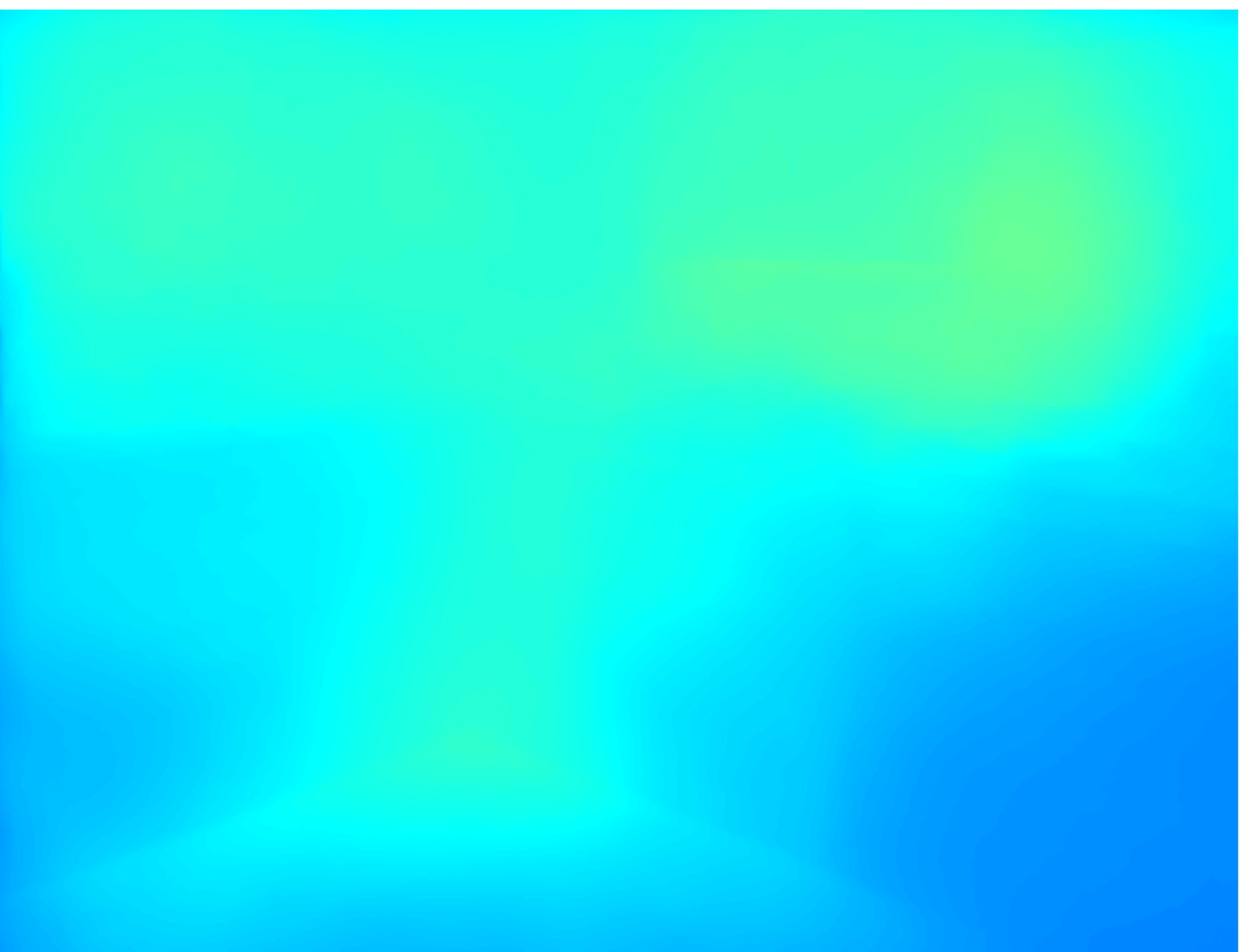} &
			\hspace{-0.37cm}\includegraphics[width=0.2\linewidth]{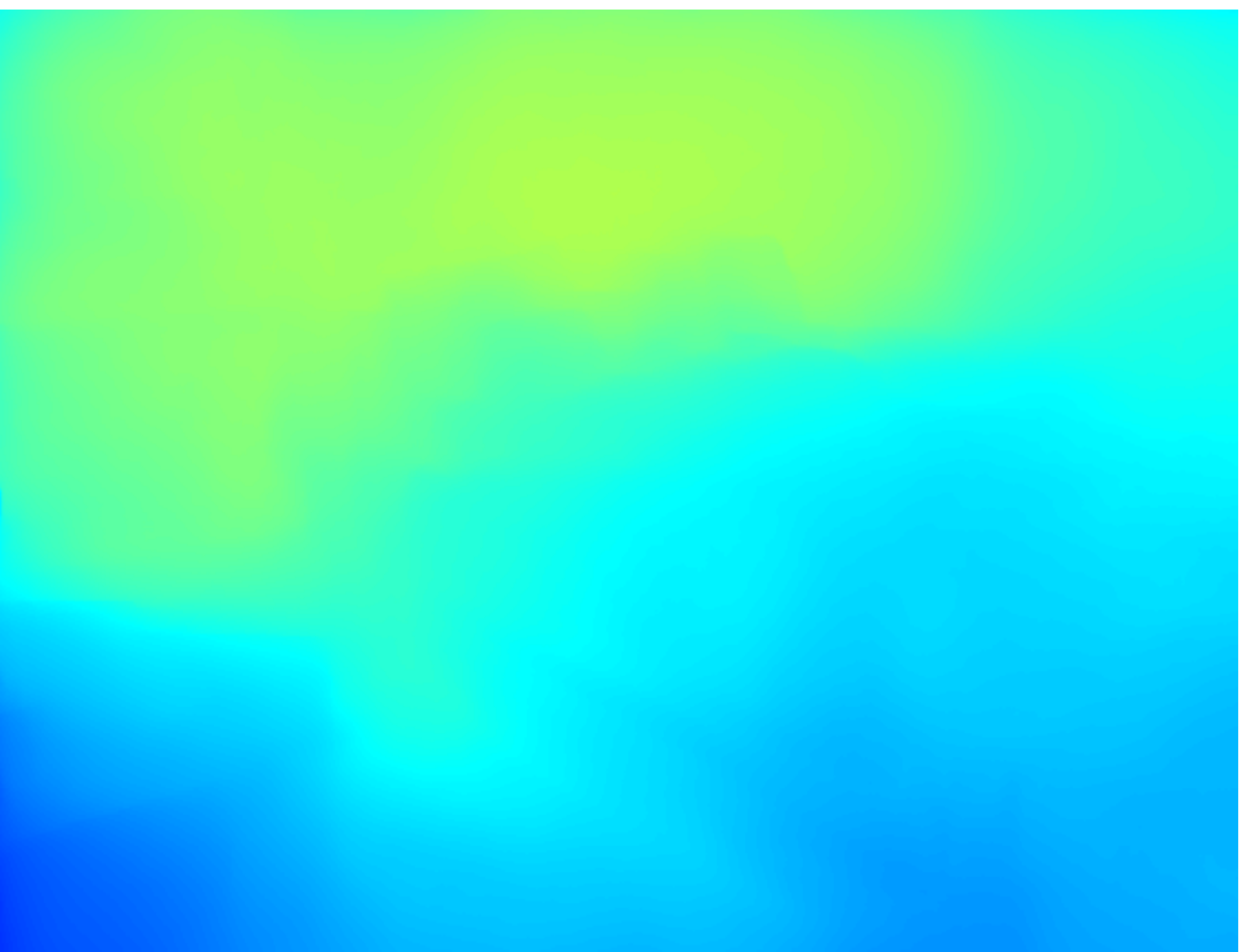} &
			\hspace{-0.37cm}\includegraphics[width=0.2\linewidth]{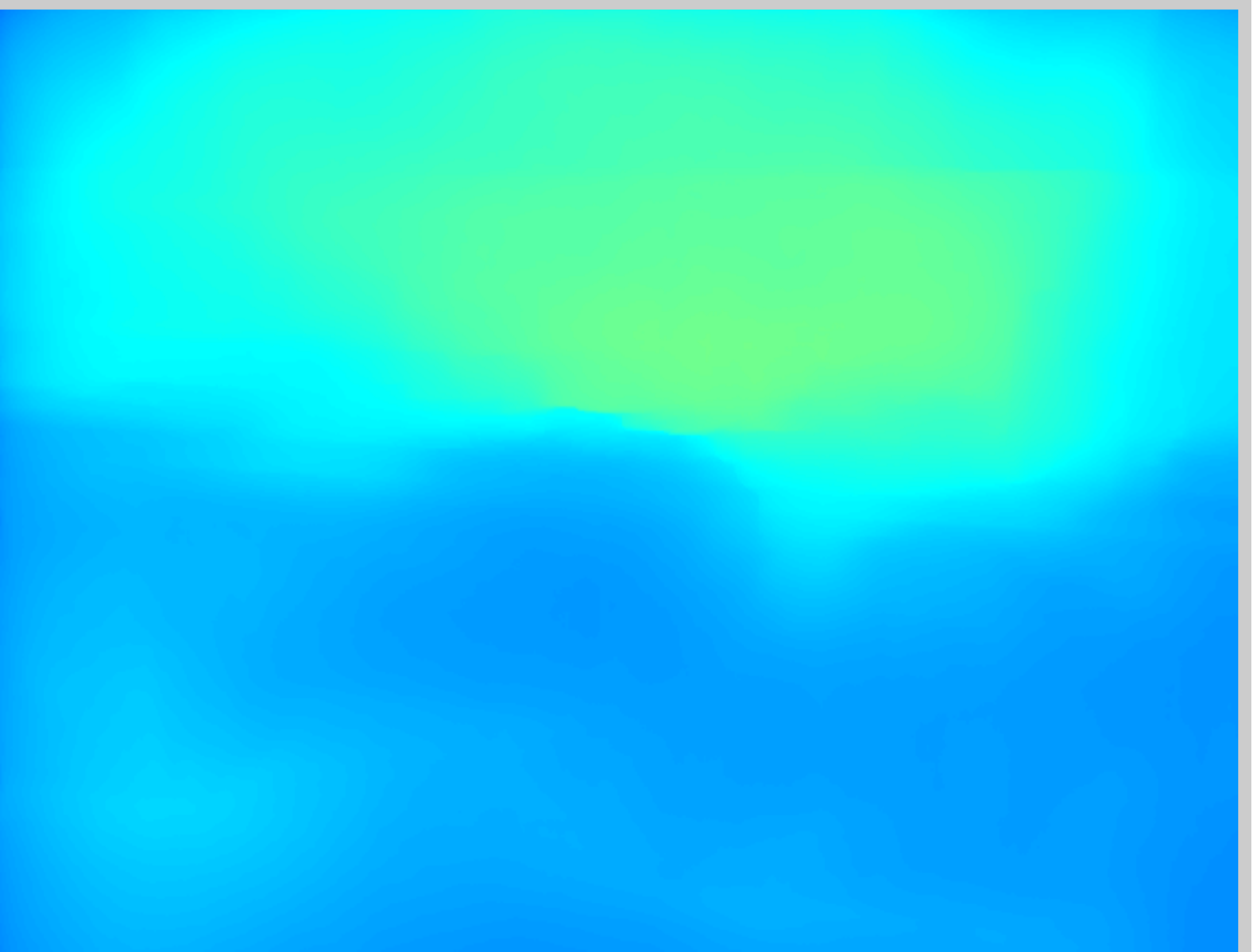} \\
			\hspace{-0.5cm}\begin{sideways}\hspace{0.7cm}{\bf Ours-1F}\end{sideways} &
			\hspace{-0.37cm}\includegraphics[width=0.2\linewidth]{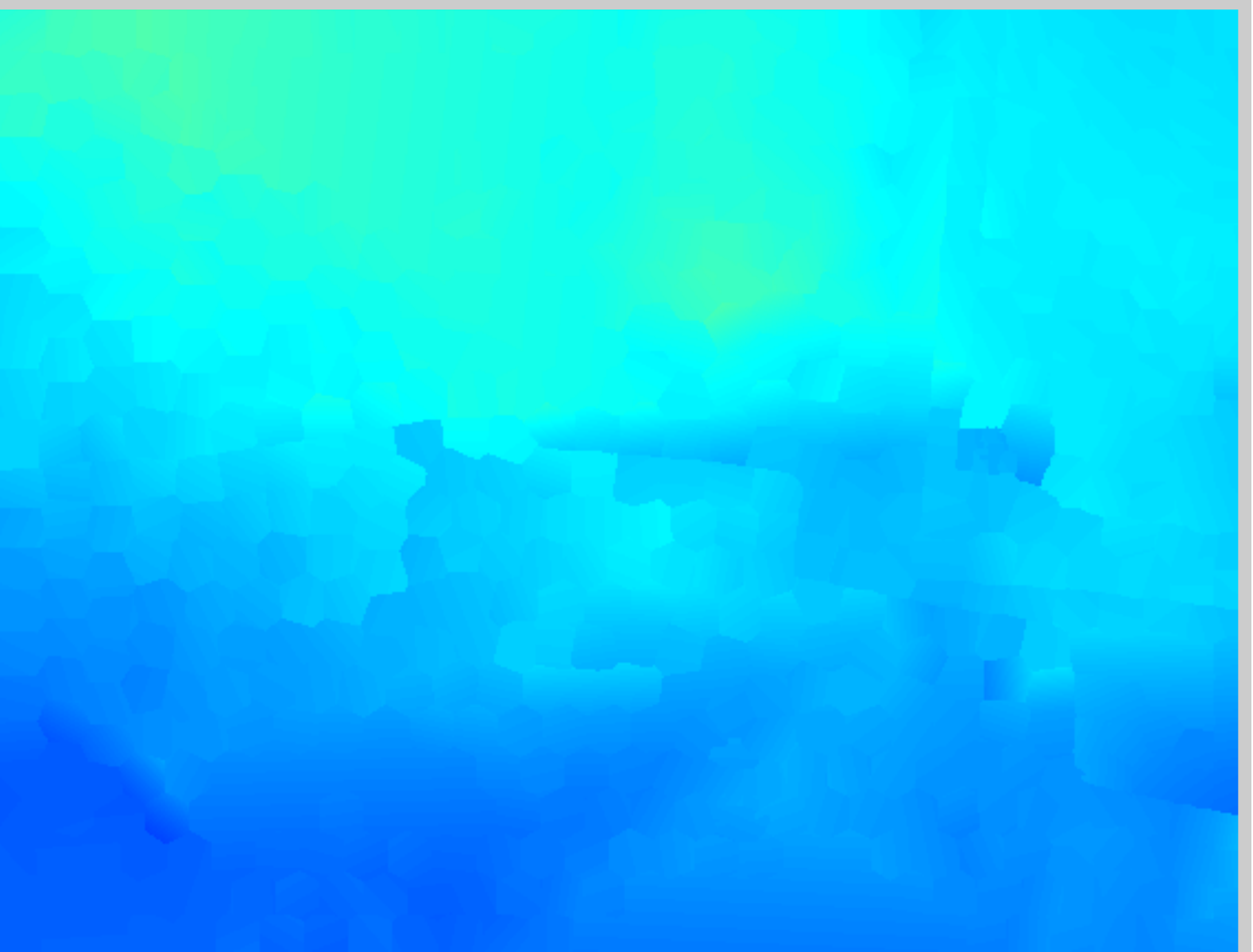} &
			\hspace{-0.37cm}\includegraphics[width=0.2\linewidth]{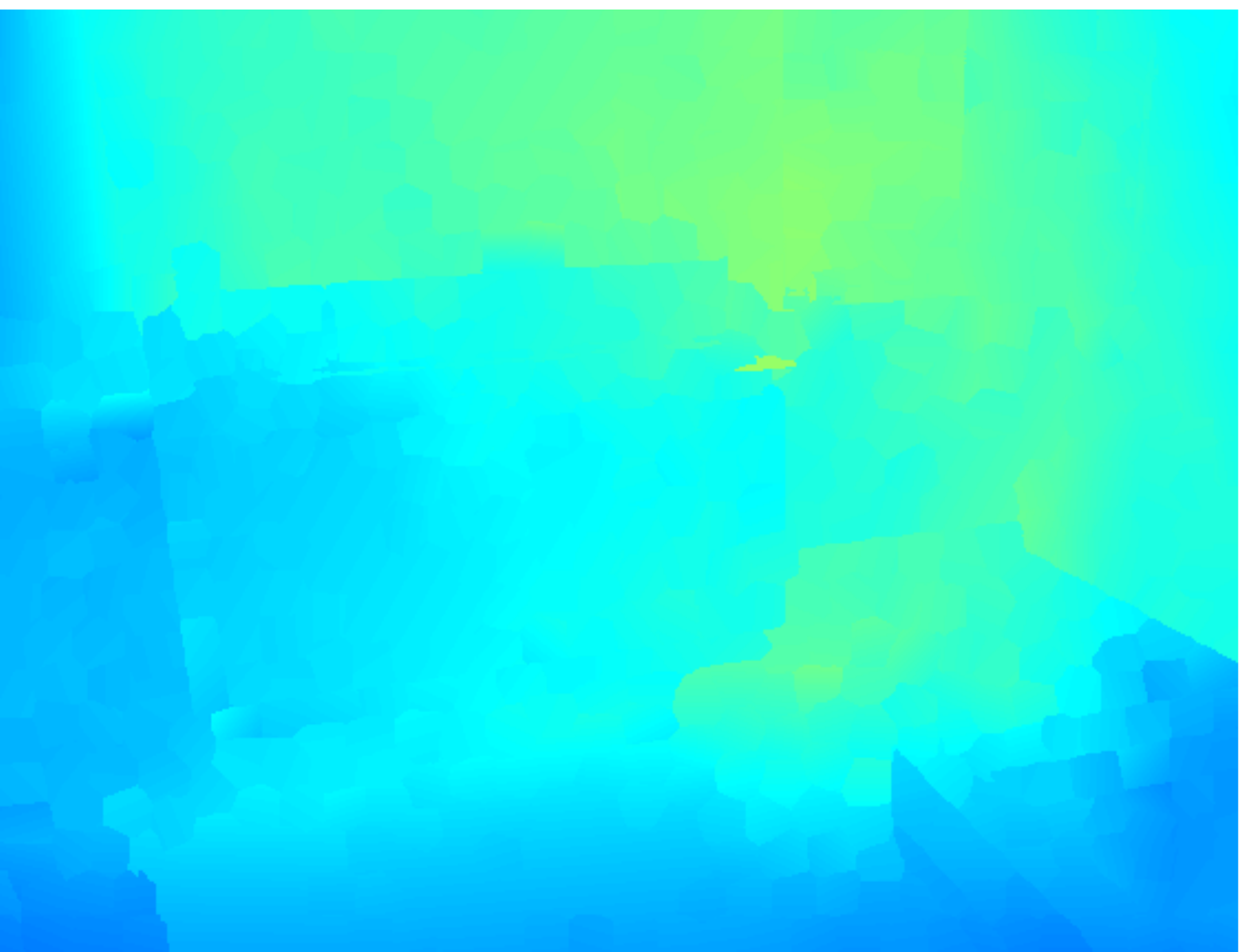} &
			\hspace{-0.37cm}\includegraphics[width=0.2\linewidth]{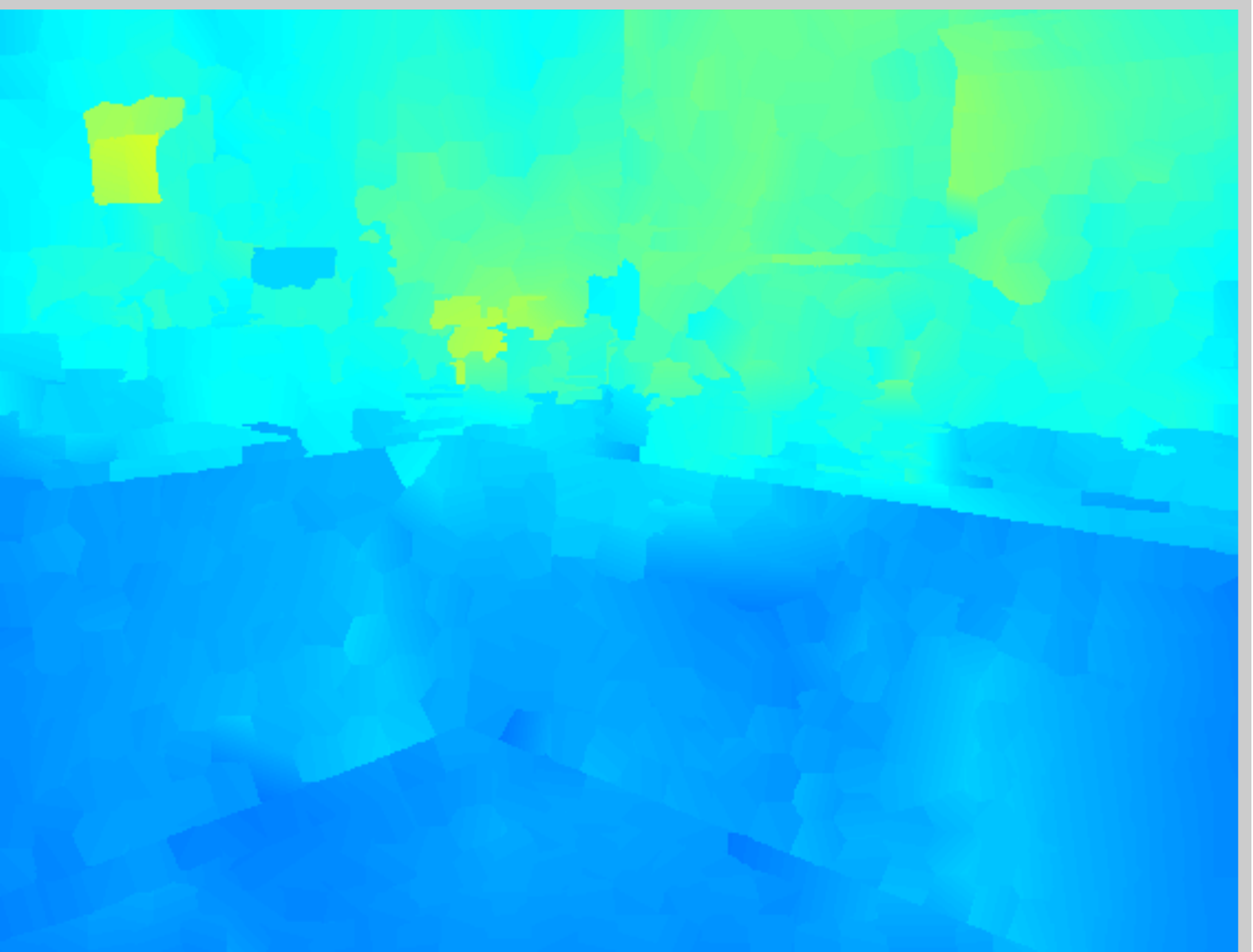} &
			\hspace{-0.37cm}\includegraphics[width=0.2\linewidth]{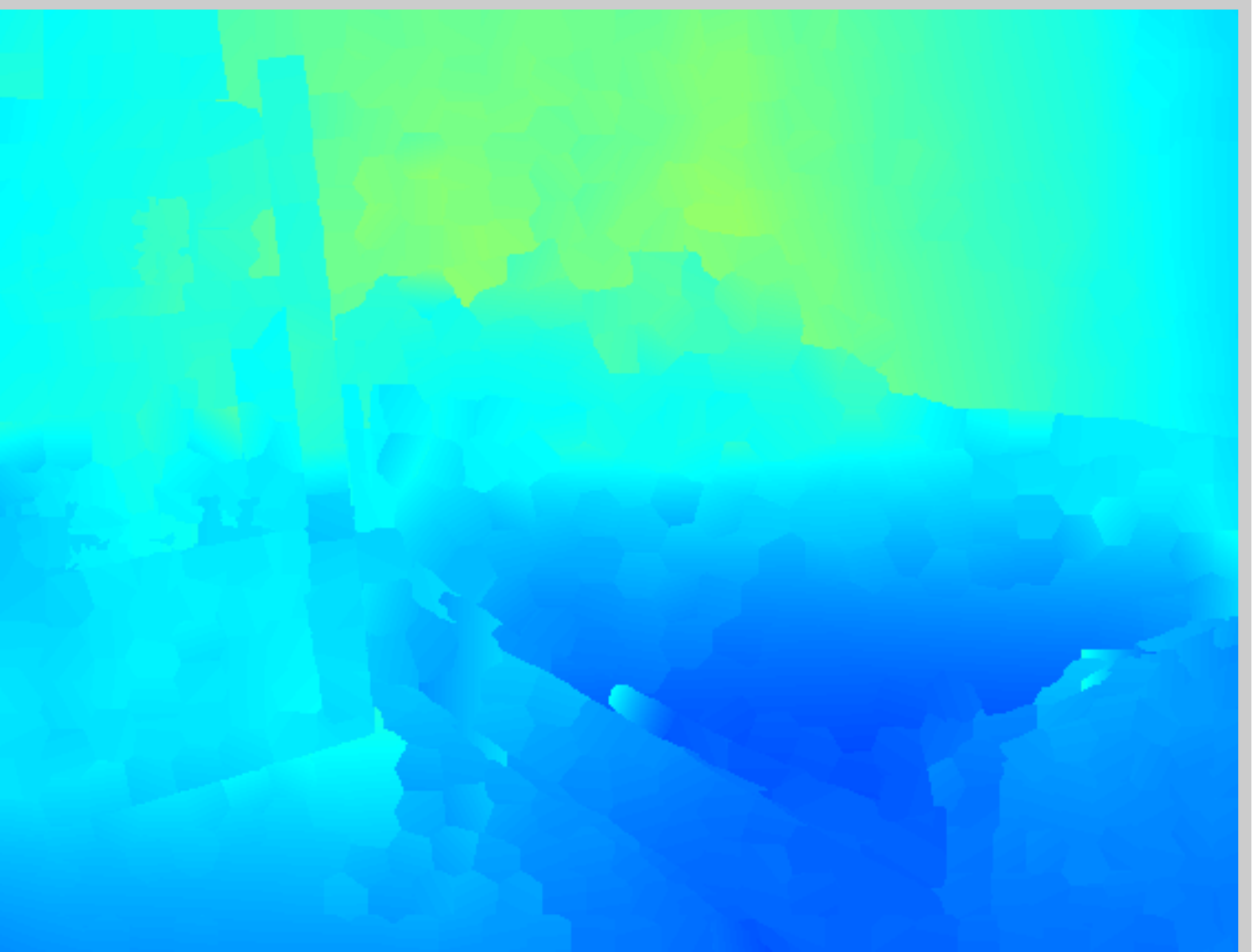} &
			\hspace{-0.37cm}\includegraphics[width=0.2\linewidth]{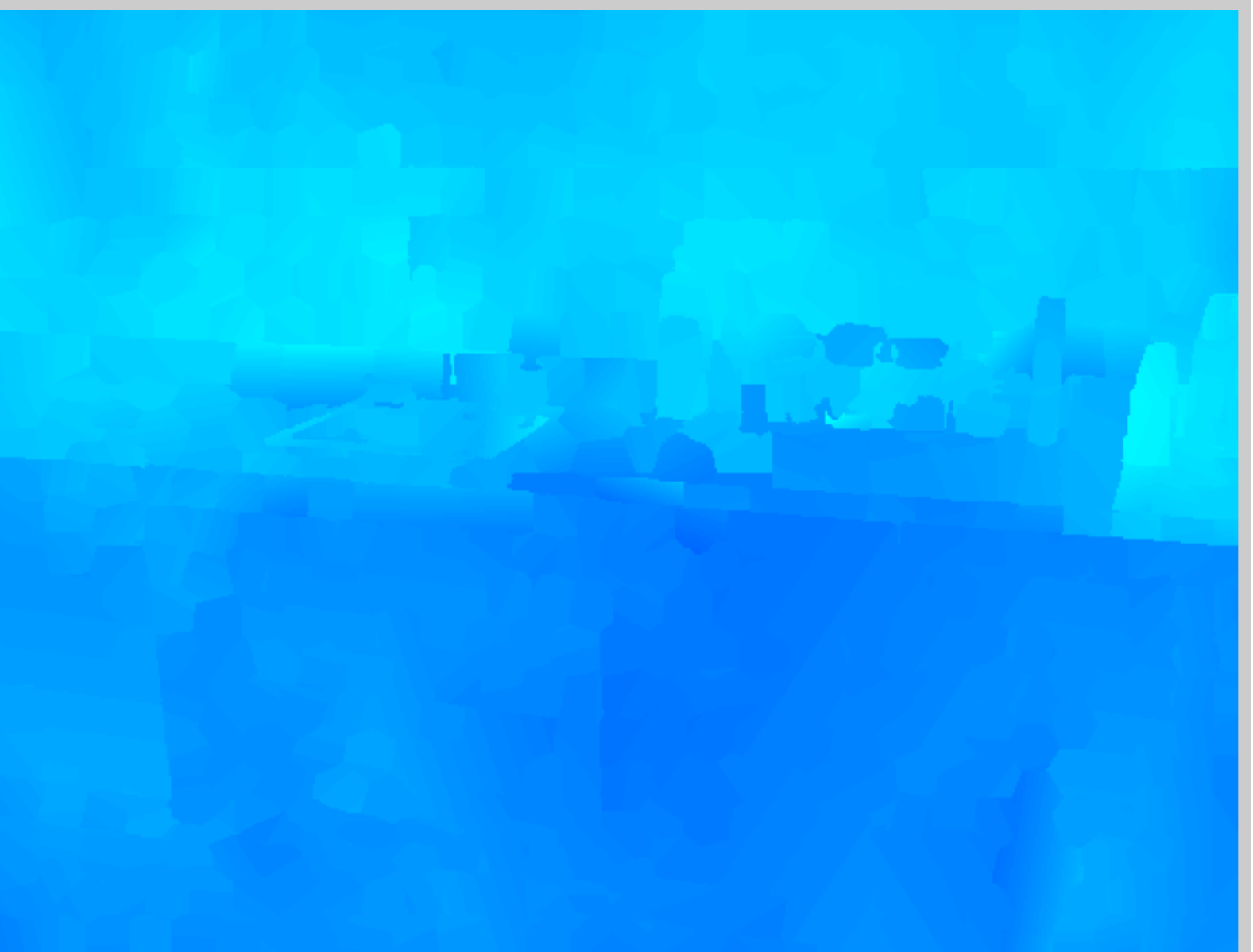} \\
		\end{tabular}
	\end{small}
	\vspace{-0.2cm}
	\caption{NYUv2: Qualitative comparison of the depths estimated with DepthTransfer~\cite{Karsch12} and with our method.}
	\label{fig:nyucomp}
\end{figure*}

\subsubsection{Indoor Scene Reconstruction: NYUv2}

The NYUv2 dataset contains 1449 images, partitioned into 795 training images and 654 test images. All the images were resized to 427$\times$561 pixels, while simultaneously respecting the masks provided with the dataset. In this case, the intrinsic camera parameters are given with the dataset. 
We evaluated the DepthTransfer code of~\cite{Karsch12} to obtain baseline results on the training/test partition provided with the dataset and compare these results with those obtained with our approach in Table~\ref{table:nyucomp}. To be able to directly compare our results with those reported in~\cite{Karsch14}, which also include the results of~\cite{KonradDCINE} and~\cite{KonradSPIE}, we also applied our method in a leave-one-out manner on the full dataset. These results are reported in Table~\ref{table:nyucomploout}. Note that, in both cases, we outperform the baselines on all metrics. These error metrics were computed over the valid pixels (non-zero depth) in the ground-truth depth maps. While better results were reported in~\cite{Eigen14}, this deep learning approach relied on a much larger training set. Note that their results could also be employed as a unary term in our model, which we expect to then further refine their depth estimate.

\begin{table}[t!]
	\centering
	\begin{tabular}{|c |c| c| c|}
		\hline
		Method & ${\bf rel}$ & ${\bf log}_{10}$ & ${\bf rms}$ \\ 
		\hline
		Depth transfer~\cite{Karsch12} & 0.374 & 0.134 & 1.12 \\ 
		Ours-1F  & {\bf 0.335} & {\bf 0.127} &{\bf 1.06} \\
		\hline
	\end{tabular}
	\vspace{0.2cm}
	\caption{NYUv2: Depth reconstruction errors for DepthTransfer~\cite{Karsch12} and for our method using the training/test partition provided with the dataset.}
	\label{table:nyucomp}
\end{table}

\begin{table}[t!]
	\centering
	\begin{tabular}{|c|c|c|c|}
		\hline
		Method & ${\bf rel}$ & ${\bf log}_{10}$ & ${\bf rms}$\\
		\hline
		Depth fusion (no warp)~\cite{KonradDCINE}&0.371 & 0.137 & 1.3\\
		Depth fusion~\cite{KonradSPIE} & 0.368 & 0.135 &1.3\\
		Depth transfer & 0.350 & 0.131 & 1.2 \\
		Ours-1F & {\bf 0.327} &{\bf 0.126} & {\bf 1.08}\\ 
		\hline
	\end{tabular}
	\vspace{0.2cm}
	\caption{NYUv2: Comparison of the depth estimation errors using a leave-one-out strategy.}
	\label{table:nyucomploout}
\end{table}

\begin{table}[t!]
	\centering
	\begin{tabular}{|c |c| c| c|}
		\hline
		Method & ${\bf rel}$ & ${\bf log}_{10}$ & ${\bf rms}$ \\ 
		\hline
		Unary & 0.350 & 0.132 & 1.11\\
		GP regression & 0.431 & 0.151 & 1.21 \\ 
		No discrete variables & 0.354 & 0.141 & 1.20\\
		No sampling  & 0.339 &  0.129 & 1.08\\
		Ours-1F  & 0.335 &  0.127 & 1.06\\
		\hline
	\end{tabular}
	\vspace{0.2cm}
	\caption{NYUv2: Comparison of our final results with those obtained with unary terms only, with our GP depth regressors only, using a model without discrete edge type variables, and after the first round of PCBP where no sampling is involved.}
	\label{table:nyu_terms}
\end{table}

\begin{figure*}[htp] 
	\begin{tabular}{ccccc}
		\hspace{-0.1cm}\includegraphics[width=0.2\linewidth]{./figures14/img_set_1_indoor/orgDepth_413} &
		\hspace{-0.3cm}\includegraphics[width=0.2\linewidth]{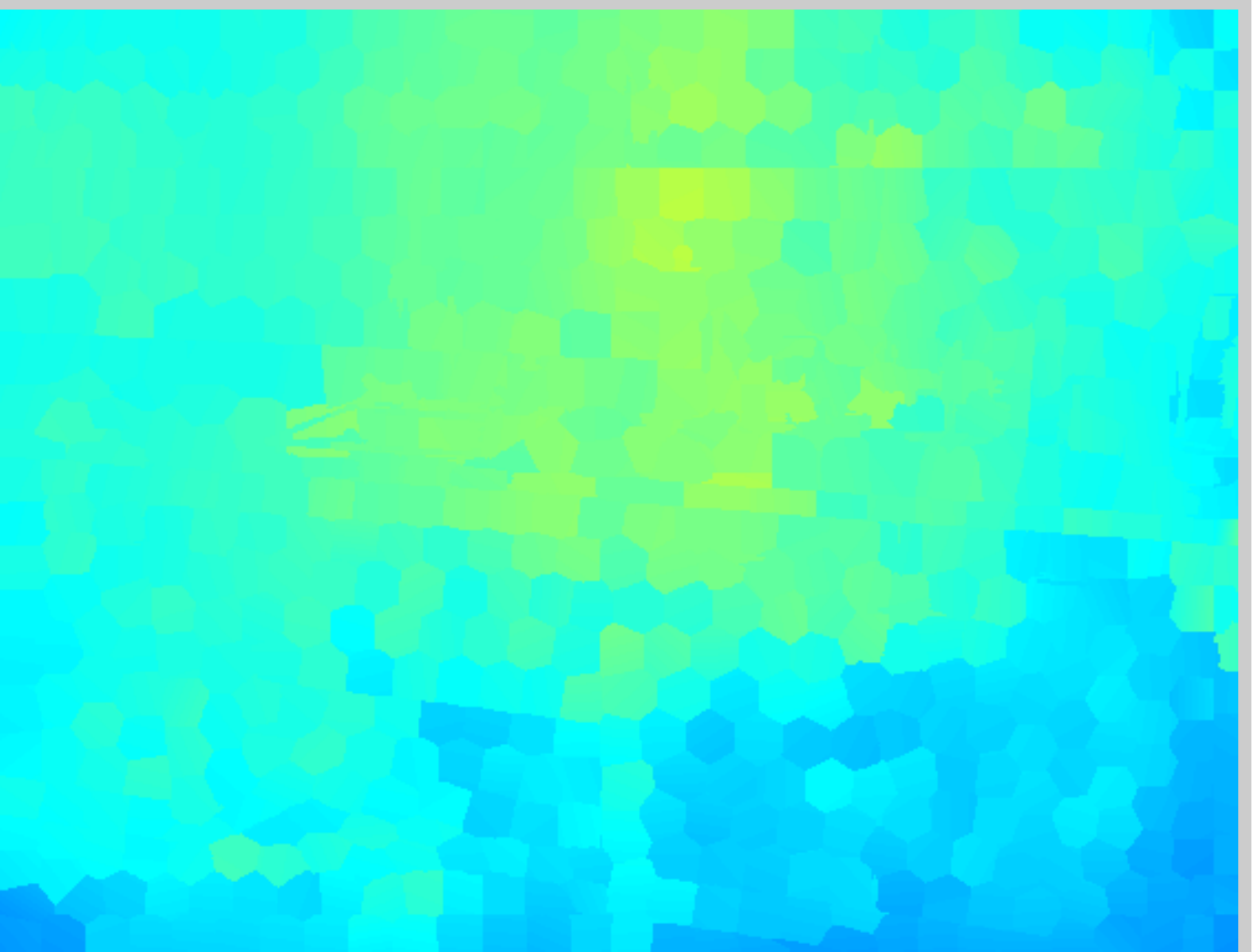} &
		\hspace{-0.3cm}\includegraphics[width=0.2\linewidth]{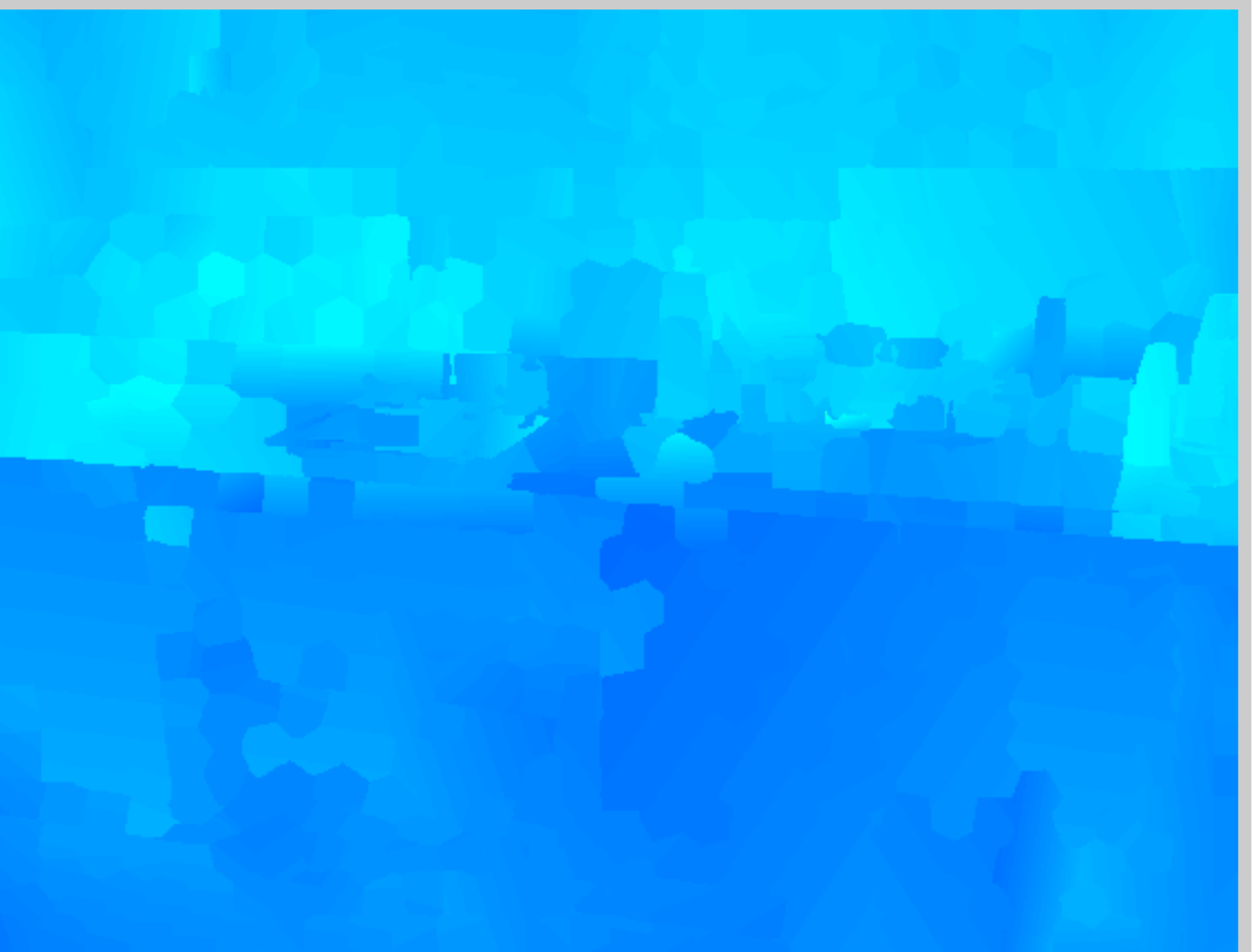} &
		\hspace{-0.3cm}\includegraphics[width=0.2\linewidth]{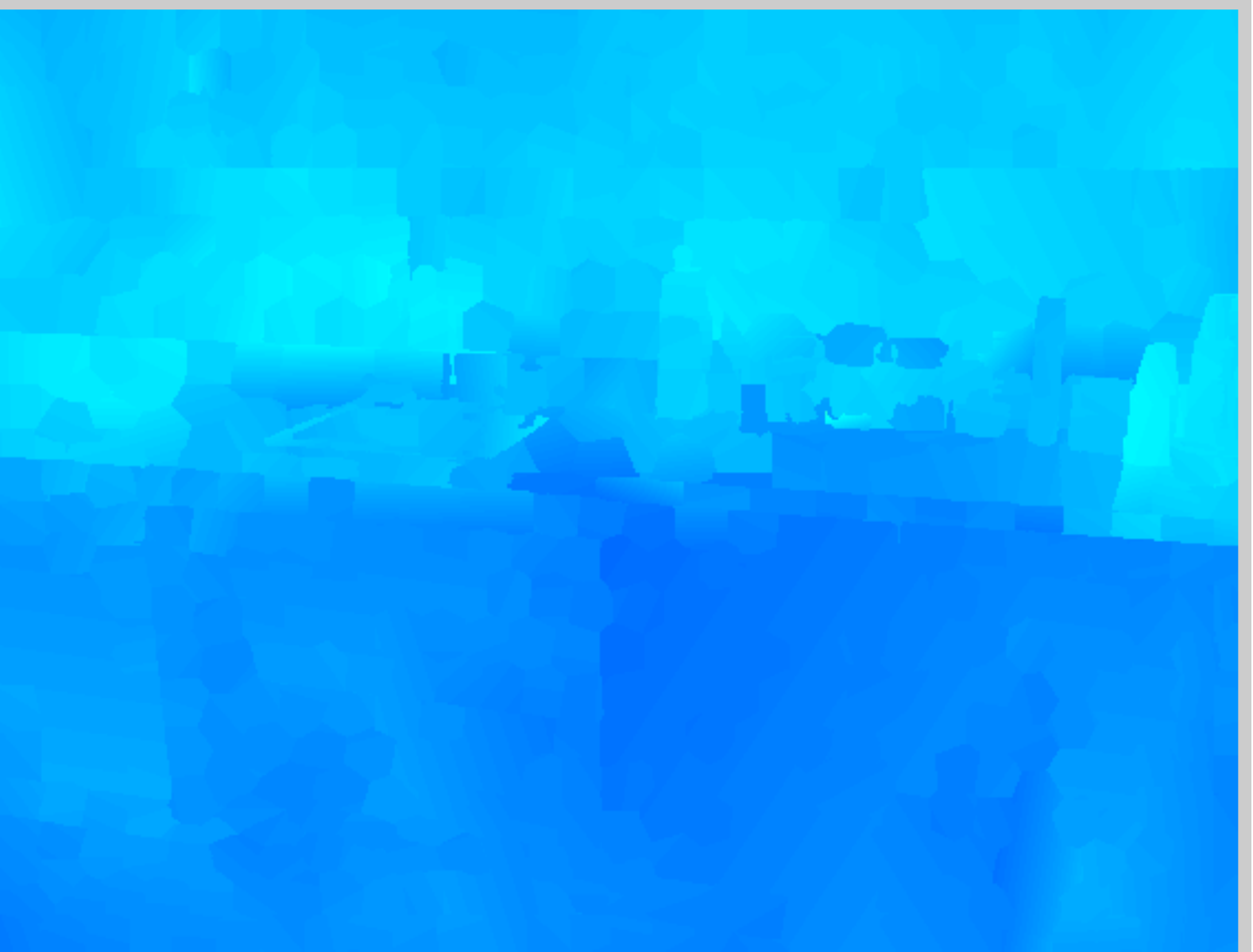} &
		\hspace{-0.3cm}\includegraphics[width=0.2\linewidth]{./figures14/img_set_1_indoor/oursEstDepth_413_samplingRound_2} \\
		\hspace{-0.1cm} Ground-truth & \hspace{-0.3cm} Regression & \hspace{-0.3cm} No sampling & \hspace{-0.3cm} Sampling round 1 & \hspace{-0.3cm}Final depth
	\end{tabular}
	\vspace{-0.2cm}
	\caption{NYUv2: Depth maps at different stages of our approach.}
	\label{fig:nyu_evolution}
\end{figure*}

In Fig.~\ref{fig:nyucomp}, we provide a qualitative comparison of our results with those of~\cite{Karsch12} for some images. Note that the over-smoothing of the depth maps generated by depth transfer is more obvious in the short depth range scenario. In contrast, our approach still yields a realistic representation of the scene.
In Table~\ref{table:nyu_terms}, we show the influence of the different parts of our model. Note that all the components contribute to our final results. Fig.~\ref{fig:nyu_evolution} depicts the depth maps at different stages of our approach. While sampling smoothes the depth map, it still respects the image discontinuities. 

\begin{figure}[t!] 
	\centering
	\begin{tabular}{cc}
		\hspace{-0.0cm}\includegraphics[height=0.4\linewidth]{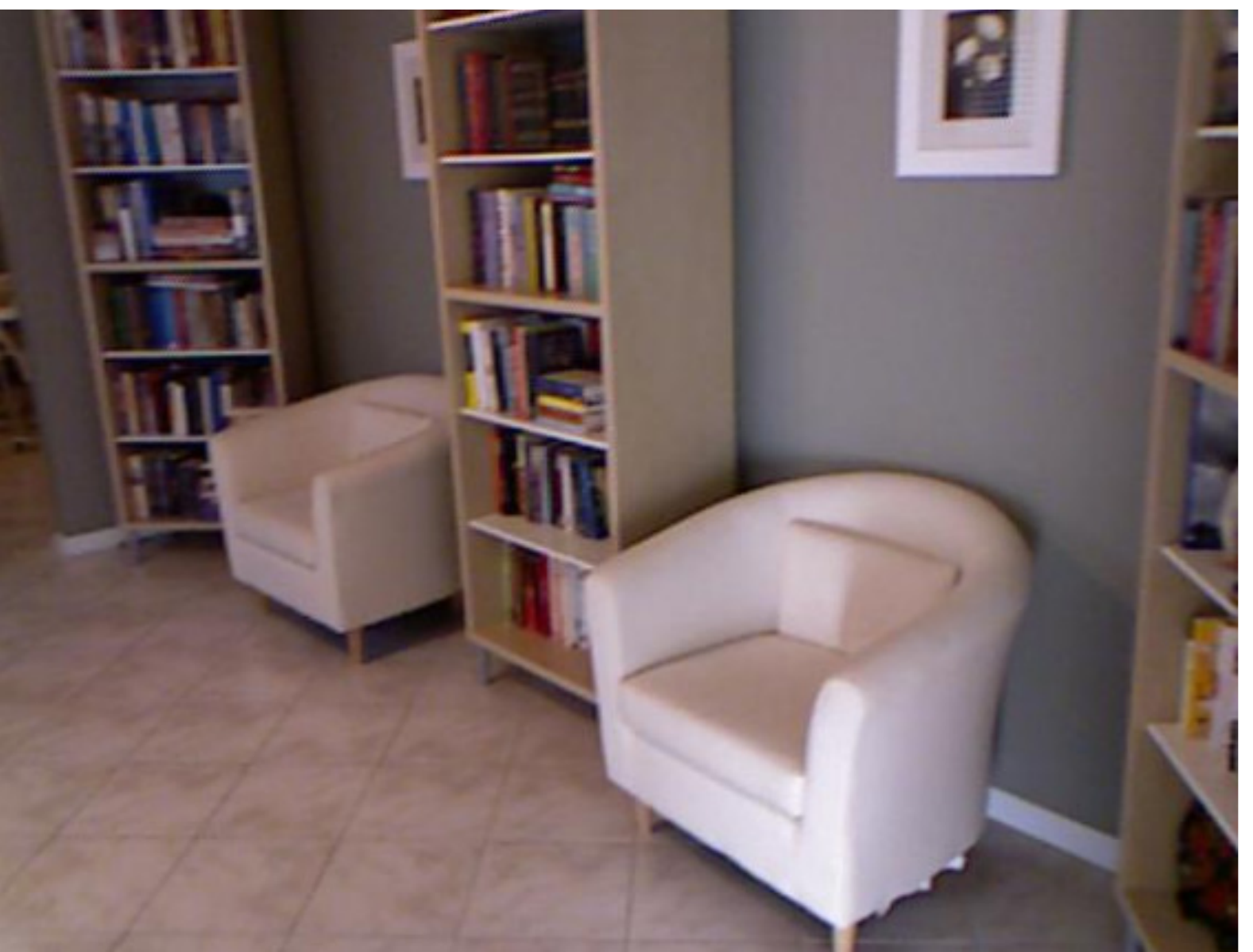} &
		\hspace{-0.0cm}\includegraphics[height=0.4\linewidth]{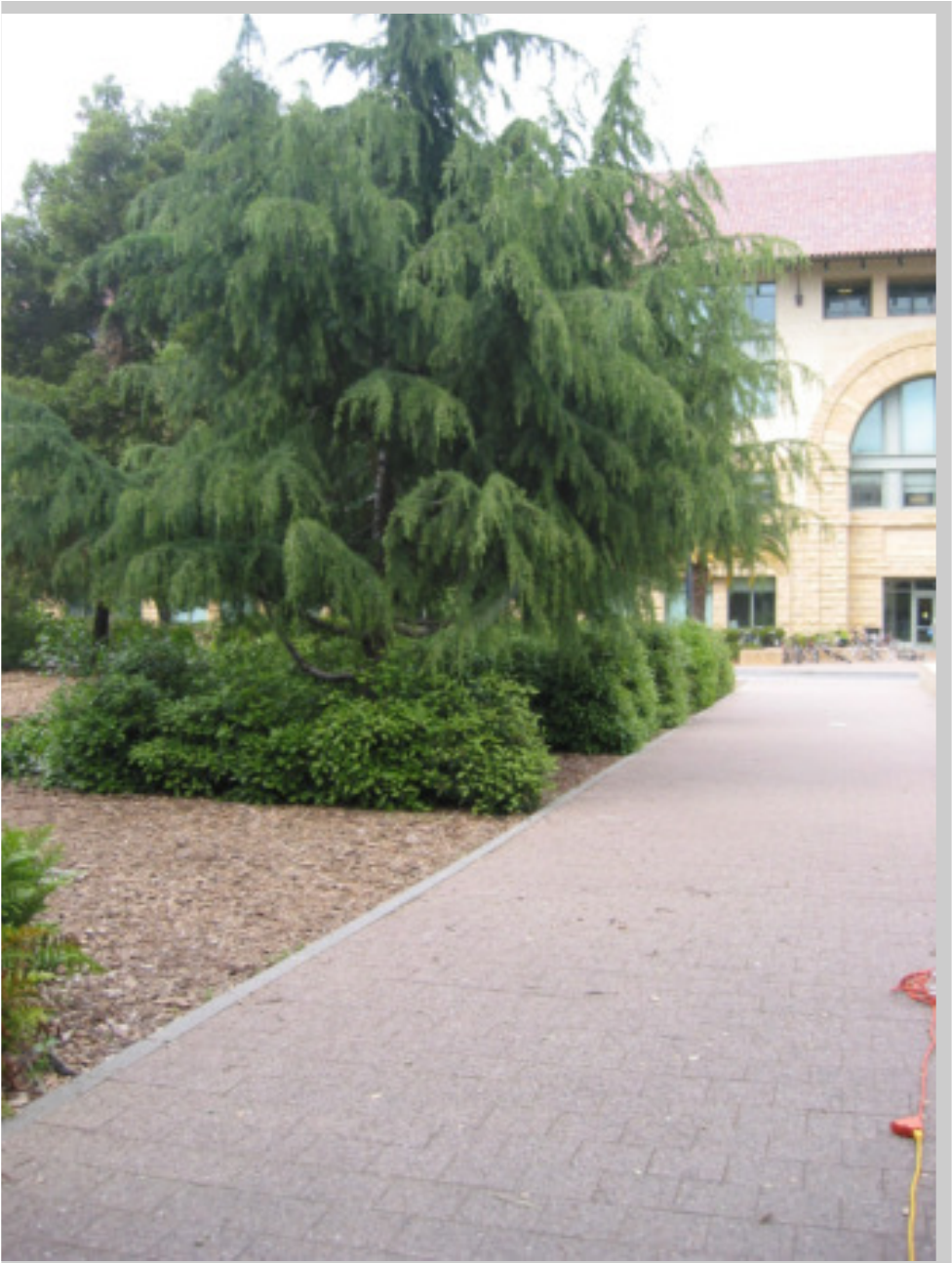} \\
		\hspace{-0.0cm}\includegraphics[height=0.4\linewidth]{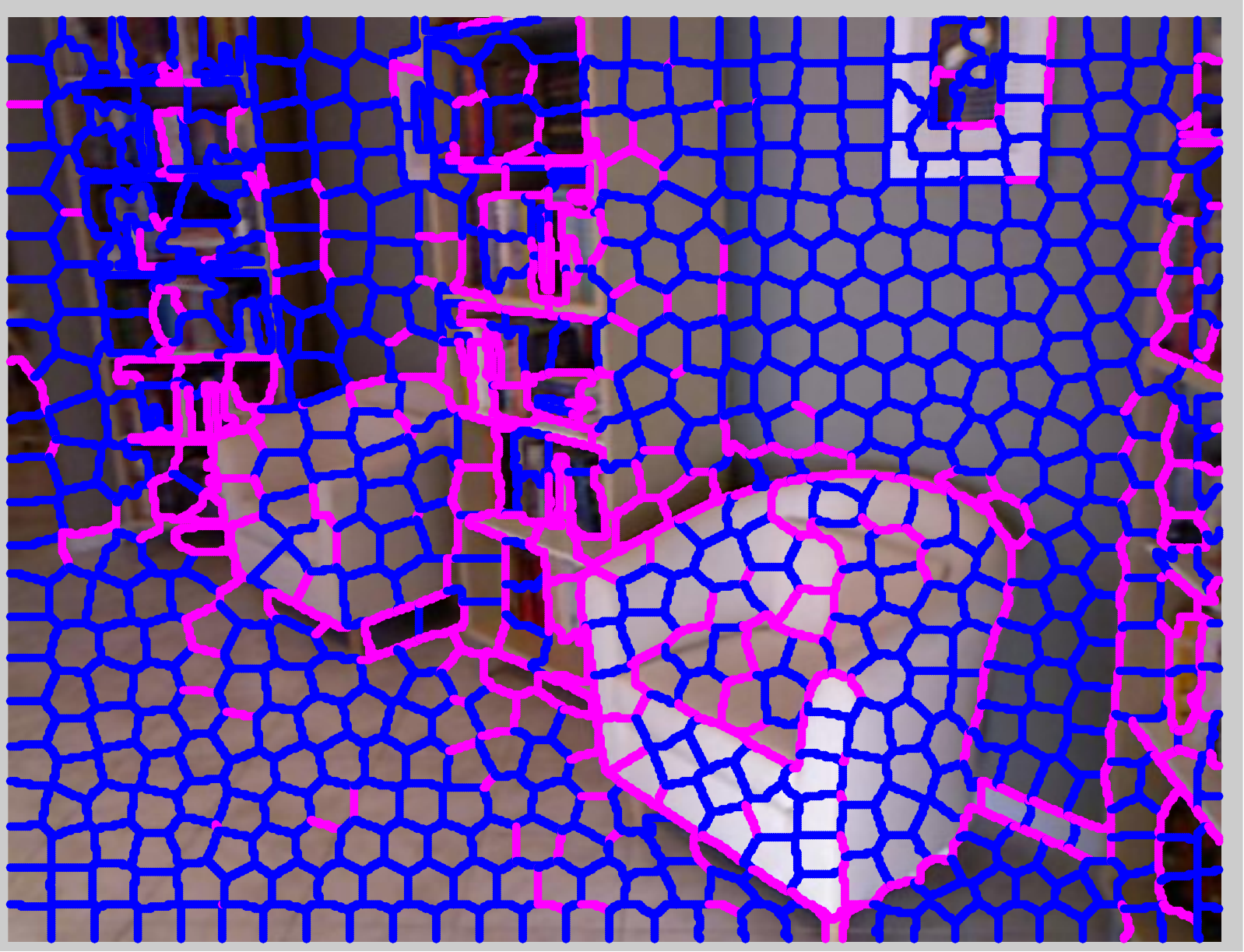} &
		\hspace{-0.0cm}\includegraphics[height=0.4\linewidth]{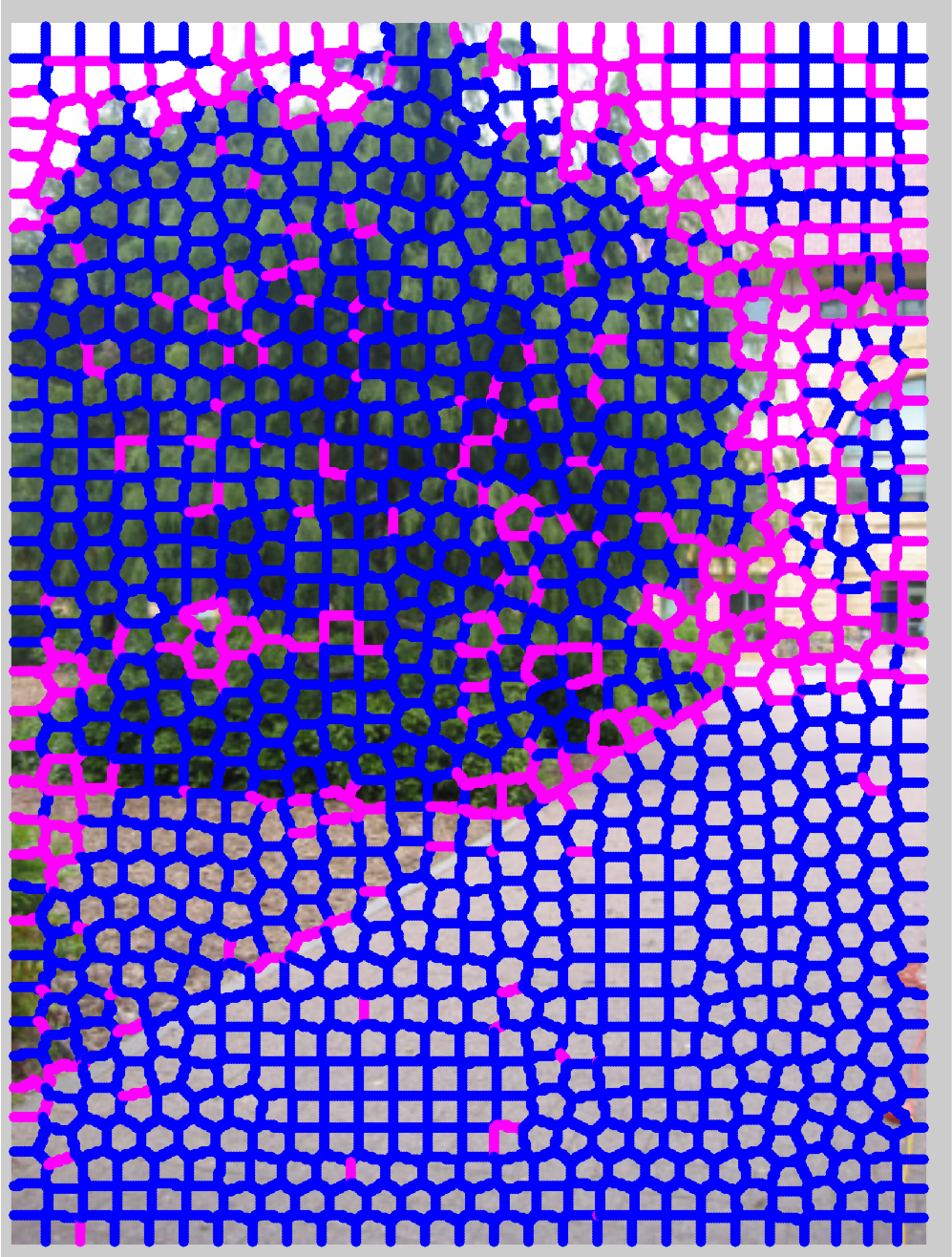} 
	\end{tabular}
	\vspace{-0.2cm}
	\caption{Estimated boundary occlusion map. The top row shows the input image and the bottom row shows the estimated boundary occlusion map. The superpixel boundaries are drawn in blue. Pixels in magenta denote the estimated occlusion boundaries.}
	\label{fig:occlusion}
\end{figure}

In addition to the estimated depth, our model can also predict the boundary type of the superpixel edges. In particular, the occlusion boundaries are useful cues for spatial reasoning. We qualitatively evaluate the occlusion boundary prediction by showing typical results in Fig.~\ref{fig:occlusion} for both indoor and outdoor scenarios. Note that our model captures most of the occlusion edges. 

\subsection{Evaluation of Video-Based Depth Estimation}
We then evaluate our video-based depth estimation method on monocular sequences containing moving foreground objects and acquired by non-translational cameras. To this end, we made use of the publicly available dataset MSR-V3D.
This dataset contains ground truth depth and image sequences for four different buildings, which are referred to as Buildings $1$, $2$, $3$ and $4$. Note that each building includes multiple sequences.  In the remainder of this section, we first compare the results of our final model (Ours-Vid) with those of~\cite{Karsch12}, which constitutes the only baseline that also tackles the video scenario. We then perform an ablation study and evaluate the influence of different parts of our model by comparing the results of our final model with those of our two-frame CRF (Ours-2F) and of our two-frame CRF without distinguishing between moving foreground and background (i.e., without moving foreground classifier) (Ours-2F-NM). In this comparison, we also show the results of our single image depth estimation method (Ours-1F).

In our experiments, we used SLIC~\cite{Achanta12} to compute superpixels. We employed the same training-test partition as~\cite{Karsch12}, i.e., we used Building $1$ as training data. For each test image, we retrieved $K=7$ image-depth candidate pairs from the training set, each of which is chosen from one different video sequence to obtain a broader variety. In practice, to speed up the approach, we applied our fully-connected CRF on overlapping blocks of consecutive frames. The parameters of our model were obtained using a validation set selected from Building $1$ (i.e., from the training data). We validated the parameters using three frames of each video sequence in the validation set, i.e. 264 images in total, and by incrementally adding terms to our energy. The parameters were found to be $w_p= \theta_r =\ w_{re} =1$, $\theta_t=w_d=w_{ye}=w_{te}=10$, $w_a = 1000$, $\theta_e=100$, and $\theta_{max}=\theta_m=20$, $\epsilon_r=\epsilon_t=0.00001$ for the two-frame model, and $\theta_{\alpha}=10$, $\theta_q = 5$, $\theta_{\beta}=30$, $\theta_{\gamma}=3$, $\omega^{(1)}=25$, $\omega^{(2)}=5$ for the fully-connected CRF. While this may seem many parameters, note that these values did not need to be very precisely tuned (mostly the order of magnitude) thanks to the robustness of our model.

\begin{figure*}[t!]
	\begin{small}
		\begin{tabular}{ccccccccc}
			&& \hspace{-2.5cm}{\bf Building 2} & & \hspace{-2.5cm}{\bf Building 3} & & \hspace{-2.5cm}{\bf Building 3} & & \hspace{-2.5cm}{\bf Building 4} \\
			\hspace{-0.5cm}\begin{sideways}\hspace{0.8cm}{\bf Image}\end{sideways} & 
			\hspace{-0.35cm}\includegraphics[width=0.12\linewidth]{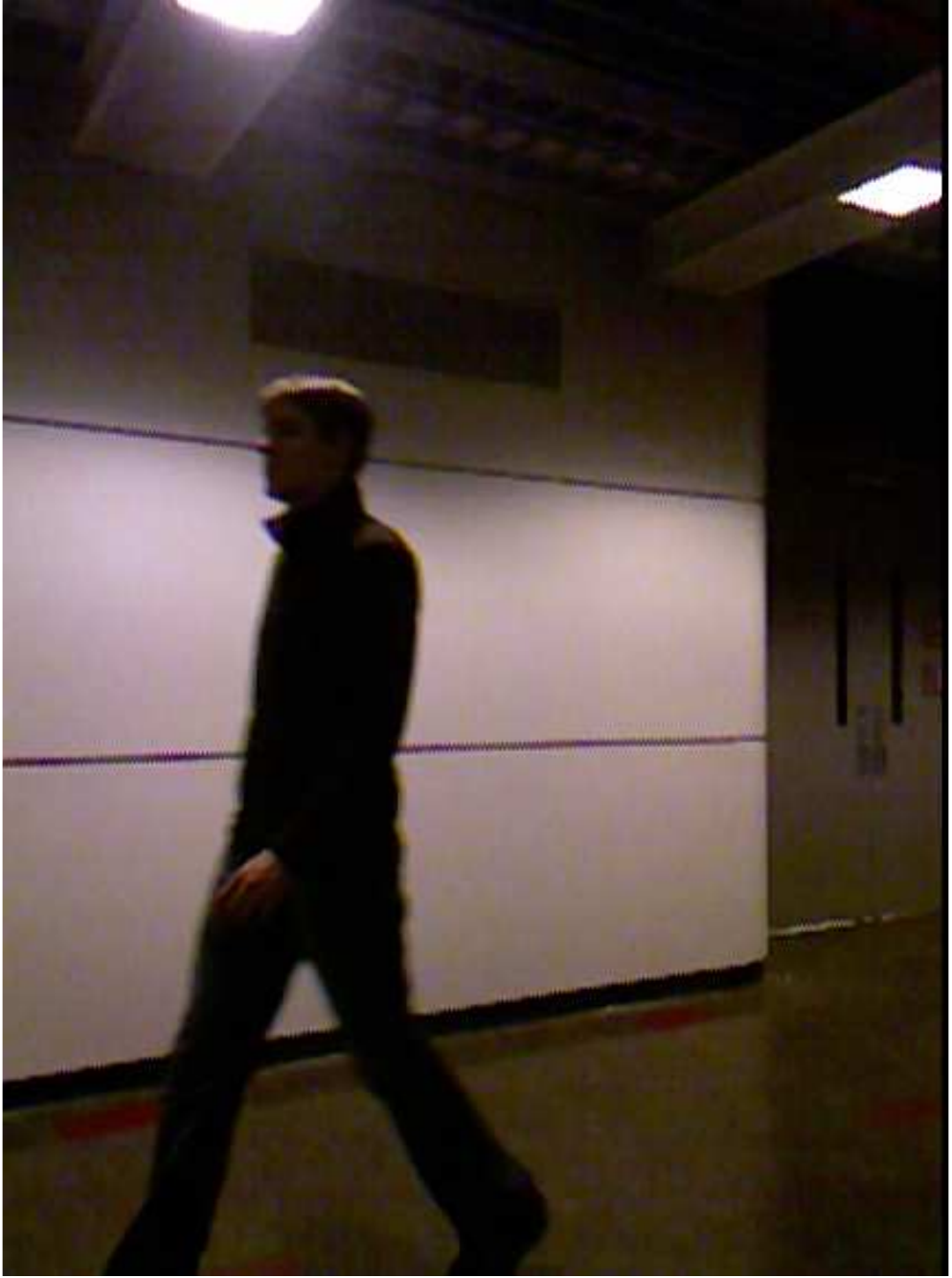} &
			\hspace{-0.35cm}\includegraphics[width=0.12\linewidth]{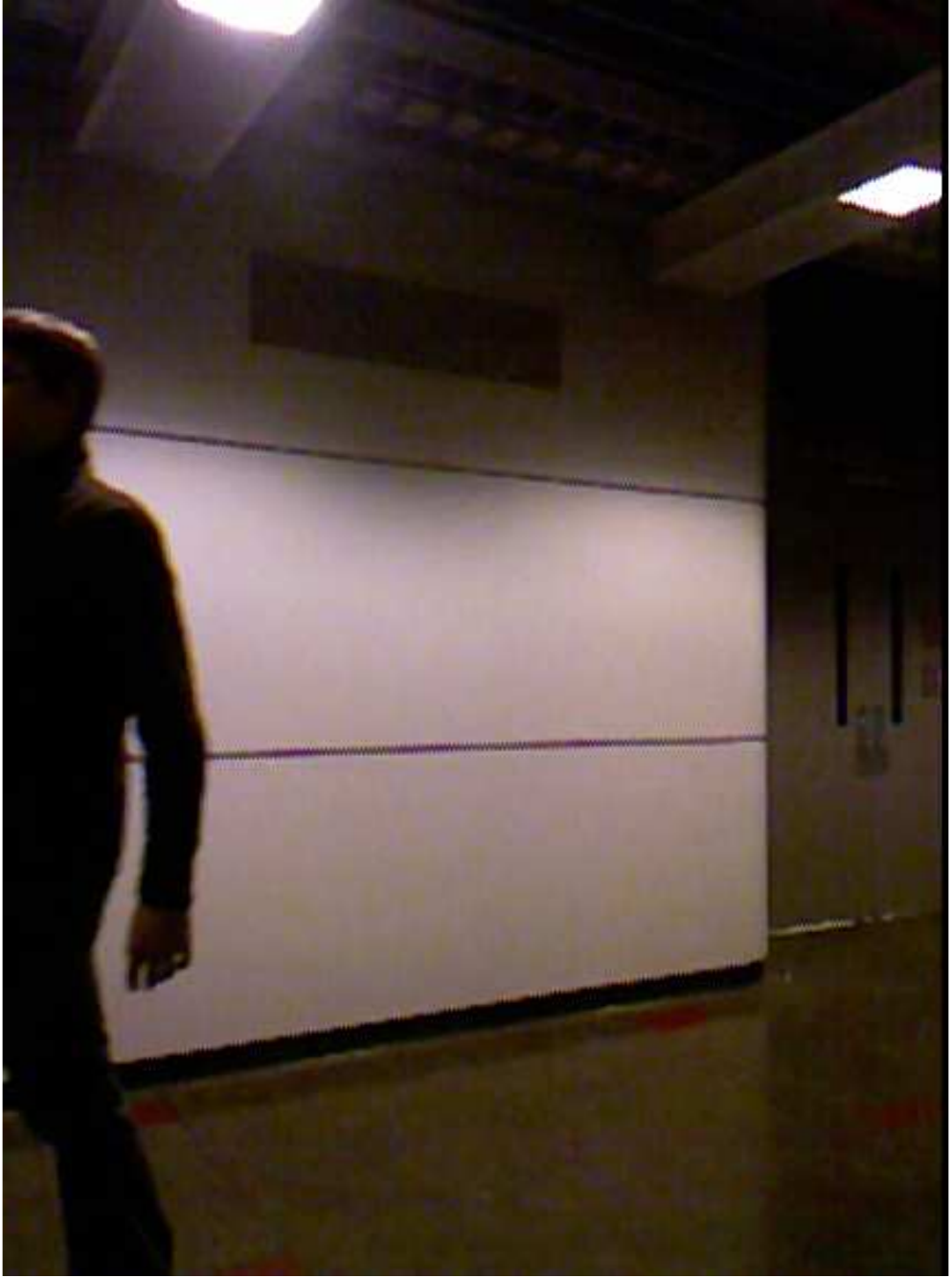} &
			\hspace{-0.35cm}\includegraphics[width=0.12\linewidth]{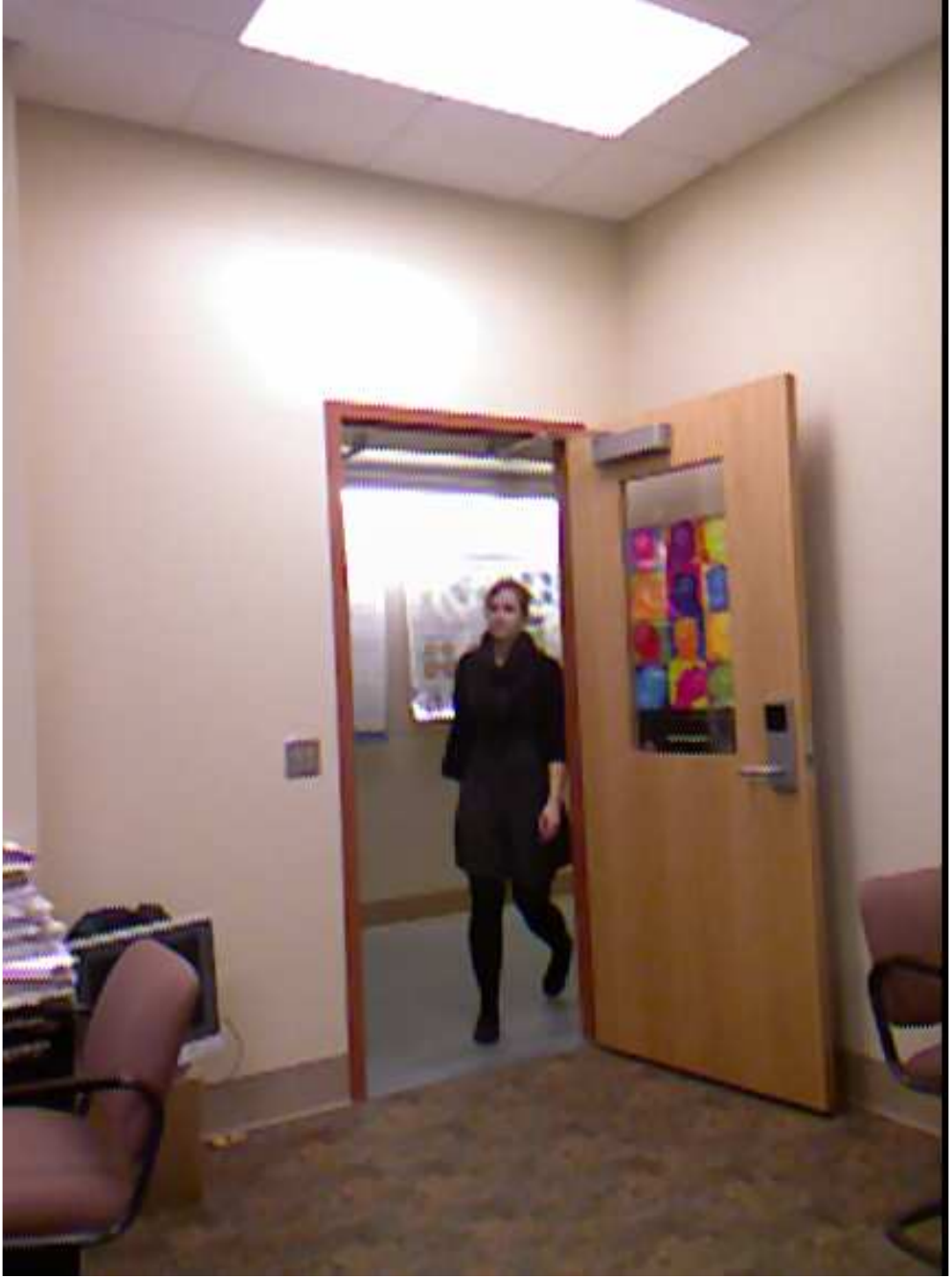} &
			\hspace{-0.35cm}\includegraphics[width=0.12\linewidth]{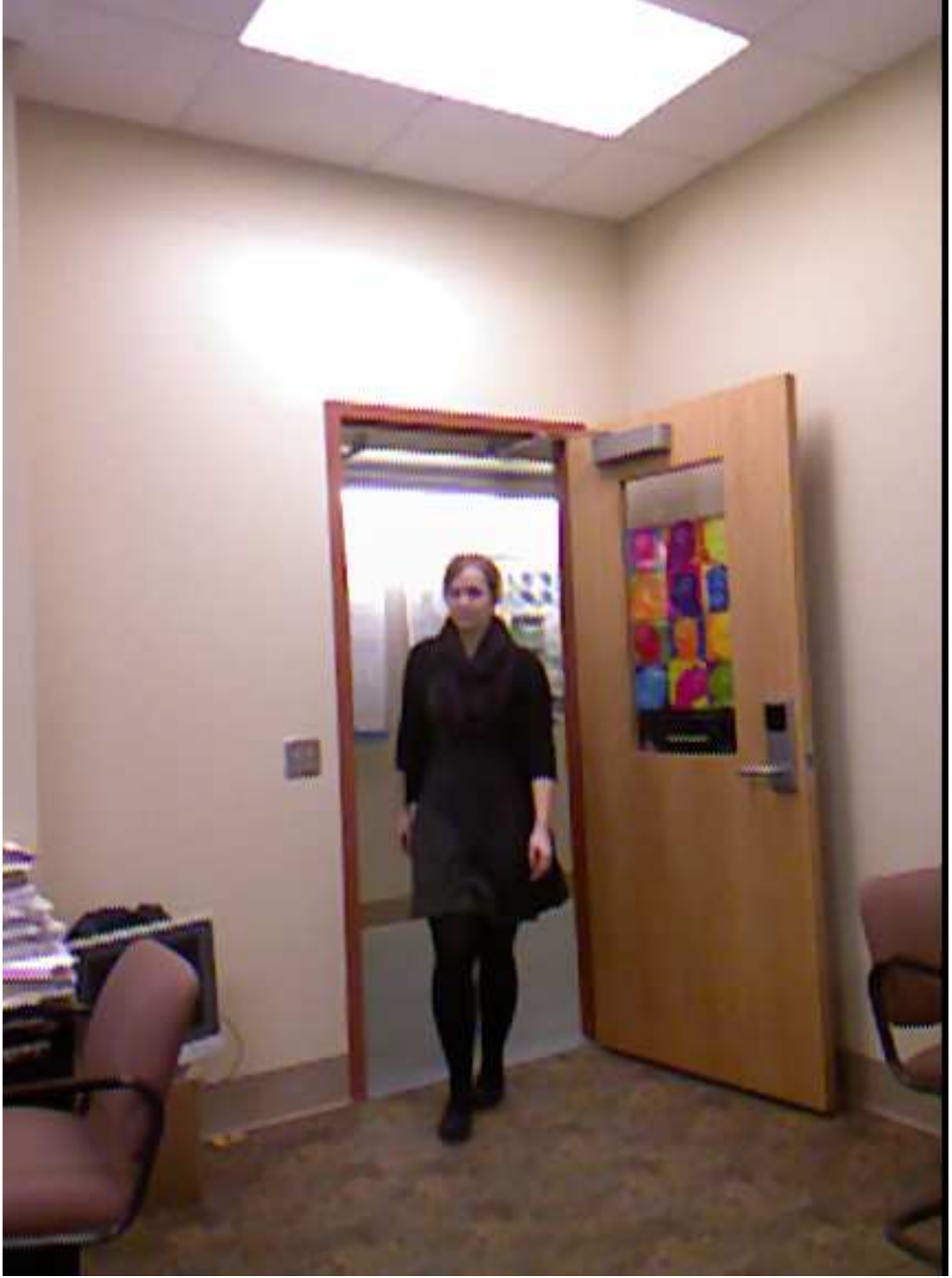} &
			\hspace{-0.35cm}\includegraphics[width=0.12\linewidth]{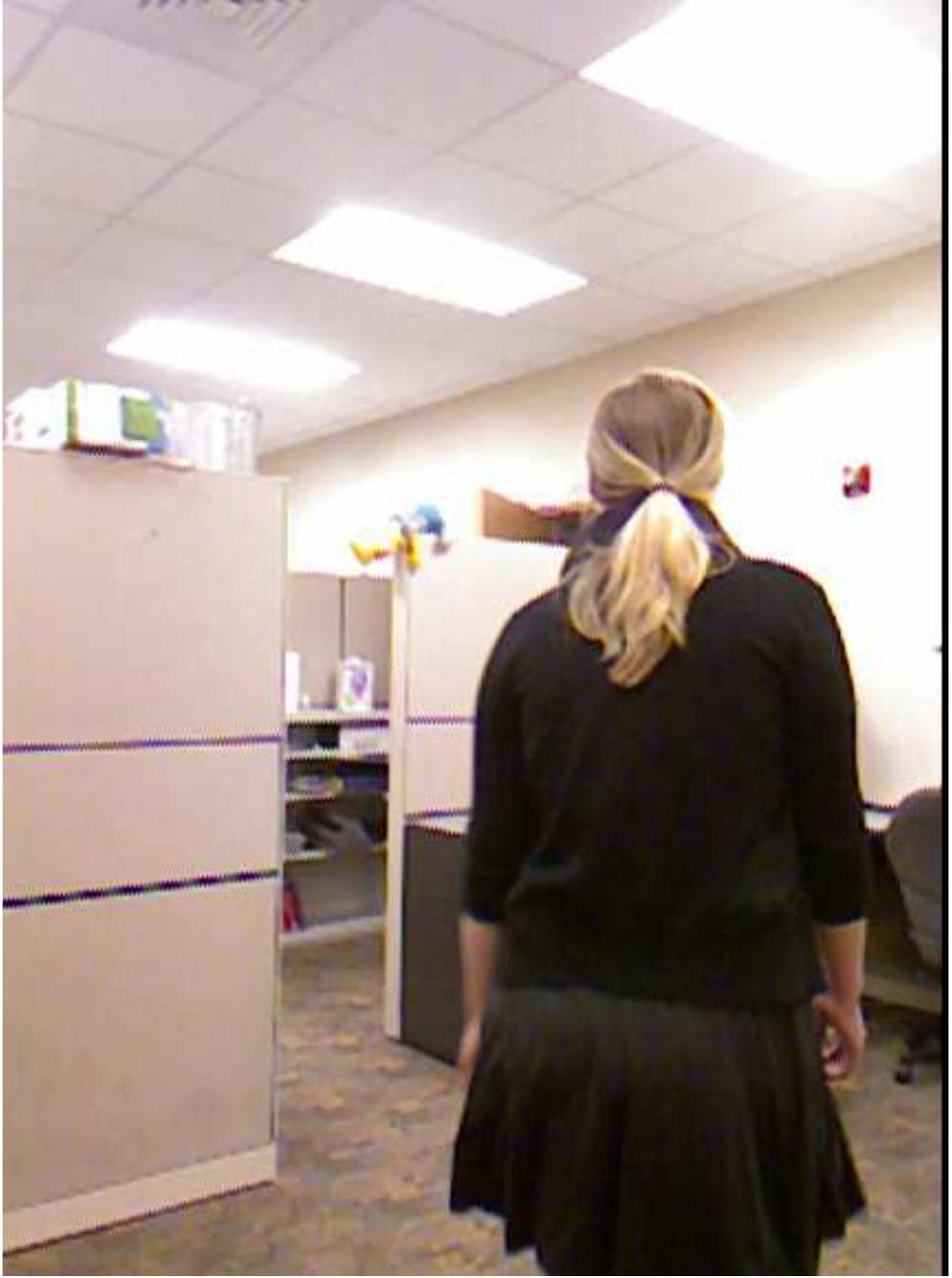} &
			\hspace{-0.35cm}\includegraphics[width=0.12\linewidth]{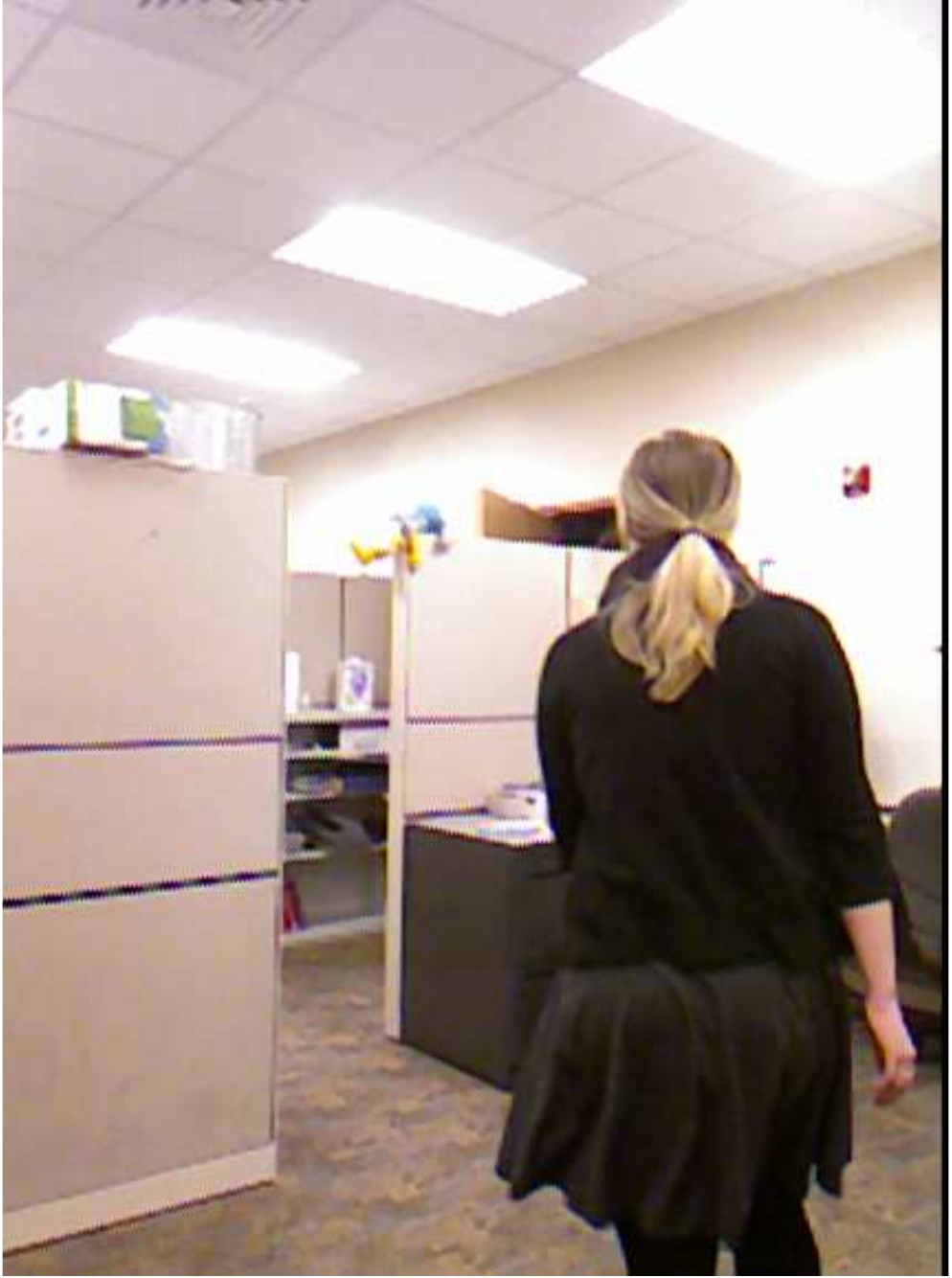} &
			\hspace{-0.35cm}\includegraphics[width=0.12\linewidth]{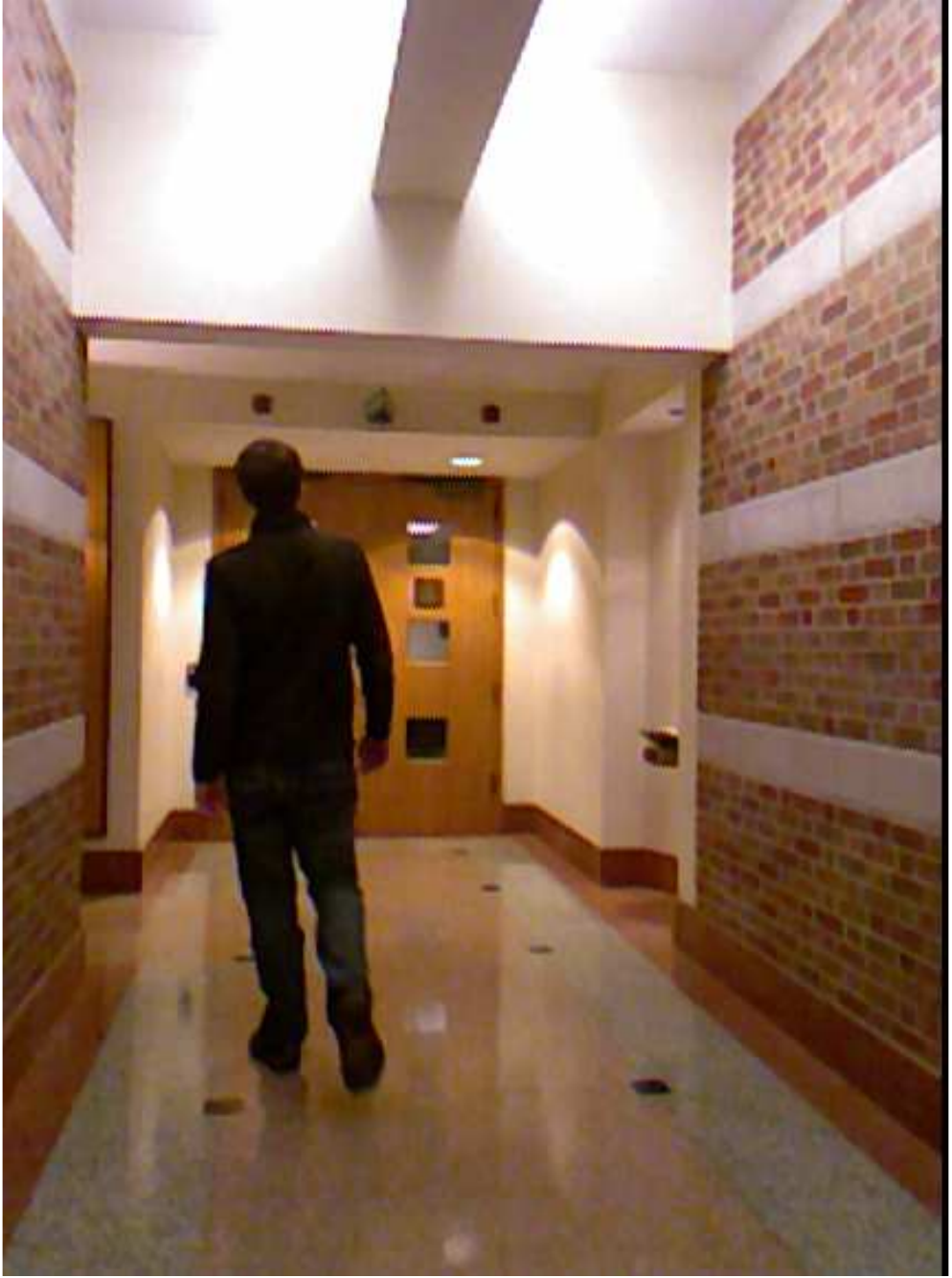}&
			\hspace{-0.35cm}\includegraphics[width=0.12\linewidth]{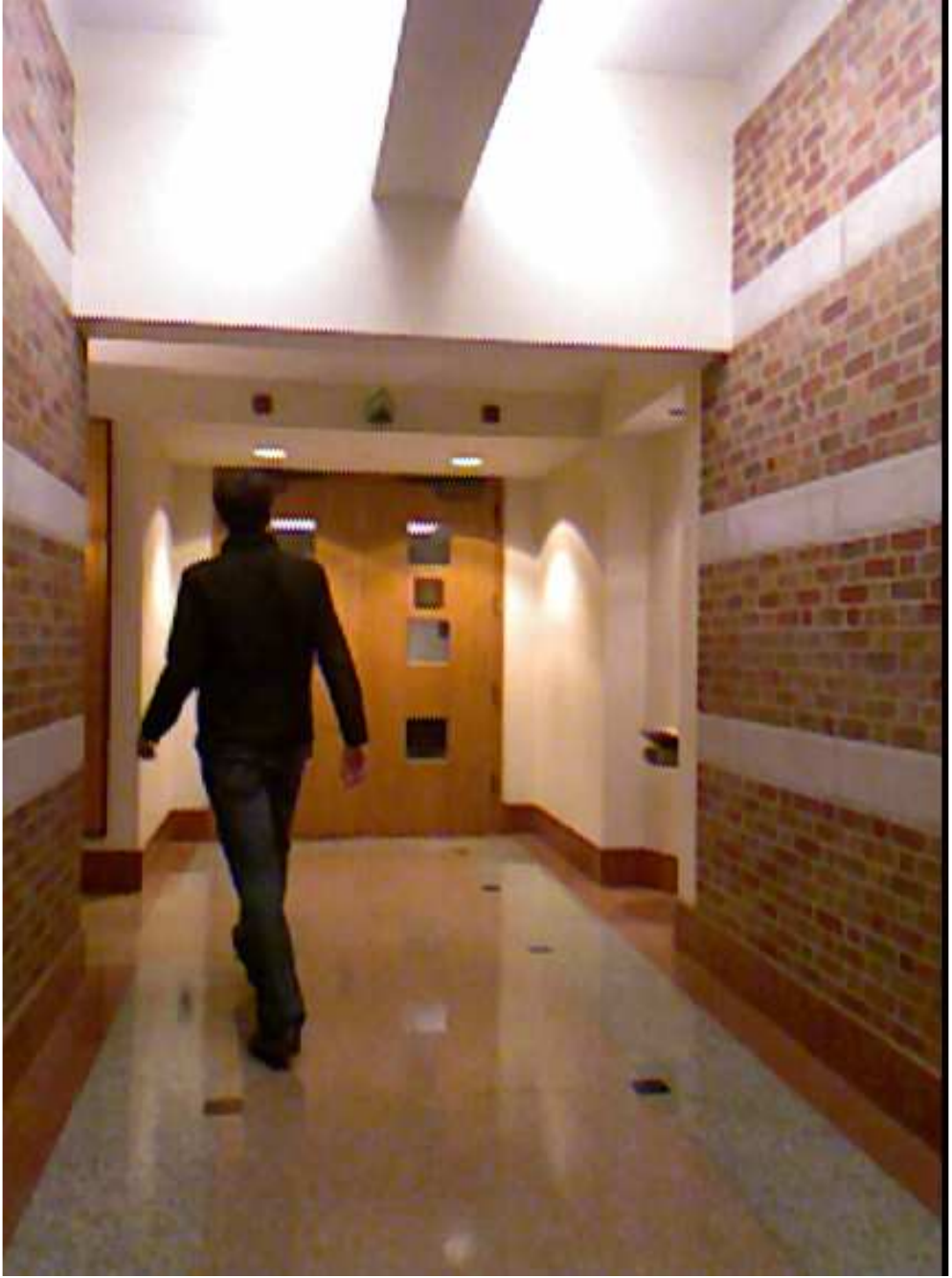}\\
			\hspace{-0.5cm}\begin{sideways}\hspace{0.4cm}{\bf Ground-Truth}\end{sideways} & 
			\hspace{-0.35cm}\includegraphics[width=0.12\linewidth]{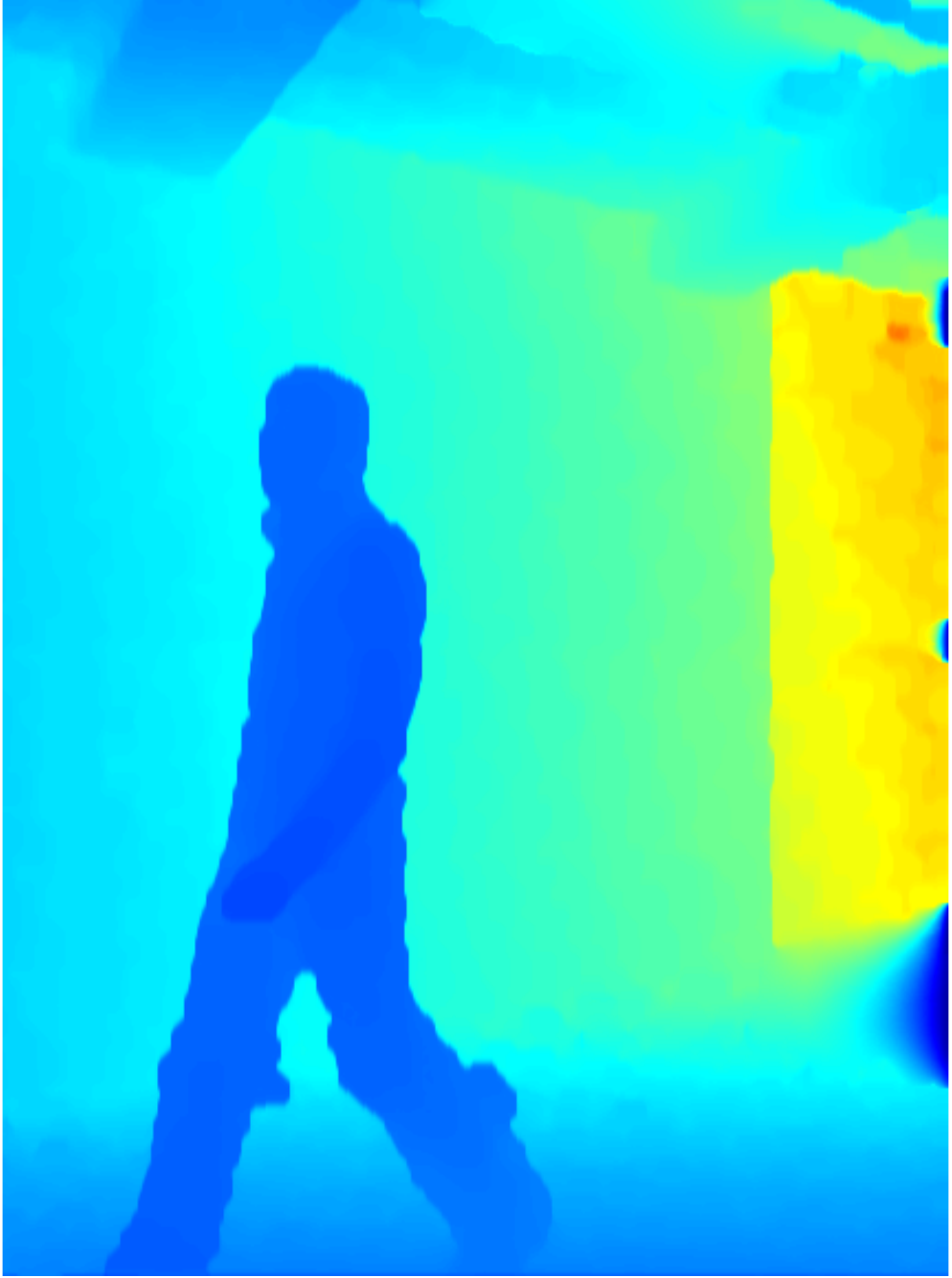} &
			\hspace{-0.35cm}\includegraphics[width=0.12\linewidth]{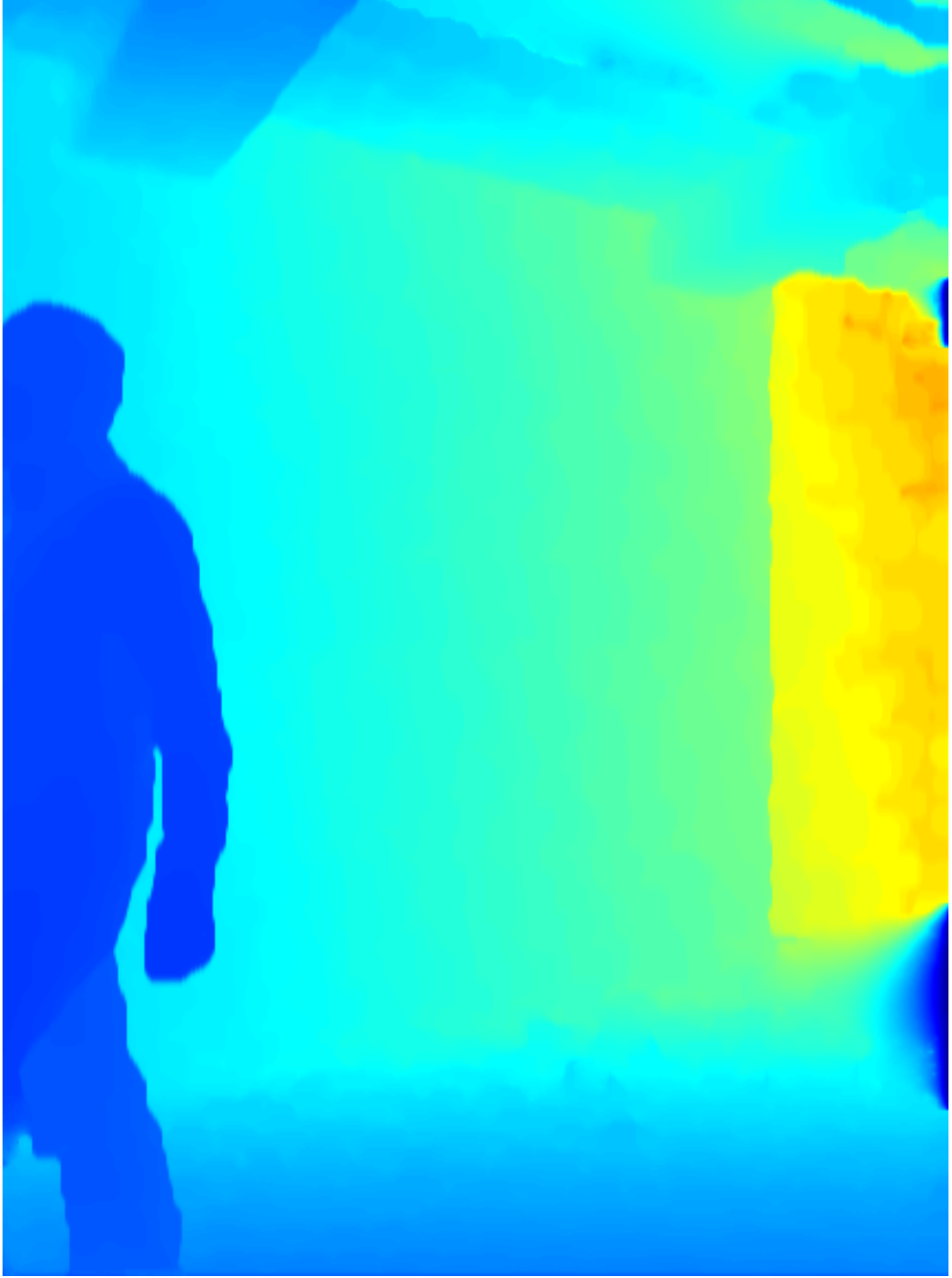} &
			\hspace{-0.35cm}\includegraphics[width=0.12\linewidth]{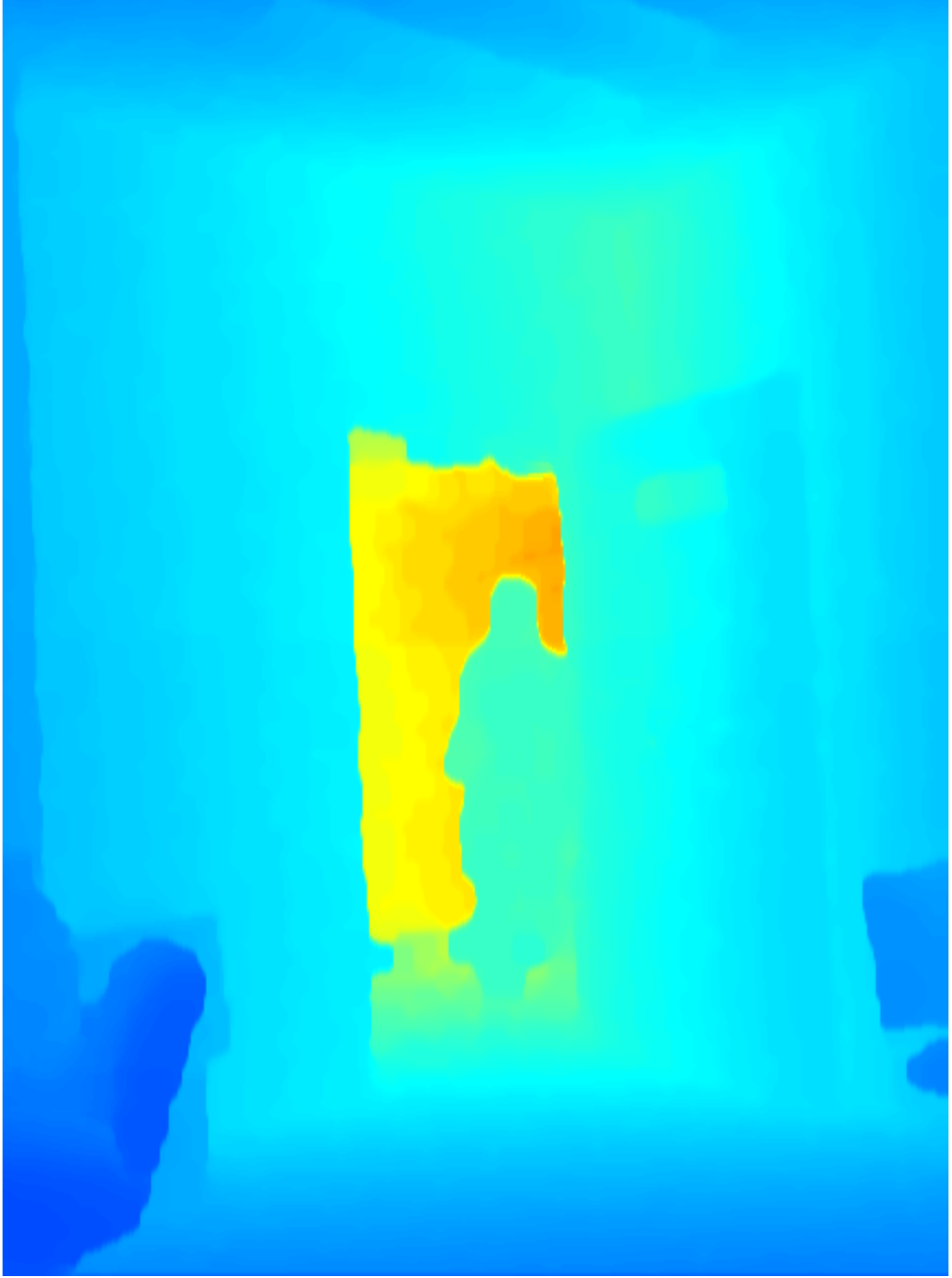} &
			\hspace{-0.35cm}\includegraphics[width=0.12\linewidth]{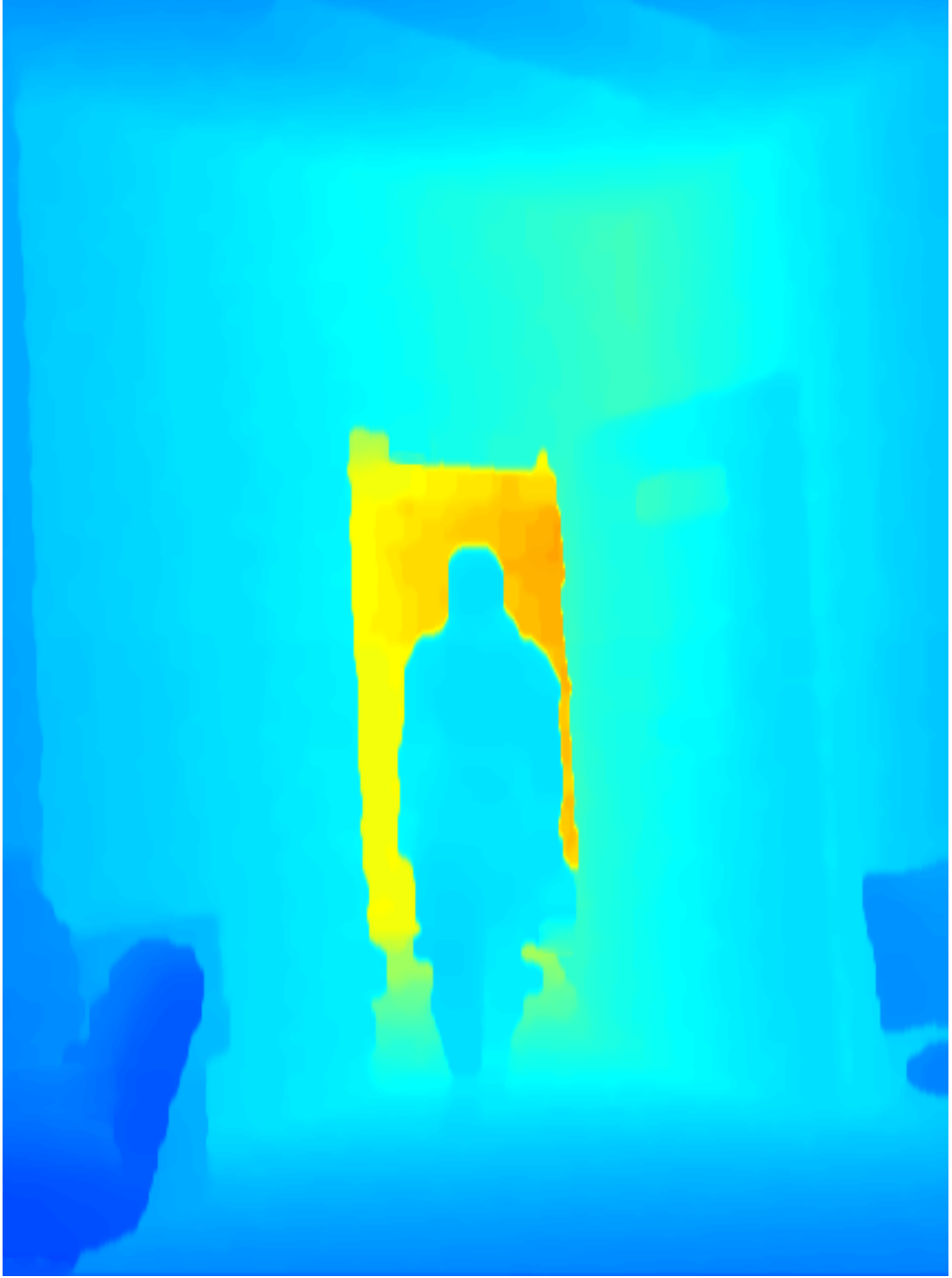} &
			\hspace{-0.35cm}\includegraphics[width=0.12\linewidth]{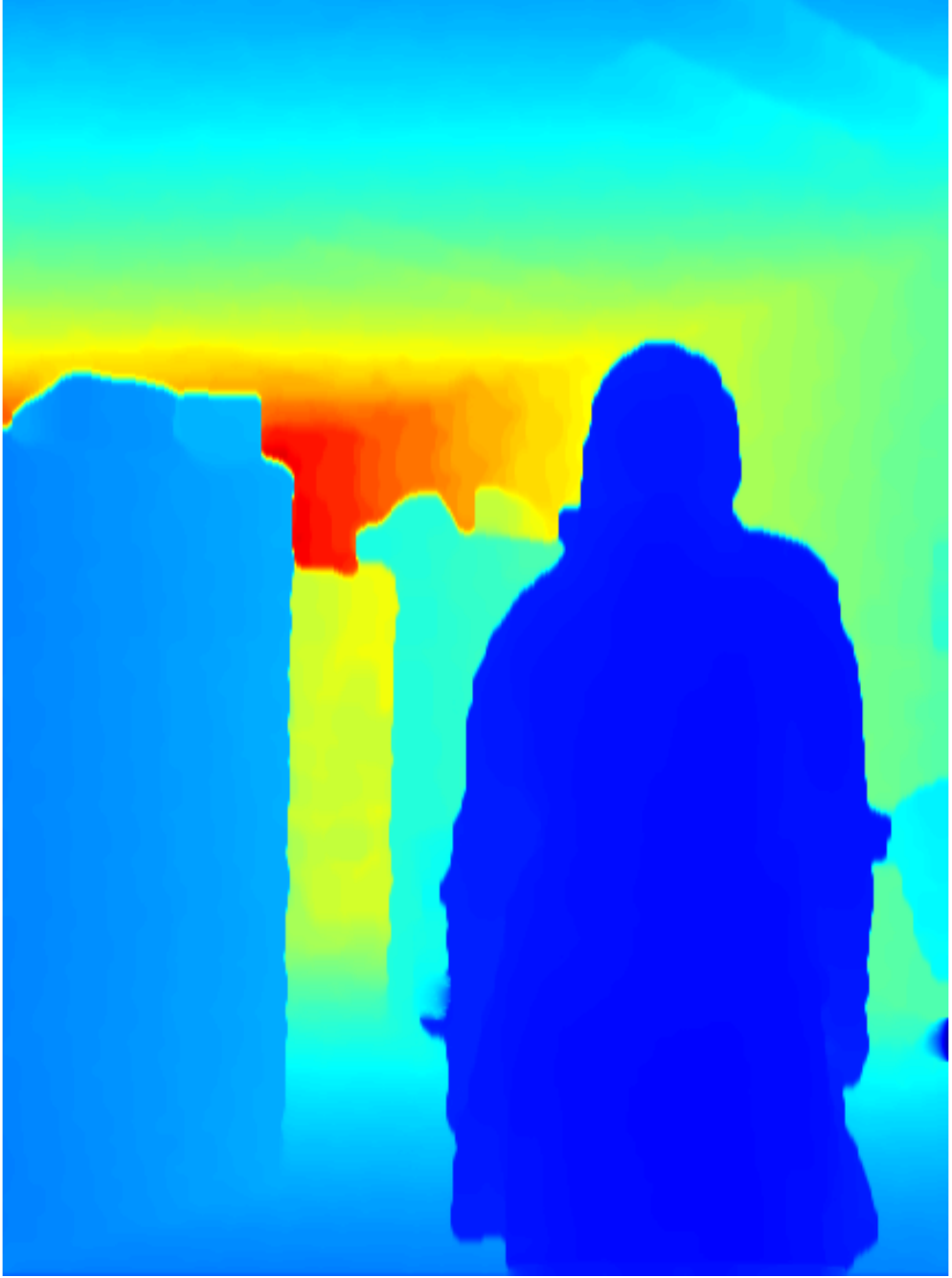} &
			\hspace{-0.35cm}\includegraphics[width=0.12\linewidth]{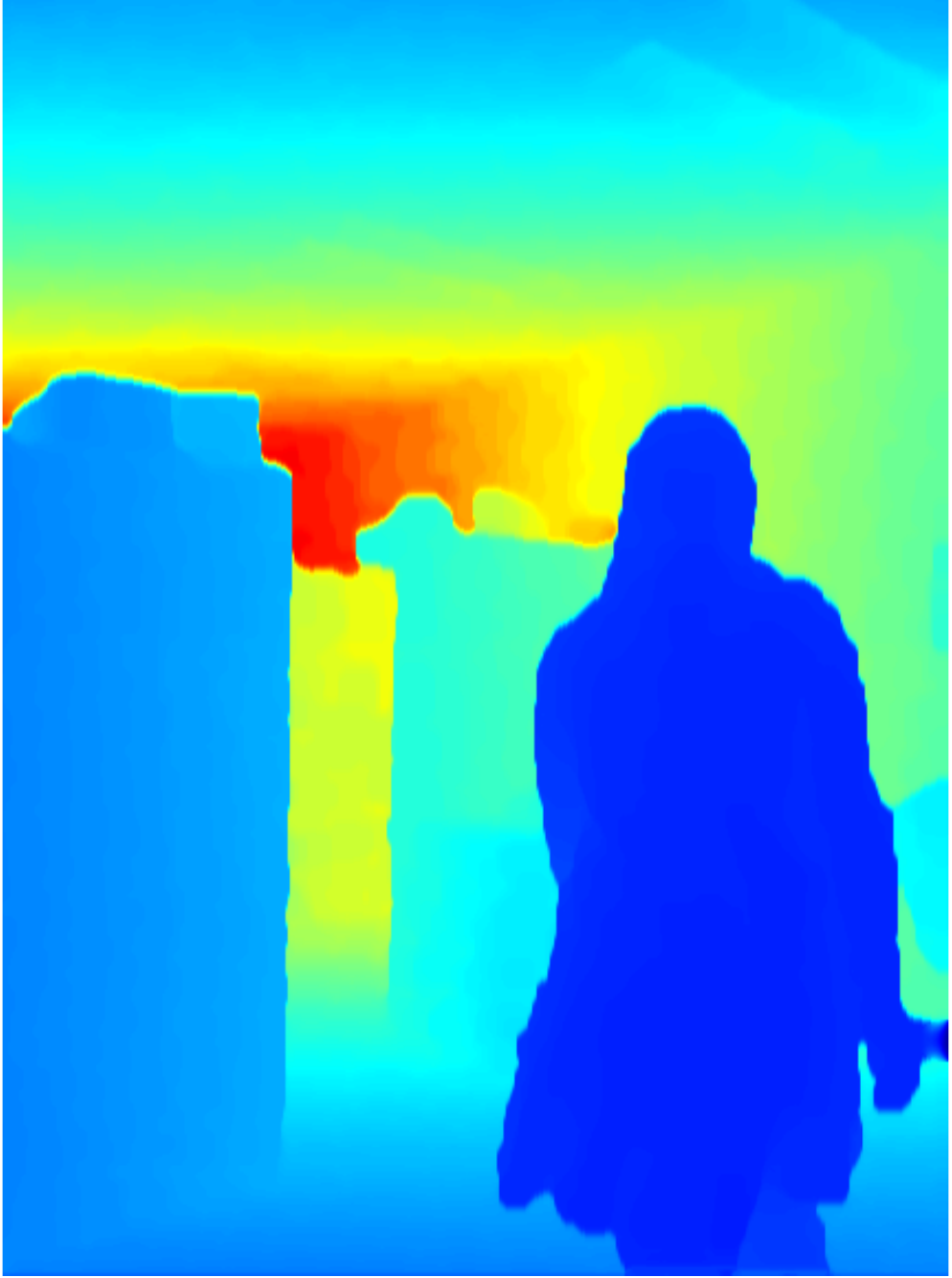} &
			\hspace{-0.35cm}\includegraphics[width=0.12\linewidth]{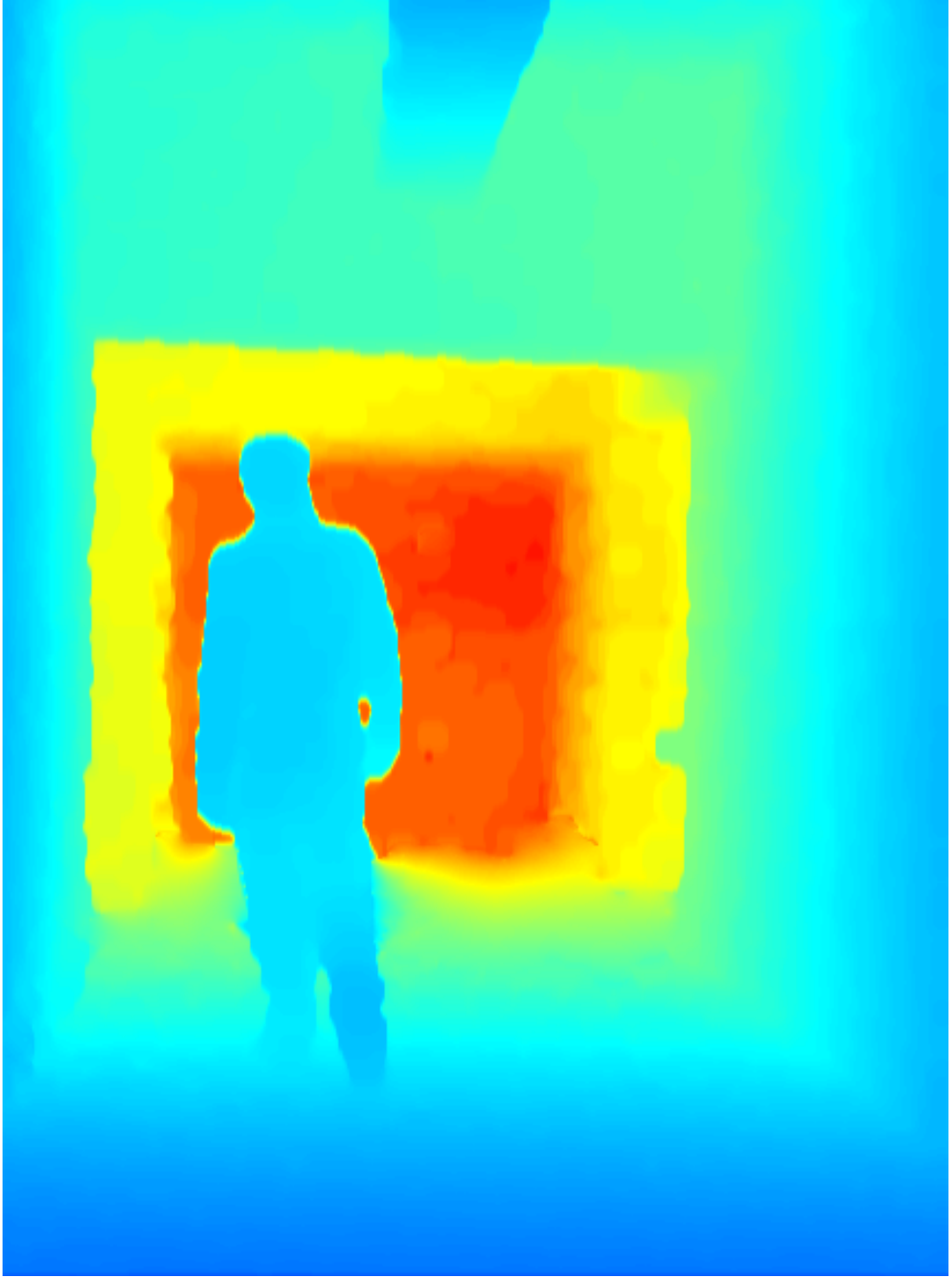}&
			\hspace{-0.35cm}\includegraphics[width=0.12\linewidth]{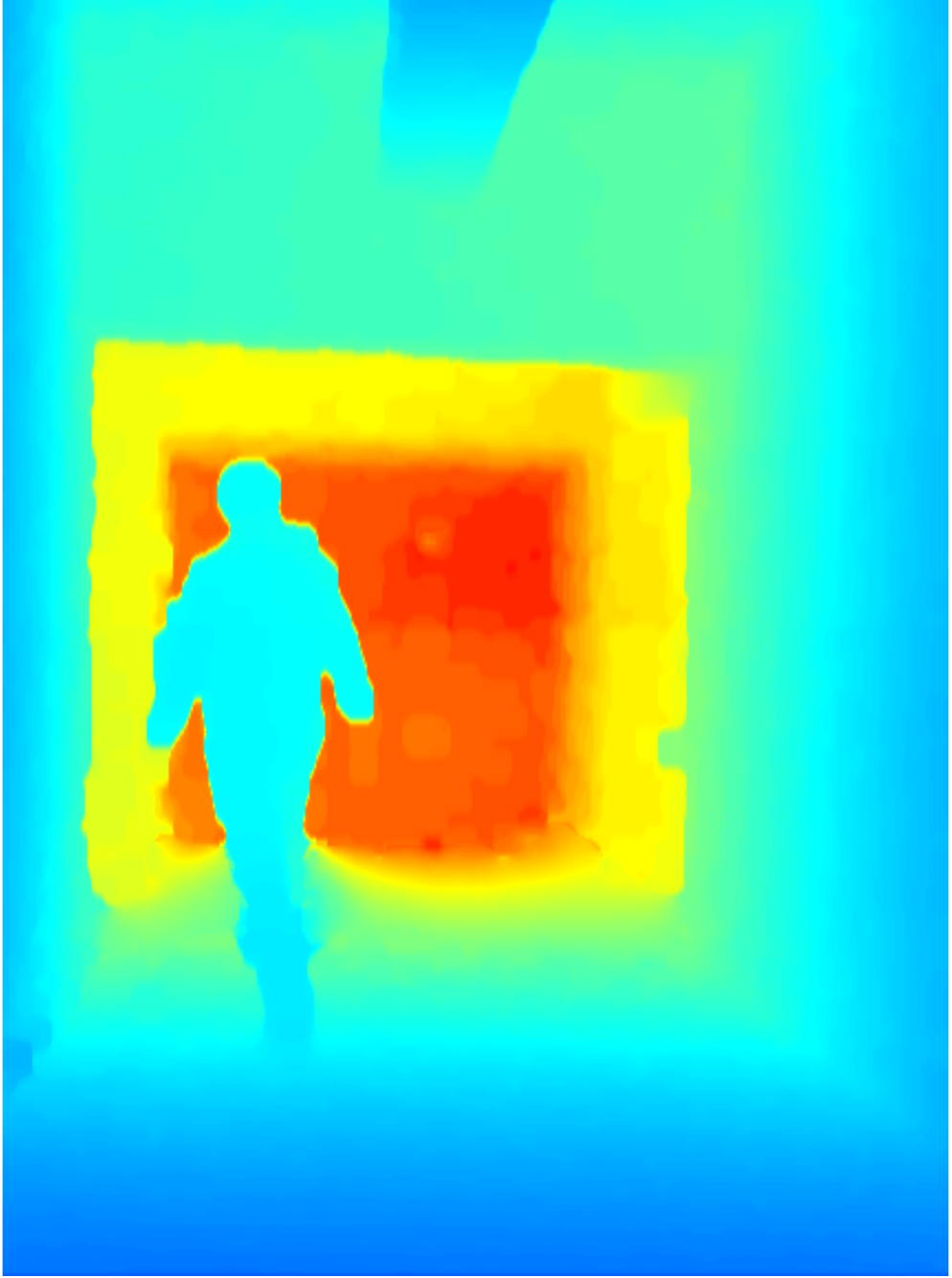}\\
			\hspace{-0.5cm}\begin{sideways}\hspace{0.7cm}{\bf Ours-1F}\end{sideways} & 
			\hspace{-0.35cm}\includegraphics[width=0.12\linewidth]{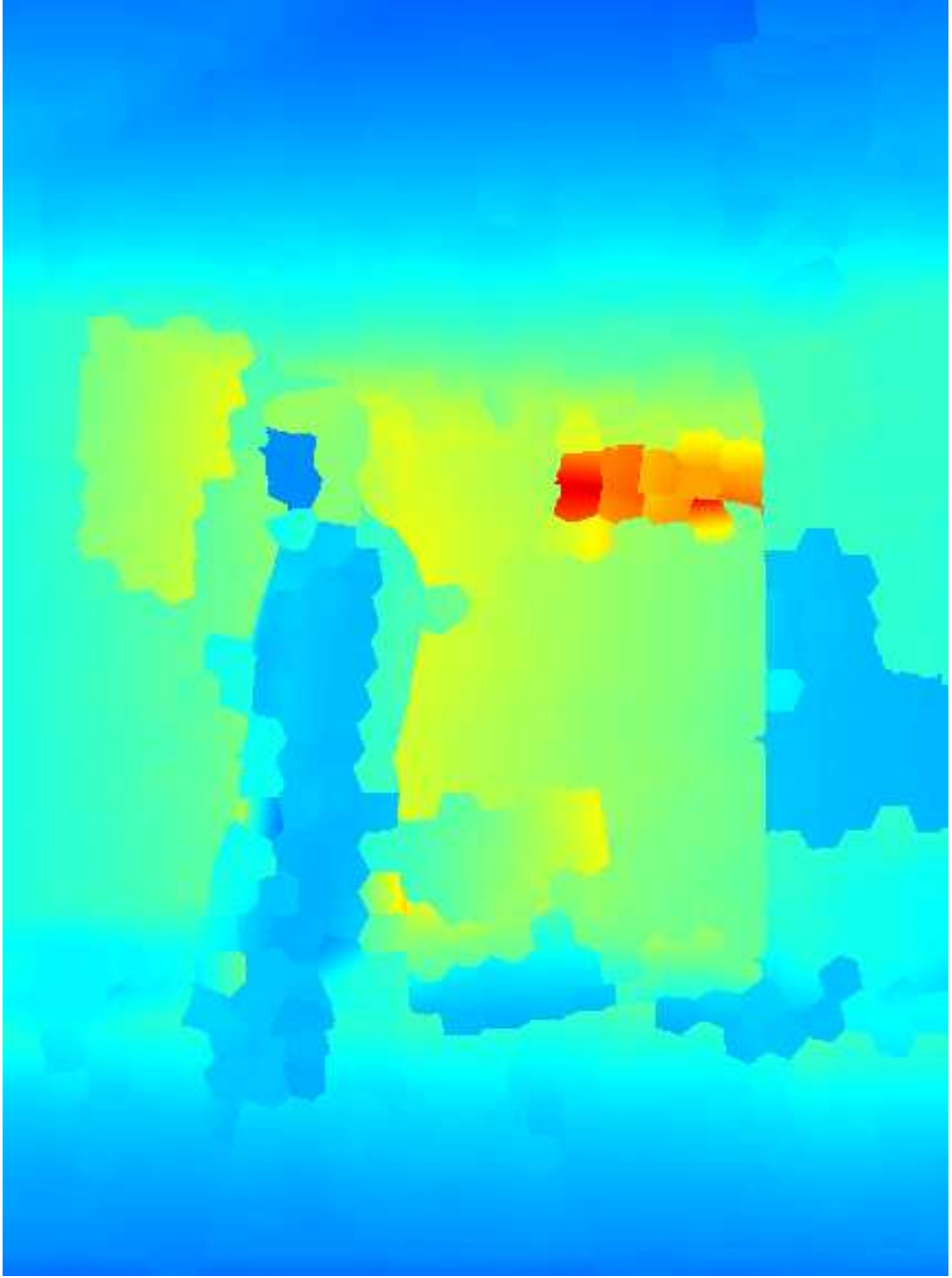} &
			\hspace{-0.35cm}\includegraphics[width=0.12\linewidth]{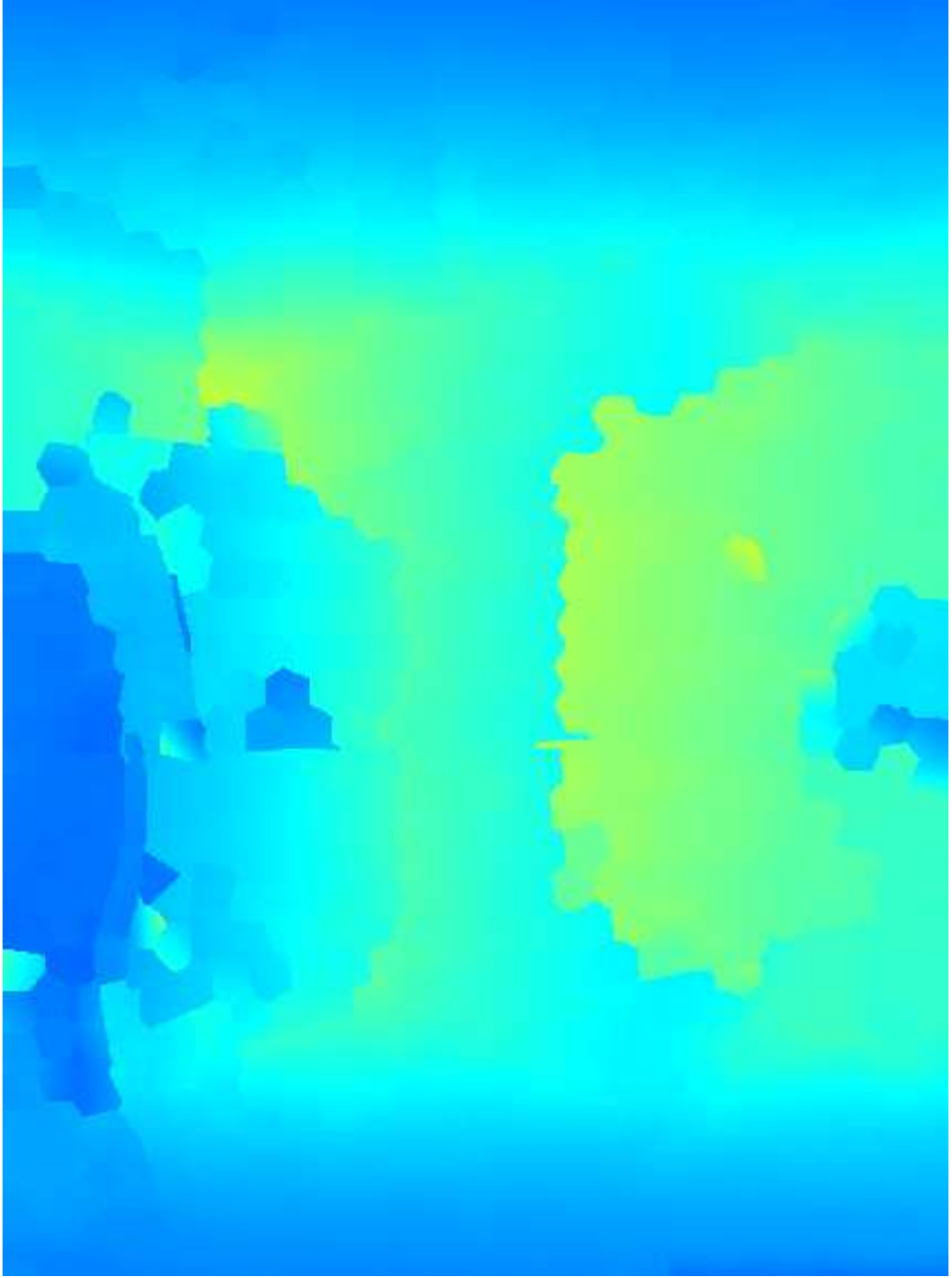} &
			\hspace{-0.35cm}\includegraphics[width=0.12\linewidth]{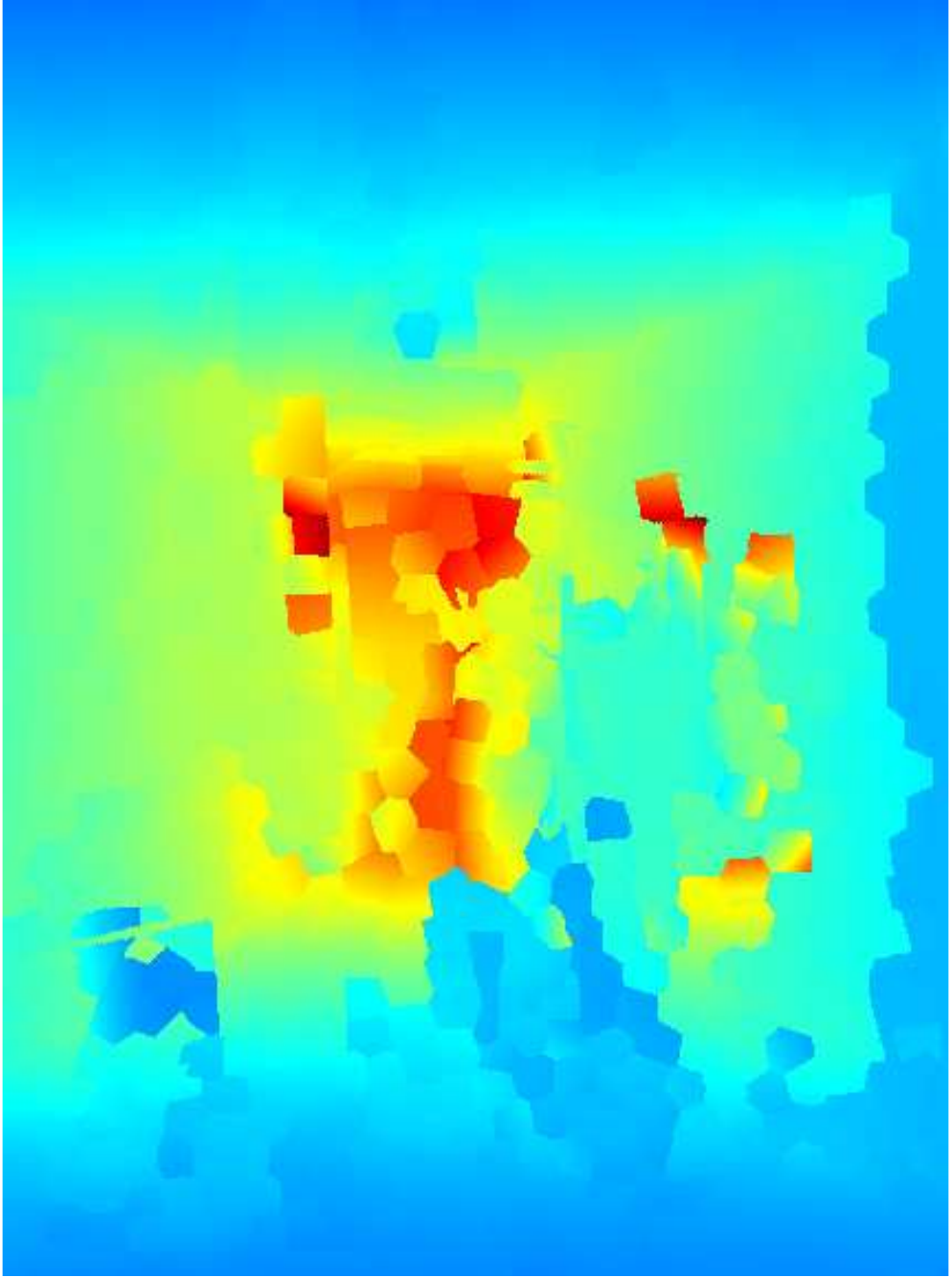} &
			\hspace{-0.35cm}\includegraphics[width=0.12\linewidth]{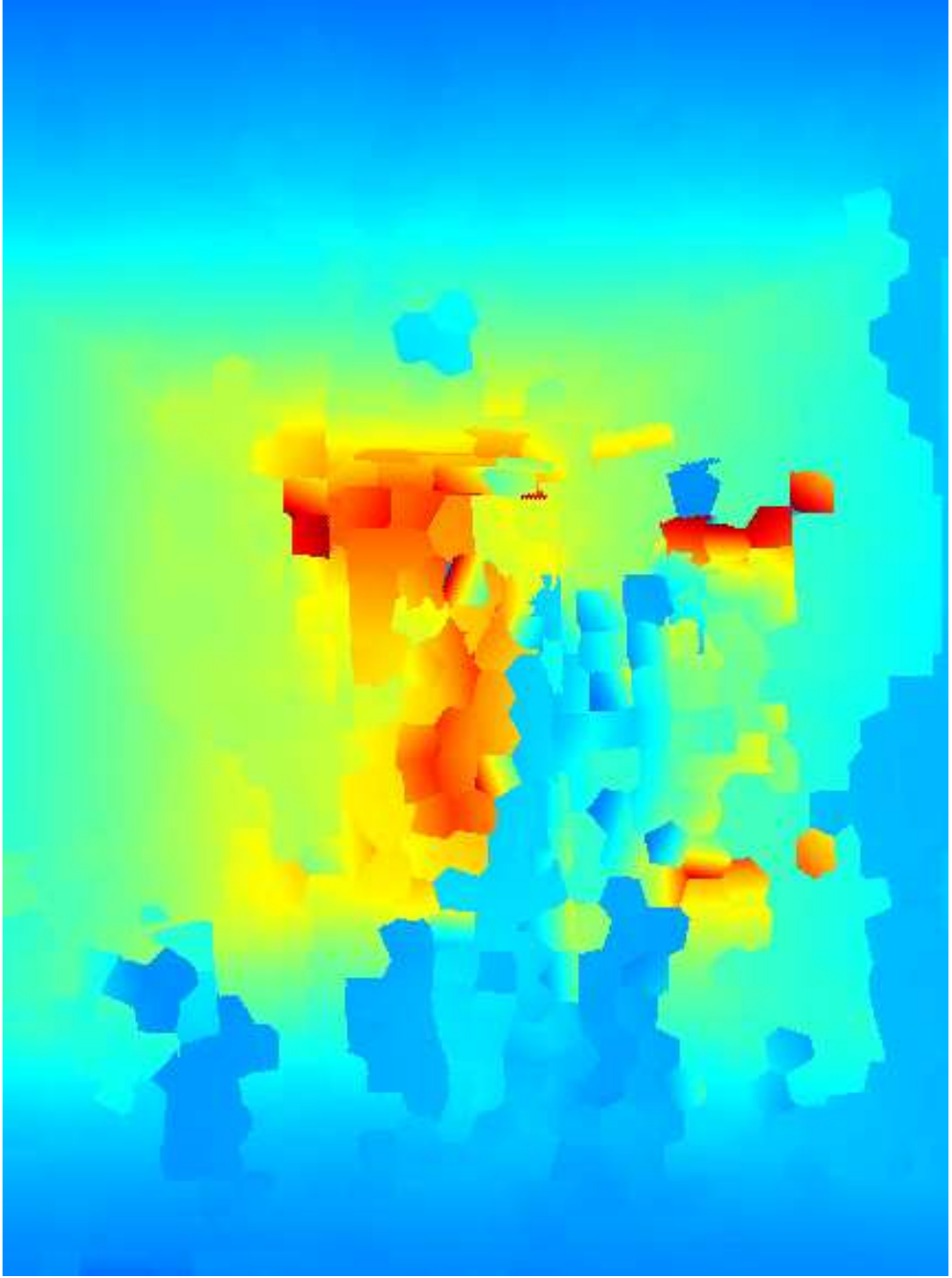} &
			\hspace{-0.35cm}\includegraphics[width=0.12\linewidth]{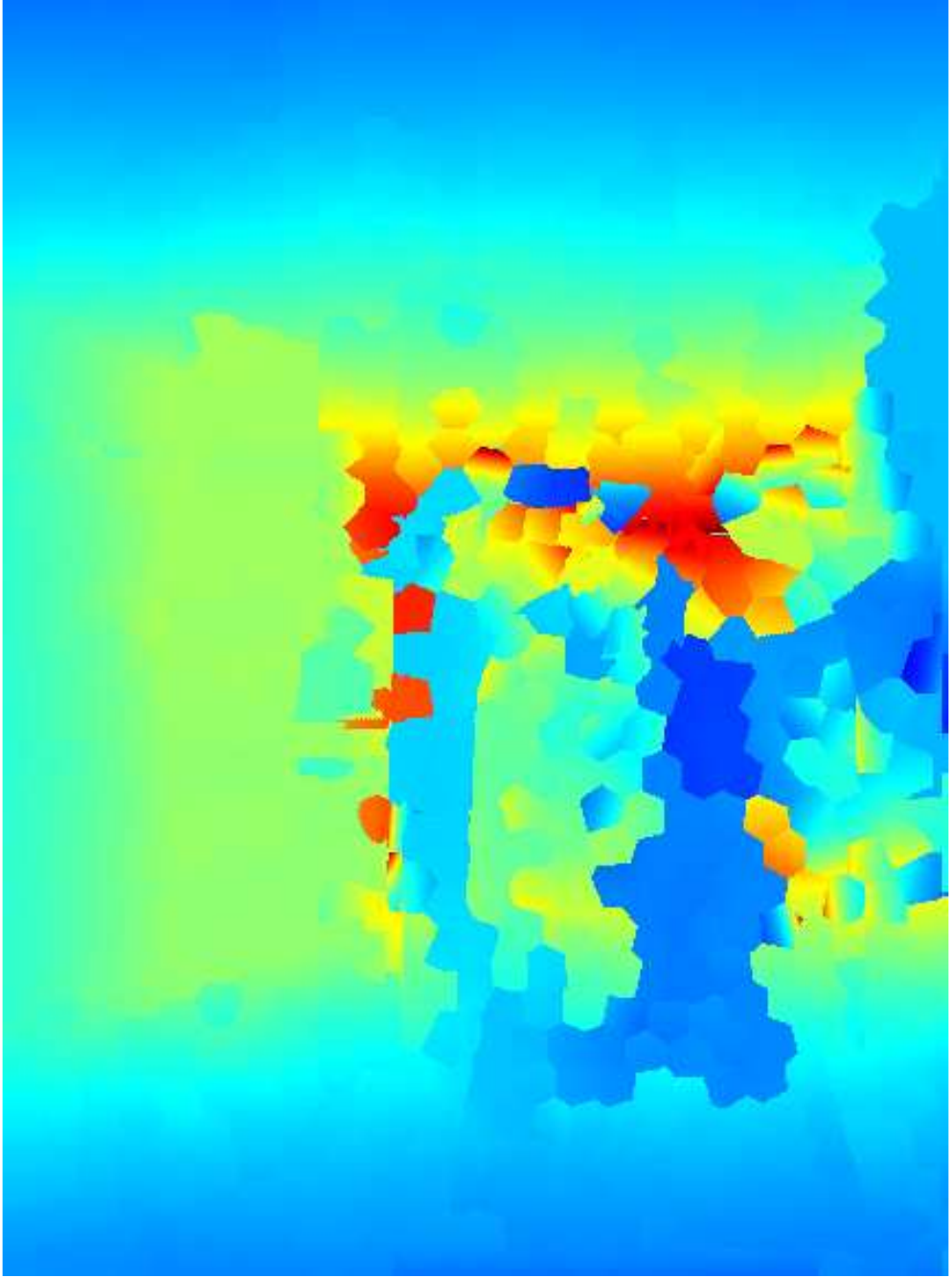} &
			\hspace{-0.35cm}\includegraphics[width=0.12\linewidth]{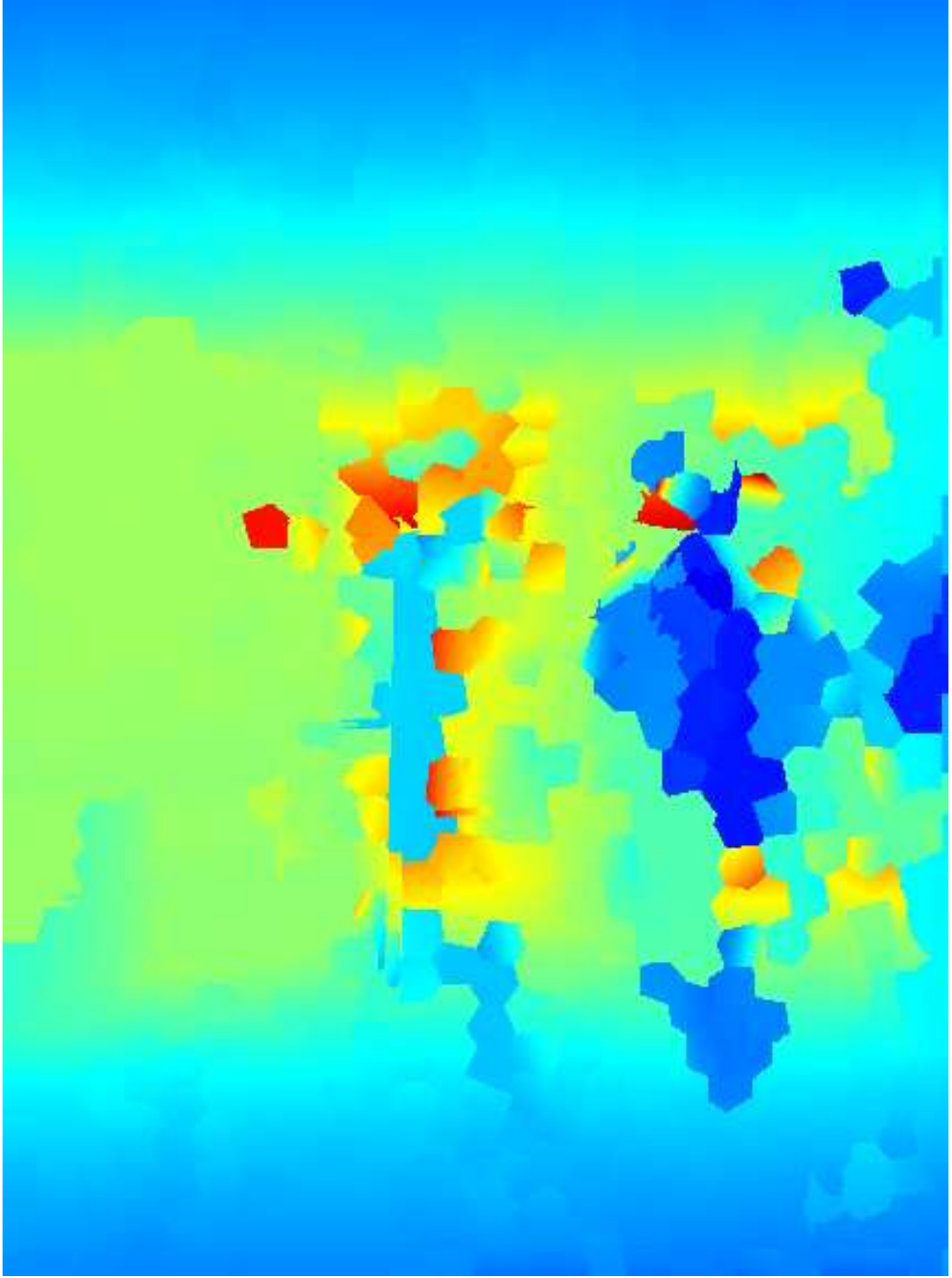} &
			\hspace{-0.35cm}\includegraphics[width=0.12\linewidth]{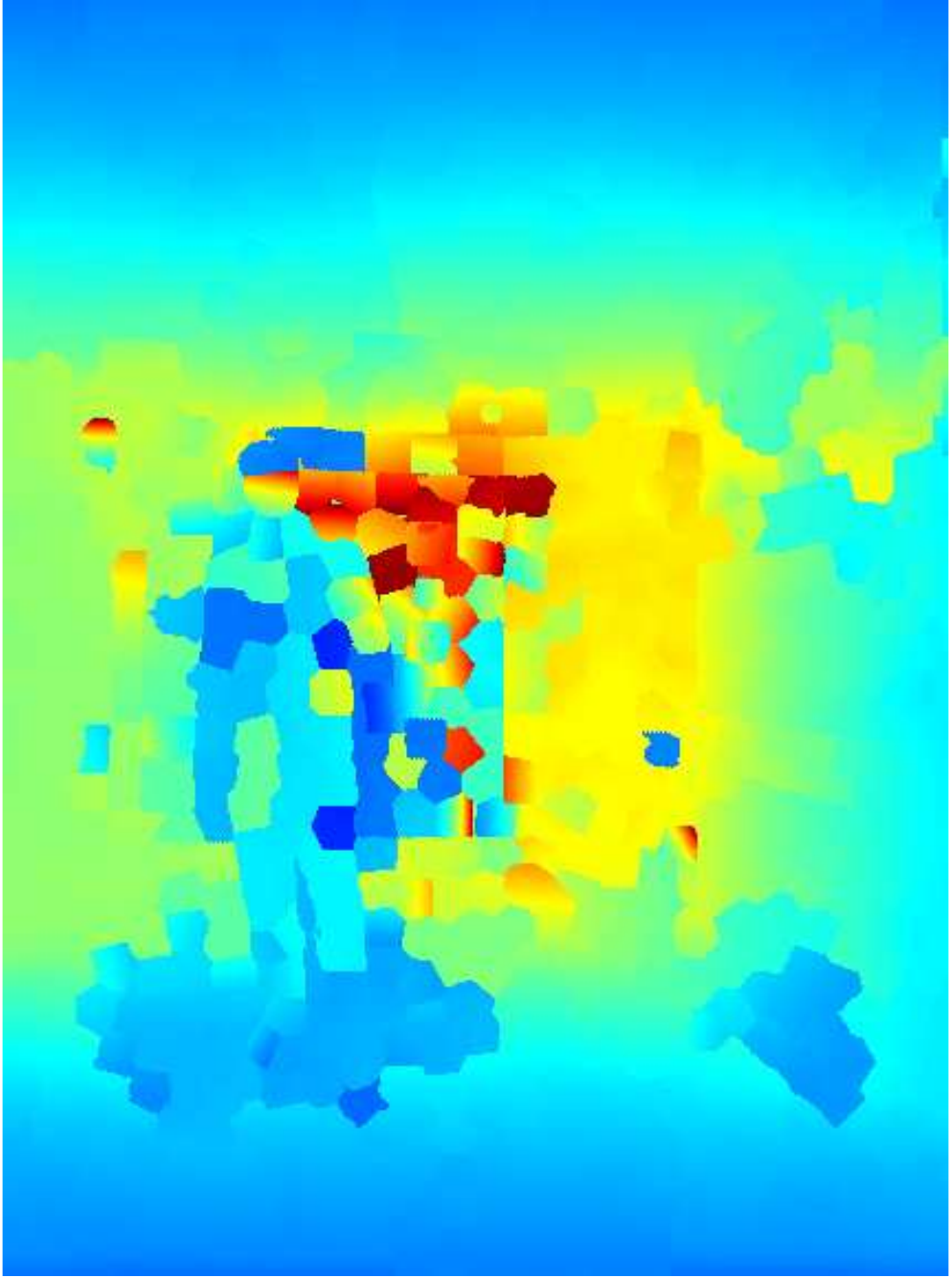}&
			\hspace{-0.35cm}\includegraphics[width=0.12\linewidth]{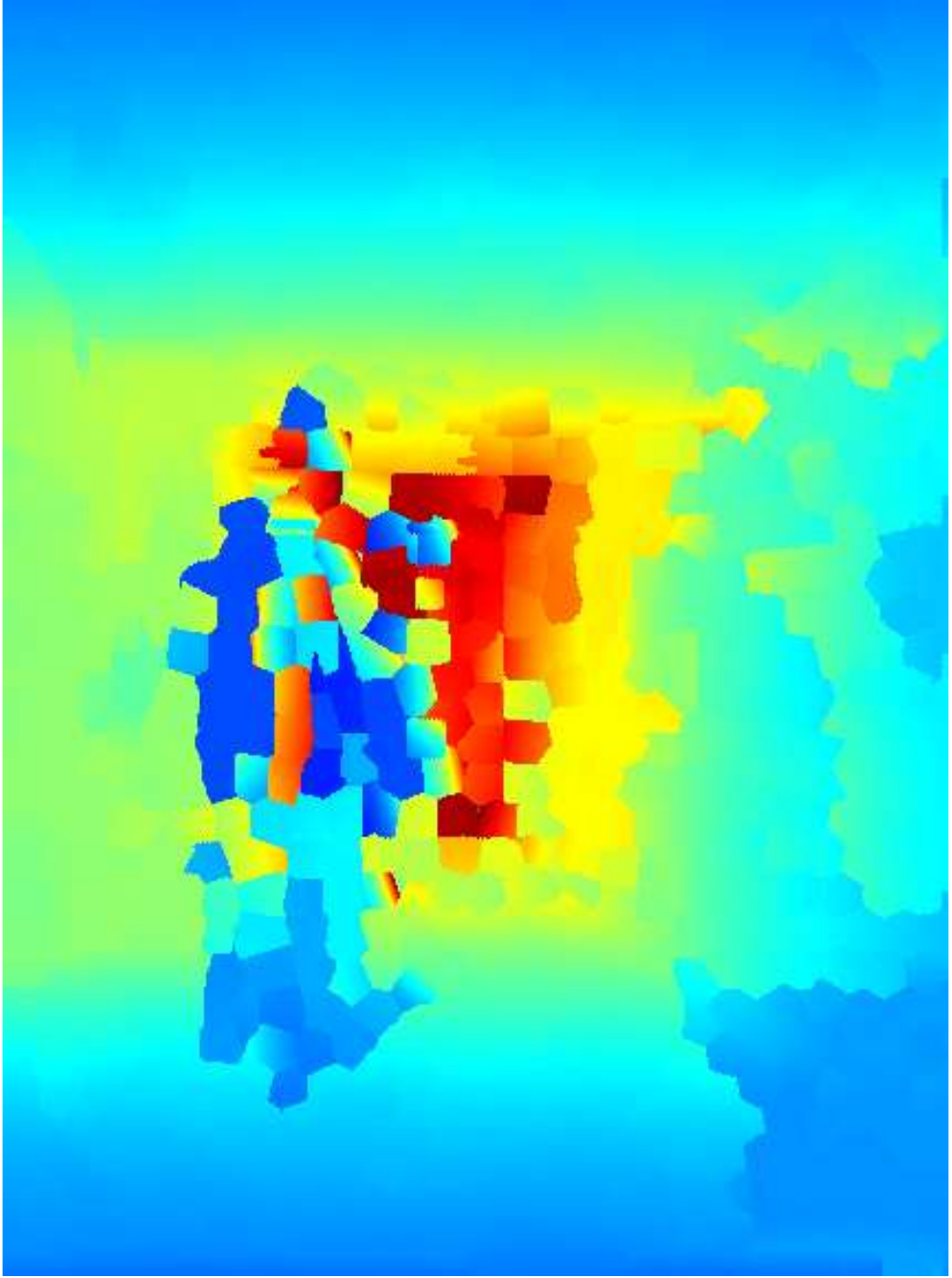}\\
			\hspace{-0.5cm}\begin{sideways}\hspace{0.1cm}{\bf DepthTransfer~\cite{Karsch12}}\end{sideways} & 
			\hspace{-0.35cm}\includegraphics[width=0.12\linewidth]{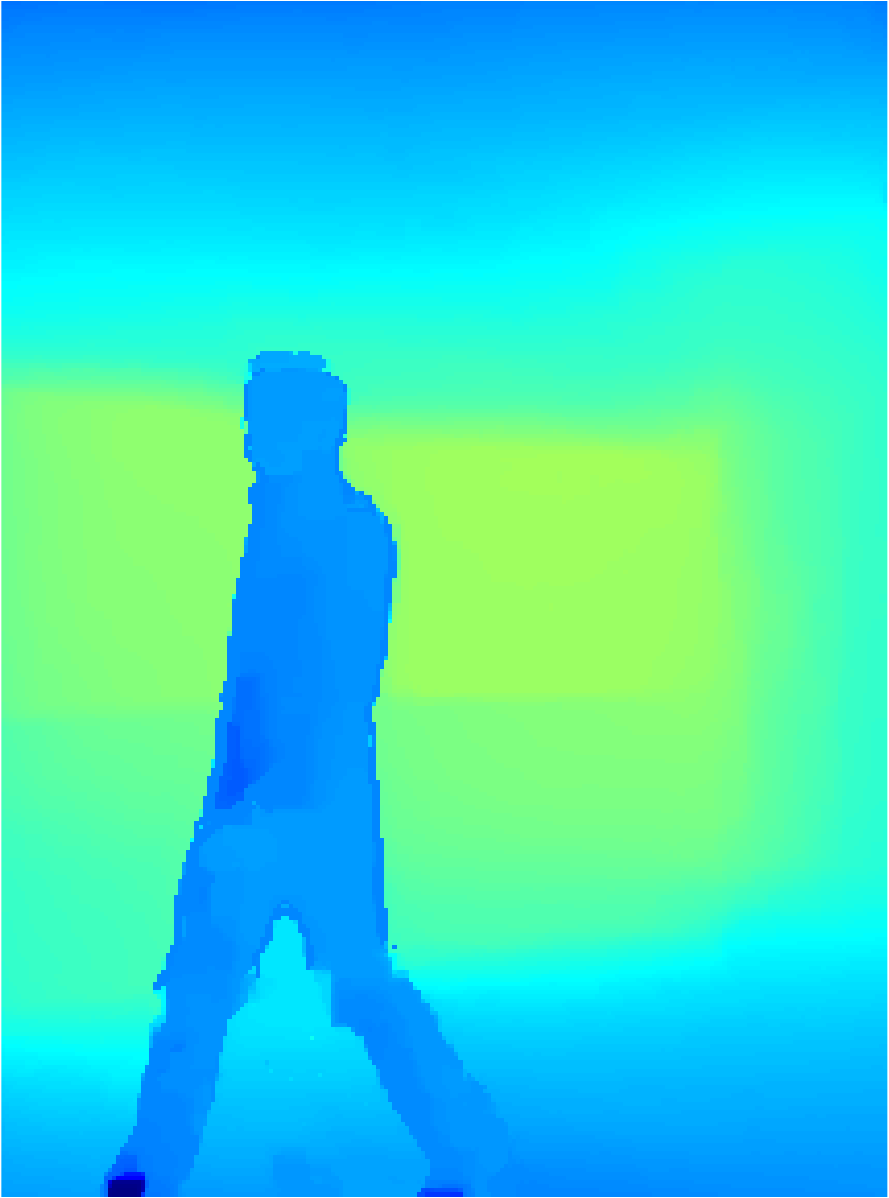} &
			\hspace{-0.35cm}\includegraphics[width=0.12\linewidth]{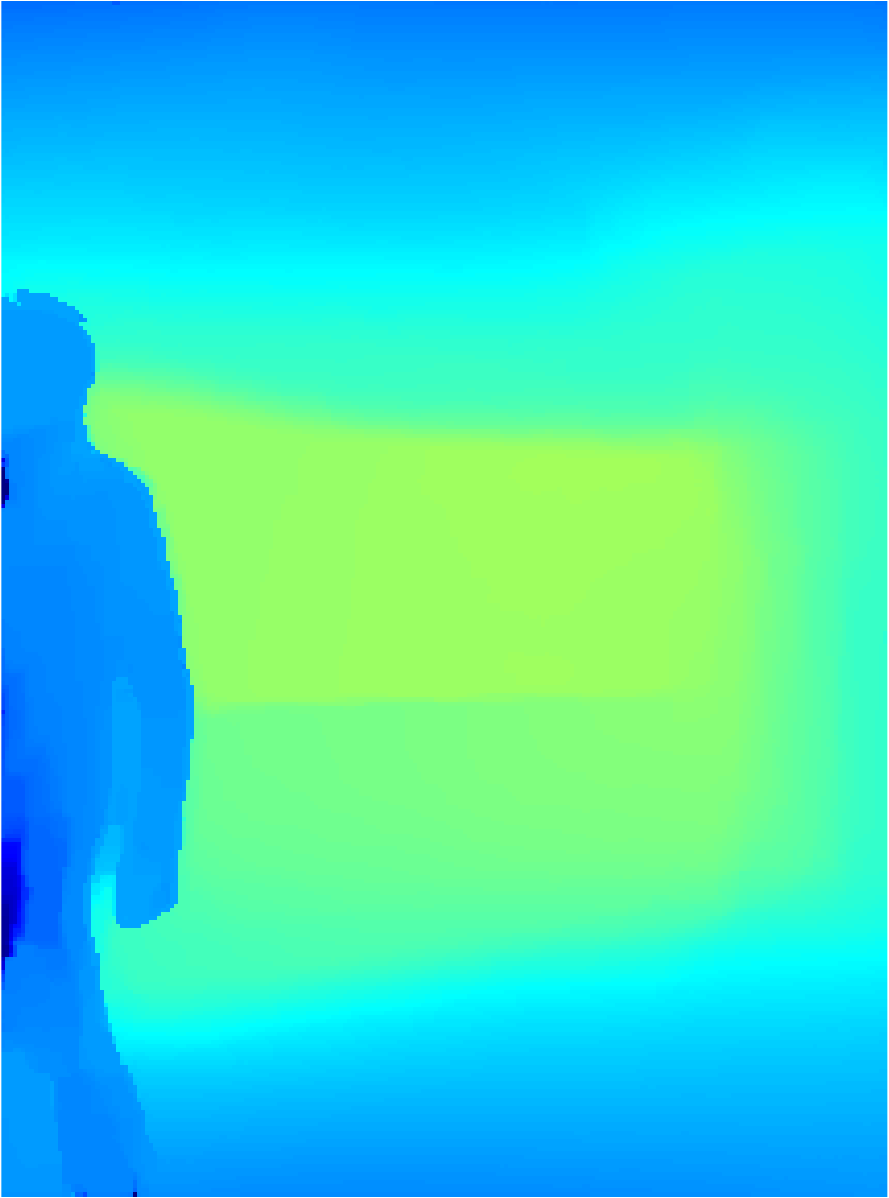} &
			\hspace{-0.35cm}\includegraphics[width=0.12\linewidth]{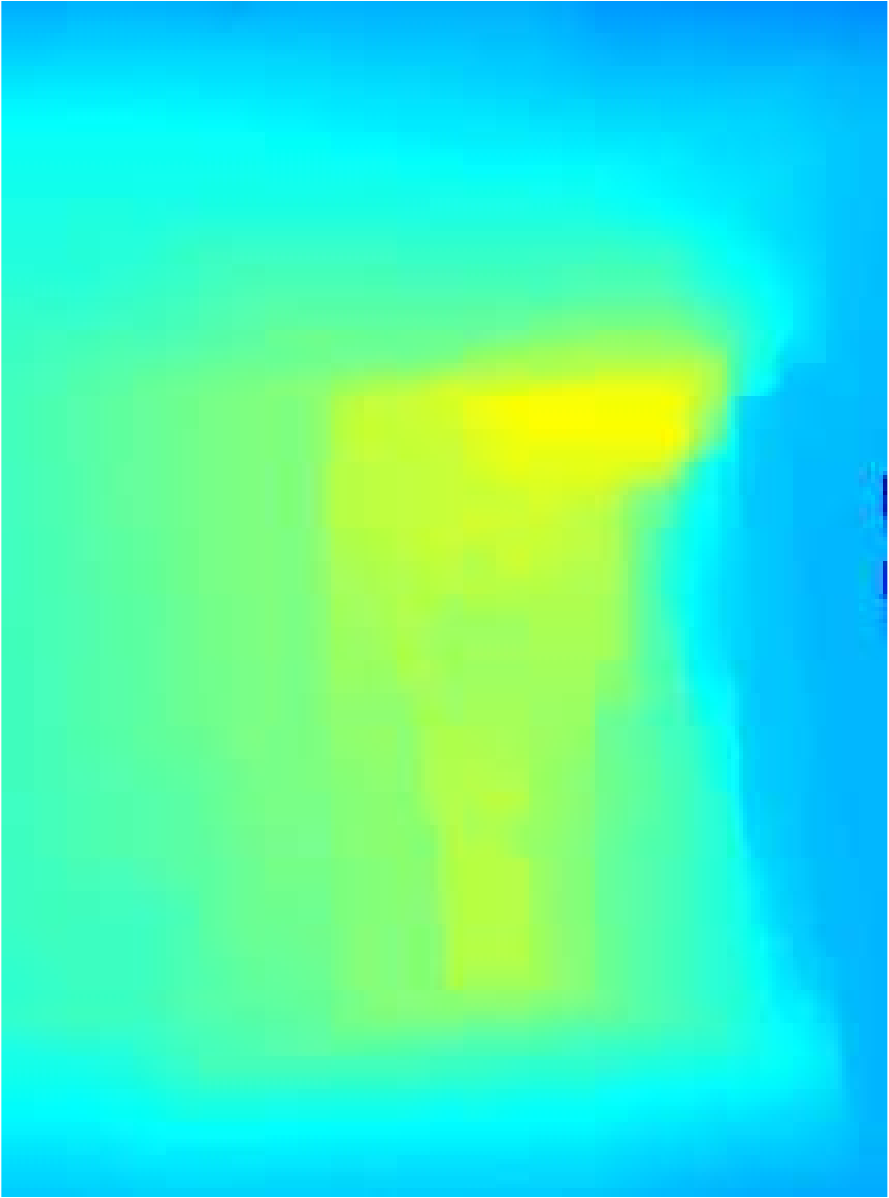} &
			\hspace{-0.35cm}\includegraphics[width=0.12\linewidth]{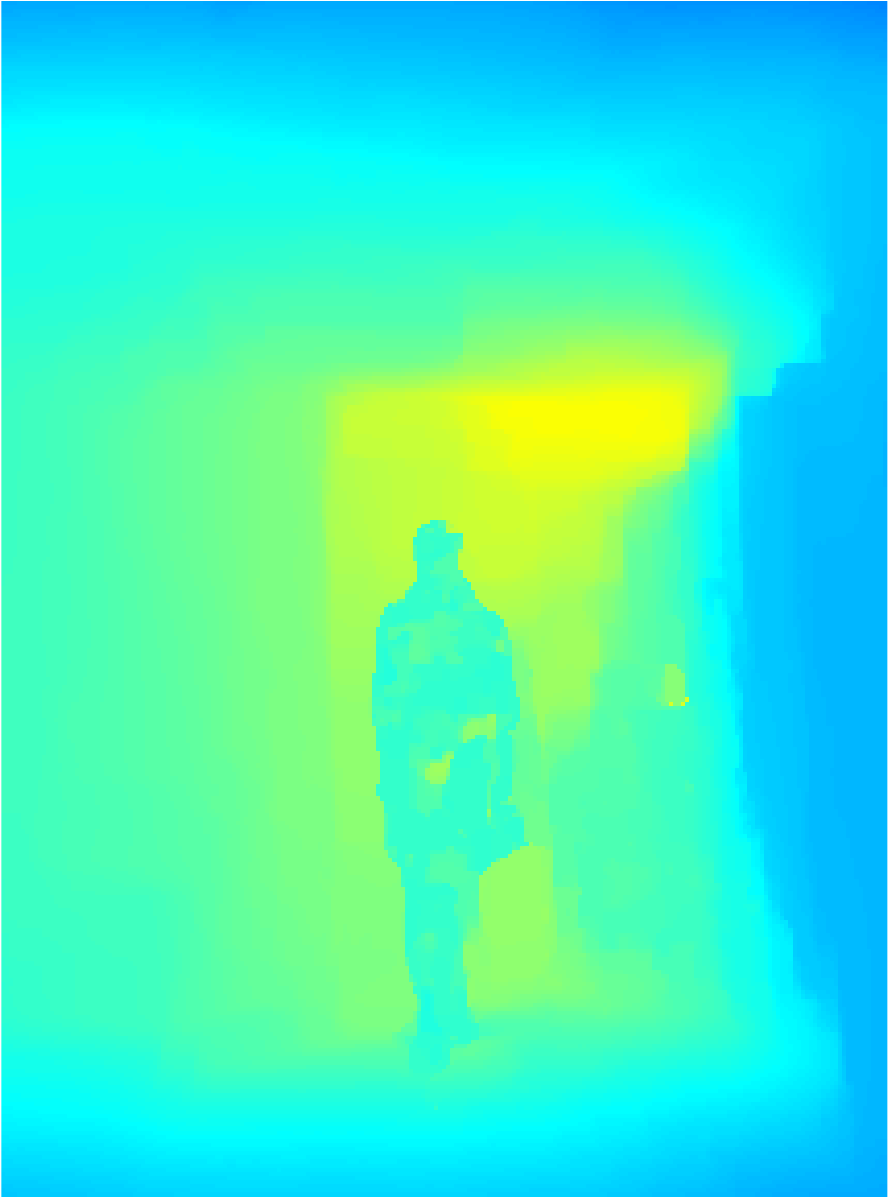} &
			\hspace{-0.35cm}\includegraphics[width=0.12\linewidth]{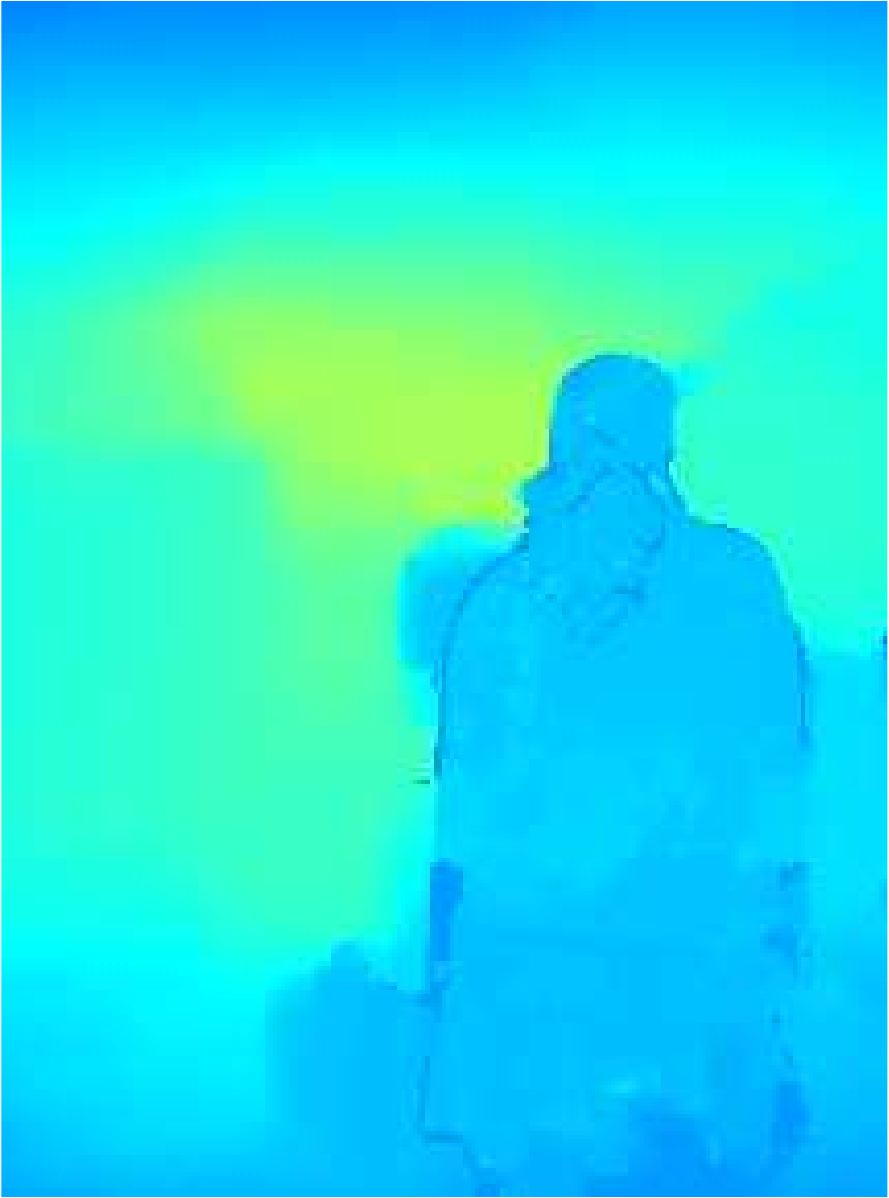} &
			\hspace{-0.35cm}\includegraphics[width=0.12\linewidth]{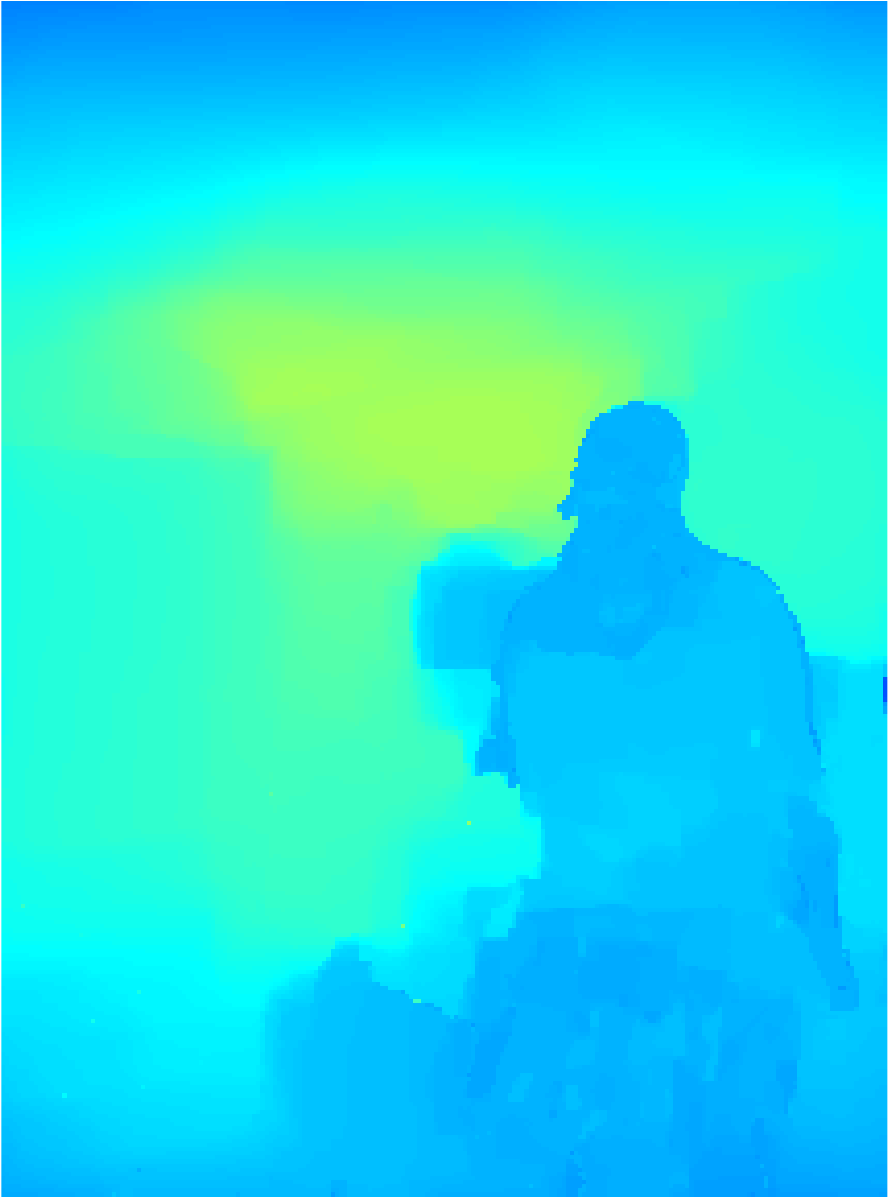} &
			\hspace{-0.35cm}\includegraphics[width=0.12\linewidth]{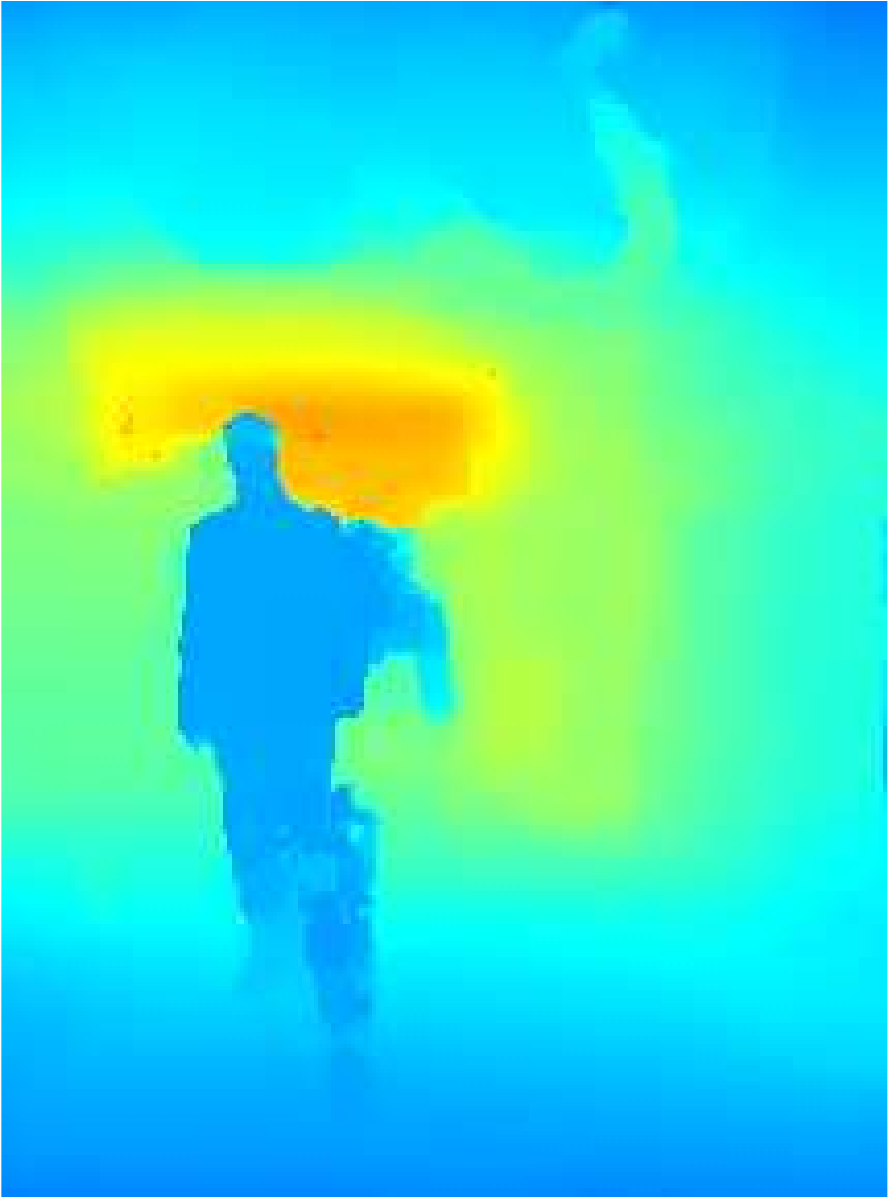}&
			\hspace{-0.35cm}\includegraphics[width=0.12\linewidth]{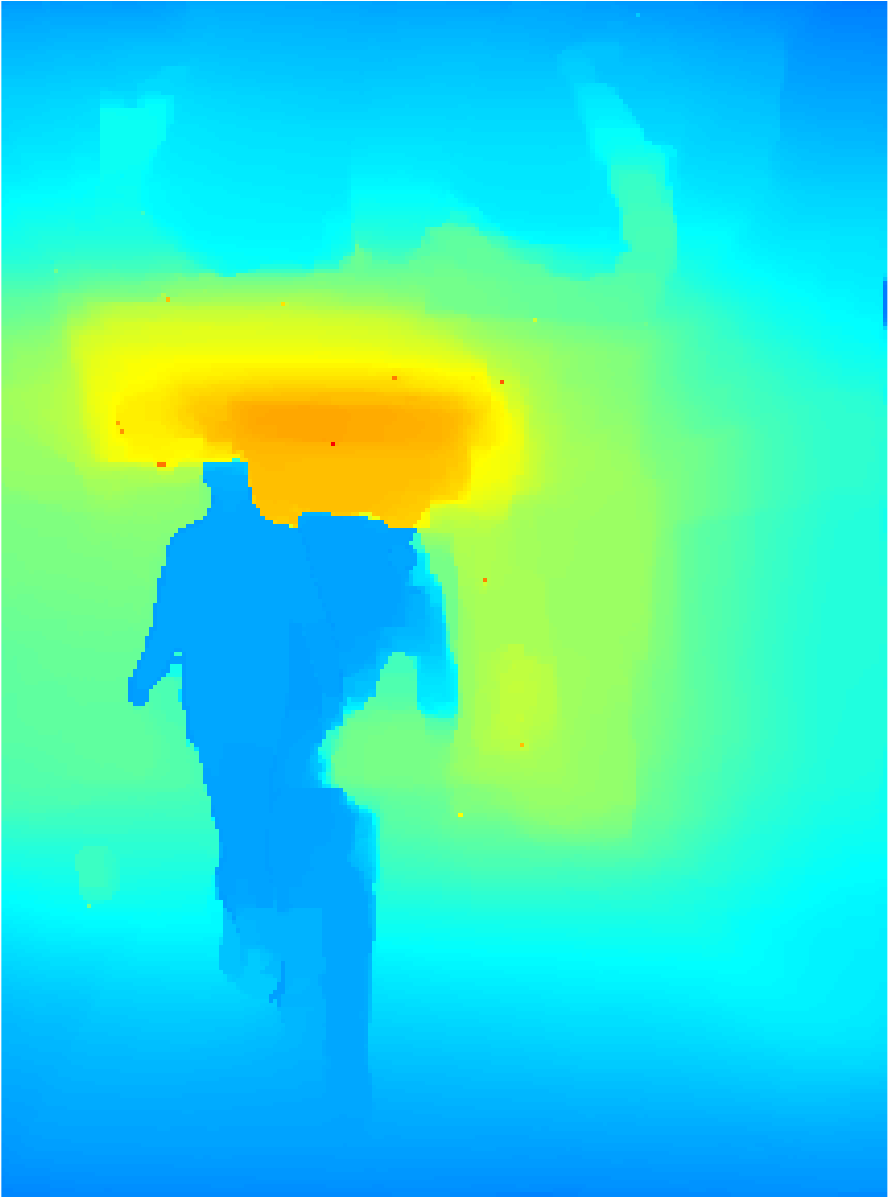}\\
			\hspace{-0.5cm}\begin{sideways}\hspace{0.7cm}{\bf Ours-2F}\end{sideways} & 
			\hspace{-0.35cm}\includegraphics[width=0.12\linewidth]{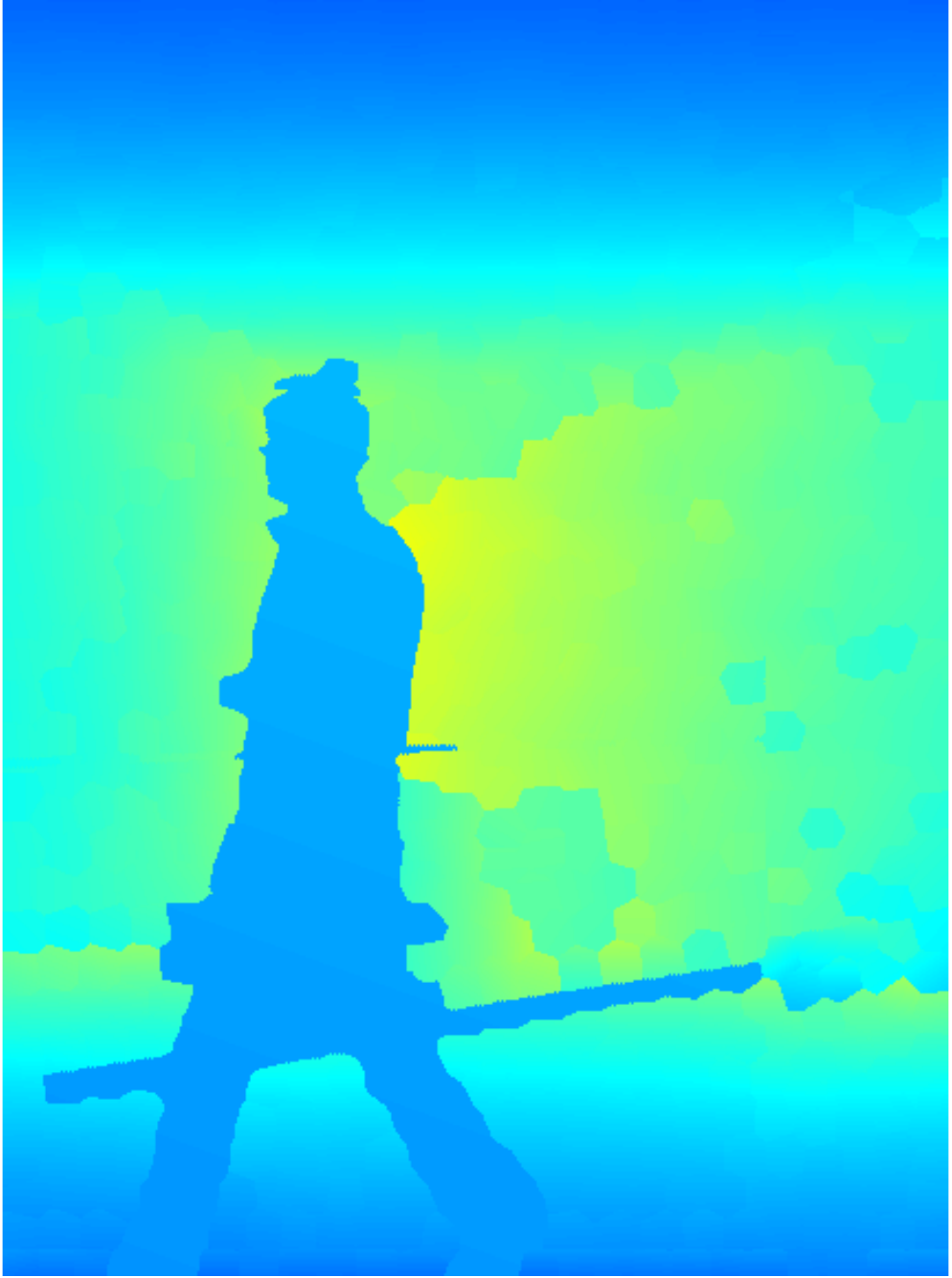} &
			\hspace{-0.35cm}\includegraphics[width=0.12\linewidth]{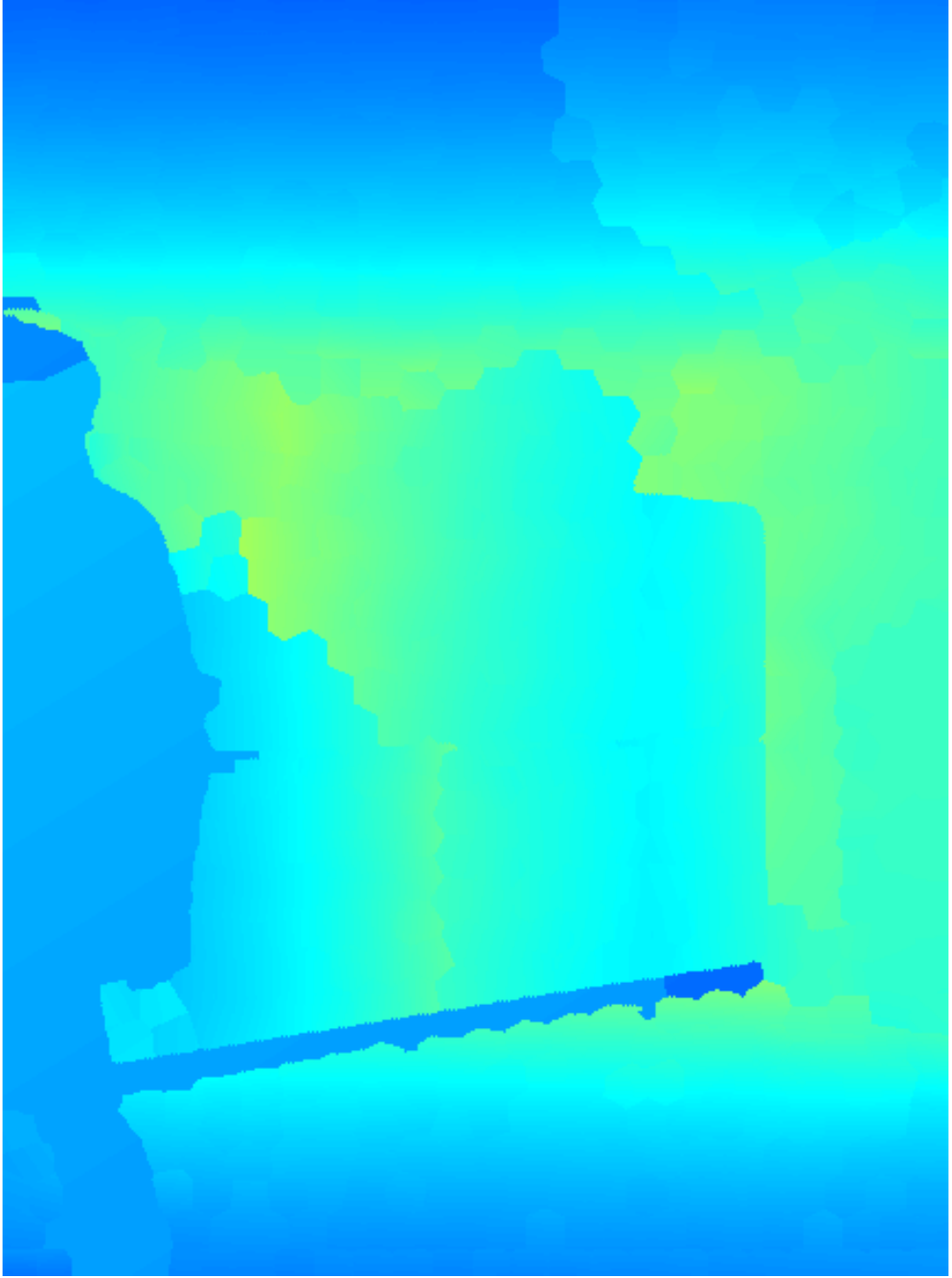} &
			\hspace{-0.35cm}\includegraphics[width=0.12\linewidth]{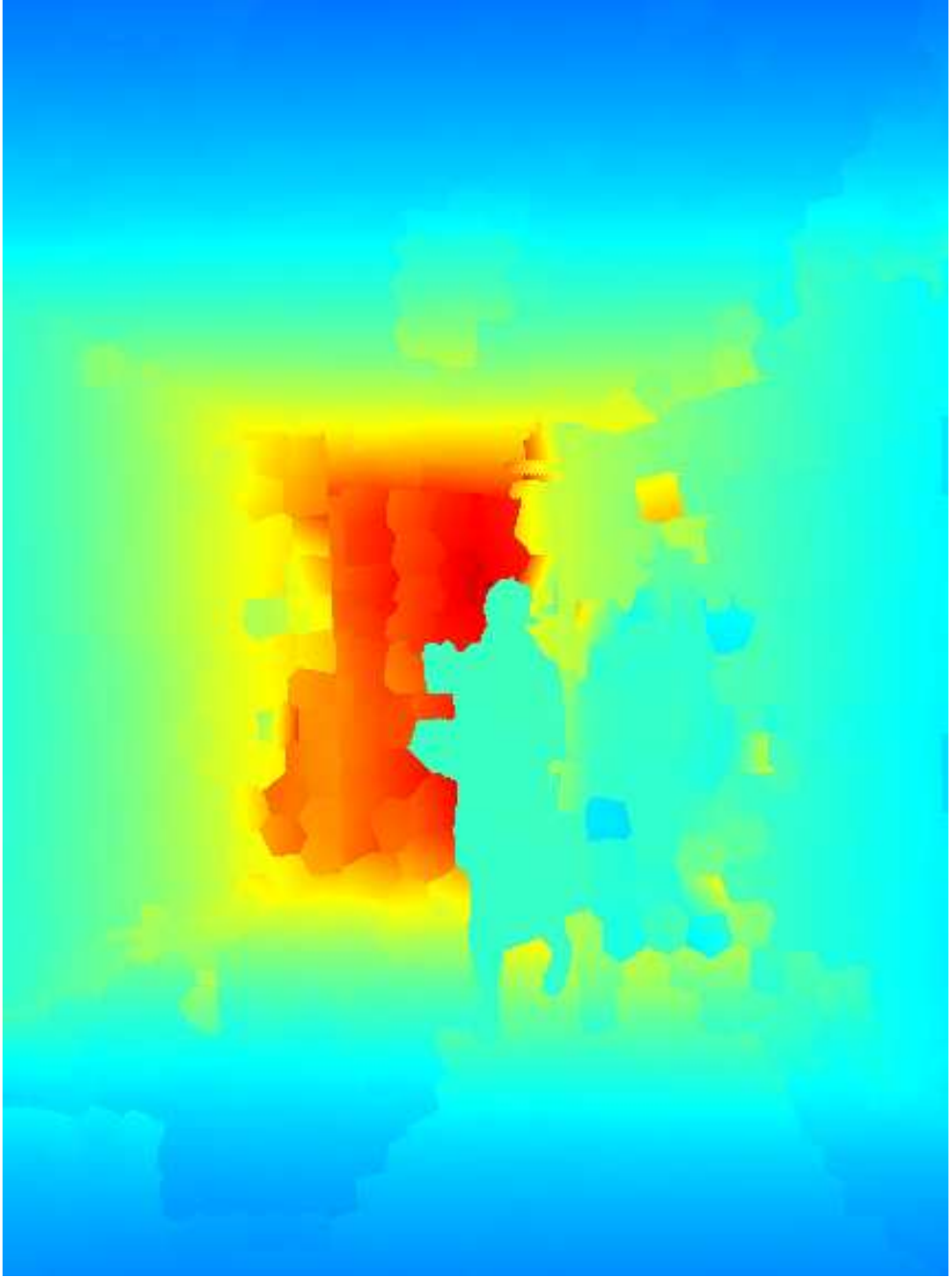} &
			\hspace{-0.35cm}\includegraphics[width=0.12\linewidth]{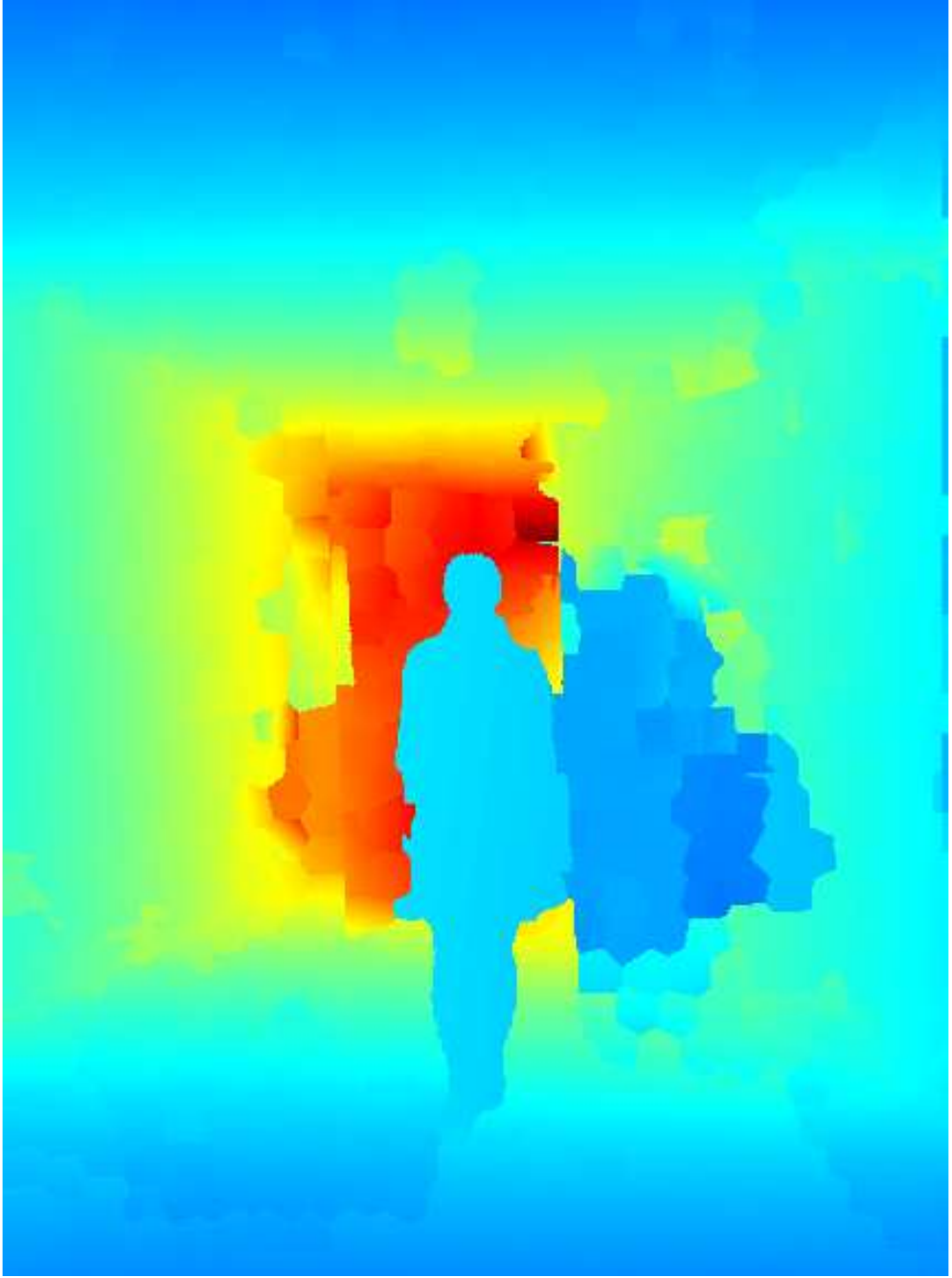} &
			\hspace{-0.35cm}\includegraphics[width=0.12\linewidth]{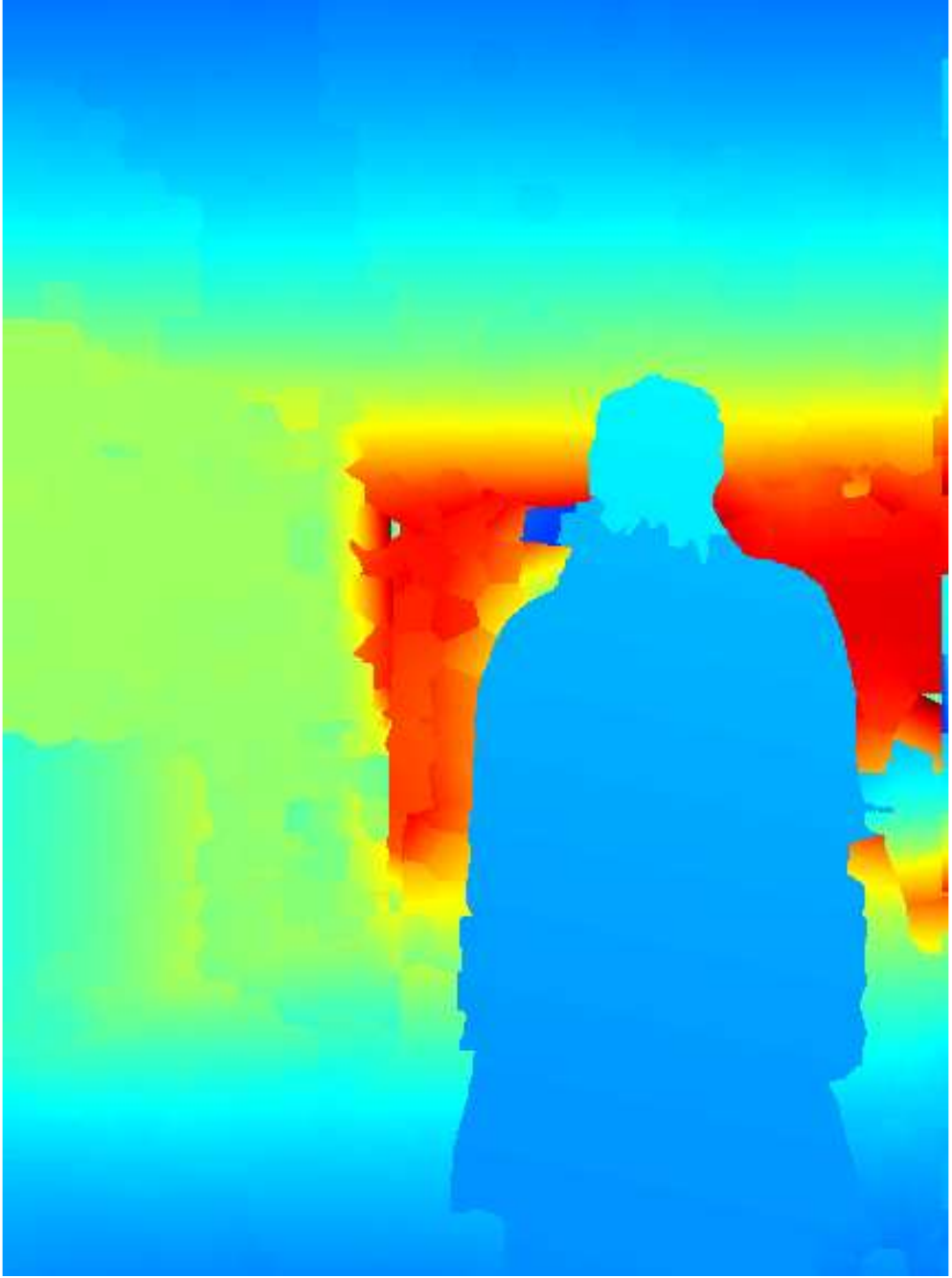} &
			\hspace{-0.35cm}\includegraphics[width=0.12\linewidth]{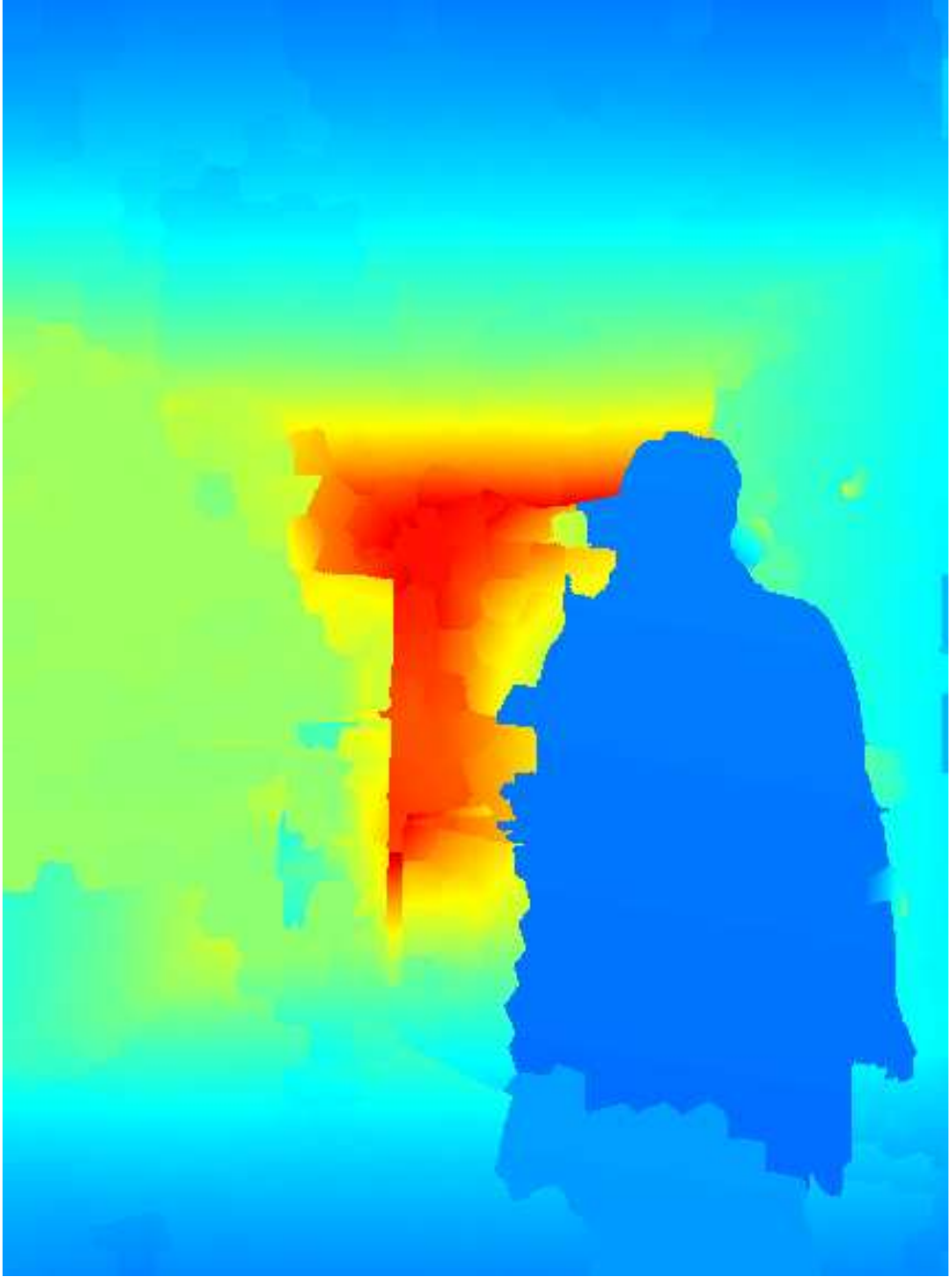} &
			\hspace{-0.35cm}\includegraphics[width=0.12\linewidth]{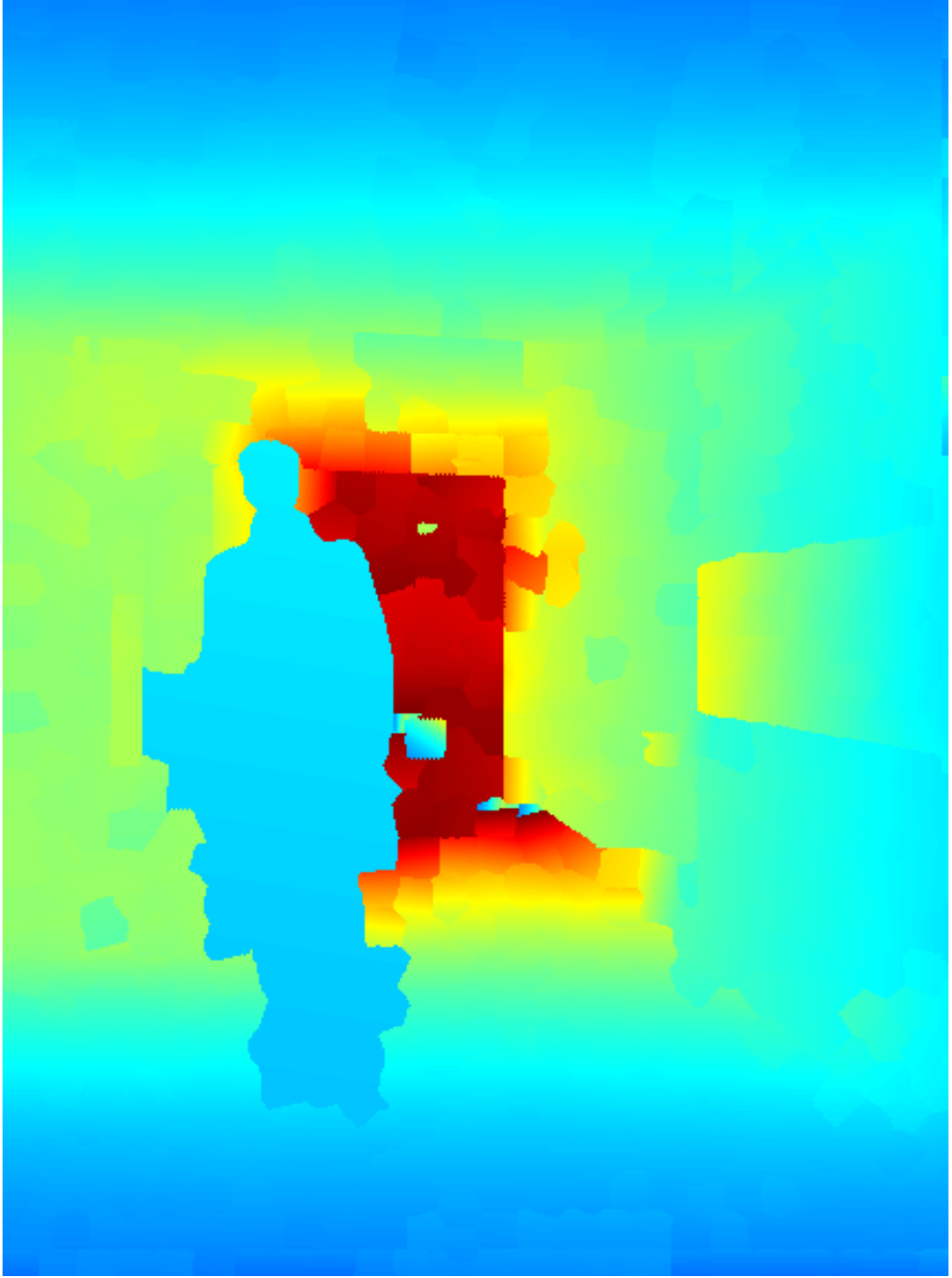}&
			\hspace{-0.35cm}\includegraphics[width=0.12\linewidth]{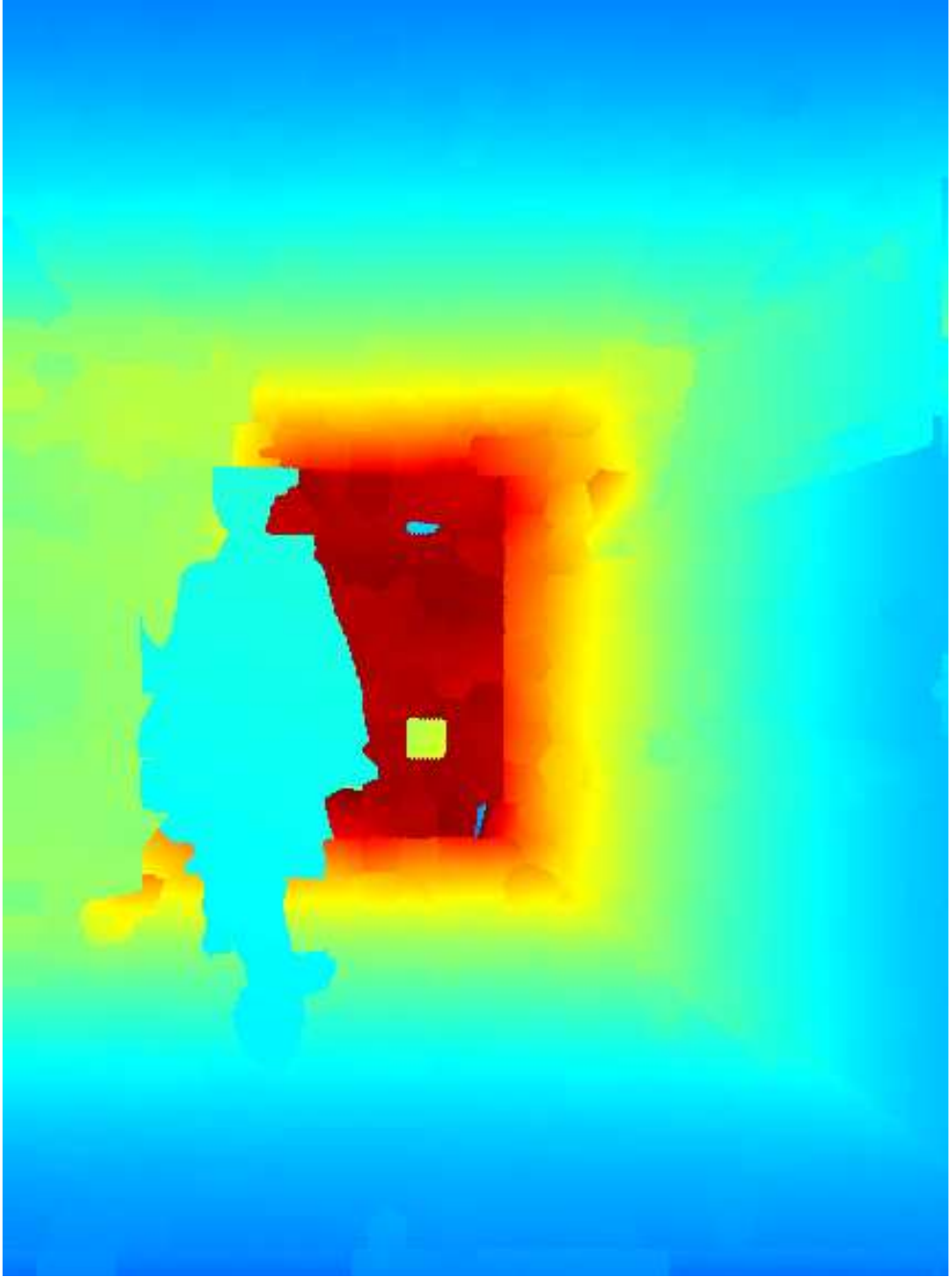}\\
			\hspace{-0.5cm}\begin{sideways}\hspace{0.9cm}{\bf Ours-Vid}\end{sideways} & 
			\hspace{-0.35cm}\includegraphics[width=0.12\linewidth]{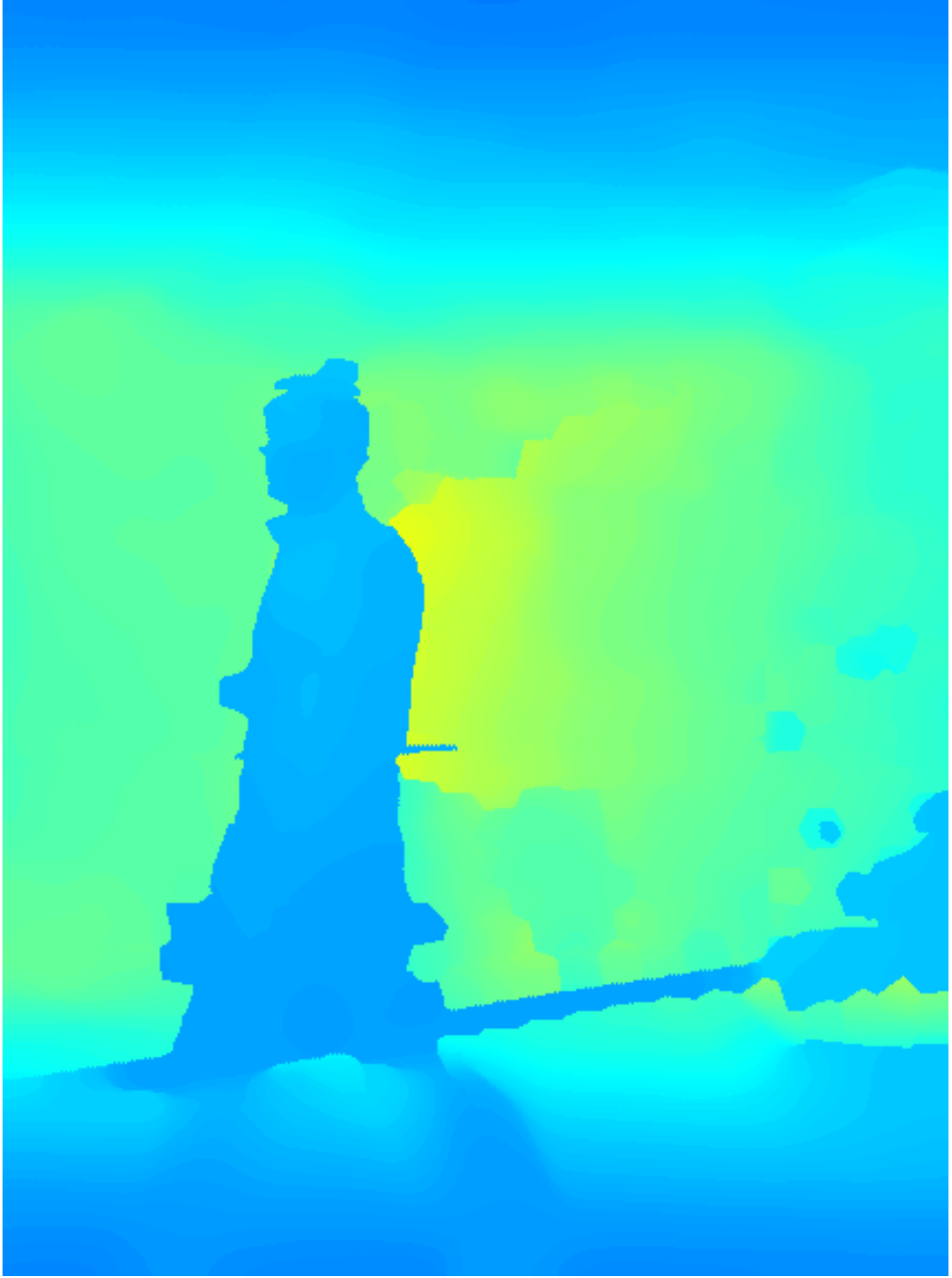} &
			\hspace{-0.35cm}\includegraphics[width=0.12\linewidth]{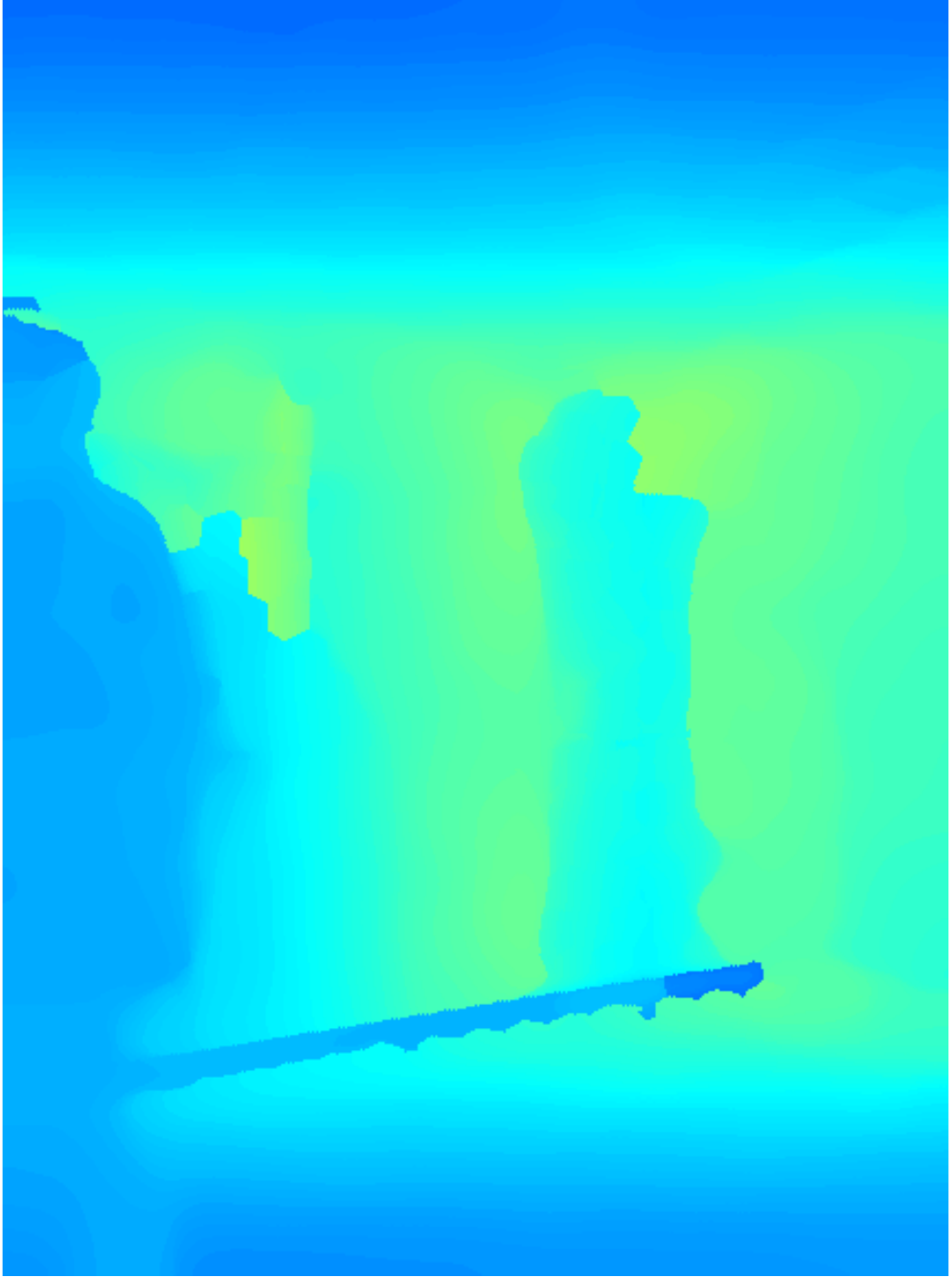} &
			\hspace{-0.35cm}\includegraphics[width=0.12\linewidth]{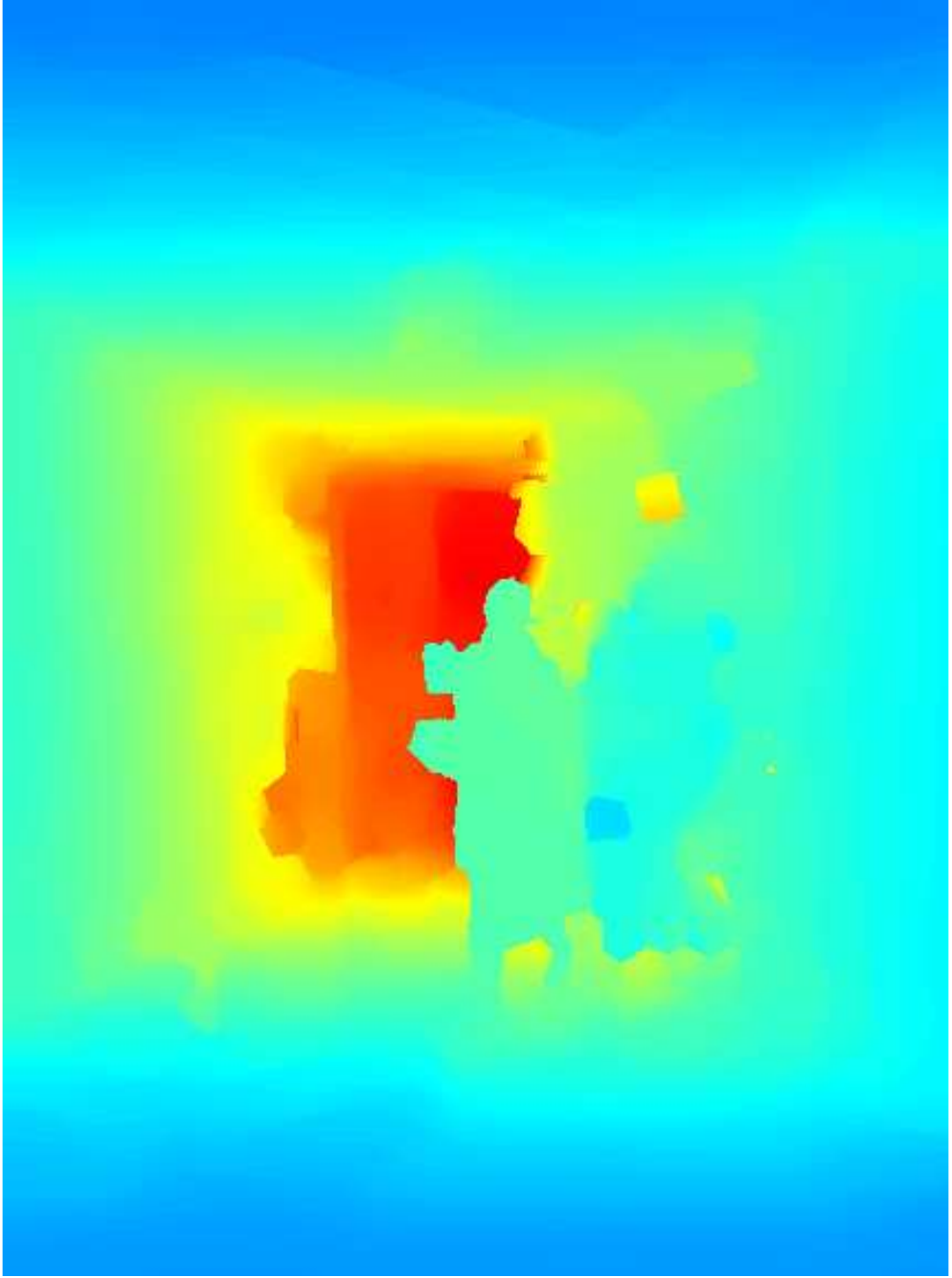} &
			\hspace{-0.35cm}\includegraphics[width=0.12\linewidth]{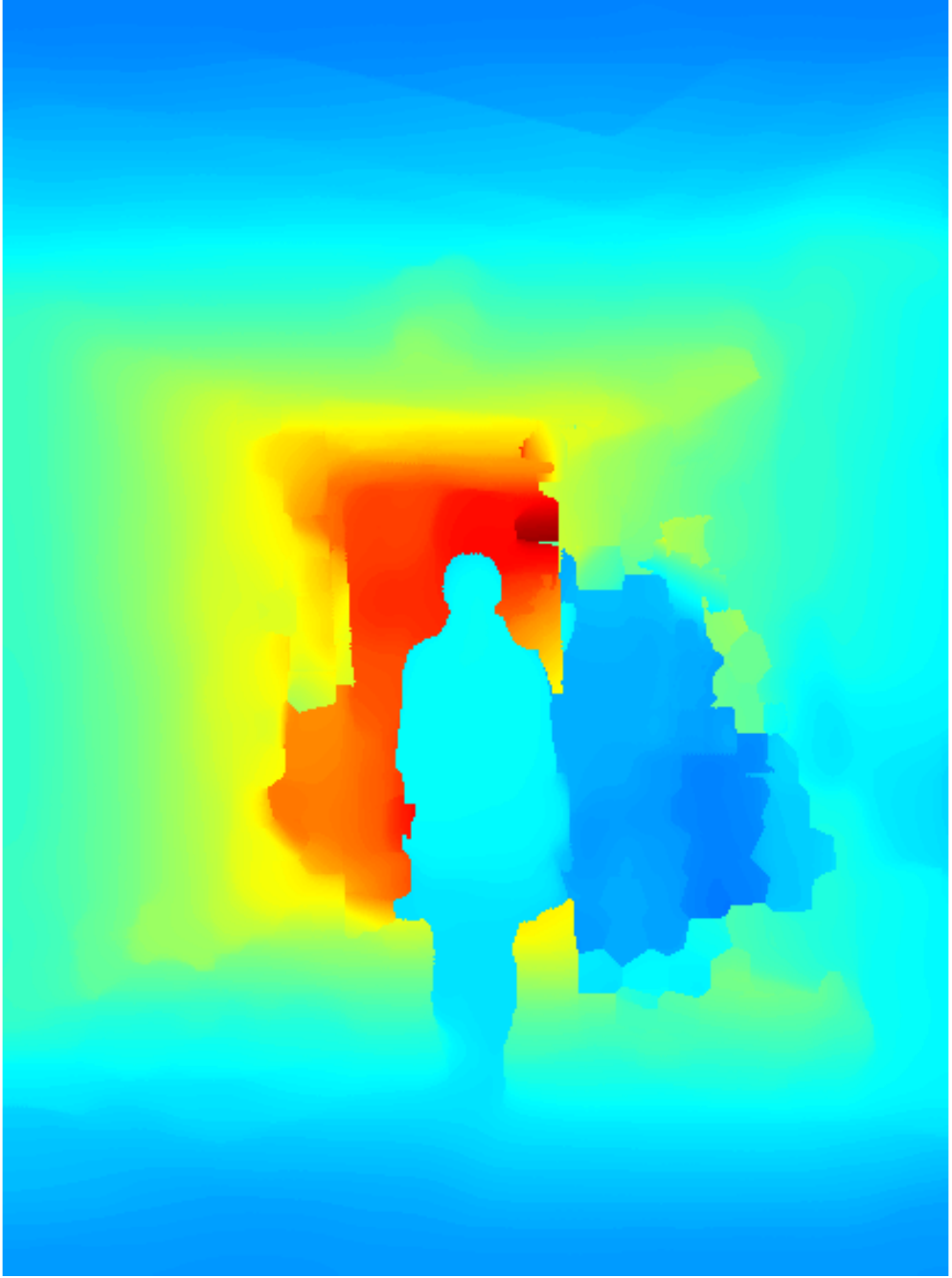} &
			\hspace{-0.35cm}\includegraphics[width=0.12\linewidth]{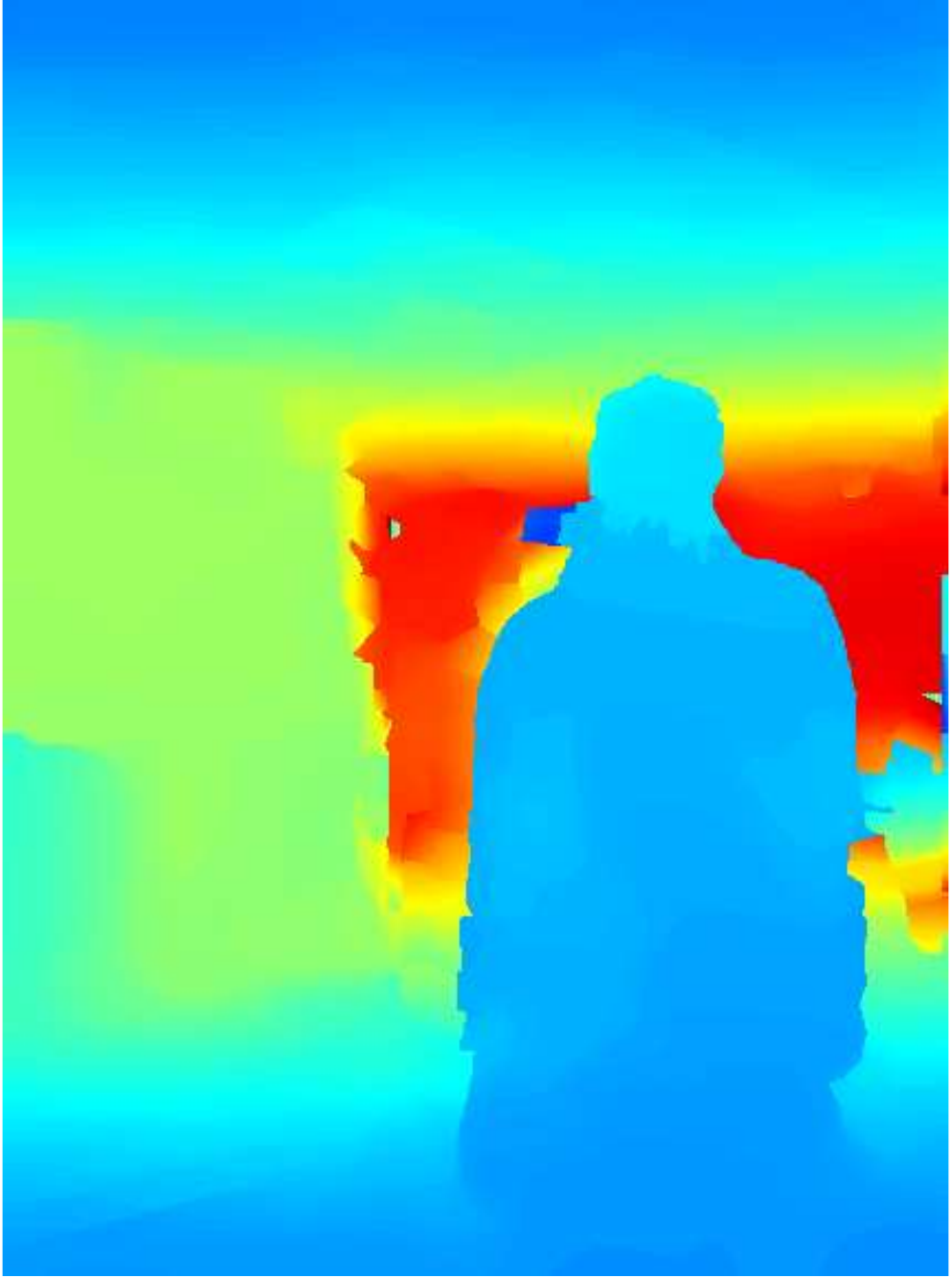} &
			\hspace{-0.35cm}\includegraphics[width=0.12\linewidth]{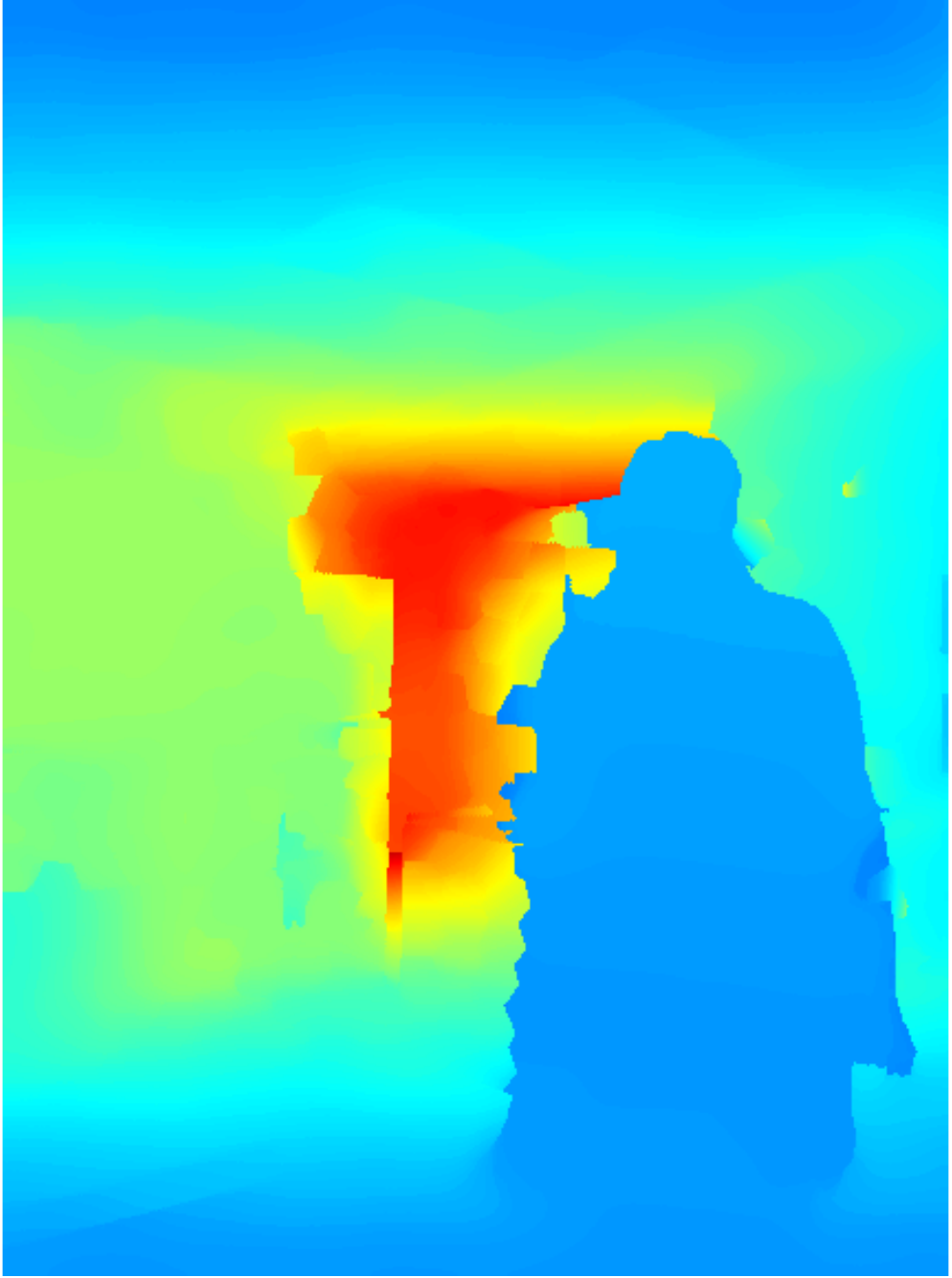} &
			\hspace{-0.35cm}\includegraphics[width=0.12\linewidth]{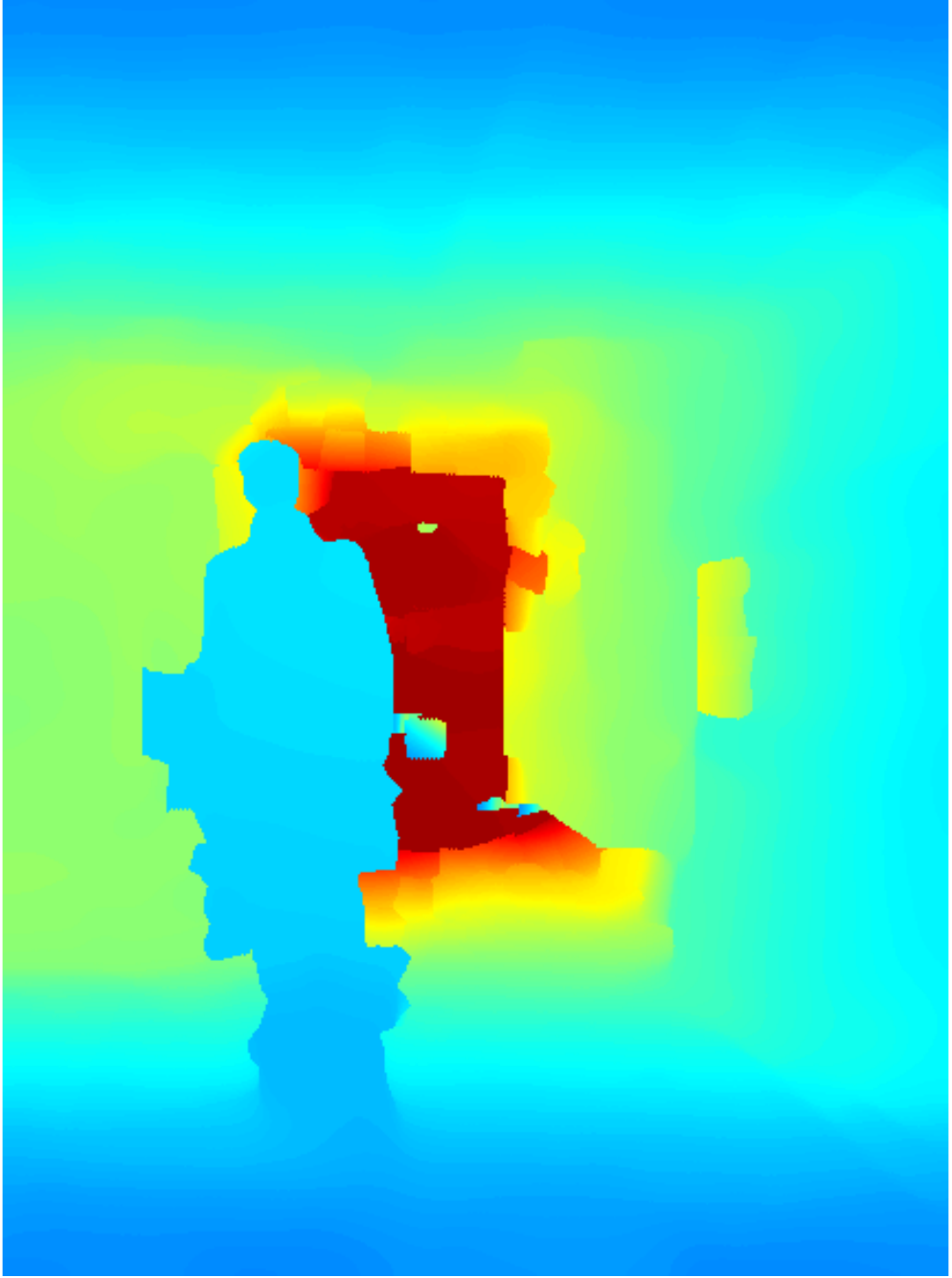}&
			\hspace{-0.35cm}\includegraphics[width=0.12\linewidth]{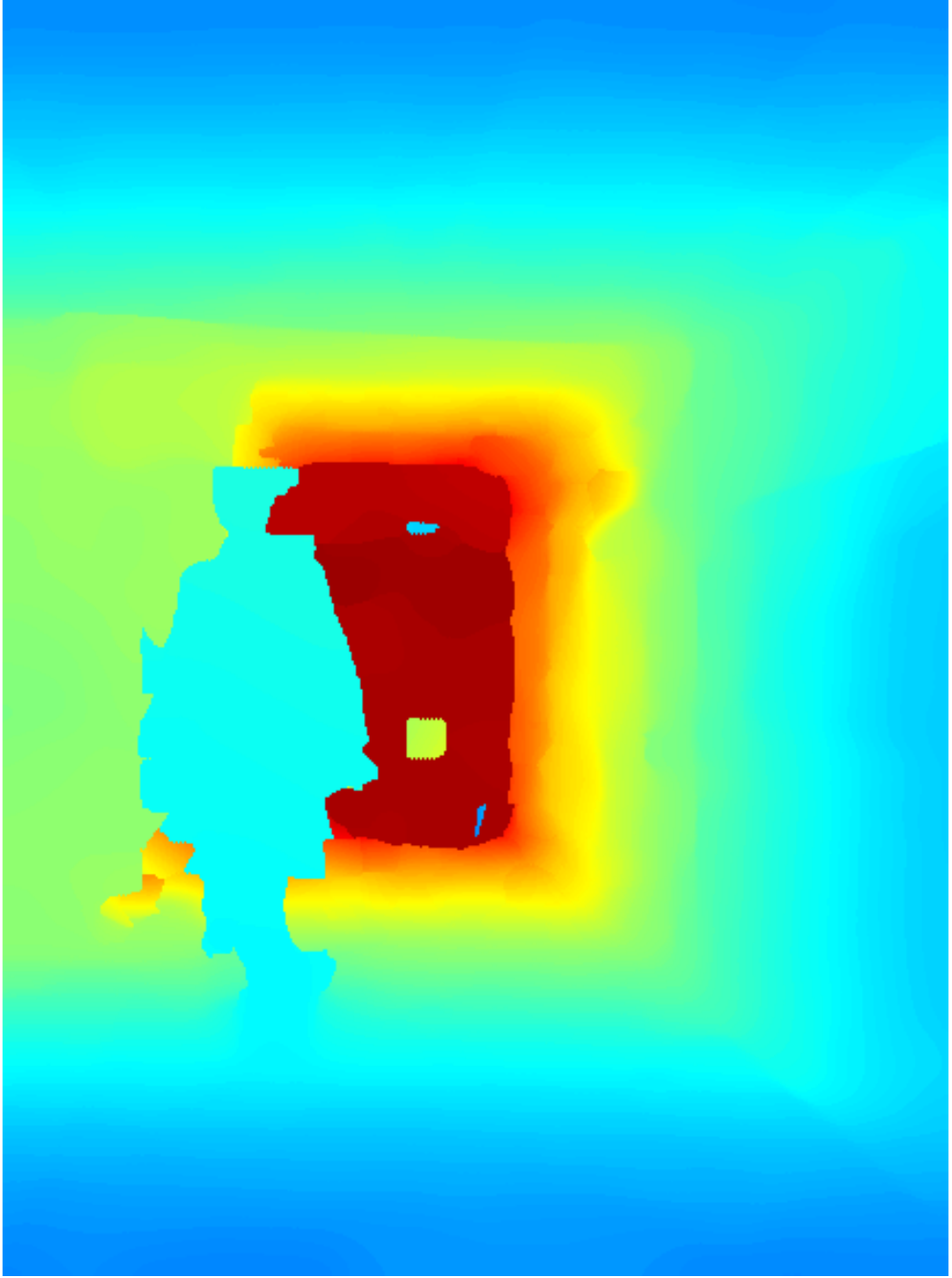}
		\end{tabular}
	\end{small}
	\vspace{-0.2cm}
	\caption{MSR-V3D: Qualitative comparison of all the methods. Note that our single-frame method lacks smoothness, both spatial and temporal. In contrast, the video-based depth transfer approach of~\cite{Karsch12} tends to oversmooth the results. Our two-frame CRF (Ours-2F) and our final model (Ours-Vid) both yield good spatial and temporal smoothness, while preserving the discontinuities of the depth maps. Note that, while not obvious at this scale, our final model does yield smoother results and better accuracy at the object boundaries.}
	\label{fig:cvprcomp}
\end{figure*}

\vspace{0.2cm}
\noindent {\bf Comparison of our final model with~\cite{Karsch12}:}\mbox{}\\
We first compare our final model with the results of the video-based depth transfer approach of~\cite{Karsch12} on the same dataset. To be able to compare our numbers directly with those published in~\cite{Karsch12}, we follow the same global rescaling procedure as reported in their paper, i.e., rescale the depth in the range 1m--81m\footnote{This rescaling procedure was introduced in~\cite{Karsch12} to be able to compare errors with those obtained on the Make3D dataset.}. The three error metrics for both methods are reported in Table~\ref{table:completeModel}. Note that we report our results without and with applying the depth validity mask provided with the dataset\footnote{It is not clearly stated in~\cite{Karsch12} whether the mask was used or not.}.  As can be observed from the table, we outperform~\cite{Karsch12} in most cases, independently of whether we consider the mask or not.
In Fig.~\ref{fig:cvprcomp}, we provide a qualitative comparison on some images of each building in the test set. Note that our approach nicely models the moving foreground objects, while still respecting the depth discontinuities.

\begin{figure}[t]
	\begin{small}
		\begin{tabular}{ccccc}
			&&& \hspace{-2.5cm}{\bf Building 1} \\
			\hspace{-0.2cm}\begin{sideways}\hspace{0.8cm}{\bf Image}\end{sideways} & 
			\hspace{-0.35cm}\includegraphics[width=0.23\linewidth]{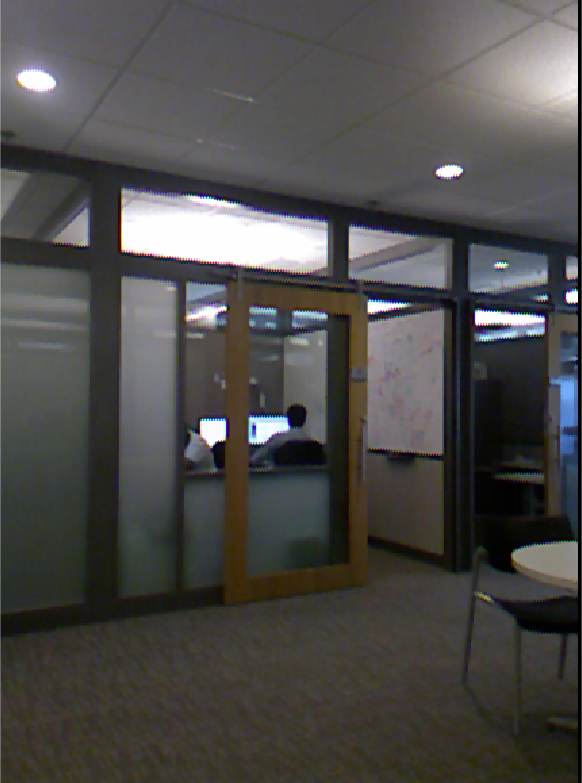} &
			\hspace{-0.35cm}\includegraphics[width=0.23\linewidth]{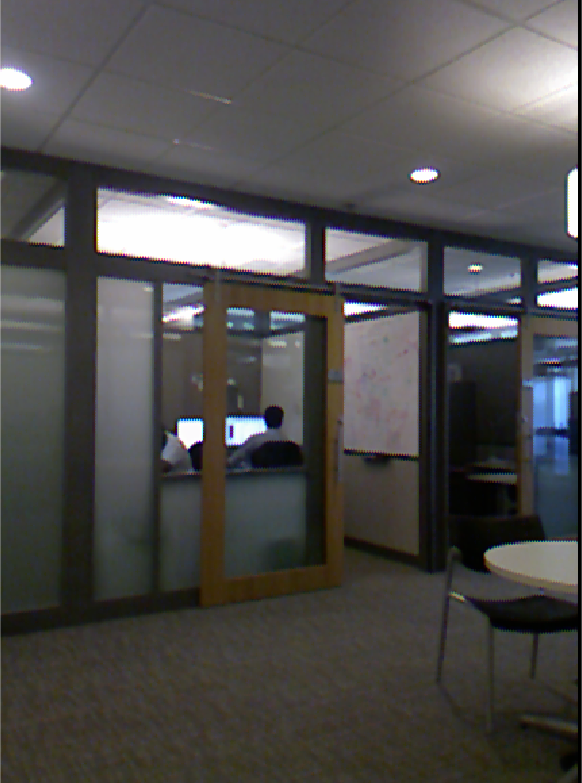} &
			\hspace{-0.35cm}\includegraphics[width=0.23\linewidth]{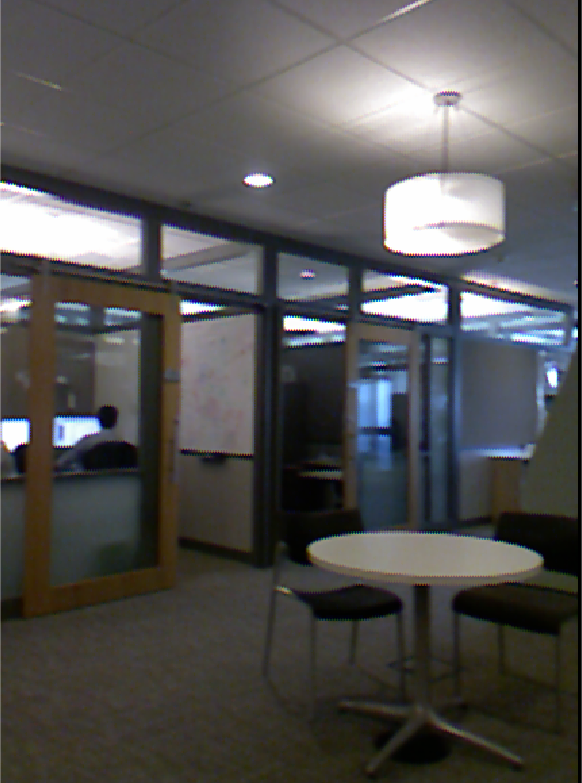} &
			\hspace{-0.35cm}\includegraphics[width=0.23\linewidth]{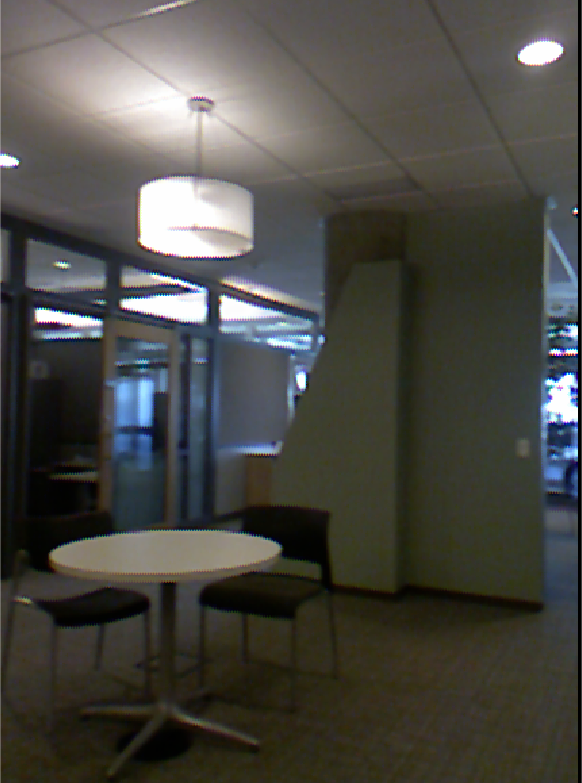}\\
			\hspace{-0.2cm}\begin{sideways}\hspace{0.4cm}{\bf Ground-Truth}\end{sideways} & 
			\hspace{-0.35cm}\includegraphics[width=0.23\linewidth]{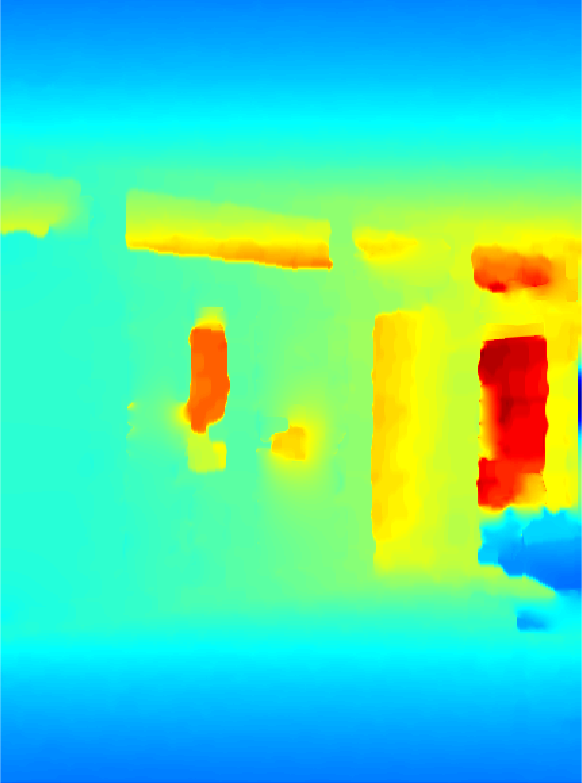} &
			\hspace{-0.35cm}\includegraphics[width=0.23\linewidth]{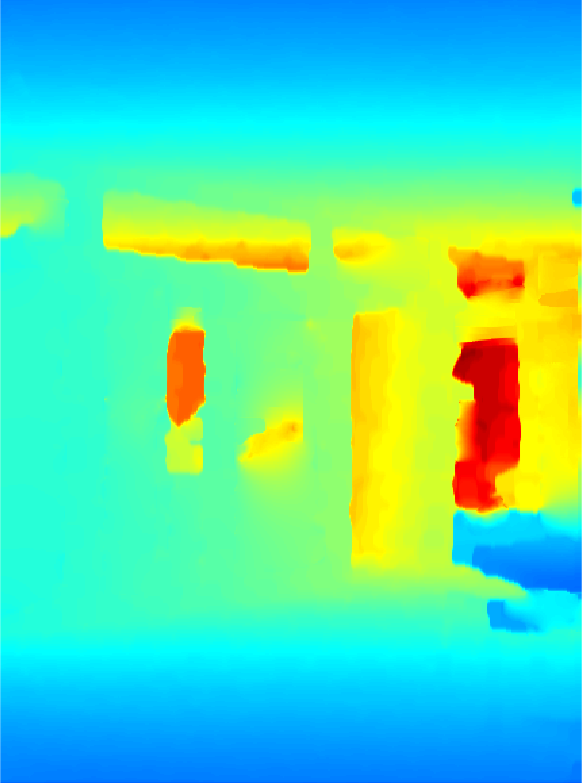} &
			\hspace{-0.35cm}\includegraphics[width=0.23\linewidth]{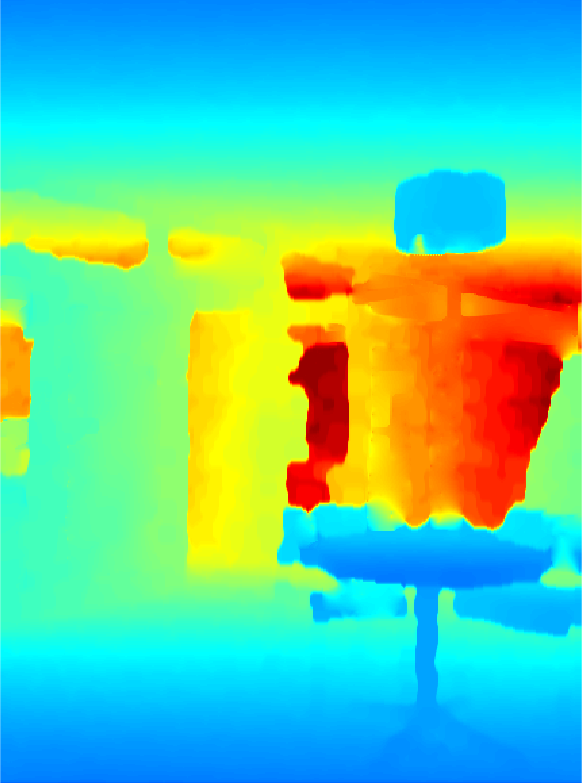} &
			\hspace{-0.35cm}\includegraphics[width=0.23\linewidth]{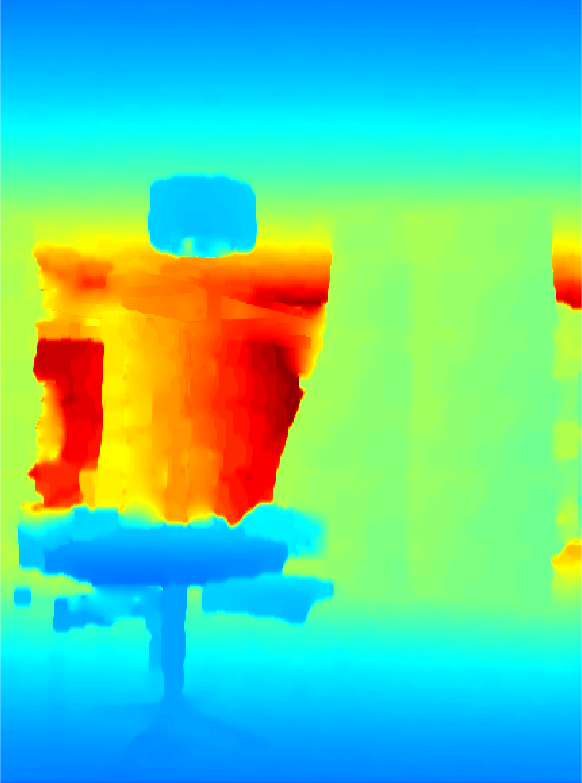} \\
			\hspace{-0.2cm}\begin{sideways}\hspace{0.1cm}{\bf DepthTransfer~\cite{Karsch12}}\end{sideways} & 
			\hspace{-0.35cm}\includegraphics[width=0.23\linewidth]{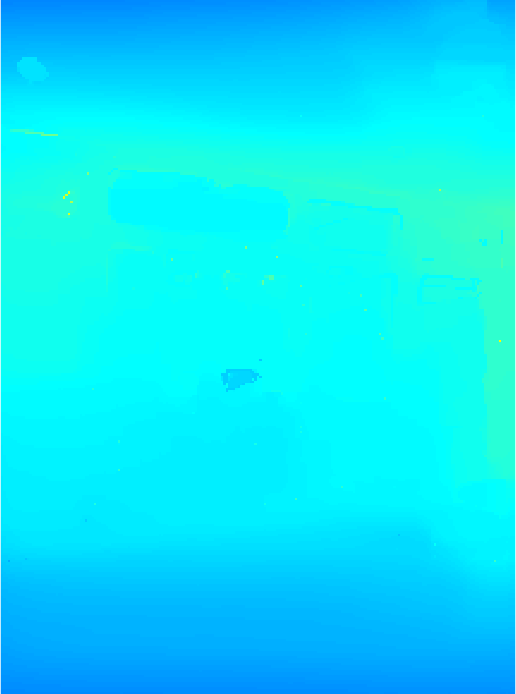} &
			\hspace{-0.35cm}\includegraphics[width=0.23\linewidth]{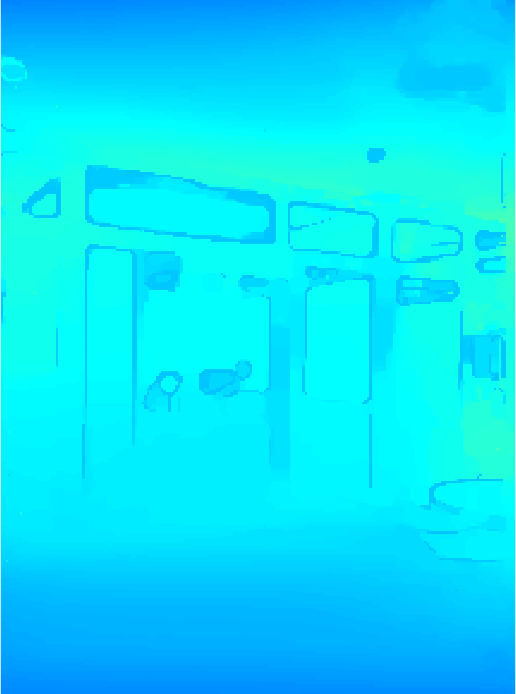} &
			\hspace{-0.35cm}\includegraphics[width=0.23\linewidth]{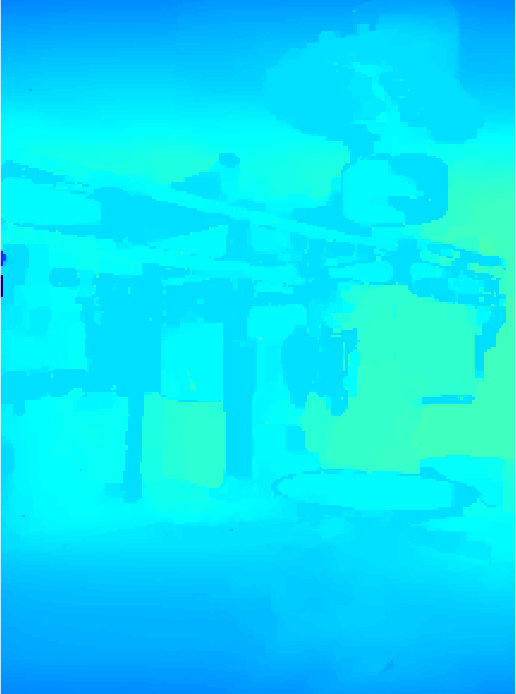} &
			\hspace{-0.35cm}\includegraphics[width=0.23\linewidth]{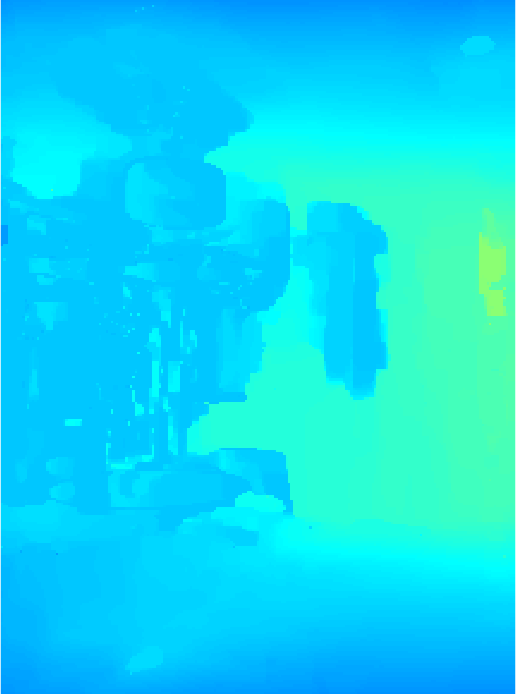}\\
			\hspace{-0.2cm}\begin{sideways}\hspace{0.6cm}{\bf Ours-Vid}\end{sideways} & 
			\hspace{-0.35cm}\includegraphics[width=0.23\linewidth]{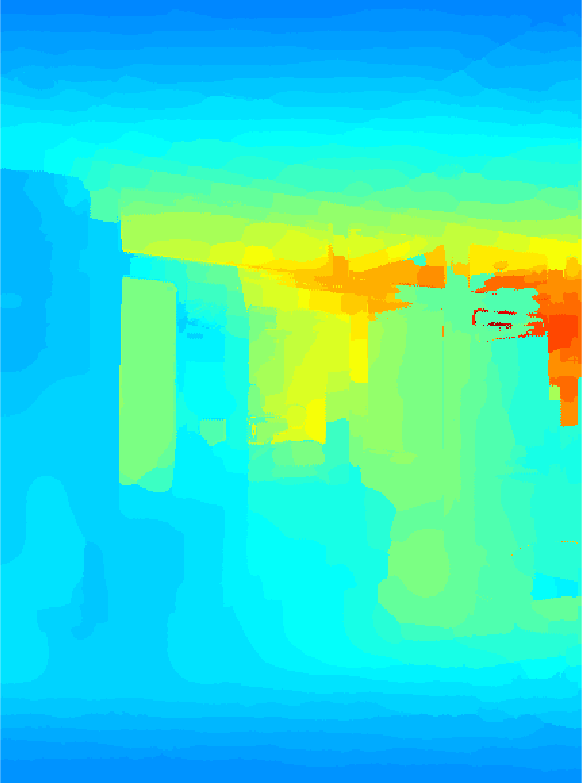} &
			\hspace{-0.35cm}\includegraphics[width=0.23\linewidth]{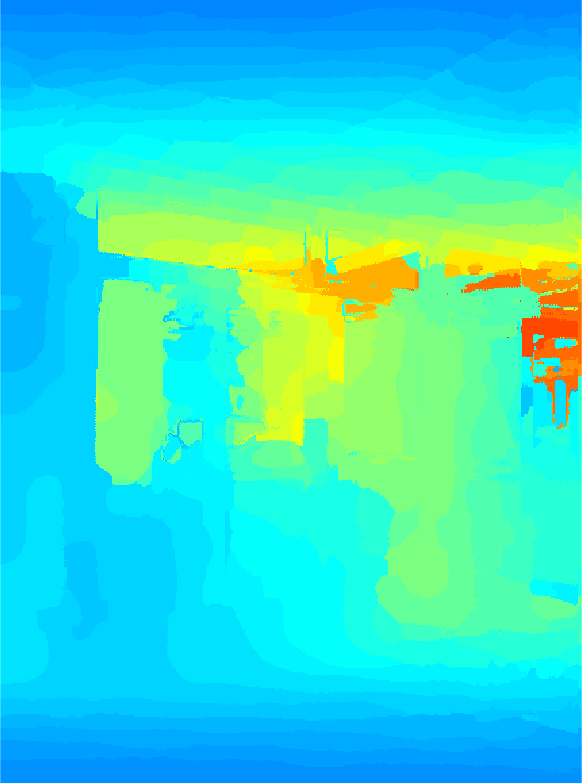} &
			\hspace{-0.35cm}\includegraphics[width=0.23\linewidth]{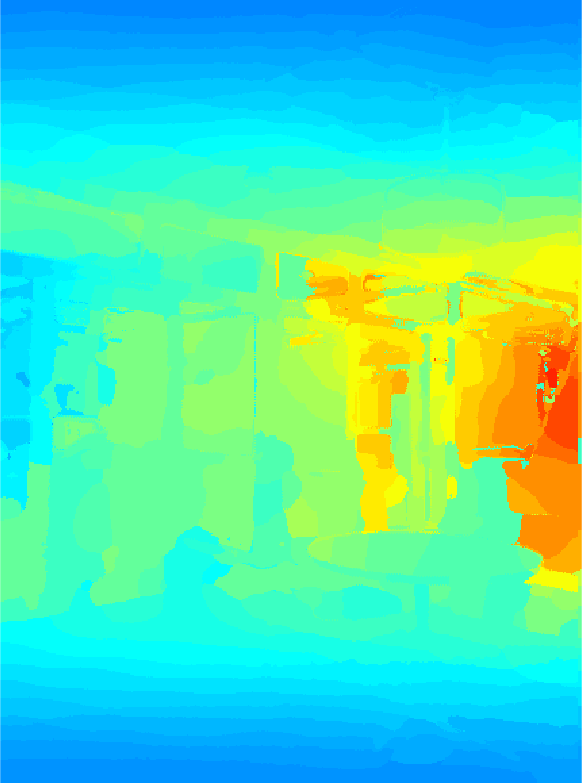} &
			\hspace{-0.35cm}\includegraphics[width=0.23\linewidth]{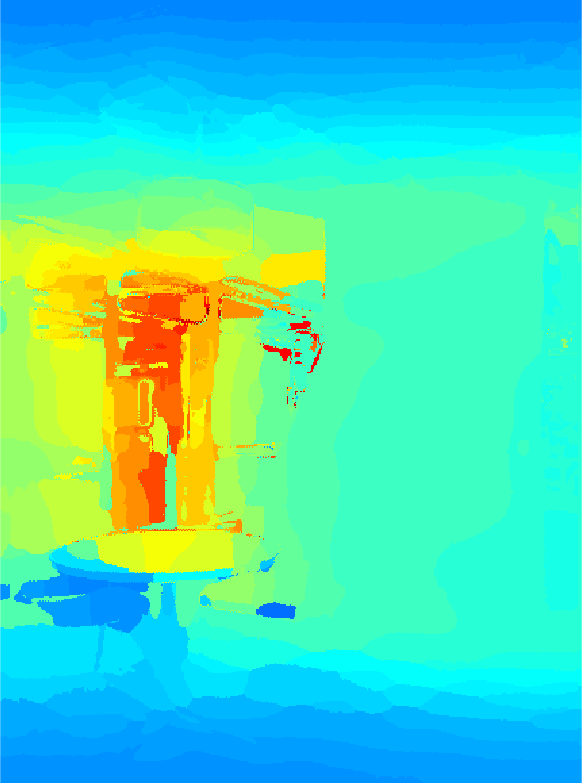}
		\end{tabular}
	\end{small}
	\vspace{-0.2cm}
	\caption{MSR-V3D: Qualitative comparison on rotational video sequences. Note that the video-based depth transfer approach of~\cite{Karsch12} tends to over-smooth the results. Our final model (Ours-Vid) yields both spatial and temporal smoothness, while preserving the discontinuities of the depth maps.}
	\label{fig:cvprcompRot}
\end{figure}

The MSR-V3D dataset has the property that the test set only contains sequences depicting dynamic scenes and acquired with a fixed camera; All the sequences acquired with a rotating camera are part of the training set (i.e., Building 1). To evaluate our approach on these sequences, we therefore followed a leave-one-sequence-out strategy on the 24 sequences of Building 1 that were acquired by a rotational camera. In Table~\ref{tab:purerotation}, we report the different error metrics on these sequences. 
 A qualitative comparison with the video DepthTransfer approach of~\cite{Karsch12} is provided in Fig.~\ref{fig:cvprcompRot}.


\begin{table}[t!]
	\centering
	\begin{tabular}{|c |c | c| c| c|}
		\hline
		Dataset & & ${\bf rel}$ & ${\bf log}_{10}$ & ${\bf rms}$ \\ 
		\hline
		\multirow{3}{*}{Building 2} &~\cite{Karsch12}&0.394 & {\bf 0.135} & {\bf 11.7}\\
		& Ours-Vid (no mask)& {\bf 0.336} &  0.255 & 25.2\\ 
		& Ours-Vid (mask)& {\bf 0.242} & 0.136 & 17.7\\ 
		\hline
		\multirow{3}{*}{Building 3}&~\cite{Karsch12}&  0.325 & 0.159 & 15.0\\
		&Ours-Vid (no mask)& {\bf 0.164} & {\bf 0.071} & {\bf 9.94} \\
		&Ours-Vid (mask)& {\bf 0.227} & {\bf 0.078} & {\bf 9.82}\\
		\hline
		\multirow{3}{*}{Building 4}&~\cite{Karsch12}& 0.251 & 0.136 & 15.3\\
		&Ours-Vid (no mask)& 0.241& {\bf 0.115} & {\bf 14.1}\\
		&Ours-Vid (mask)& 0.253&  {\bf 0.111} & {\bf 13.3}\\
		\hline
		\multirow{3}{*}{All}&~\cite{Karsch12}& 0.323 & 0.143 & 14.0\\
		&Ours-Vid (no mask)& {\bf 0.247}& 0.147 & 16.4\\
		&Ours-Vid (mask)& {\bf 0.240} & 0.108 &  13.6\\
		\hline
	\end{tabular}
	\vspace{0.2cm}
	\caption{MSR-V3D: Comparison of our final model with the video DepthTransfer of~\cite{Karsch12}. We directly report the number published in~\cite{Karsch12}. Following~\cite{Karsch12}, our depth maps were rescaled in the range 1m--81m. For our approach, (no mask) and (mask) indicate that the errors were computed without and with the pixel validity mask provided with the dataset, respectively (not stated in~\cite{Karsch12}). Note that our approach outperforms~\cite{Karsch12} in most cases.}
	\label{table:completeModel}
\end{table}

\begin{figure}[t!]
	\begin{small}
		\begin{tabular}{cccc}
			\hspace{-0.3cm}\includegraphics[width=0.23\linewidth]{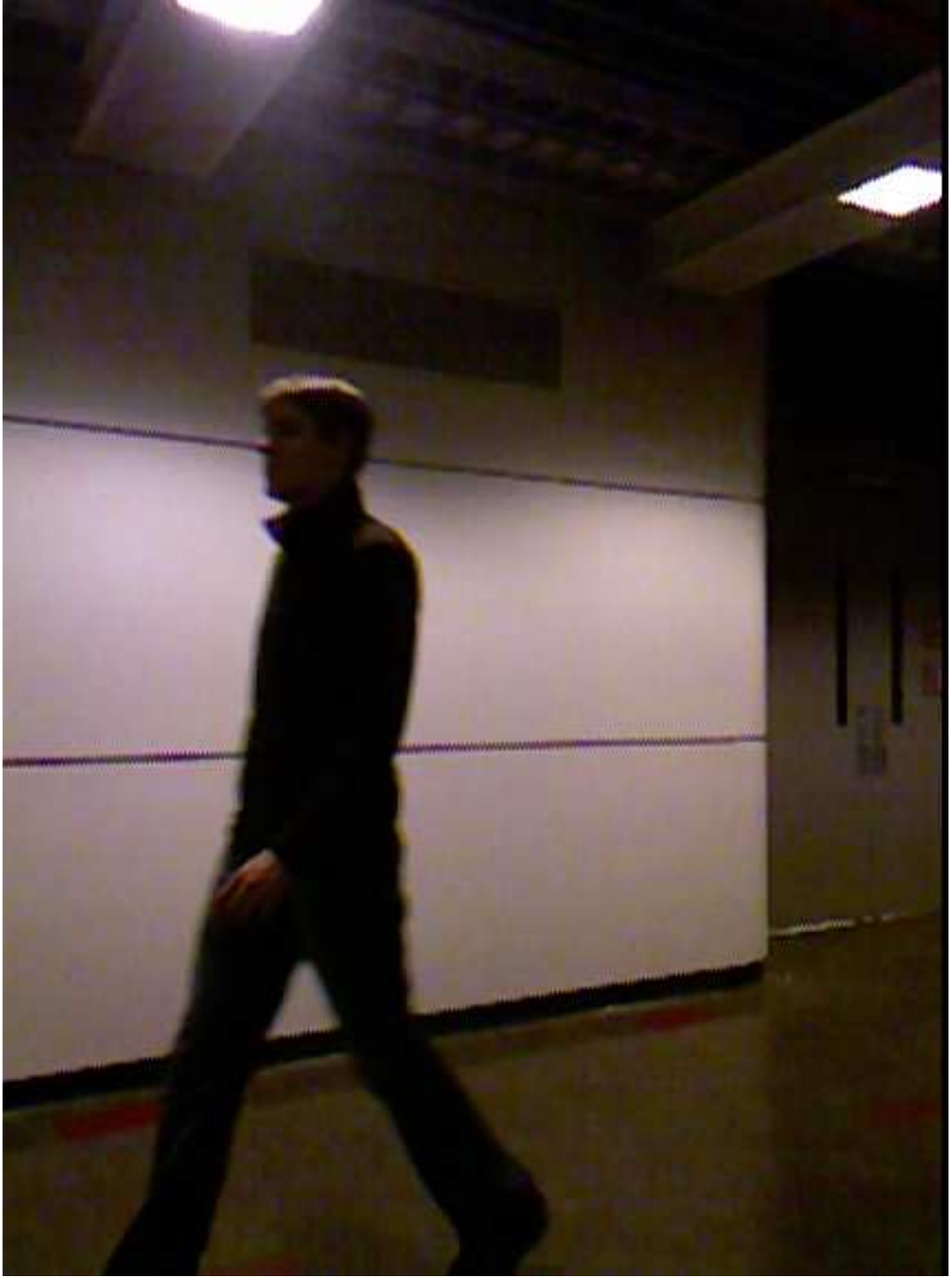} &
			\hspace{-0.3cm}\includegraphics[width=0.23\linewidth]{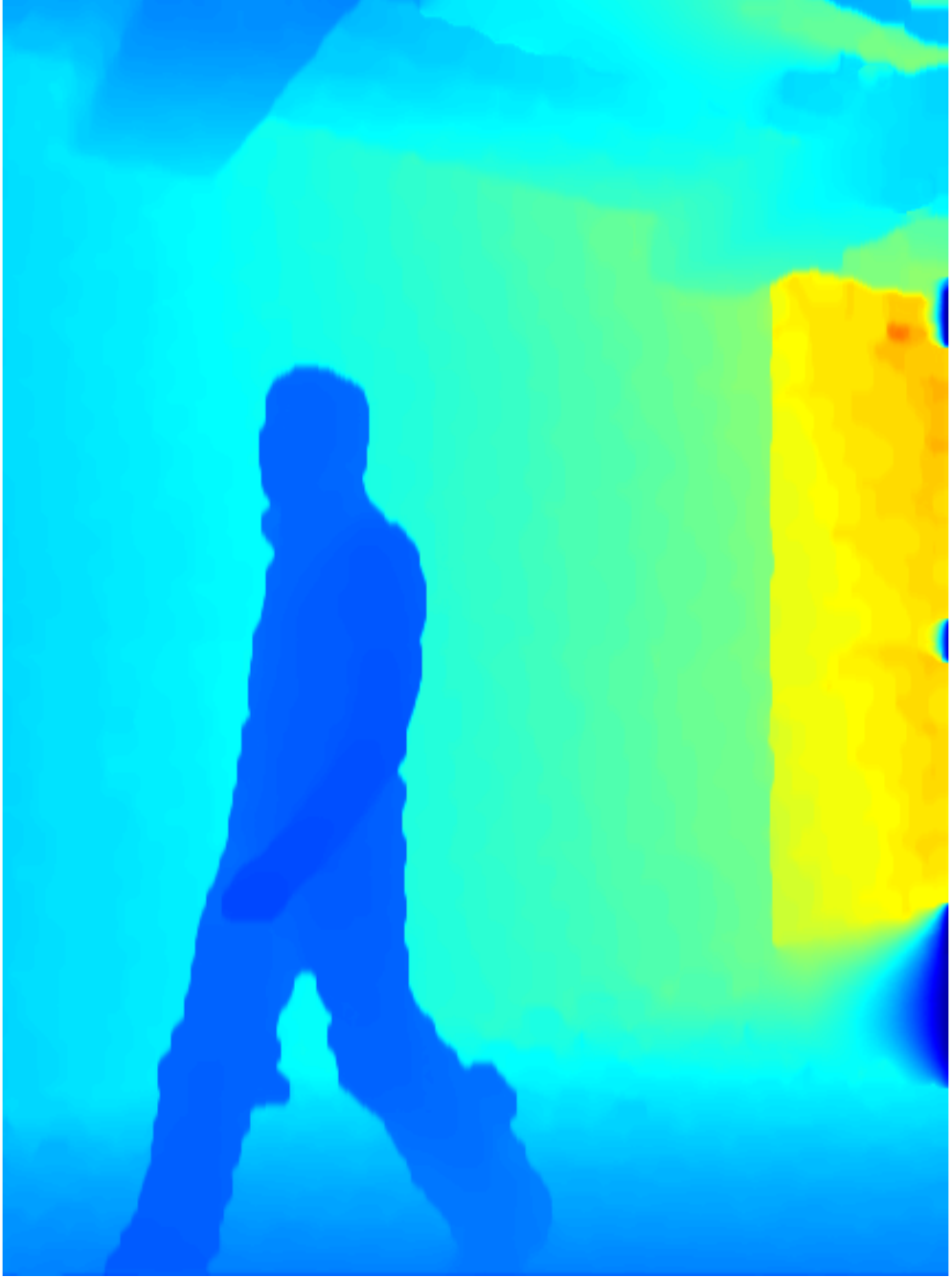} &
			\hspace{-0.3cm}\includegraphics[width=0.23\linewidth]{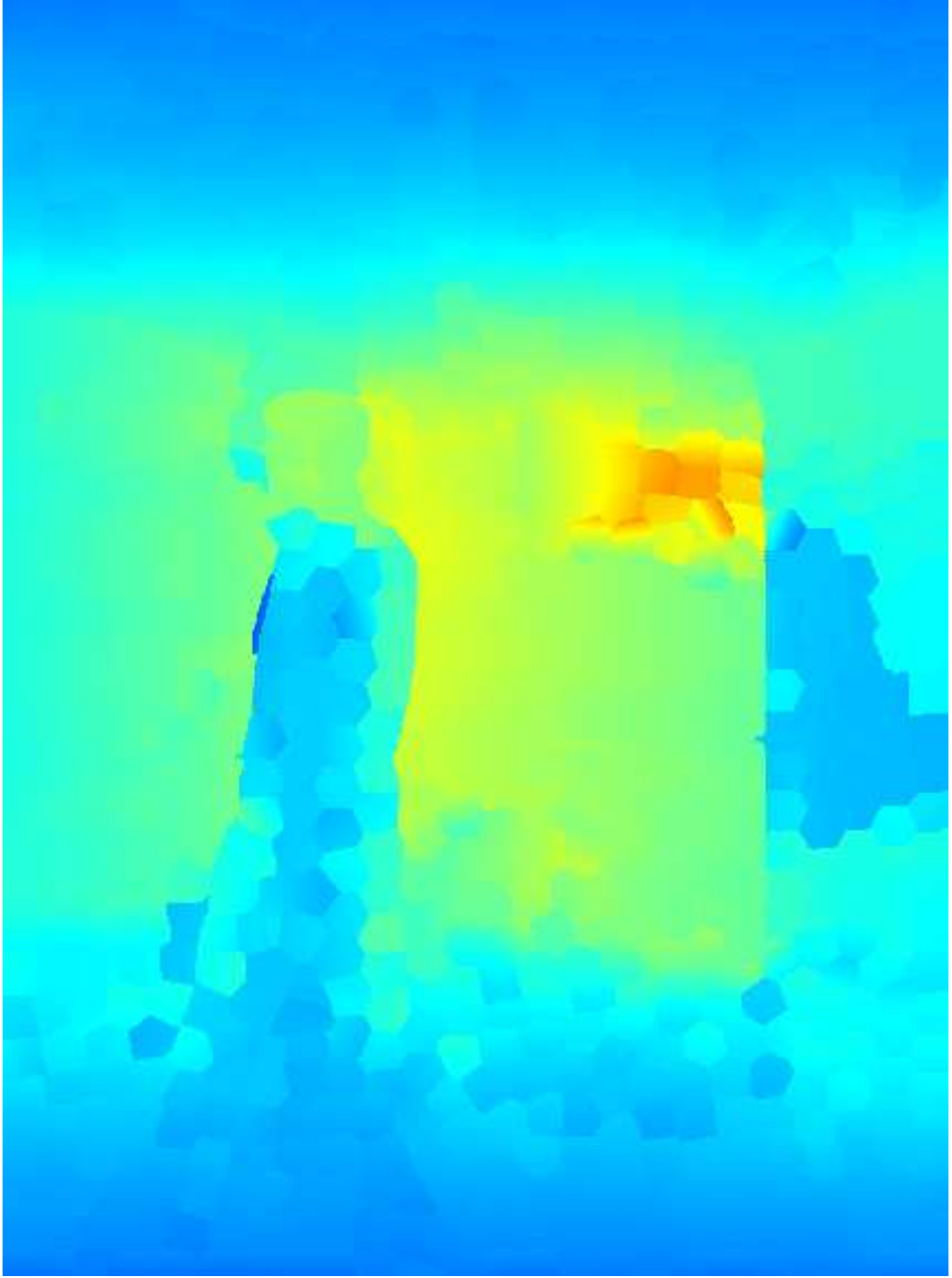} &
			\hspace{-0.3cm}\includegraphics[width=0.23\linewidth]{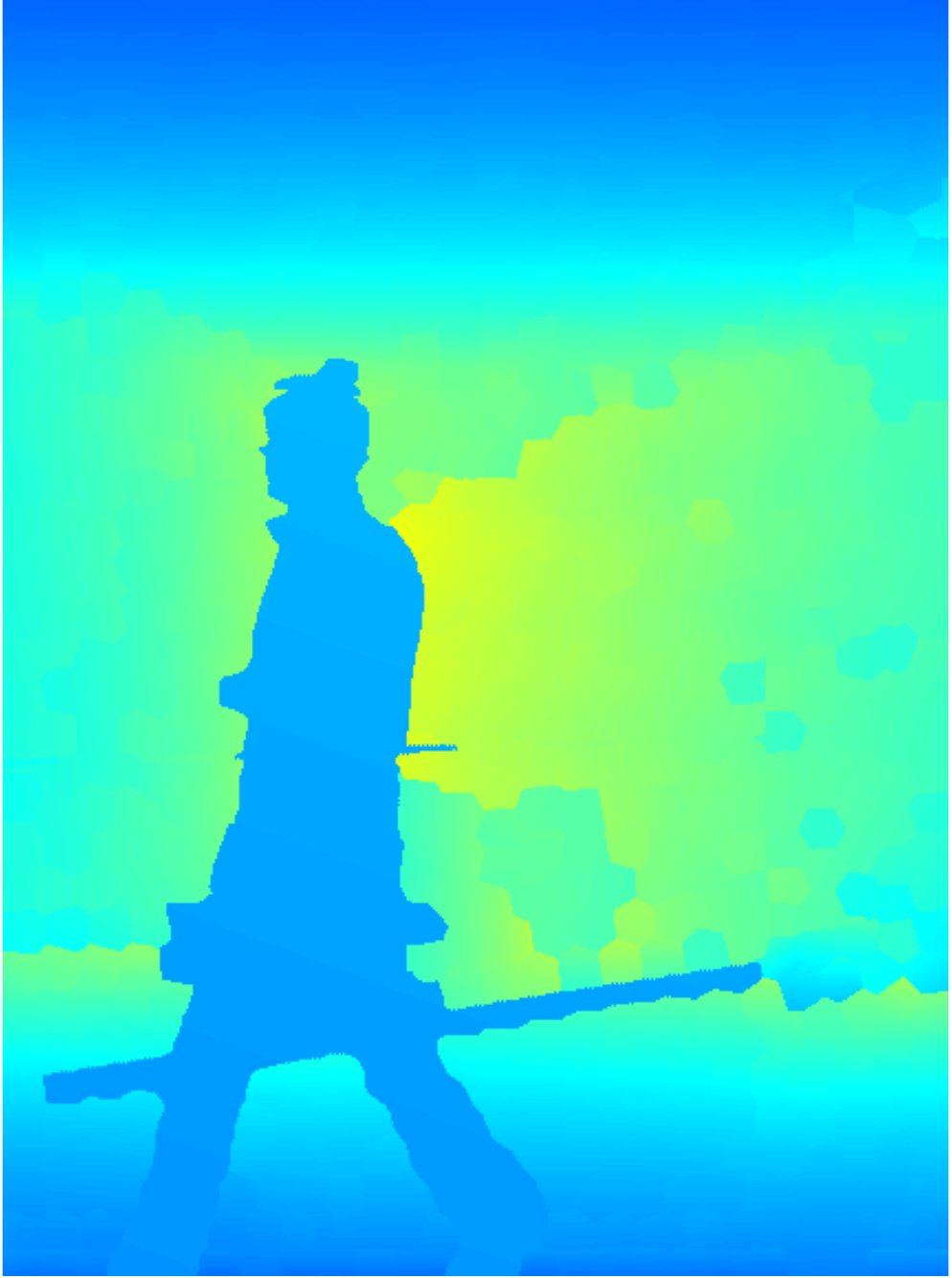}\\
			\hspace{-0.3cm}\includegraphics[width=0.23\linewidth]{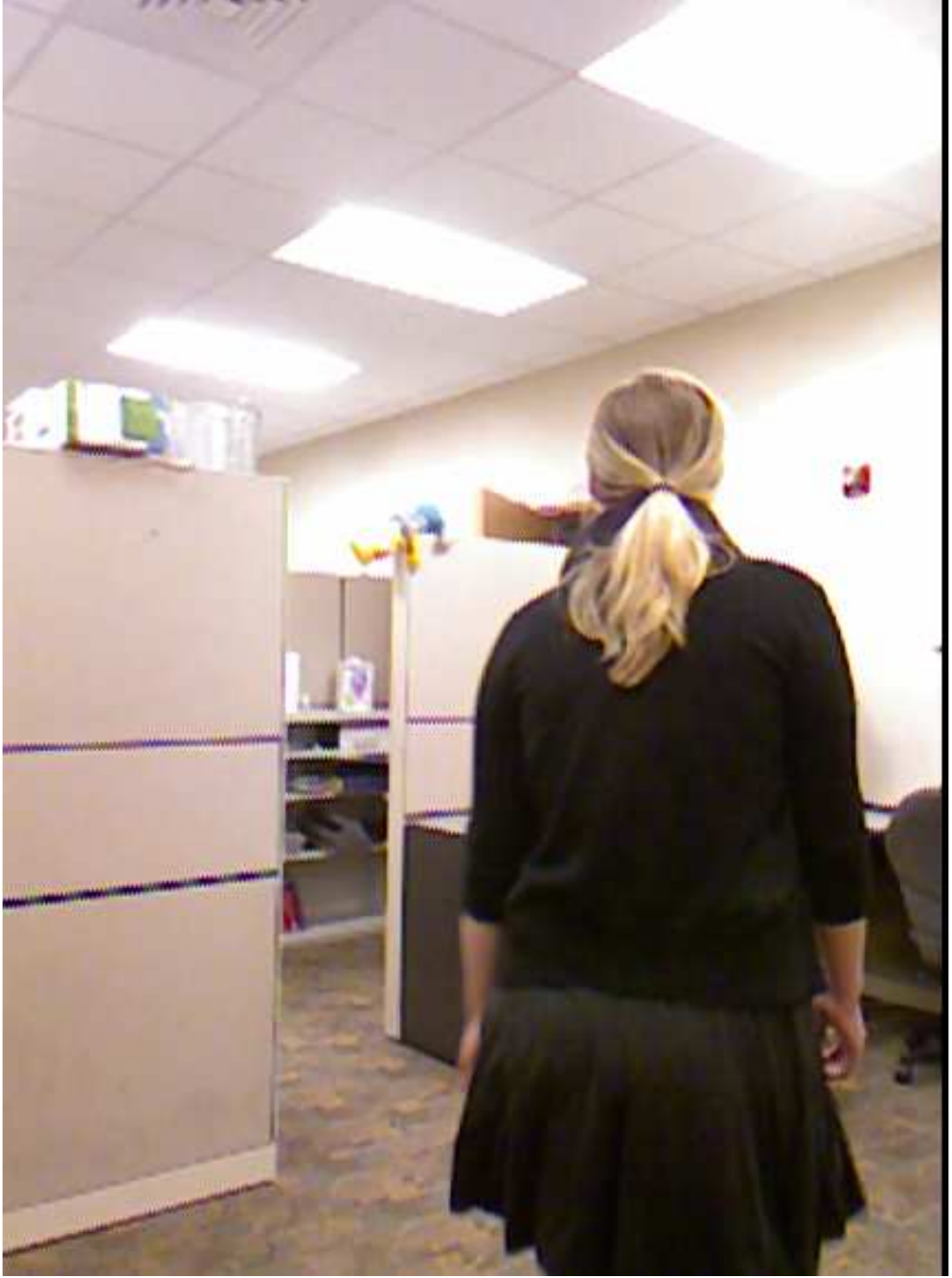} &
			\hspace{-0.3cm}\includegraphics[width=0.23\linewidth]{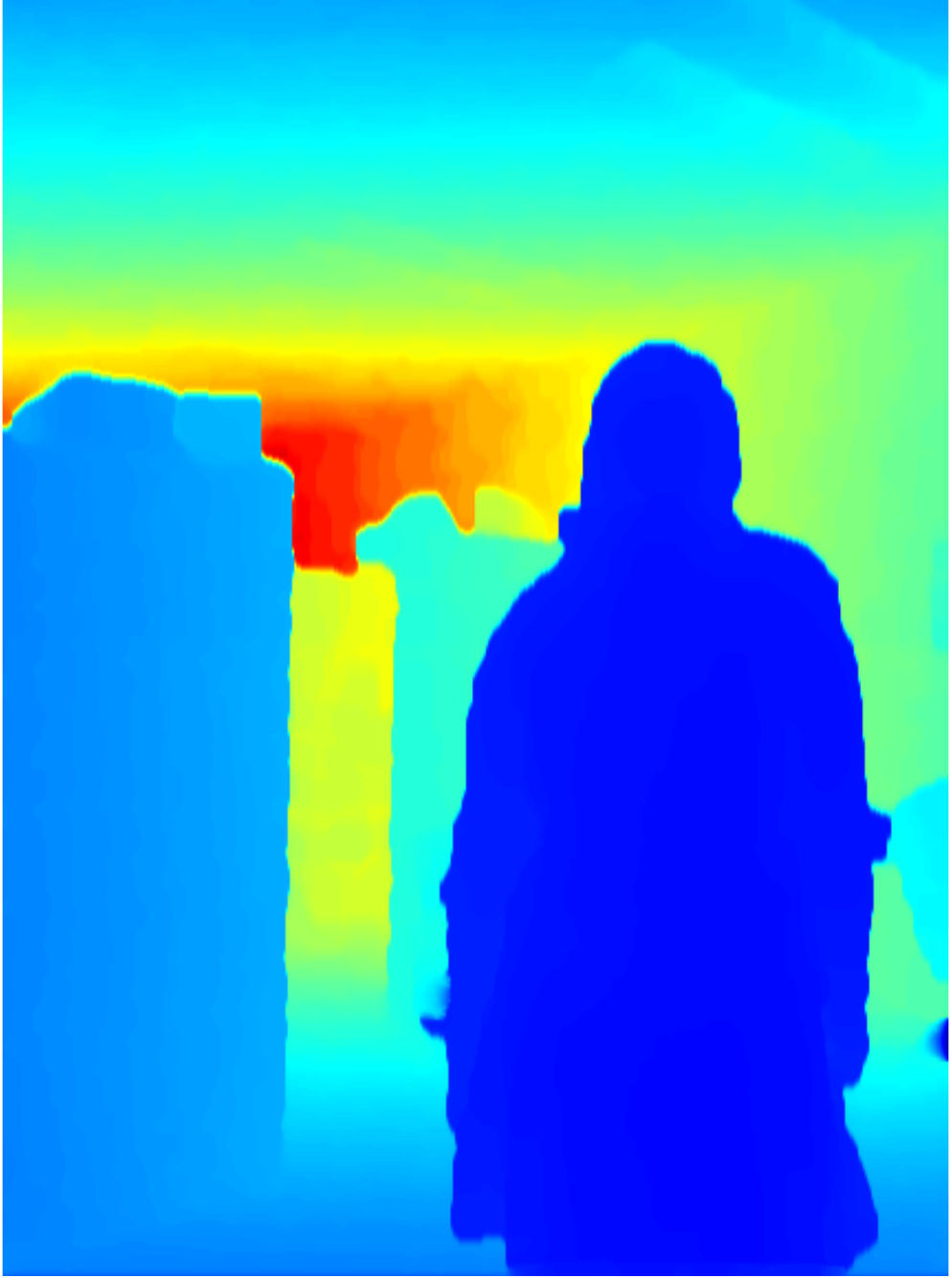} &
			\hspace{-0.3cm}\includegraphics[width=0.23\linewidth]{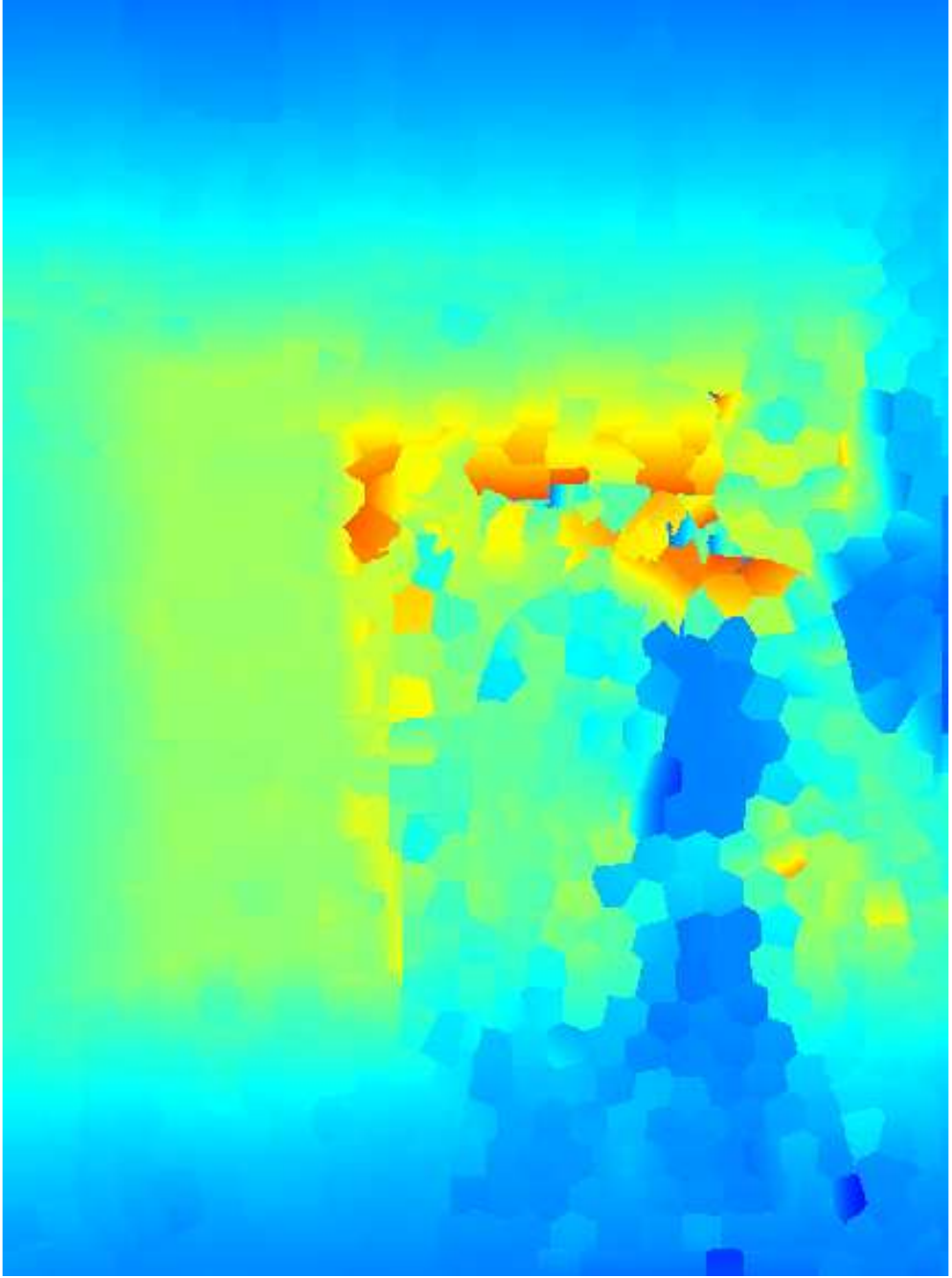} &
			\hspace{-0.3cm}\includegraphics[width=0.23\linewidth]{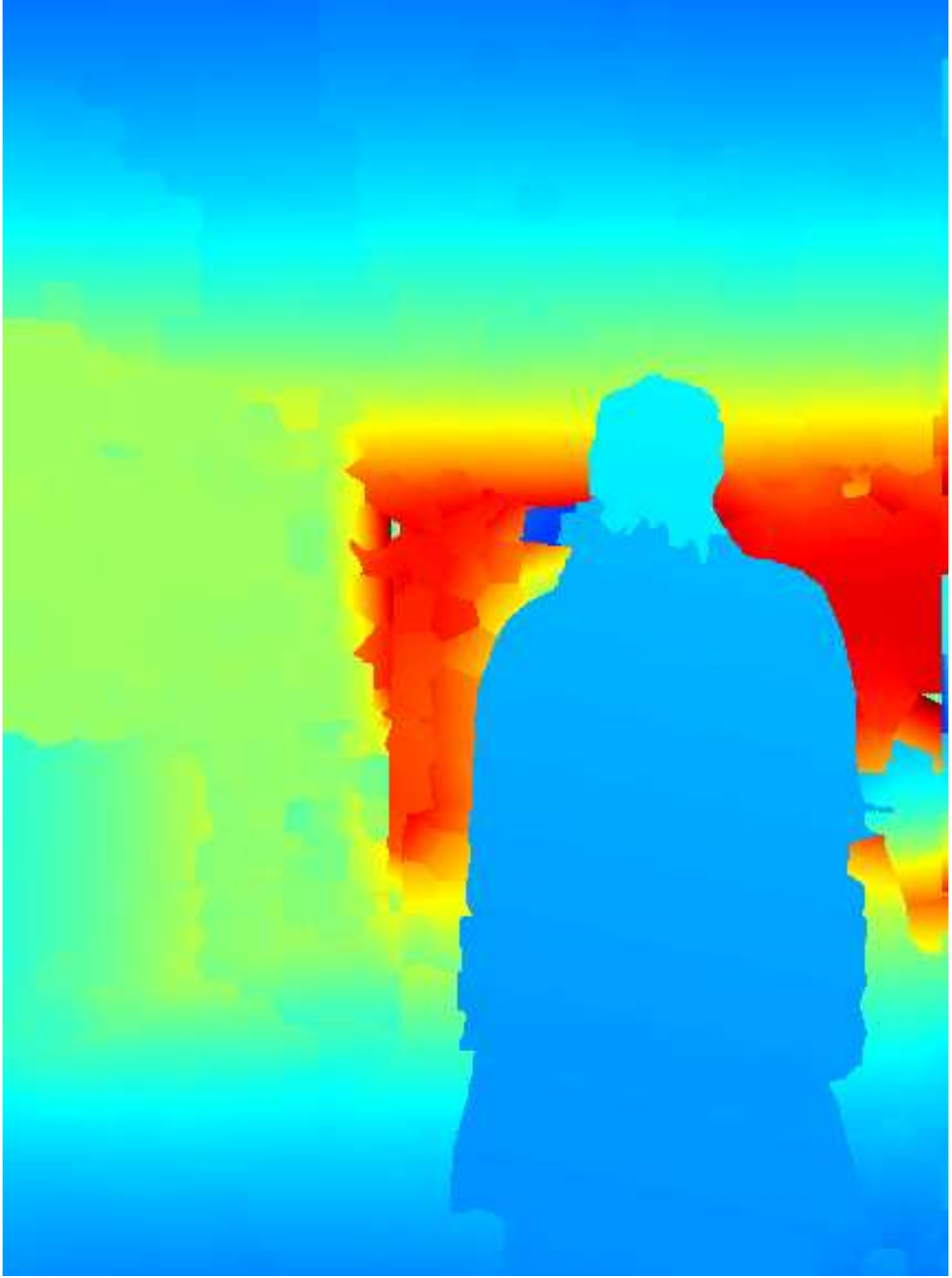}\\
			\hspace{-0.3cm}{\small Images}&\hspace{-0.3cm}{\small GT Depth}&\hspace{-0.3cm}{\small Ours-2F-NM}&\hspace{-0.3cm}{\small Ours-2F}
		\end{tabular}
	\end{small}
	\vspace{-0.2cm}
	\caption{MSR-V3D: Importance of the moving foreground classifier. While numerical errors appear to be very similar, our two-frame CRF with moving foreground classifier clearly yields more realistic results than without it.}
	\label{fig:motioncomp}
\end{figure}

\vspace{0.2cm}
\noindent {\bf Ablation study and comparison with Ours-1F:}\mbox{}\\
We then compare different parts of our model to evaluate their importance in our final results. Furthermore, we compare these results with those obtained by applying our single image model on the individual frames of the sequences. Note that, here, we did not apply the rescaling procedure of~\cite{Karsch12}, thus more clearly evaluating the actual results of the approaches. In Table~\ref{table:compcvprMotion}, we provide the three error metrics for our single frame approach, our two-frame CRF without (Ours-2F-NM) and with (Ours-2F) moving foreground classifier and our final model (Ours-Vid). All errors are reported after applying the pixel validity mask. Note that all the different versions of our video-based approach outperform our single frame model, thus showing the benefits of accounting for temporal information. While, from the numerical values, the benefits of our moving foreground classifier may not be obvious, looking at the qualitative results of Fig.~\ref{fig:motioncomp} clearly evidences its importance; the classifier is crucial for the realism of the recovered depth. Similarly, the numerical values do not evidence the benefits of our final model. However, zooming in on the resulting depth maps of Fig.~\ref{fig:cvprcomp} shows the smoothing effect of our final model, both spatially and temporally, which nonetheless still preserves this realism of the two-frame CRF. 


\begin{table}[t!]
	\centering
	\begin{tabular}{|c |c | c| c| c|}
		\hline
		Dataset & & ${\bf rel}$ & ${\bf log}_{10}$ & ${\bf rms}$ \\ 
		\hline
		\multirow{3}{*}{Building 2} &~\cite{Liu14}&0.246 & 0.134 & 2.05\\
		& Ours-2F-NM&{\bf 0.240}&{\bf 0.127}&{\bf 1.98}\\ 
		& Ours-2F&0.243 &0.139&2.11 \\ 
		& Ours-Vid&0.244 &0.138&2.12 \\ 
		\hline
		\multirow{3}{*}{Building 3}&~\cite{Liu14}&  0.258 & 0.091 & 1.34\\
		& Ours-2F-NM& 0.256 & 0.088 & 1.31 \\
		& Ours-2F& {\bf 0.240}&{\bf 0.083} &{\bf 1.23} \\
		& Ours-Vid& 0.241 & {\bf 0.083}  & {\bf 1.22} \\
		\hline
		\multirow{3}{*}{Building 4}&~\cite{Liu14}& 0.280 & 0.120 & 1.62\\
		& Ours-2F-NM& 0.282& 0.121 &{\bf 1.57}\\
		&Ours-2F& {\bf 0.268}&{\bf 0.111} &{\bf 1.57}\\
		& Ours-Vid& 0.275& 0.114 & 1.61\\
		\hline
		\multirow{3}{*}{All}&~\cite{Liu14}& 0.261 & 0.115 & 1.67\\
		&Ours-2F-NM& 0.259& 0.112 &{\bf 1.62}\\
		&Ours-2F& {\bf 0.250}&{\bf 0.110} & 1.63\\
		&Ours-Vid& 0.253& 0.112 &1.65\\
		\hline
	\end{tabular}
	\vspace{0.2cm}
	\caption{MSR-V3D: Comparison of the different parts of our model and of our single-frame method. Note that all the parts of our video-based approach outperform our single frame model. While the numerical values do not always evidence the benefits of the different components, the qualitative results show them more clearly.}
	\label{table:compcvprMotion}
\end{table}

\begin{figure}[t!]
	\begin{tabular}{cccc}
	 \hspace{-0.3cm}\includegraphics[width=0.23\linewidth]{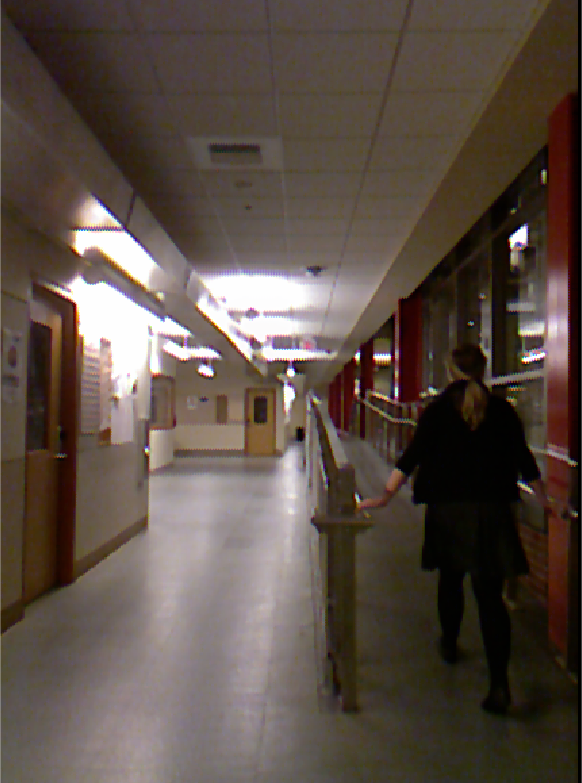}&\hspace{-0.3cm} \includegraphics[width=0.23\linewidth]{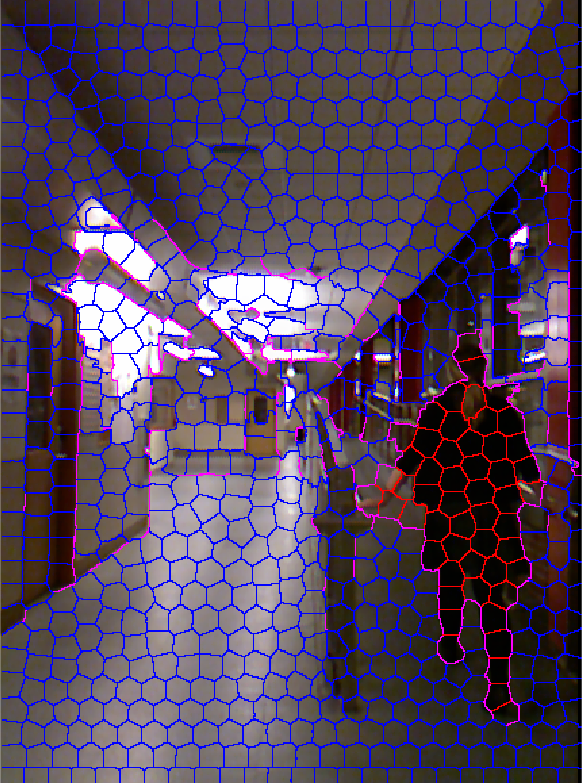}& \hspace{-0.25cm}\includegraphics[width=0.23\linewidth]{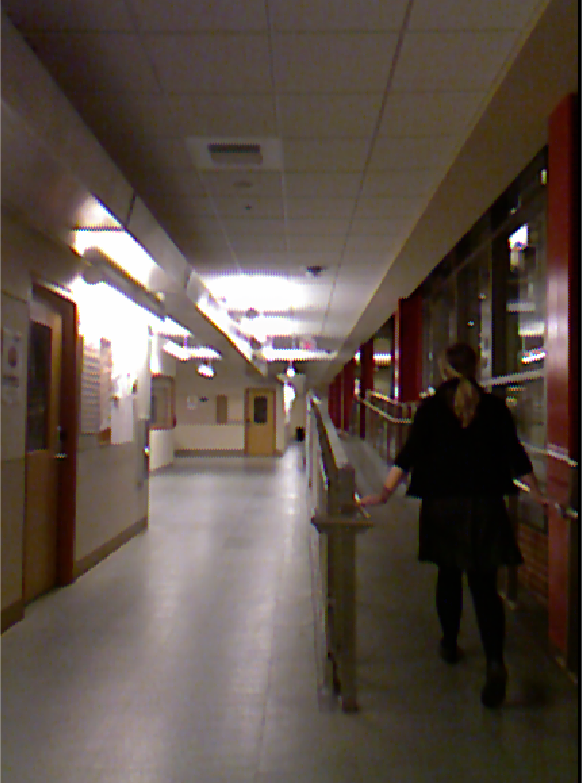}&\hspace{-0.3cm} \includegraphics[width=0.23\linewidth]{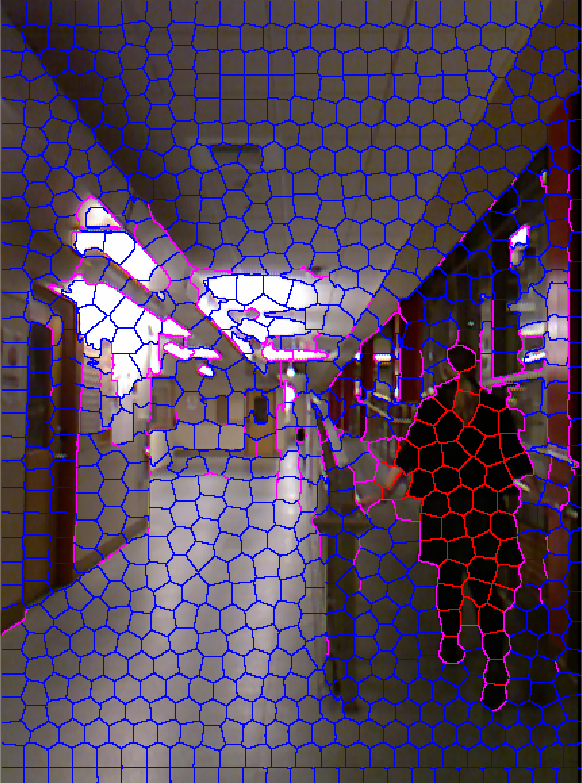}
	\end{tabular}
	\caption{Estimated occlusion boundary map for two consecutive frames from our two-frame CRF model (Ours-2F). The superpixel boundaries for the background are shown in blue. Pixels in magenta denote the occlusion boundaries for the background. Pixels in red show the superpixel boundaries belonging to the foreground.}
\end{figure}

\begin{table}[t!]
	\centering
	\begin{tabular}{|c |c c| c| c|}
		\hline
		Method & & ${\bf rel}$ & ${\bf log}_{10}$ & ${\bf rms}$ \\ 
		\hline
	\multirow{2}{*}{DepthTransfer~\cite{Karsch12}}&${\bf C_1}$&0.331 & 0.123 &  14.1\\
	&${\bf C_2}$&0.234 &  0.100 & 12.0\\
	\hline
		\multirow{2}{*}{Ours}&${\bf C_1}$&0.168 &  0.095 &  11.0\\
		&${\bf C_2}$&0.168 & 0.069 & 6.54\\
		\hline
	\end{tabular}
	\vspace{0.2cm}
	\caption{MSR-V3D: Depth reconstruction errors for DepthTransfer~\cite{Karsch12} and for our method evaluated according to two criteria (${\bf C_1}$ and ${\bf C_2}$) on pure rotational sequences in Building 1.}
	\label{tab:purerotation}

\end{table}

\section{Conclusion}
In this paper, we have addressed the problem of estimating depth from monocular video sequences. In particular, we have tackled the scenario of video sequences acquired by non-translational cameras or depicting dynamic scene. To this end, we have first introduced an approach to estimating the depth from a single image, which relies on continuous variables to represent the depth of image superpixels, and discrete ones to encode relationships between neighboring superpixels. This has let us formulate depth estimation as inference in a higher-order, discrete-continuous graphical model, which we have performed using particle belief propagation. Our experiments have shown that this model lets us effectively reconstruct general scenes from still images in both the indoor and outdoor scenarios. To achieve our ultimate goal of video-based depth estimation, we have then extended our single image model to an approach that estimates depth in two consecutive frames, which has allowed us to inherently model short-range temporal relationships, as well as to handle moving objects. Finally, we have modeled longer-range temporal dependencies by making use of a fully-connected CRF. Our experiments have shown that our approach can effectively estimate the depth maps of challenging video sequences, thus yielding state-of-the-art results. In the future, we intend to explore how information about the scene structure, the objects in the scene and the pixel semantic labels can be leveraged to further improve monocular depth estimation.


%

\ifCLASSOPTIONcompsoc
  \section*{Acknowledgments}  
\else
  \section*{Acknowledgment}
\fi
\vspace{-0.12cm}
NICTA is funded by the Australian Government as represented by the Department of Broadband, Communications and the Digital Economy and the Australian Research Council through the ICT Centre of Excellence program.
\ifCLASSOPTIONcaptionsoff
  \newpage
\fi



%

\bibliographystyle{ieee}

\end{document}